\documentclass[]{fairmeta}

\title{Omnilingual MT: Machine Translation for \TOTALlanguages Languages}

\author[]{Omnilingual MT Team}
\author[\dagger]{Belen Alastruey} 
\author[\dagger]{Niyati Bafna}
\author[\dagger]{Andrea Caciolai} 
\author[\dagger]{Kevin Heffernan} 
\author[\dagger]{Artyom Kozhevnikov} 
\author[\dagger]{Christophe Ropers} 
\author[\dagger]{Eduardo S\'anchez} 
\author[\dagger]{Charles-Eric Saint-James} 
\author[\dagger]{Ioannis Tsiamas}

\author[\S]{Xiang ``Tony'' Cao}
\author[\S]{Chierh Cheng}
\author[\S]{Joe Chuang}
\author[\S]{Paul-Ambroise Duquenne}
\author[\S]{Mark Duppenthaler}
\author[\S]{Nate Ekberg}
\author[\S]{Cynthia Gao}
\author[\S]{Pere Lluís Huguet Cabot}
\author[\S]{João Maria Janeiro}
\author[\S]{Jean Maillard}
\author[\S]{Gabriel Mejia Gonzalez}
\author[\S]{Holger Schwenk}
\author[\S]{Edan Toledo}
\author[\S]{Arina Turkatenko}
\author[\S]{Albert Ventayol-Boada}

\author[\ddagger]{Rashel Moritz}
\author[\ddagger]{Alexandre Mourachko}
\author[\ddagger]{Surya Parimi}
\author[\ddagger]{Mary Williamson}
\author[\ddagger]{Shireen Yates}

\author[\perp]{David Dale}
\author[\perp]{Marta R. Costa-jussà}

\affiliation[]{FAIR at Meta}
\contribution[\dagger]{Core Contributors, alphabetical order}
\contribution[\S]{Other Contributors, alphabetical order}
\contribution[\ddagger]{Project Management, alphabetical order}
\contribution[\perp]{Technical Leadership, alphabetical order}

\abstract{Advances made through No Language Left Behind (NLLB) have demonstrated that high-quality machine translation (MT) scale to 200 languages. Later Large Language Models (LLMs) have been adopted for MT, increasing in quality but not necessarily extending language coverage. Current systems remain constrained by limited coverage and a persistent generation bottleneck: while cross-lingual transfer enables models to somehow understand many undersupported languages, they often cannot generate them reliably, leaving most of the world’s 7,000 languages—especially endangered and marginalized ones—outside the reach of modern MT. Early explorations in extreme scaling offered promising proofs of concept but did not yield sustained solutions.

We present Omnilingual Machine Translation (OMT), the first MT system supporting more than \TOTALlanguages languages. This scale is enabled by a comprehensive data strategy that integrates large public multilingual corpora with newly created datasets, including manually curated MeDLEY bitext, synthetic backtranslation, and mining, substantially expanding coverage across long-tail languages, domains, and registers. To ensure both reliable and expansive evaluation, we combined standard metrics with a suite of evaluation artifacts: \blaser quality estimation model (reference-free), \omnitox toxicity classifier, \bouquet dataset (a newly created, largest-to-date multilingual evaluation collection built from scratch and manually extended across a wide range of linguistic families), and Met-\bouquet dataset (faithful multilingual quality estimation at scale).
We explore two ways of specializing an LLM for machine translation: as a decoder-only model (\omtllama) or as a module in an encoder–decoder architecture (\nllbtwo). The former consists of a model built on \llamathree, with multilingual continual pretraining and retrieval-augmented translation for inference-time adaptation. The latter is a model built on top of a multilingual aligned space (\sonaromni, itself also based on \llamathree), and introduces a training methodology that can exploit non-parallel data, allowing us to incorporate the decoder-only continuous pretraining data into the training of an encoder–decoder architecture.
Notably, all our 1B to 8B parameter models match or exceed the MT performance of a 70B LLM baseline, revealing a clear specialization advantage and enabling strong translation quality in low-compute settings. Moreover, our evaluation of English-to-\TOTALlanguages translations further shows that while baseline models can interpret undersupported languages, they frequently fail to generate them with meaningful fidelity; \omtllama models substantially expand the set of languages for which coherent generation is feasible. Additionally, OMT models improve in cross-lingual transfer, being close to solving the “understanding” part of the puzzle in MT for the \TOTALlanguages evaluated.  Beyond strong out-of-the-box performance, we find that finetuning and retrieval-augmented generation offer additional pathways to improve quality for the given subset of languages when targeted data or domain knowledge is available.
Our leaderboard and main humanly created evaluation datasets (\bouquet and Met-\bouquet) are dynamically evolving towards Omnilinguality and freely available.
} %

\correspondence{Marta R. Costa-juss\`a at \email{costajussa@meta.com}, David Dale at \email{daviddale@meta.com}}

\metadata[Leaderboard and Available Evaluation]{\url{https://huggingface.co/spaces/facebook/bouquet}}

\usepackage{xspace}
\usepackage{longtable}
\usepackage{booktabs}
\usepackage{multirow}
\usepackage{xspace}
\usepackage{xspace}
\usepackage[utf8]{inputenc}
\usepackage{textgreek}

\usepackage{CJKutf8}
\usepackage{latexsym}
\usepackage{algorithm}
\usepackage{algpseudocode}
\usepackage{tikz}
\usetikzlibrary{calc}
\def\checkmark{\tikz\fill[scale=0.4](0,.35) -- (.25,0) -- (1,.7) -- (.25,.15) -- cycle;} 
\usepackage{rotating} %
\usepackage{tablefootnote}
\usepackage[utf8]{inputenc}
\usepackage[T2A,T1]{fontenc}
\usepackage{comment}
\usepackage[export]{adjustbox}

\usepackage{microtype}
\usepackage{tabto}
\usepackage{inconsolata}
\usepackage{graphicx}
\usepackage{booktabs}

\usepackage{amssymb}
\usepackage[table,dvipsnames]{xcolor}

\usepackage{chngcntr}
\counterwithin{table}{section}
\counterwithin{figure}{section}

\usepackage[T1]{fontenc}
\usepackage{lmodern}

\usepackage{tabularx}
\usepackage{longtable}

\usepackage{gb4e,cgloss}
\noautomath

\usepackage[mcolblock, section=section]{leipzig}

\usepackage{makecell}

\usepackage{caption}
\usepackage{subcaption}
\usepackage{newfloat}

\newcommand{\OmniMT}{{Omnilingual MT}\xspace}

\newcommand{\TOTALlanguages}{{1,600}\xspace} 
\newcommand{\biblelanguages}{{1,560}\xspace} %
\newcommand{\bouquet}{BOUQuET\xspace}
\newcommand{\metbouquet}{Met-BOUQuET\xspace}
\newcommand{\omnitox}{OmniTOX\xspace}
\newcommand{\sonaromni}{\textsc{OmniSONAR}\xspace}
\newcommand{\floresplus}{FLoRes+\xspace}
\newcommand{\floreshrl}{FLoRes-HRL\xspace}
\newcommand{\floreshard}{FLoRes-Hard\xspace}
\newcommand{\florestwoh}{FLoRes-200\xspace}
\newcommand{\blaser}{BLASER 3\xspace}

\newcommand{\bible}{Bible\xspace}
\newcommand{\chrf}{ChrF++\xspace}
\newcommand{\xcomet}{xCOMET\xspace}
\newcommand{\metricx}{MetricX\xspace}
\newcommand{\metricX}{MetricX\xspace}
\newcommand{\Xeng}{XX-En\xspace}
\newcommand{\engX}{En-YY\xspace}
\newcommand{\XX}{XX-YY\xspace}

\newcommand{\medley}{MeDLEy\xspace}
\newcommand{\seed}{\medley}

\newcommand{\primary}{\textsc{OMT Primary}\xspace}
\newcommand{\langwise}{\textsc{OMT Langwise}\xspace}
\newcommand{\mined}{\textsc{OMT Mined Data}\xspace}
\newcommand{\backtranslation}{\textsc{OMT Backtranslated Data}\xspace}

\newcommand{\ccweb}{\textsc{CC-2000-Web}\xspace}
\newcommand{\ccpdf}{\textsc{CC-2000-Pdf}\xspace}

\newcommand{\nllbdata}{\textsc{CC-NLLB-200}\xspace}

\newcommand{\dclmedu}{\textsc{DCLM-Edu}\xspace}
\newcommand{\nllb}{\textsc{NLLB}\xspace}
\newcommand{\llama}{\textsc{LLaMA}\xspace}
\newcommand{\llamathree}{\textsc{LLaMA3}\xspace}
\newcommand{\glotlid}{\textsc{GlotLID}\xspace}
\newcommand{\ourblocks}{\textsc{OMT-base-FTdata}\xspace}

\newcommand{\omtllama}{\textsc{OMT-LLaMA}\xspace} 
\newcommand{\omtnllb}{\textsc{OMT-NLLB}\xspace} 
\newcommand{\nllbtwo}{\omtnllb}

\makeatletter
\AddToHook{cmd/appendix/before}{\def\cref@section@alias{appendix}}
\AddToHook{cmd/appendix/before}{\def\cref@subsection@alias{appendix}}
\AddToHook{cmd/appendix/before}{\def\cref@subsubsection@alias{appendix}}
\makeatother

\newcommand{\ttt}[1]{\texttt{#1}}
\newcommand{\datasetname}{MeDLEy\xspace}

\newcommand{\numberofLRLs}{109\xspace}

\newcommand{\datasetnamesource}{\datasetname-source\xspace}

\newcommand{\nllbseed}{\nllb-Seed\xspace}

\newleipzig{ger}{ger}{gerund}
\newleipzig{prn}{prn}{pronoun}
\newleipzig{sjv}{sjv}{subjunctive}

\newcommand{\rus}[1]{\foreignlanguage{russian}{#1}}

\DeclareFloatingEnvironment{boxes} %
\newcommand{\boxref}[1]{\hyperref[{#1}]{Box~\ref*{#1}}}

\begin{document}
\enlargethispage{3cm}
\maketitle

\clearpage
\newpage
\tableofcontents
\clearpage
\newpage

\section{Introduction}
\label{sec:introduction}

The recent success of No Language Left Behind (NLLB) \citep{nllb-24} marked a turning point in multilingual translation. By demonstrating that high-quality MT could be extended to 200 languages, NLLB reshaped the research landscape and set a new standard for linguistic inclusion. It catalyzed new data pipelines, evaluation frameworks, and community partnerships that continue to benefit the entire field—including the work we present here. But NLLB also revealed a deeper asymmetry in multilingual MT. Modern models can often recognize or interpret long-tail languages through cross-lingual transfer, yet they struggle to produce them reliably. This generation bottleneck is compounded by a static, training-time definition of coverage: languages with little or no data simply never enter the system. Together, these constraints leave most of the world’s 7,000 languages—especially endangered and underdocumented ones—effectively outside the reach of current MT technology. Early attempts to explore extreme scaling, such as Google’s Massively Multilingual Translation project \citep{siddhant2022towards}, demonstrated the feasibility of reaching toward 1,000 languages, but these efforts did not evolve into sustained work, and progress toward broader global coverage has stalled. However, notable progress has been made towards improving quality for top priority languages with decoder-only architectures (Large Language Models, LLMs) e.g.  \citep{tower_llm_2024,dang2024ayaexpansecombiningresearch, gemma3}. 

In this work, we introduce \textbf{Omnilingual Machine Translation (Omnilingual MT)}, a family of multilingual translation systems that extend support to more than \TOTALlanguages languages, the broadest coverage of any benchmarked MT system to date. To start, our data efforts included assembling and curating one of the largest and most diverse multilingual corpora to date, drawing from prior massive collections while substantially expanding coverage through new human-curated and synthetic data pipelines. More specifically, we integrated material from large-scale public sources and augmented them with newly created resources—including manually curated seed datasets and synthetic backtranslation—to address long-tail gaps in domains, registers, and under-documented languages. 

Omnilingual MT explores two complementary ways of specializing LLMs for translation: as a standalone decoder-only model and as a block within an encoder–decoder architecture.
In the first approach, we extend \llamathree–based decoder-only models with a multilingual continual-pretraining recipe and retrieval-augmented translation for inference-time adaptation. In the second approach, we employ a cross-lingually aligned encoder (\sonaromni \citep{sonaromni} built itself on top of \llamathree) to build an encoder–decoder architecture that maintains the size of the original NLLB model while expanding its language coverage through a novel training methodology that exploits non-parallel data, reusing the continual-pretraining data from the decoder-only model.
Both approaches share an expanded 256K-token vocabulary and improved pre-tokenization for underserved scripts, enabling large-scale language expansion to cover over 1,000 languages.

To ensure both reliable and expansive evaluation, we combined standard metrics such as MetricX and ChrF with a suite of evaluation artifacts developed for this effort. These include \blaser quality estimation model (reference-free), \omnitox toxicity classier, \bouquet dataset (a newly created, largest-to-date multilingual evaluation collection built from scratch and manually extended across a wide range of linguistic families), and Met-\bouquet dataset (which provides faithful multilingual quality estimation at scale).

Omnilingual MT expands the number of languages that modern models ``understand sufficiently well'' twofold, from about 200 to over 400 languages. Moreover, it offers non-trivial performance when translating from \TOTALlanguages and into about 1,200 languages, outperforming all competitive translation systems by a large margin and establishing new (and often first) state-of-the-art (SOTA) results for the majority of these \TOTALlanguages languages.

 Notably, we show that specialized MT models offer superior efficiency–performance tradeoffs compared to general-purpose LLMs. More specifically, our 1B to 8B parameter models match or exceed the MT performance of a 70B-parameter LLM baseline, revealing a clear Pareto advantage: specialization, not scale, is perhaps a more reliable path to high-quality multilingual translation. This efficiency extends the practical reach of the model, enabling strong MT performance in real-world, low-compute contexts. In addition, our systematic evaluation of Omnilingual MT on English-to-\biblelanguages Bible translations reveals a striking pattern: many baseline models can interpret undersupported languages, yet they often fail to generate them with even remote similarity to the target. Omnilingual MT substantially widens the set of languages for which coherent generation is possible, reinforcing the central claim of this work—that large-scale MT coverage requires not only cross-lingual understanding but robust language generation, which current baselines do not reliably provide. Beyond strong out-of-the-box performance, we analyze how targeted techniques, such as finetuning and retrieval-augmented generation, can further boost translation quality for individual languages. %
With this, Omnilingual MT not only provides broad coverage but also offers flexible pathways for further improving performance when additional data or domain knowledge is available.

Although Omnilingual MT is primarily designed for translation, we consider it as a general-purpose multilingual base model. Its architecture can be further trained to build multilingual LLMs, enabling future research that integrates translation, reasoning, dialog, and multimodal capabilities in thousands of languages. Moreover, with the recent release of Omnilingual ASR \citep{omnilingualasrteam2025omnilingualasropensourcemultilingual}, Omnilingual MT can be cascaded with large-scale speech recognition to produce speech-to-text translation systems operating at a scale previously unattainable. 
The Omnilingual MT recipes for building models with unprecedented language support can in principle be reproduced on top of diverse base language models, and we hope that they will inspire communities, researchers, and practitioners to build systems that evolve alongside the world’s languages.

The main claims of our line of work are outlined in \Cref{tab:opensource}. Our translation models are built on top of freely available models. \bouquet and \metbouquet and the adjacent leaderboard are freely available\footnote{\url{https://huggingface.co/spaces/facebook/bouquet}}.

\begin{table*}
\centering
    \begin{tabular}{llp{3.2cm}p{7.5cm}}
        \toprule
        Name & Type & Description & Claim\\
        \midrule
       \omtllama & models family & \llamathree-based models for MT (different sizes) + extendable recipes %
       & \TOTALlanguages MT coverage with non-trivial performance and above baselines (Section \ref{sec:deconly}).  \\
        \nllbtwo & model & Encoder-decoder MT model built on top of \sonaromni &  \TOTALlanguages to 250 MT coverage with non-trivial performance and above baselines (Section \ref{sec:encdec}). \\
        \midrule
        \blaser & model & \sonaromni-based model for MT quality estimation & Highest multilingual coverage in reference-free MT quality estimation (\TOTALlanguages+ languages in the source-side). Outperforming previous metrics by several points in correlation with human judgments on 
        Met-BOUQuET 
        (Section \ref{sec:blaser}). \\
        \omnitox & model & \sonaromni-based model for toxicity detection & Largest language coverage in toxicity detection (\TOTALlanguages languages) while outperforming its predecessor 200-language MuTox model by +0.06 mean per-language ROC AUC, achieving 0.86 across 30 diverse languages (Section \ref{sec:omnitox}).\\
        \midrule
        \medley & dataset & MT training dataset & Large-language scale training data collection created from scratch in multiple languages and manually extended finally covering 108 extremely low-resource languages (Section \ref{sec:seed}) \\
        \bouquet & dataset & MT benchmark & Largest language scale evaluation data collection created from scratch in multiple languages and manually extended finally covering 275 languages including a wide variety of linguistic families and resources (Section \ref{sec:bouquet})\\
        \metbouquet & dataset & Human judgments of MT quality & Largest language scale human annotations dataset designed to cover 161 language directions (Section \ref{sec:metbouquet}).  \\
        \bottomrule
    \end{tabular}
    \caption{A summary of the corresponding claims of our line of work.}  \label{tab:opensource}
\end{table*}

\section{Expanding Machine Translation}
\label{sec:expandingMT}

Recent advances in multilingual MT have demonstrated that high-quality translation can extend far beyond high-resource languages. Most notably, the trajectory is best represented by NLLB \citep{nllb-24}, which demonstrated that it is possible to deliver strong translation quality for 200 languages, setting a new standard for linguistic inclusion. More specifically, NLLB illustrated that the long-standing curse of multilinguality-the tendency for quality to degrade as the number of languages increases—was not an insurmountable barrier. Through large-scale data curation, targeted architecture choices, and multilingual optimization strategies, NLLB achieved both breadth and quality, overturning the assumption that scaling coverage inevitably sacrifices performance. Since the release of NLLB, the work has reshaped the multilingual MT ecosystem in several ways, including establishing \florestwoh as the de facto evaluation standard, catalyzing new data pipelines across academia and industry, and enabling dozens of downstream models, fine-tuning efforts, and community adaptation projects that continue to rely on its multilingual backbone.

Despite the impact of this effort, NLLB and other projects in this current landscape continue to leave the vast majority of the world’s 7,000 languages, especially endangered or marginalized, largely absent from technological representation. As a result, coverage plateaus at roughly the same frontier across systems. Compounding this issue, many models exhibit cross-lingual understanding of underserved languages through transfer, yet consistently fail to generate them with meaningful fidelity, revealing a generation bottleneck that further limits practical support for long-tail languages.

That said, several projects have started to explore scaling MT beyond the 200-language ceiling. Google’s Massively Multilingual Translation work investigated models covering up to 1,000 languages, offering an early proof of concept that extreme multilingual scaling was technically possible \citep{siddhant2022towards}. These efforts demonstrated that multilingual transfer can be leveraged even in very low-resource conditions. However, they did not yield sustained, or extensible systems, and none produced a practical path for continual expansion. Subsequent multilingual MT systems, including large decoder-only LLMs, mostly reporting MT quality improvements on top priority languages, have also increased language coverage indirectly. Their big-scale pretraining exposes them to a broader, if uneven, range of languages than purpose-built MT systems, allowing them to exhibit surprising zero-shot and few-shot translation abilities. Improvements in reasoning, instruction-following, and cross-lingual representations have also provided new avenues for multilingual transfer. Yet these gains remain dominated by high-resource languages, and large LLMs remain inefficient MT systems—often requiring tens of billions of parameters to match the MT performance of much smaller specialized models (see Section \ref{sec:evaluations}). %

The expansion of MT is compounded by other problems. Long-tail languages, for one, bring substantial linguistic diversity ranging from rich morphological systems and agglutinative patterns to unique orthographic and script traditions—with available written data often distributed across different formats and community contexts \citep{magueresse2020low}. These linguistic and sociocultural features expose the brittleness of closed-coverage systems: adding a language requires far more than acquiring data; it requires modeling choices that account for typological diversity and social context. Furthermore, while large-scale multilingual corpora—including Bible-based datasets, Gatitos \citep{jones-etal-2023-gatitos} and SMOL \citep{caswell2025smol} with parallel texts, and large-scale web datasets like FineWeb2 \citep{penedo2025fineweb2pipelinescale} or HPLT 3.0 \citep{oepen2025hplt30largescalemultilingual}—have expanded the availability of multilingual training text, these corpora exhibit systematic gaps.%
They disproportionately represent formal registers, religious domains, and well-documented language families while underrepresenting dialect variation, colloquial styles, and many of the world’s marginalized languages. As a result, increasing dataset size does not reliably translate into broader or more equitable coverage. However, recent work using synthetic data \citep{zebaze2025topxgentopicdiverseparalleldata}, bitext mining, and multilingual transfer \citep{oepen2025hplt30largescalemultilingual} has proven to be helpful in extending coverage.

In addition, large-scale evaluation remains a major bottleneck for multilingual MT. \floresplus \citep{goyal2022flores}, %
and the Aya benchmark \citep{singh-etal-2024-aya} provide high-quality evaluation for hundreds of languages, but none provide coverage beyond 200–300 languages (there are few recent exceptions to this \citep{chang2025chikhapolargescalemultilingualbenchmark}) . Reference-based metrics also struggle at scale: BLEU and \chrf fail to capture meaning adequacy, while reference-free metrics such as COMET, \blaser and MetricX require careful calibration and validation in typologically diverse languages (see metric cards in Appendix \ref{app:cards}). 
Without reliable evaluation for long-tail languages, progress becomes difficult to measure and even harder to compare across systems. This problem becomes acute when scaling to 1,000+ languages, where many systems can produce outputs that appear fluent yet remain unintelligible or unrelated to the target language, making generation accuracy particularly challenging to assess. The field requires multilingual quality-estimation frameworks that scale to thousands of languages while preserving metric fidelity.

Taken together, these limitations highlight that progress in massively multilingual MT %
now depends less on marginal model improvements than on rethinking how systems can grow, adapt, and represent the world’s linguistic diversity. What is needed is not only broader coverage but deeper support—models that can generate underserved languages robustly, operate efficiently at smaller scales, and provide reliable evaluation mechanisms for long-tail settings. Our goal is to operationalize this shift. Rather than building another large model centered on high-resource performance, we design \OmniMT to address the structural challenges of extreme coverage: data scarcity, typological diversity, long-tailed language generation, efficiency–performance tradeoffs, and the absence of evaluation frameworks for \TOTALlanguages+ languages.

This perspective also motivates how we organize the remainder of the paper. In \Cref{sec:languages}, we move from the structural challenges outlined above to the linguistic realities of scaling to \TOTALlanguages+ languages. This section is specially relevant to inform about the concept of language; and related language features such as what does it take to qualify as pivot language, relevance of context, how to determine resource-levels. %

\Cref{sec:newdata} presents the data contributions in this work with special focus on under-represented languages. This section reports several well-known directions to expand data for pretraining MT models. Additionally, it reports more innovative diverse and representative post-training and evaluation datasets.

The three subsequent sections---\Cref{sec:tmoverview}, \Cref{sec:deconly}, and \Cref{sec:encdec}---describe the translation model architectures that we propose. We report several ablations to motivate our modeling decisions.

Next, \Cref{sec:mtmetrics} is dedicated to the contributions that we make towards expanding the MT metrics to Omnilinguality.  We propose a variation of human evaluation protocol (XSTS+R+P) to better represent languages outside of English, build the largest human annotations collection on language coverage on MT quality (Met-\bouquet), propose the largest multilingual MT quality metric (\blaser), and the largest multilingual toxicity detector (\omnitox).

\Cref{sec:evaluations} reports the final results of our MT models evaluation focusing on answering questions such as language coverage and relative performance to external baselines.

The final sections focus on key features of the MT adoption problem space. \Cref{sec:familyofmodels} demonstrates how our smaller models achieve performance improvements over, or parity with, larger models. \Cref{sec:extensibility} addresses the growing trend of researchers fine-tuning NLLB for machine translation in their languages and adapting smaller \llama models to various language-specific tasks, including translation. Building on this momentum, we demonstrate that our models are architecturally designed to facilitate such extensions and adaptations.
The findings presented in this paper underscore the importance of continued investment in specialized models to enhance translation quality and expand language coverage in MT. Finally, Section \ref{sec:conclusions} summarizes the conclusions and discusses the social impact of our work.

\section{Languages}
\label{sec:languages}

\subsection{Referring to languages}
In the absence of a strict scientific definition of what constitutes a \textit{language}, we arbitrarily started considering as language candidates, and referring to those candidates as languages, those linguistic entities—or \textit{languoids}, following \citet{good2006modeling}—that have been assigned their own ISO 639-3 codes.

We acknowledge that language classification in general, and the attribution of ISO 639-3 codes in particular, is a complex process, subject to limitations and disagreements, and not always aligned with how native speakers themselves conceptualize their languages. To allow for greater granularity when warranted, ISO 639-3 codes can be complemented with Glottolog languoid codes \citep{hammerstrom2024glottolog}. 

Additionally, as some languages can typically be written using more than a single writing system, all languages supported by our model are associated with the relevant ISO 15924 script code. For example, we use \texttt{cmn\_Hant} to denote Mandarin Chinese written in traditional Han script and \texttt{cmn\_Hans} for the same language written in simplified Han script. 
When counting languages throughout this paper, we typically count the distinct combinations of the language and the writing system, identified by the pair of ISO 639-3 and ISO 15924 codes. 

Finally, the use of the phrases \textit{long-tail languages} 
and \textit{underserved languages} also needs further defining. There are over 7,000 languages used in the world today used by over 8 billion human beings. The number of users is not evenly distributed among those languages, far from it. It is estimated \citep{ethnologue2025} that slightly less than half of the world's population uses as their native languages (or L1) one of the 20 most used languages, which means that the other half uses as their L1 one of the remaining 7,000+ languages. The same authors\footnote{\url{https://www.ethnologue.com/insights/ethnologue200/}, last accessed 2026-02-18} estimate that 88\% of the world's population use as their L1 or L2 one of the 200 most used languages. Overall, we can see that the distribution of L1 users per language is quasi-zipfian, and therefore displays a conspicuous long tail (hence, our use of the phrase \textit{long-tail languages}). It is not uncommon for many of the long-tail languages to be considered underserved, as defined in the following section.

\subsection{Quality translations from or into underserved languages}
In this section we discuss the main impediments to the creation of high-quality training or evaluation data that could partially offset the lack of existing data for underserved languages, and present non-English-centric solutions as an alternative to existing translation workflows. We use the phrase \textit{underserved languages} as a synecdoche referring to communities of language users who do not have access to the full gamut of language technologies—and more specifically here to machine translation—in their respective native languages. The language technology industry often refers to those languages as \textit{low-resource languages} because of the small amount of available data. We discuss resource-level classification at greater length in the next section, as this kind of classification carries some degree of arbitrariness that warrants further explanations. 

The problem of low-quality translations into or out of underserved languages can be approached from at least two angles: training data and evaluation. On the training data front, mitigation strategies for observed quality issues entail creating additional parallel data; this is most often done by commissioning translations into underserved languages. From the standpoint of evaluation, quality issues can stem from the lack of evaluation datasets or the lack of useful human evaluation annotations. The common denominator to training data and evaluation shortcomings is the difficulty faced by the research community to commission high-quality work from proficient translators or bilingual speakers. 

\paragraph{\textbf{Determining pivot languages}} Receiving high-quality work products from proficient translators or bilingual speakers implies, firstly, having access to said speakers and, secondly, creating optimal conditions for quality work. When it comes to underserved languages, it is important to consider that the vast majority of those languages score high on the intergenerational disruption scale~\footnote{For additional information on language disruption and disruption scoring, please see \cite{lewis2010assessing}.}. High disruption typically occurs when different generations of speakers become geographically estranged due to drastic changes in labor and macroeconomic settings (e.g., when a country's economy shifts its primary source of production from the primary sector to the secondary or tertiary sector). Corollary to this shift is a massive displacement of younger generations from rural areas to urban business and higher education centers. As a result, the linguistic profiles of those generations become differentiated. The older generations are proficient native speakers of the underserved language and, in most cases, proficient second-language speakers of an official language of the country where they reside. The younger generations are native or near-native proficient speakers of the official language and of a business or research lingua franca (more often than not, English) but they are not as proficient in the underserved language. For the above reasons, pairing underserved languages with English in human translation work is not always the optimal solution. Alternatively, we also need to provide for the pairing of underserved languages with high-resource languages at which native speakers are proficient. In this project, we refer to those alternate high-resource languages as \textit{pivot languages} (e.g., Spanish used as a pivot language for translations into or out of Mískito [\texttt{miq}], or K'iche' [\texttt{quc}]).

\paragraph{\textbf{Providing contextual information}} Even when English is a possible—or the only available—pivot option, its lack of explicit grammatical markings is a constant reminder that sentences rarely speak for themselves, and that translators need a good amount of contextual information to produce quality translations, especially when moving away from the formal textual domain and closer to the conversational domain. For example, one of the many differences between those two domains is a shift from a predominance of unspecified third grammatical persons to first and second grammatical persons (often in the singular). In English, the pronouns \textit{I} and \textit{you} do not provide any intrinsic information about grammatical gender; in fact, \textit{you} does not even provide any distinctive information about grammatical number, which is not complemented either by any form of verbal or adjectival inflection. This causes ambiguities that translators cannot resolve on their own, which in turn may lead to mistranslations that are not due to lack of proficiency but rather lack of relevant information. The same is true of information about language register and formality. In the conversational domain, English provides very few formality markers, and identifying language register markers may require a very high level of proficiency, which is only accessible to translators with extensive cultural experience. To mitigate these problems, we first ensured that all sentences to be translated be included in a paragraph (or what would be the equivalent of a paragraph in speech). We also provided translators with additional information about the following: the overall domain in which the paragraphs are most likely to be found, the protagonists depicted or referred to in the paragraphs, the language register most likely to be used in such situations, and the overall tone of the paragraphs (if any specific tones were to be conveyed).

\subsection{Resource levels} 
\label{sec:resourcelevels}
Historically, languages in MT have typically been classified as either high-resource or low-resource (e.g., see WMT evaluations \citep{kocmi-etal-2024-findings,kocmi-etal-2025-findings}). This classification facilitates the analysis of MT performance in highly multilingual and massively multilingual settings, among other applications.

The definition of low-resource languages is somewhat arbitrary, or at the very least, highly dynamic, as additional resources may be created at any time. More broadly in NLP, this classification is based on the availability of corpora, dictionaries, grammars, and overall research attention. One widely used definition in the field of MT originates from the NLLB work \citep{nllb-24}, in which the authors differentiate between high- and low-resource languages based on the amount of parallel data available for each language (with "documents" predominantly consisting of single sentences). Specifically, the threshold is set at 1 million parallel documents, above which a language is considered high-resource.

We want to revisit this definition because we are dealing with a much larger amount of languages than NLLB; and works with similar amount of languages \citep{bapna2022buildingmachinetranslationsystems} do not provide an explicit definition; we want to optimize for this definition to correlate with MT quality; and given the large amount of languages that we are covering, we want to further fine-grain our language resource classification by splitting low resource languages into low and extremely low resource, and by distinguishing high- and medium-resourced languages. 

\begin{figure}[h!]
\centering
   \includegraphics[scale=1]{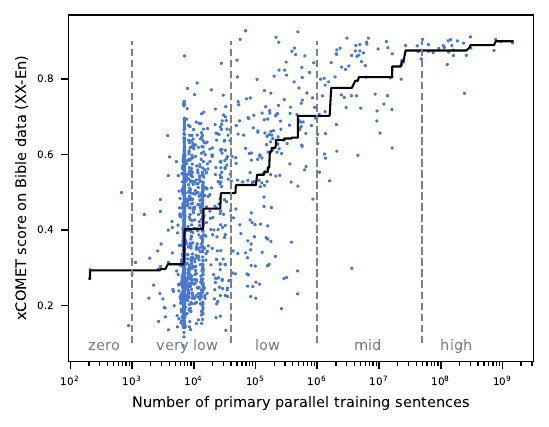} 
  \caption{\label{fig:resourcelevel} Correlation between translation quality (\omtllama model, Bible benchmark of \biblelanguages languages, mean xCOMET score) and amount of parallel documents from primary sources (not mined or synthetic). We fit an isotonic regression to show the global trend.
}
\end{figure}

Based on our experiments (see Figure \ref{fig:resourcelevel}), we confirm a correlation between translation quality and the amount of parallel documents available. A clear shift in translation quality is observed for languages with more than 1 million parallel documents, which, following the NLLB convention, we establish as the threshold for the "low-resource" designation. An additional qualitative change is observed at approximately 40K parallel documents: this corresponds to a corpus size comparable to that of the Bible, supplemented by at least one additional source of parallel training data.

\begin{figure}[h!]
\centering
   \includegraphics[scale=1]{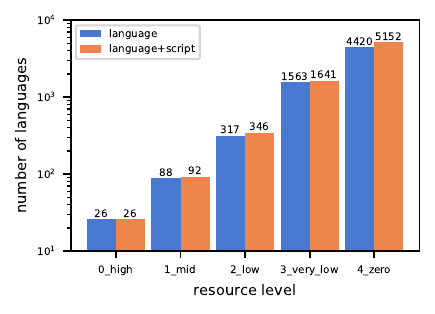} 
  \caption{\label{fig:graphdistributionoflanguages} Graph with distribution of languages per resource bucket. Note that we count all languages for which we have some data (including monolingual data and word-level parallel data like Panlex), but the buckets are determined based on the parallel data that is at least (and predominantly) sentence-level.
}
\end{figure}

Therefore, the final classification relies on parallel documents available, with a language considered high resource if we have more than 50M document pairs (for such languages, MT quality of most systems is predictably high), mid resource above 1M, low resource if we have parallel documents between 40K and 1M, extremely low resource if we have between 40K and 1K parallel documents, and zero-resource below 1K (mostly to indicate that their training data size is much lower than even a typical Bible translation or a seed corpus, often represented only by a few sentences in a multilingual resource like Tatoeba). See the graph with distribution of languages per resource bucket in Figure \ref{fig:graphdistributionoflanguages}.

This definition comes with some limitations. Most anomalies come from languages with less than 1k parallel documents on which we observe equally high translation quality. By doing a manual inspection on those languages, we could hypothesize that this bucket contains languages which are highly similar to other high resource languages and benefit from positive knowledge transfer. Another anomaly, but much more rare, is low-performing languages in the bucket of languages with more than 1M documents. In this case, again, by manual analysis we could hypothesize that there are languages for which we have extremely low quality data or very narrow domain distribution of it. Or, complimentary, rare scripts that are not well represented by the tokenizer and as a consequence low quality MT performance. Finally, sometimes we misattribute the available training data to other languages, due to loosely defined language boundaries (e.g. some data for a dialectal Arabic language could be identified simply with the \texttt{ara\_Arab} code, pointing to the Arabic macro-language without specifying the language).

\subsection{Describing languages in prompts} 
\label{sec:language_names}
When prompting both \omtllama models and instruction-following external baselines to translate, the precise format of describing the target language may affect the generation results. Different organizations prefer different language code formats, and to ensure interoperability of our prompts between diverse models, we opted for natural-language descriptions of the language varieties. 

Our template for language names includes the language name itself, followed by optional brackets with the script, locale, or a dialect: for example, \texttt{spa\_Latn} becomes ``Spanish'', \texttt{cmn\_Hans} becomes ``Mandarin Chinese (Simplified script)'',  \texttt{eng\_Latn-GB} turns into ``English (a variety from United Kingdom)'', and \texttt{twi\_Latn\_akua1239}, into ``Twi (Akuapem dialect)''. We omit the script description for the languages that are ``well-known'' (using inclusion into FLORES-200 as a criterion) and that are expected to be using one single script in an overwhelming majority of scenarios. 

For mapping the codes of languages, scripts and locales into their English names, we mostly rely on the Langcodes package,\footnote{\url{https://github.com/rspeer/langcodes}}, which in turn relies on the IANA language tag registry. For referring to dialects, we use their names from the Glottolog database\footnote{From the languoids table in \url{https://glottolog.org/meta/downloads}; currently we are using version 4.8.}, but employ them only for disambiguating otherwise identical language varieties in \floresplus.

\section{Creating High-Quality Datasets}\label{sec:newdata}

Access to high quality data is crucial to develop a high quality translation system. We give special focus to creating, selecting, and curating high-quality data for thousands of languages both for training and evaluation.  In this section, regarding training, we mainly discuss continual pretraining (CPT) data, while main data for postraining is directly discussed in the corresponding section (Section \ref{sec:post_training}).
For CPT, we leverage mainly datasets in Table \ref{tab:CPTtrainingdetails} and presented in Section \ref{sec:CPTtrainingdata}.
For evaluation, we rely on a subset of the Bible (\Cref{sec:biblebenchmarking}) and several standard ones—e.g., \floresplus~\citep{nllb-24}. 
 
 Beyond this, and to compensate for existing limitations in the existing data such as lack of long-tail languages, domains, registers and others, we curate new training datasets, both monolingual (inspired by~\citet{penedo2025fineweb2pipelinescale} and~\citet{kydlicek2025finepdfs}) and aligned (manual \medley, parallel data inspired by~\citet{nllb-24}, Section \ref{sec:seed}, and synthetic data, Section \ref{sec:syntheticdata}) as well as  the \bouquet evaluation dataset (Section \ref{sec:bouquet}).

\subsection{Main CPT Training Data Collection}\label{sec:CPTtrainingdata}

As follows, we mention and briefly categorize some highly multilingual text resources of various kinds: parallel and non-parallel, word-, sentence-, and document-level. Additionally, Table \ref{tab:CPTtrainingdetails} summarises the main sources and volumes used to train our systems.

\begin{table*}
\centering
\begin{tabular}{llll}
    \toprule
    \textbf{Type} & 
    \textbf{Source} &
    \textbf{\# Sentences} 
    & \textbf{\# Languages}  
    \\
    \midrule
    \multirow{2}{*}{Monolingual} & \ccweb & $\approx$20M & >2000 \\
    & \ccpdf & $\approx$5M & $\approx$1700 \\
    \midrule
     \multirow{6}{*}{Translation} 
     & Bible & $\approx$600M & $\approx$1600 \\
     & Panlex & $\approx$2B & $\approx$1000  \\
     & Tatoeba & $\approx$25M & $\approx$500  \\
     & \nllbdata & $\approx$500M & $\approx$200  \\
     & \primary & $\approx$55M & $\approx$500  \\
     & \langwise & $\approx$8M & $\approx$100  \\
    \midrule 
    Synthetic Translation & \backtranslation & $\approx$135M & >2000  \\
    Synthetic Aligned & \mined & $\approx$100K & $\approx$60  \\
    \bottomrule
    \end{tabular}
    \caption{A summary of the main sources and volumes (in number of sentences) used for CPT as detailed in Section \ref{sec:cptmodel}.}
    \label{tab:CPTtrainingdetails} 
\end{table*}

\paragraph{Monolingual Datasets} 
We collect and curate two massively monolingual corpora, starting from snapshots of Common Crawl,\footnote{\url{https://commoncrawl.org/}} inspired by the methodology and motivated by the results of~\citet{penedo2025fineweb2pipelinescale} and~\citet{kydlicek2025finepdfs}. We apply filters based on original URLs to discard low-quality pages, resulting in a URL-filtered version of Common Crawl. From this we create two datasets, that we refer to as~\ccweb and~\ccpdf, that collectively contain monolingual textual data sourced from web-pages or PDF documents spanning more than 2000 identifiable languages, as per the GlotLID model \citep{kargaran-etal-2023-glotlid}. Since our scope is continual pretraining (not full pretraining) and gathering more data for lower resource languages, we assign a fixed budget of at most 50 thousands documents per language, randomly sampling from the upstream corpora.

\paragraph{Bible texts} are used as one of the main parallel dataset for language analysis, for training and evaluation of MT systems.  The Bible has the large language coverage (over 2000 languages) and many Bible translations are publicly available under permissive licenses. Finally, the Bible books are explicitly segmented into chapters and verses which are always preserved during translation, so aligning the translated text across the languages is trivial. Due to these reasons, Bible has already been used as the primary training set in several research works \citep{mms,omnilingualasrteam2025omnilingualasropensourcemultilingual,ma-etal-2025-taxi1500,janeiro-etal-2025-mexma}. Additionally, we use the Bible to do part of our evaluation in order to have a reference-based benchmarking with large language coverage. We suggest using the Gospel by John as the test set, because the Gospels are the most translated from the Bible books, and John is considered to be the most different from the other Gospels. Training the Bible data has its caveats: its domain coverage is very limited, and language is often very old and formal. While training a model to understand such language might be alright, generating it would result in very unnatural style and various translation errors. Evaluating with the Bible shares the caviat of narrow domain and adds the contamination issue. %
We accept these risks and still use Bible both for training and evaluating, but we mitigate them using other sources (as explained in this section). We compile our Bible dataset from multiple sources\footnote{With the prevailing majority of texts being downloaded using the eBible tool: \url{https://github.com/BibleNLP/ebible}.}

\paragraph{Panlex} is a project collecting various dictionaries in a unified format\footnote{\url{https://panlex.org/}}. A processed dump of its database has 1012 languages containing at least 1,000 entries, as well as over 6000 languages with at least one entry. This makes it probably the most multilingual publicly available dictionary.

\paragraph{Tatoeba} \citep{TIEDEMANN12.463} is a dataset of aprox 400 languages and 11M sententences. Overall, it is a large, open‑source collection of example sentences and their translations, built collaboratively by volunteers around the world. Its main goals are to provide a freely available multilingual resource for language learning, research, and the development of natural‑language‑processing tools.

\paragraph{\nllbdata} We aim at building a system that improves upon NLLB-200~\citep{nllb-24}, at least retaining the performance on the 202 language varieties it covered. As a consequence, we apply the same URL filtering as we used to create~\ccweb and~\ccpdf to a mixture of primary and mined datasets roughly reproducing the original data composition used to train NLLB-200 models, which we refer to as~\nllbdata.

\paragraph{\primary}
is a group of several massively multilingual datasets, some of which we describe as follows. SMOL \citep{caswell2025smol} dataset includes sentences and small documents manually translated from English into 100+ languages. The docs are present both fully and by individual sentence pairs 2.4M rows. 
This dataset includes Gatitos \citep{jones-etal-2023-gatitos}, which is a dataset of 4000 words and short phrases translated from English into 173 low-resourced languages. %
BPCC~\citep{gala2023indictrans} is a collection of various human-translated and mined texts in Indic languages, parallel with English.
KreyolMT \citep{robinson2024kreyol} contains bitexts for 41 Creole languages from all over the world.
The dataset from the AmericasNLP shared task \citep{de-gibert-etal-2025-findings} represents 14 diverse Indigenous languages of the Americas.
AfroLingu-MT \citep{elmadany-etal-2024-toucan} covers 46 African languages.

\paragraph{\langwise} 
This dataset groups a set of less multilingual datasets, usually focused on a single low-resourced language or a group of related languages. A few examples of this compilation include ZenaMT \citep{haberland-etal-2024-italian} focused on Ligurian language,  the Feriji dataset \citep{Feriji} for Zarma, and a dataset from \citet{yankovskaya-etal-2023-machine} covering 6 low-resourced Finno-Ugric languages. %

\paragraph{LTPP} Part of our training data mix constituted an extremely valuable parallel data from the Language Technology Partnership Program\footnote{\url{https://about.fb.com/news/2025/02/announcing-language-technology-partner-program/}} which was launched with the purpose of expanding the support of underserved languages in AI models. Specifically, the compilation of parallel data from LTPP that we were able to use comprises 18 sources of various sizes and about 1.4M sentence pairs in total.

\paragraph{Limitations} Although we did a relevant effort to collect data, we are not exhaustive and detailed on its description and this inhibits replicability of our training, which is mitigated by the fact that we are sharing the model. More importantly, our current version of the model still misses many relevant sources. %

\subsection{Synthetic Data for CPT} 
\label{sec:syntheticdata}

For a significant portion of the languages we aim to support with our MT systems, there simply is no parallel data available, beside the Bible, and for some of them, not even the Bible has been translated yet\footnote{\href{https://wycliffe.org.uk/statistics}{Bible translations statistics}} or is not available for MT use. However, for several of them, publicly available monolingual corpora do exist and can be leveraged to generate synthetic parallel data via \emph{backtranslation} and \emph{bitext mining}, resulting in~\backtranslation and~\mined.

\subsubsection{Backtranslation}
\label{subsubsection:backtranslation}

\paragraph{Motivation and related work} Backtranslation has become a standard technique to do data‑augmentation strategies for MT by translating  monolingual target‑language data back into the source language ~\citep{sennrich-haddow-birch_2016_improving}. Since then, there have been several works exploring variations of this strategy. Edunov~\textit{et~al.} ~\citeyear{edunov2018understanding} showed that iterative back‑translation, where the augmented data are repeatedly re‑translated, yields further gains and helps the model learn more robust representations.
Subsequent work has extended the technique to low‑resource settings.  \citet{currey2017copied} proposed copying monolingual sentences directly into the training data.  \citet{liu2020multilingual} demonstrated that multilingual back‑translation can simultaneously improve translation across many language pairs by sharing a single encoder‑decoder architecture. \cite{nllb-24} focused on efficiently in massively multilingual settings and they used a combination of neural and statistical MT translated data similarly to \citep{soto-etal-2020-selecting}.  More recently, \citep{zebaze2025topxgentopicdiverseparalleldata} propose to use LLM‑based technique that generates topic‑diverse data in multiple low‑resource languages (LRLs) and back‑translate the resulting data.
Several studies have investigated how to best filter~\citep{seamless2025} back‑translated sentences. Recently, the approach has even been proved useful in speech translation \citep{wang-etal-2025-tens}.

\paragraph{Methodology} 

To produce backtranslation data we mainly rely on the two massively monolingual datasets obtained from Common Crawl:~\ccweb and~\ccpdf. Furthermore, to increase domain diversity of our backtranslation data mix, we also
rely on \dclmedu~\citep{allal2025smollm2smolgoesbig} for educational-level forward-translated (out of English) data.

The backtranslation pipeline we build extracts clean monolingual texts from the monolingual corpora above, produces source- or target-side translations, and estimates the translation quality of the resulting synthetic bitext. Several of these steps are model-based, including but not limited to the translation step itself. 

The first step consists of text segmentation, for which we use a fine-tuned version \citep{sonaromni} of the \textsc{sat-12l-sm} model~\citep{frohmann2024segment}, trained to predict the probability of a newline occurring at a given point in the text. Both sentence and paragraph boundaries can be obtained directly by tweaking the decision threshold. However, we find that resorting to heuristics to further refine these splits, e.g. re-splitting sentences deemed too long into smaller units, is beneficial.

After extracting textual units, the following step aims at removing \emph{noisy} monolingual samples, i.e. units that are either too short or too long, and those for which we struggle to identify the language with enough certainty. For the language identification task we resort to~\glotlid~\citep{kargaran2023glotlid}, supporting 1,880 languages at the time of writing. Empirically we find that \glotlid top-1 score aligns well with human judgement on \emph{sample quality}, with texts falling below certain thresholds either containing artifacts (e.g. HTML tags) or otherwise appearing as nonsensical text. We also find that this threshold is language-dependent, with a negative correlation between resourcefulness of the language and the average \glotlid score of positive samples, when tested on annotated data. This suggests that, in line with intuition, texts in lower-resource languages are not just harder to translate but also to identify. We generalize this by calibrating \glotlid scores on the aligned Bible, and define language-dependent thresholds for rejecting samples. This helps balance the competing objectives of keeping more data and rejecting lower quality samples.

For the translation step, we rely on two base MT systems: \nllb~\citep{nllb-24} and \llama 3~\citep{grattafiori2024llama}. The former is used as-is with no further fine-tuning to translate out of (or into) the 200 languages it supports, while the latter is used with no restriction, taking the best CPT and FT 8b model we were able to produce thus far. Notably, this model has been trained on both monolingual texts sampled from \ccweb itself and bitext from the Bible belonging to more than 1,700 languages. This is crucial since the base model has not been explicitly optimized for tasks (e.g. translation) in languages, present in the original pre-training corpora, but beyond 8 high resource languages. Given the more demanding nature of producing translations with \llama compared to \nllb, and that we already have data for languages covered by~\nllb, we only run the former on a stratified sample of the monolingual corpora, down-sampling languages already supported by \nllb.

Finally, we estimate translation quality of the produced synthetic bitext with a mixture of model-based and model-free signals. For the model-based signals we rely on omnilingual latent space (\sonaromni) similarity of source and target text~\citep{sonaromni}. We find that other model-free signals such as unique character ratio are helpful to complement the model-based ones, as they are strong predictors of particular failure cases, e.g. repetition issues or MT systems producing translations that are just a copy of the source text.

\paragraph{Ablations}

We run a series of ablations to understand how to effectively filter the produced data and incorporate it along other non-synthetic pre-existing corpora during continual pre-training. 

First, we study the downstream effect of filtering the backtranslation data according to cosine similarity of the translation in \sonaromni space. We first naively calibrate \sonaromni scores on the Bible development set, assuming perfectly uniform similarity estimation across languages, and find the mean latent space cosine similarity between aligned sentences: $\mu_{sim}$. Then, we define three thresholds, one standard deviation below ($LQ := \mu_{sim} - \sigma_{sim}$), at the mean ($MQ:= \mu_{sim}$) and above the mean ($HQ:= \mu_{sim} + \sigma_{sim}$). Then, we run an ablation training~\llama 3.2 3B Instruct on a stratified sample of~\ccweb, producing backtranslation data and then filtering according to these thresholds. 
We evaluate on~\floresplus, measuring translation quality over different language buckets (see section \ref{sec:MTevaldata}) and comparing against a baseline fine-tuned on the same data mix but without backtranslation data. The results, summarized in~\cref{tab:4newdata:backtranslation:filtering-ablation}, indicate that a uniform filtering strategy across language groups yield the best results, although filtering more aggressively on high-resource languages can yield even better results.

\begin{table}[ht]
\centering
\begin{tabular}{@{}l|cccc|cccc@{}}
\toprule
\multirow{2}{*}{\textbf{System}} & 
\multicolumn{4}{c}{\textbf{\engX}} & \multicolumn{4}{c}{\textbf{\Xeng}}
\\ 
& \textbf{High} & \textbf{Mid} & \textbf{Low} & \textbf{Very Low} & \textbf{High} & \textbf{Mid} & \textbf{Low} & \textbf{Very Low}
\\ 
\midrule
Baseline Data Mix & 44.68 & 23.97 & 16.98 & 19.97
& 56.96 & 42.29 & 35.71 & 38.21  \\
+ BT Data  (LQ)  & 46.1  & 26.11 & 20.32 & 23.6 
& 58.02 & 43.81 & 38.12 & 40.93 \\
+ BT Data  (MQ) & 46.33 & \textbf{26.6}  & \textbf{20.82} & \textbf{24.34}
& 58.17 & \textbf{44.01} & \textbf{38.26} & \textbf{41.29} \\
+ BT Data  (HQ) & \textbf{46.85} & 26.19 & 20.73 & 24.08   
& \textbf{58.52} & 43.64 & 37.72 & 40.98 \\
\bottomrule
\end{tabular}
\caption{\chrf when evaluating MT systems trained with different backtranslation data mixes.}
\label{tab:4newdata:backtranslation:filtering-ablation}
\end{table}

Second, after establishing a filtering strategy, we investigate the effect that mixing backtranslation data in different proportions along with pre-existing training corpora has on the downstream MT system performance. Given a fixed token budget for a training batch, we allocate $x\%$ of those tokens to examples sampled from backtranslation data, exploring a range of $5\%$ (ratio of 1:19 with respect to natural bitext) up to $75\%$ (ratio of 3:1 with respect to natural bitext). The results reported in~\cref{tab:4newdata:backtranslation:mixing-ablation} show how the optimal performance for lower-resource languages is achieved when maintaining a ratio of 1:9 or 1:4 with natural bitext, as performance increases up to that point and then starts decreasing again. On the other hand, all the other buckets see increased performance as we increase the amount of backtranslation data.

\begin{table}[ht]
\centering
\begin{tabular}{@{}l|cccc|cccc@{}}
\toprule
\multirow{2}{*}{\textbf{System}} & 
\multicolumn{4}{c}{\textbf{\engX}} & \multicolumn{4}{c}{\textbf{\Xeng}}
\\ 
& \textbf{High} 
& \textbf{Mid} 
& \textbf{Low} 
& \textbf{Very Low} 
& \textbf{High} 
& \textbf{Mid} 
& \textbf{Low} 
& \textbf{Very Low} 
\\ 
\midrule
Baseline Data Mix & 44.68 & 23.97 & 16.98 & 19.97
& 56.96 & 42.29 & 35.71 & 38.21 \\
+ BT Data (5\%) & 46.54 & 25.16 & 18.41 & 21.1 
& 59.68 & 43.1 & 37.53 & 39.67 \\
+ BT Data (10\%) & 46.67 & 25.46 & 18.84 & \textbf{21.3}
& 59.84 & 43.37 & 37.83 & 40.05 \\
+ BT Data (20\%) & 46.93 & 25.78 & 19.25 & 21.13
& 60.11 & 43.78 & 38.31 & \textbf{41.11} \\
+ BT Data (50\%) & 47.41 & 26.7 & 20.13 & 20.72
& 60.49 & 44.34 & 38.95 & 40.82 \\
+ BT Data (75\%) & \textbf{47.73} & \textbf{26.92} & \textbf{21.16} & 21.16 & \textbf{60.71} & \textbf{44.62} & \textbf{39.31} & 39.88 \\
\bottomrule
\end{tabular}
\caption{\chrf when evaluating MT systems trained with different backtranslation data mixes.}
\label{tab:4newdata:backtranslation:mixing-ablation}
\end{table}

\paragraph{Dataset statistics}

In~\cref{tab:4newdata:backtranslation:statistics} we summarize the resulting dataset obtained with the methodology outlined above. The dataset contains roughly 270 million sentences spanning more than 2,000 languoids, that we divide in three buckets: \emph{high resource} and \emph{low resource} indicate languoids that were described as such in~\citet{nllb-24}, while \emph{very low resource} indicate any languoid not included among the ones supported by~\nllb. The stratified sampling by languoid group at the source results in an artificially balanced distribution, with high resource languoids taking up 38\% of the unfiltered data, low resource languoids taking up 35\% and very low resource languoids the remaining 23\%. The progressively more relaxed filtering strategy leads to a final distribution where 51\% of the data is taken up by sentences belonging to low resource languoids, 26\% belonging to very low resource languoids, and 23\% belonging to high resource languoids.

\begin{table}[htb]
\centering
\begin{tabular}{@{}lcccc@{}}
\toprule
\textbf{Languoid Group}    
& \textbf{\# Languages }
& \textbf{\# Sentences }
&\textbf{ \# Sentences after filtering }
\\ 
\midrule
High
& $\approx$ 30
& $\approx$ 100M
& $\approx$ 30M
\\
Mid
& $\approx$ 100
& $\approx$ 150M
& $\approx$ 100M
\\
Low
& $\approx$ 300
& $\approx$ 100M
& $\approx$ 70M
\\
Very low
& $\approx$ 1300
& $\approx$ 40M
& $\approx$ 20M
\\
Zero
& $\approx$ 400
& $\approx$ 10M
& $\approx$ 10M
\\ 
All
& $\approx$ 2000
& $\approx$ 400M
& $\approx$ 230M
\\ 
\bottomrule
\end{tabular}
\caption{Statistics about resulting backtranslation data.}
\label{tab:4newdata:backtranslation:statistics}
\end{table}

\subsubsection{Bitext Mining}
\label{subsubsection:bitextmining}

\paragraph{Motivation and related work} Complementary to backtranslation, bitext mining is another method for data augmentation which expands parallel corpora by automatically aligning pairs of text spans with semantic equivalence from collections of monolingual text. In order to find semantic equivalence, early works such as \citet{resnik1999mining} attempted to find parallel text at the document level by examining an article's macro information such as the metadata and overall structure. Later works have applied more focus on the textual content within articles, leveraging methods such as bag-of-words \citep{buck2016findings} or Jaccard similarity \citep{azpeitia2017weighted}. With the advances of representation learning, more recent approaches have begun to employ the use of embedding spaces by encoding texts and applying distance metrics within the space to determine similarity, moving beyond the surface-form structure. Works such as \citet{hassan2018achieving} and \citet{yang2019improving} used bilingual embedding spaces. However, a drawback to this approach is that custom embedding spaces are needed for each possible language pair, limiting the permissibility to scale. Alternatively, encoding texts with a massively multilingual embedding space allows for any possible pair to be encoded and subsequently mined, and has become the adopted backbone for many large-scale mining approaches \citep{suryanarayanan2025pralekhacrosslingualdocumentalignment, ramesh2022samanantar, artetxe2019margin}. Generally, within this setting there are two main approaches: \emph{global mining} and \emph{hierarchical mining}. The latter focuses on first finding potential document pairs using methods such as URL matching, and then limiting the mining scope within each document pair only. Examples of such approaches are the European ParaCrawl project \citep{banon2020paracrawl, al2023exploring}. Alternatively \emph{global mining} disregards any potential document pair as a first filtering step, and instead considers all possible text pairs across available sources of monolingual corpora \citep{schwenk2021ccmatrix, schwenk2021wikimatrix}. This approach has yielded considerable success in supplementing existing parallel data for translation systems \citep{nllb-24, seamless2025}.

\paragraph{Methodology} We adopt the \emph{global mining} approach in this work. For our source of non-English monolingual corpora, we used \ccweb and FineWeb-Edu \citep{lozhkov2024fineweb-edu} as our source of English articles. We also considered \dclmedu as an option for English texts. However, as \dclmedu contains less articles than Fineweb-Edu, and given that the likelihood of a possible alignment increases as a function of the dataset size, we opted for the latter. We begin by first pre-processing our monoglingual data using the same sentence segmentation and LID methods as our backtranslation pipeline (see \Cref{subsubsection:backtranslation}). Subsequently, we encode the resulting data into the massively multilingual \sonaromni embedding space. In order to help accelerate our approach, we use the FAISS library to perform quantization over our representations, and enable fast KNN search \citep{johnson2019billion}. We first train our quantizers on a sample of 50M embeddings for each language using product quantization \citep{jegou2010product}, and then populate each FAISS index with all available quantized data. For our KNN search we set the number of neighbours fixed to $k = 3$, and to apply our approach at scale we leverage the stopes mining library\footnote{\url{https://github.com/facebookresearch/stopes}} \citep{andrews2022stopes}.

\paragraph{Ablation} In order to measure the effect of our resulting mined data, we performed a controlled ablation experiment. We choose the LLaMA3.2 3B Instruct model, and continuously pretrain it with two different data mixtures: one without the mined alignments, and a second supplemented with the mined data. In order to control for possible confounding variables, we fix the effective batch size, number of training steps, and all other hyperparameters for both models. Similar to our backtranslation ablations, we evaluate performance with the \floresplus benchmark using the metric~\chrf. Results are shown below in \cref{tab:5newdata:mining:ablation}. Overall, we see improvements when adding in mined alignments to the data mixture showing the effectiveness across both high and low-resource settings. For example, langoids such as Greek and Turkish both see good relative improvements of 2.95\% (47.4 $\rightarrow$ 48.8) and 2.74\% (43.7 $\rightarrow$ 44.9) respectively. Similarly, we observe a 5.12\% relative increase for the low-resource langoid N'Ko (13.00 $\rightarrow$ 13.67).

\begin{table}[ht]
\centering
\begin{tabular}{l|cccc|c|cccc|c}
\toprule
\textbf{Direction} & \multicolumn{5}{c|}{\textbf{\engX}} & \multicolumn{5}{c}{\textbf{\Xeng}} \\
\textbf{Resource level} & \textbf{high} & \textbf{mid} & \textbf{low} & \textbf{very low} & \textbf{all} & \textbf{high} & \textbf{mid} & \textbf{low} & \textbf{very low} & \textbf{all} \\
\midrule
Baseline Data Mix & 45.99 & 22.64 & 16.98 & 18.32 & 23.34 & 59.25 & 42.04 & 36.86 & 37.71 & 42.09 \\
+ Mined Data & 46.24 & 22.68 & 17.08 & 18.52 & 23.45 & 59.48 & 42.13 & 36.89 & 37.67 & 42.16 \\
\bottomrule
\end{tabular}

\caption{\chrf on \floresplus when evaluating MT systems continuously pretrained with and without mined data, split by whether English is the target or the source language and by the resource level of the other language.}
\label{tab:5newdata:mining:ablation}
\end{table}

\subsubsection{Conclusions and limitations}

The synthetic data we produce plays an important role in boosting MT system performance for lower-resource languages. Here, we briefly discuss some limitations and potential future work to further improve the impact of synthetic data.

We work from a limited collection of Common Crawl snapshots, that cover only a portion of the human spoken languages. Furthermore, since we rely on resource-hungry models and algorithms for both backtranslation and mining, scaling up the approaches is expensive and we limit the production of synthetic data to stratified samples of those snapshots. A more thorough investigation of the relationship between synthetic data quantity and downstream MT performance might reveal scaling laws that can be used to take more informed sampling decisions.

The backtranslation approach we employ could be improved both in the generation and filtering phase. In the generation phase, previous work (e.g.~\citet{hoang2018iterative, brimacombe2023quick}) often employ backtranslation in an iterative fashion, using a base system to backtranslate monolingual data, using the synthetic bitext to build a system better than the base one, then using the new system to produce higher quality synthetic data, and repeating the cycle for a number of steps. 
In the filtering phase, we could complement latent space similarity metrics with LLM-as-a-judge approaches similar to~\citet{kocmi2023gemba}; if the base model is itself a LLM, we could investigate the ability of the model to effectively score its own translations, and the interference between this ability and translation ability as CPT on new backtranslated data progresses.

The mining approach could be significantly scaled up by considering alignment with pivot languages beyond English. For instance, aligning languages within the same family or group such as Spanish and Portuguese can enhance cross-lingual transfer by leveraging their structural and lexical similarities. This strategy not only facilitates more effective knowledge transfer between related languages but also helps to reduce the model's bias toward English-centric data, promoting greater linguistic diversity and inclusivity in multilingual applications.

\subsection{Seed Data for Post-Training: \medley}
\label{sec:seed}

In this section we present \textbf{\medley}, a \textit{m}ulticentric, multiway, \textit{d}omain-diverse, \textit{l}inguistically-diverse, and \textit{e}asy-to-translate seed dataset. \medley is a large scale data collection effort covering \numberofLRLs~ LRLs. It  \datasetnamesource and \datasetname-\numberofLRLs.
\textbf{\datasetnamesource} consists of $605$ manually constructed paragraphs with roughly $2200$ sentences and $34K$ words (counted in English). It is \emph{multicentric}: source paragraphs are written in five source languages, thus including styles and cultural perspectives as well as topical subjects from a few different cultures. Each paragraph is accompanied with notes on any additional context required for its translation. It is then manually \emph{multiway} parallelized across 8 pivot languages, increasing its accessibility to bilingual communities around the world. 
It is \emph{domain-diverse} and \emph{grammatically diverse}: it covers 5 domains and provides coverage to 61 cross-linguistic functional grammatical features that aim to cover a broad range of grammatical features in any arbitrary language that the dataset may be translated to. 
Further, we ensure that it is \emph{easy to translate}: i.e., that it uses accessible, jargon-free language for lay community translators.
\textbf{\datasetname-\numberofLRLs~} provides professional translations of the dataset into \numberofLRLs~ low-resource languages, as can be seen in~\cref{tab:languages}. More details can be found in~\cref{app:medley}, with examples from the dataset in \cref{medley:app:examples_dataset}.

\begin{figure*}[ht]
    \centering
    \includegraphics[width=0.9\linewidth]{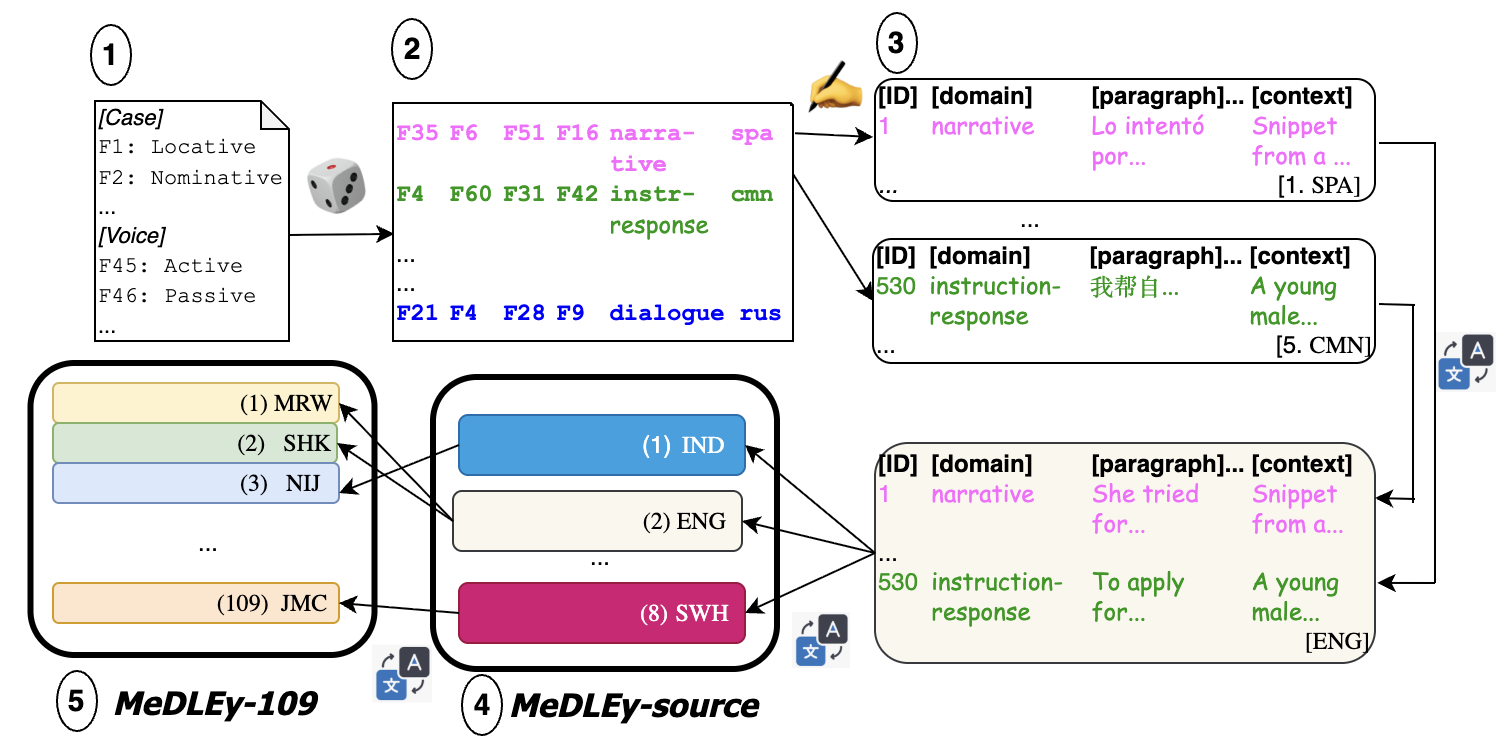}
    \caption{Steps in the creation of \datasetnamesource and \datasetname-\numberofLRLs. This includes (1) enumeration of grammatical features, (2) template generation including domain and source language assignment, (3) manual creation of paragraphs in 5 source languages: English, Mandarin, Spanish, Russian, and German, and (4) n-way parallelization (via English) across 8 pivot languages: English, Mandarin, Spanish, Russian, Hindi, Indonesian, Swahili, and French, resulting in \datasetnamesource. This is then (5) translated into \numberofLRLs~ low-resource languages, each from a convenient pivot depending on the translator, resulting in \datasetname-\numberofLRLs.}
    \label{medley:fig:medley_creation}
\end{figure*}

\subsubsection{Motivation and related work}
\label{medley:sec:intro}

It is infeasible to manually curate data for LRLs at a large scale.
Previous work emphasizes quality over quantity in the context of data collection for low-resource languages \citep{yu-etal-2022-beyond,de2022quality,talukdar2023influence}, and previous efforts seek to curate a small high-quality set of examples in these languages \citep{nllb-24,caswell2025smol}.
Such a ``seed'' dataset has various uses, such as training LID systems that can be used for data mining \citep{kargaran2023glotlid}, or providing high-quality examples for few-shot learning strategies \citep{lin-etal-2022-shot,pmlr-v202-garcia23a}. Importantly, while high-quality MT systems in both directions typically require training data at a much larger scale, seed datasets can be used to train models to translate into English with reasonable quality, which can then be used for bootstrapping synthetic bitext and better MT systems using monolingual data in LRLs \citep{sennrich-haddow-birch_2016_improving,nllb-24}.

While there exist web-crawled monolingual and parallel datasets with low-resource languages such as \textsc{MADLAD} \citep{kudugunta2023madlad}, \textsc{Glot500} \citep{imani2023glot500}, and \textsc{NLLB} \citep{nllb-24}, these may be noisy and of unclear quality due to the scarcity of high-quality LRL content on the web \citep{kreutzerquality} as well as LID quality issues for LRLs \citep{kargaran2023glotlid}.
There have been manual data collection efforts focusing on particular language groups, such as Masakhane \citep{nekoto2020participatory}, Turkish Interlingua \citep{mirzakhalov-etal-2021-large}, Kreyol-MT \citep{robinson2024kreyol}, HinDialect \citep{bafna-etal-2022-combining}, as well as efforts for particular languages, such as Bhojpuri \citep{kumar2023machine}, Yoruba \citep{adelani-etal-2021-effect,akpobi-2025-yankari}, Quechua \citep{ahmed-etal-2023-enhancing}, among many others. \nllbseed is a highly-multilingual, professionally-translated parallel dataset, containing 6000 sentences from the Wikipedia domain translated into 44 languages \citep{nllb-24}. 
However, the most comparable effort to \datasetname, in terms of scale, in collecting high-quality, professionally-translated parallel datasets is \textsc{SMOL} suite~\citep{caswell2025smol}. It consists of the \textsc{SmolSent} and \textsc{SmolDoc} datasets. The former consists of sentence-level source samples selected from web-crawled data translated into 88 language pairs, focusing on covering common English words. The latter consists of automatically generated source documents designed to cover a diverse range of topics and then translated into \numberofLRLs languages. 
\datasetname covers 92 languages not present in \textsc{SMOL} or \nllbseed, contributing to the language coverage of existing datasets.
\datasetname also differs significantly in design considerations from the above, and it is the first such effort to focus on the coverage of grammatical phenomena in an arbitrary target language.

\subsubsection{Approach}
\label{medley:sec:approach}

The goal of \datasetname is to provide a bitext corpus that is domain-diverse and grammatically diverse in a large number of included languages.
Given that a seed dataset is limited in size, it becomes crucial to include diverse and representative examples in it, so as to gain as much information as possible about the language.
In this work, we focus on grammatical and domain diversity. 
The knowledge of a language's \emph{grammar} is crucial to navigating the translation of basic situations into or out of that language. 
In order for an MT system to be flexible across various registers, domains, and sociopragmatic situations, it needs to be exposed to a variety of grammatical mechanisms used in those conditions.

\paragraph{What is grammar?} 
\label{medley:sec:what_is_grammar}

A language uses its grammar to systematically express certain kinds of information (for example, case is a grammatical mechanism used to express information about the role of a noun). In this work, we call the underlying meaning of a grammatical mechanism a grammatical \emph{function}, and the actual shape of the grammatical mechanism used in the language the grammatical \emph{form}. We show examples of functions and their forms in various languages in \cref{medley:tab:examples_gram_features}. Note that, as these examples show, these function-form pairs may be at all levels of linguistic structure, including morphology, syntax, and information structure. To refer to particular functions in this paper, we use canonical names associated with them for convenience. We refer to these as \emph{grammatical features}. We construct our grammar schema in terms of these features.

\paragraph{Cross-linguistic variation}
\label{medley:sec:lang_wise_variation}

It is important to stress that languages vary extensively in terms of a) the set of forms they use to codify grammatical functions, b) in the manner of codification of a function (i.e. what form a particular function takes), and c) the mapping between form and function.
First, the set of grammatical forms found in each language are not the same. For example, while some languages have honorifics to convey esteem or respect to address their interlocutors, other languages may not use any grammatical mechanism for this at all. Secondly, the same grammatical function may be codified into different grammatical mechanisms depending on the language, as in the example of the locative case (see~\cref{medley:tab:examples_gram_features}). Finally, forms and functions often follow many-to-many relationships across languages.
For example, the same feature can cover slightly different functions in two languages despite each having forms that share a core meaning with the other: English allows the so-called present continuous to express future events, while Spanish does not (\ref{tense-aspect}).

{
\small
\begin{exe}
    \ex \label{tense-aspect} \begin{xlist}
        \ex English: I'm leaving tomorrow.
        \ex \gll Spanish: *Estoy saliendo mañana. \\
         {} be.\Prs.\Fsg{} leave.\Ger{} tomorrow \\ 
    \end{xlist}
\end{exe}
}

\paragraph{Building a grammatically-diverse corpus}
\label{medley:sec:gram_div_corpus}

\begin{table*}[ht]
    \centering
    \includegraphics[width=0.8\linewidth]{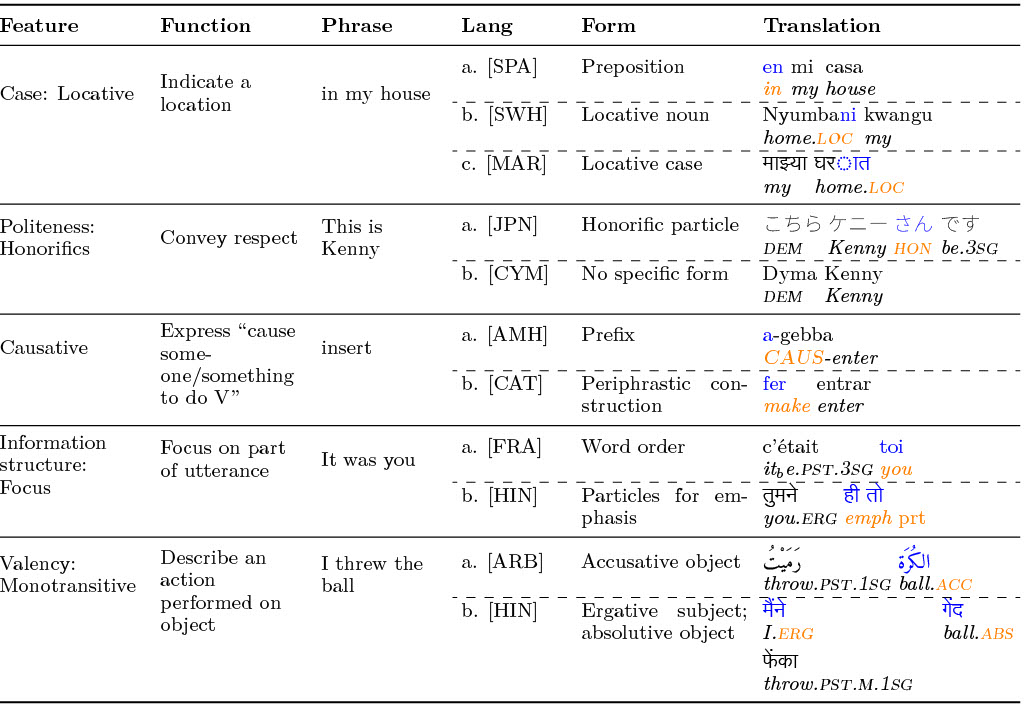}
    \caption{Examples of grammatical features and their associated functions or meanings, with various language dependent forms (specific mechanisms used to express that feature).}
    \label{medley:tab:examples_gram_features}
\end{table*}

Despite these differences, many grammatical functions codified in grammars tend to be shared across languages. For example, most languages have ways to differentiate which participants perform the action of an event and which ones experience it, the time at which an event occurred, how many referents there are, or whether an event is conditional upon another event taking place, to name a few.

Thus, achieving coverage over grammatical functions is a reasonable proxy for achieving coverage of grammatical phenomena at different levels of linguistic structure in a particular language.
These grammatical functions can be enumerated up to a required degree of fine-grainedness at all linguistic levels with broad cross-linguistic coverage.
Also note that, broadly speaking, most functions can be expressed in \textit{any} language, regardless of the grammar of that language. For example, even though Spanish and English don't have a case system, it is certainly possible to express location in these languages akin to the Marathi locative case (see \cref{medley:tab:examples_gram_features}). 
Since the function is likely to be retained across translation, we can construct a source corpus that has high coverage over our grammatical features (in any source language), and expect that when it is translated into an arbitrary language, it will cover many grammatical phenomena in that language. For example, when we translate the English phrase ``in my house'' to Marathi, we gain coverage of the locative case in Marathi. We do not expect that each grammatical feature will be realized in the same manner across languages.
However, we do expect that in many cases, the function associated with a feature will be manifested in some manner in a text or its context regardless of the language of the text.
This allows us to build a grammatically-diverse source corpus that achieves broad grammatical coverage when translated into arbitrary target languages.
This forms our cross-linguistic framework of grammatical diversity.

\paragraph{Dataset construction} Here we summarize the dataset construction process. The creation process involves: (1) curating a list of cross-linguistic grammatical features, as described above; (2) selecting domains (informative, dialogue, casual, narrative, and instruction-response) and source languages (English, Mandarin, Russian, Spanish, and German); (3) having expert native-speaker linguists craft natural, accessible source paragraphs based on templates combining grammatical features and domains, along with contextual notes; (4) translating these source paragraphs into eight pivot languages (English, Mandarin, Hindi, Indonesian, Modern Standard Arabic, Swahili, Spanish, and French) chosen to represent common L2 languages of low-resource language communities; and (5) commissioning professional translations from these pivots into numerous low-resource target languages selected for translator availability, prior coverage gaps, and language family diversity, resulting in grammatically diverse parallel text for underrepresented languages.
See the list of grammatical features used, annotator guidelines, and more details about our approach in~\cref{medley:sec:approach_details} and~\cref{medley:sec:data_creation}.

\paragraph{Grammatical features transfer and retention}

Given the aim of covering naturally rarer features in our dataset, we compare the entropy of grammatical feature distributions in our dataset versus other seed datasets, across 9 categories (e.g. tense, formality). We find that \medley shows the highest entropy in 5 out of 9 categories, indicating that \medley often has higher proportions of rarer features in a paradigm. 
Furthermore, given our assumptions of feature transfer via translation during the creation of \medley, we measure the extent of feature transfer and the extent to which features are preserved across translation hops. We thus conduct a qualitative feature transfer analysis looking both at single-hop and 2-hop translations. We find that most morphosyntactic features have transfer rates above 50\%, and interestingly, forms that do not surface in a target translation can resurface in next hop from that language, indicating that grammatical diversity is preserved in a language-dependent manner via translation. 
The grammatical feature distribution as well as feature retention analyses  are detailed in~\cref{medley:app:grammatical_feature_analyses}.

\subsubsection{Experiments}
\label{sec:exp_setup}

\datasetname may have several uses given its grammatical feature coverage and n-way parallel nature. 
We demonstrate its general utility for fine-tuning MT models for LRLs.

\paragraph{Experiment Setup}
In particular, we perform a \textbf{Token-controlled comparison} and also measure \textbf{Absolute and combined gains} for models fine-tuned on \datasetname versus other datasets. 
On the former, given that larger datasets are more expensive to annotate, we compare randomly sampled equally-sized training subsets of \datasetname and baseline datasets in terms of number of tokens for a fair comparison, using the size of the smallest dataset as the token budget. 
For the latter, we report absolute gains from training on the entire dataset. We also look at additive gains from combining seed datasets, which may help inform decisions about language coverage in future seed datasets.

We evaluate the performance of the fine-tuned models on \floresplus\citep{nllb-24} and \bouquet~\citep{bouquet}, considering 5 languages that are in the intersection of all the datasets. In particular, we experiment with \textsc{\nllb-200-3.3B}\footnote{\href{facebook/nllb-200-3.3B}{https://huggingface.co/facebook/nllb-200-3.3B}} as a representative of sequence-to-sequence (seq2seq) models~\citep{nllb-24}, and \textsc{\llama-3.1-8B-Instruct}\footnote{\href{meta-llama/Llama-3.1-8B-Instruct}{https://huggingface.co/meta-llama/Llama-3.1-8B-Instruct}} representing LLM-based MT, and fine-tune them to obtain language-specific checkpoints, considering into- and out-of-English separately. More precise details about the experiments setup can be found in~\cref{medley:app:experiment_setup_details}.

\paragraph{Experiment Results}
\label{sec:results}

\begin{table}[!t]
\centering
\small
\begin{tabular}{@{}rrrrrrr@{}}
\toprule
    \multirow{2}{*}{\textbf{Model}} &
    \multirow{2}{*}{\textbf{Seed Dataset}} & 
    \multirow{2}{*}{\textbf{\#tokens}} & 
    \multicolumn{2}{c}{\textbf{\bouquet}} & \multicolumn{2}{c}{\textbf{\floresplus}} 
    \\
    & & & 
    \multicolumn{1}{l}{\textbf{\engX}} &
    \multicolumn{1}{l}{\textbf{\Xeng}} &
    \multicolumn{1}{l}{\textbf{\engX}} &
    \multicolumn{1}{l}{\textbf{\Xeng}} \\
    \midrule
    \multicolumn{7}{c}{\textit{Token-controlled Experiment}} \\
    \midrule
\multirow{4}{*}{LLaMA} & No Seed               & 0   & 8.49           & 14.45          & 10.07          & 16.81          \\
                       & SmolDoc               & 215K   & \textbf{20.08} & 18.74          & \textbf{21.53} & 22.25          \\
                       & SmolSent              & 215K   & 18.16          & 18.74          & 18.46          & 22.42          \\
                       & \medley                & 215K   & 19.60          & \textbf{20.39} & 20.69          & \textbf{23.73} \\
\midrule
\multirow{4}{*}{NLLB}  & No Seed               & 0   & 31.75          & 39.43          & 29.01          & 39.75          \\
                       & SmolDoc               & 215K   & 31.70          & 40.88          & 29.85          & 39.34          \\
                       & SmolSent              & 215K   & \textbf{32.54} & 40.88          & \textbf{30.27} & 39.31          \\
                       & \medley                & 215K   & 30.58          & \textbf{43.05} & 29.35          & \textbf{40.72} \\
    \midrule
    \multicolumn{7}{c}{\textit{Direct Comparison Experiment}} \\
    \midrule
\multirow{7}{*}{LLaMA} & No Seed               & 0     & 8.49           & 14.45          & 10.07          & 16.81          \\
                       & SmolDoc (D)           & 800K & 25.07          & 23.06          & 25.03          & 25.09          \\
                       & SmolSent (S)          & 225K & 18.69          & 19.47          & 19.70          & 23.87          \\
                       & \medley (M)            & 525K & 22.30          & 23.18          & 22.43          & 25.71          \\
                       \cline{2-7}
                       & M+D                   & 1.31M & 26.98          & 27.44          & 26.10          & 26.94          \\
                       & M+S                   & 745K & 24.33          & 25.98          & 23.27          & 26.43          \\
                       & M+D+S                 & 1.54M & \textbf{28.12} & \textbf{29.04} & \textbf{26.79} & \textbf{27.90} \\
\midrule
\multirow{7}{*}{NLLB}  & No Seed               & 0     & 31.75          & 39.43          & 29.01          & 39.75          \\
                       & SmolDoc (D)           & 800K & 32.92          & 41.24          & 30.05          & 39.36          \\
                       & SmolSent (S)          & 225K & 32.29          & 41.77          & 30.10          & 39.94          \\
                       & \medley (M)            & 525K & 30.60          & \textbf{43.40} & 29.81          & \textbf{40.94} \\
                       \cline{2-7}
                       & M+D                   & 1.31M & 33.43          & 42.67          & 30.69          & 40.17          \\
                       & M+S                   & 745K & 32.92          & 41.98          & 31.02          & 39.83          \\
                       & M+D+S                 & 1.54M & \textbf{34.73} & 42.90          & \textbf{31.60} & 40.19          \\
\bottomrule
\end{tabular}
\caption{Token-controlled and Direct comparison: reporting number of \llama tokens on involved languages (Bambara, Mossi, Wolof, Yoruba, and Ganda, for both evaluation datasets) and \chrf numbers.}
\label{tab:experiments:medley:unified}
\end{table}

The overall results of our experiments are reported in~\cref{tab:experiments:medley:unified}. A more detailed breakdown of these results can be found in~\cref{app:experiment_results_details}.
We see that \datasetname  matches or outperforms baseline datasets in the token-controlled setting, and shows gains in the into-English direction, while adding \medley to existing datasets yield generally modest gains. We also show similar findings on a comparison on \nllbseed on a separate set of intersection languages\footnote{In addition, we also show that \nllbseed contains a high proportion of difficult-to-translate texts potentially due to technical or obscure terminology (54\% as compared to 10.41\%), which may hinder lay community translators.}, see~\cref{tab:medley:nllb-seed:results-overall}. We confirm these trends over various other MT evaluation metrics such as \xcomet and \metricx~\citep{guerreiro2024xcomet,juraska-etal-2024-metricx}, see~\cref{fig:medley:results:metrics-correlations}.
This supports a major application of seed datasets, i.e., synthetic data generation from monolingual LRL data via better \ttt{xx-en} systems as discussed in \cref{medley:sec:intro}.

\subsubsection{Conclusions and limitations}

\medley has been a large scale MT training data collection effort across \numberofLRLs low-resource languages, culminated in a multi-centric, domain-diverse, and multi-way parallel seed dataset, that showed both a broader grammatical diversity and a larger impact when used to fine-tune MT models, compared to other pre-existing seed datasets. Indeed, \medley proved to be an essential component of our post-training recipe, as can be seen in~\cref{sec:post_training}. Nevertheless, the iterative nature of the data collection effort impacted the scope of the experiments we could perform with the dataset, both in isolation and as part of the broader MT recipe; furthermore, we identify several limitations that we mention below.

The grammatical coverage of \medley is limited by both budget constraints and by intrinsically  language-specific source-side grammatical functions, that may not reliable transfer into target languages. As a consequence, \medley targets common, cross-lingual, function-oriented features rather than language-specific phenomena. Furthermore, the lack of labeled evaluation data in low-resource languages prevents us from performing more fine-grained evaluations, at the level of single grammatical phenomena. Finally, as both translation and quality assurance mainly relies on external vendors for low resource languages, inaccurate translations may occur more frequently in these languages than in higher resource ones, where in-house expertise allows further quality checks.

\subsection{Evaluation Data}
\label{sec:MTevaldata}

MT evaluation has been driven by a series of publicly available test collections that enable reproducible comparison of systems. The Workshop on Machine Translation (WMT) series introduced large, community‑curated benchmarks that have become the de‑facto standard for both automatic and human evaluation \cite{kocmi-etal-2025-findings, deutsch-etal-2025-wmt24}. Alternative efforts have focused on multilingual, low‑resource, and cross‑domain evaluation.  FLORES benchmarks extended evaluation to 200 languages, providing expert‑translated reference sentences for a curated set of English sentences~\citep{nllb-24}.  In this work, we report results with our proposed datasets (\bouquet, which has been manually created from scratch, and a subset of the Bible that we explicitly reserved for evaluation) described in this section, and exisiting ones like \floresplus \citep{maillard-etal-2024-findings}, which covers 220+ languages, in 3 domains (wikipedia, travel guides and news). The complete list of evaluation datasets is summarised in Table \ref{tab:eval_datasets_language_groups_coverage}.

\subsubsection{\bouquet} 
\label{sec:bouquet}

\paragraph{Description} To evaluate translation systems that purport to be massively multilingual or omnilingual, we need a multi-way parallel evaluation dataset. Prior to \bouquet, such datasets as those derived from FLoRes-101 \citep{goyal2022flores} and FLORES-200 \citep{nllb-24} (e.g. 2M-FLoRes \citep{costajussa20242m}, or FLoRes+ \citep{maillard-etal-2024-findings})  existed but came with various shortcomings in that they represented a narrow selection of domains and registers, were prone to contamination \citep{sainz-etal-2023-nlp} due mainly to automatic construction, and proved difficult to translate accurately because of their English-centric nature or their lack of helpful context needed by translators (e.g., context about grammatical gender when referring to human beings only mentioned by proper nouns or titles). Some of this context could have been inferred through paragraph-level parsing if there were not missing metadata on existing paragraph structures. 

With the introduction of \bouquet, we aim to address the above limitations and progress towards a more culturally-diverse MT evaluation \citep{oh-etal-2025-culture}. \bouquet was created (as opposed to crawled or mined) from scratch in eight non-English languages\footnote{\texttt{arz\_Arab}, \texttt{cmn\_Hans}, \texttt{deu\_Latn}, \texttt{fra\_Latn}, \texttt{hin\_Deva}, \texttt{ind\_Latn}, \texttt{rus\_Cyrl}, and \texttt{spa\_Latn}} by linguists, who provided gold-standard English translations, contextual information, as well as register labels to facilitate accurate translations into a large number of languages. The sentences that compose \bouquet are all part of clearly delineated paragraphs of various lengths, and they represent eight domains that are not represented in FLoRes-derived datasets.
The construction and evaluation of \bouquet are described in further detail in \citep{bouquet}.

\begin{figure}[H]
    \centering
    \includegraphics[width=1\linewidth]{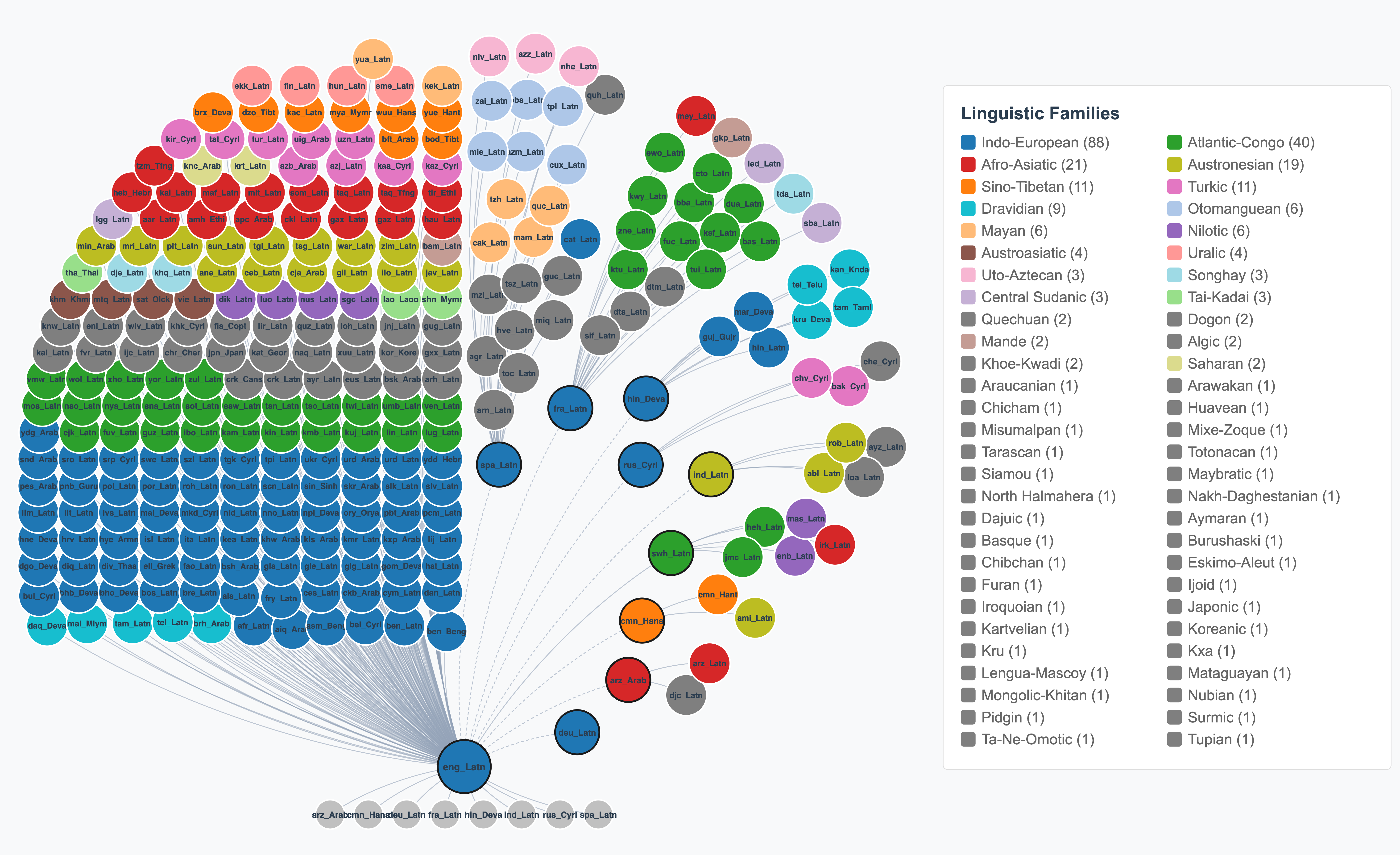}
    \caption{Bouquet Language Expansion Visualization. Details on pivots that were used to translate each language show that most useful ones were fra\_Latn, ind\_latn, swh\_latn and spa\_latn. }
    \label{fig:pivots}
\end{figure}

\paragraph{Expansion Analysis} \bouquet started early 2025, with 9 pivot languages, since then it has been expanded through vendors, partnering (Mozilla Common Voice\footnote{\url{https://commonvoice.mozilla.org/en}}) and the open-initiative\footnote{\url{https://bouquet.metademolab.com/}}. To date of this paper, \bouquet is available in 275 languages (see Appendix \ref{app:comparelangs}) covering 56 distinct language families and 33 scripts. 18 languages have been totally translated through community efforts (16 by Mozilla Common Voice and 2 through the \bouquet open-initiative) and the rest have been commissioned through vendors. Regarding pivots, we learned that some pivot languages (French, Indonesian, Swahili, Spanish) appear to ease resourcing and translation more than others (German, Hindi); vendors that rely exclusively on English deliver translations of lower quality, drive up costs, and fail to deliver a significant amount of work we commission. Figure \ref{fig:pivots} details the pivots that were used for each language.

\subsubsection{Bible Evaluation Partition} 
\label{sec:biblebenchmarking}

In order to have validation/evaluation signals, we suggest keeping some data from the highest multilingual sources aside. From the Bible, we suggest using the Gospel by John as the test set, because the Gospels are the most translated from the Bible books, and John is considered to be the most different from the other Gospels. The Gospel of John still contains about 30\% of verses that have a high semantic overlap with other books (such as ``If you will ask anything in my name, I will do it.'' in John vs ``All things, whatever you ask in prayer, believing, you will receive.'' in Matthew). This benchmark is multi-way parallel and it allows to compare performance across languages and systems. The content and partitioning of our Bible dataset are the same as in \citep{mms,omnilingualasrteam2025omnilingualasropensourcemultilingual}  %

A clear limitation of these datasets is the training contamination (since models are likely to have ingested the entire Bible). However, these are the best efforts to have a validation signal for the long-tail of languages in our process to constructing Omnilingual datasets (\bouquet) and Omnilingual quality estimation metrics (\blaser, see section \ref{sec:blaser}).

\subsubsection{Benchmarks Specialization}
\label{sec:benchmarkspe}

Why use more than one evaluation dataset? The Bible benchmark is used for providing direct evidence on 1561 language varieties, albeit a single domain. \floresplus\footnote{In all evaluations throughout the paper, we use version 2.1 of \floresplus, corresponding to its state in early 2025. Unless otherwise specified, we use its \texttt{devtest} split, which has a slightly different set of languages from the \texttt{dev} split.} and \bouquet provide a more varied coverage of domains in languages of different resource levels, with \bouquet including a rich representation of extremely low-resourced languages. 
Languages and resources of several of these datasets are reported in Table \ref{tab:bouquetlangs}.
For ablations, we also define two subsets of~\floresplus, that we call~\floreshrl and~\floreshard. \floreshrl consists of a selection of 54 languages chosen to represent higher resource languages, in accordance with the definition provided in~\citep{nllb-24}. \floreshard consists of a selection of 20 languages chosen to represent lower resource languages with particularly low performance from baseline MT systems.\footnote{The codes of the selected languages are \texttt{ayr\_Latn, brx\_Deva, chv\_Cyrl, dar\_Cyrl, dgo\_Deva, dik\_Latn, dzo\_Tibt, gom\_Deva, knc\_Arab, mhr\_Cyrl, min\_Arab, mos\_Latn, myv\_Cyrl, nqo\_Nkoo, nus\_Latn, quy\_Latn, sat\_Olck, taq\_Tfng, tyv\_Cyrl, vmw\_Latn}. For selection, we prioritized the languages added to \floresplus by the community, as well as  languages from the FLORES-200 list representing diverse language families and scripts. For experiments with \floreshard, we are using the \texttt{dev} split, as some of its languages are not included in \texttt{devtest}.}

\begin{table}[H]
    \centering
    \begin{tabular}{l|rrrrr|r|c}
        \toprule
       \textbf{ Resource level} & \textbf{high} & \textbf{mid} & \textbf{low} & \textbf{very low} & \textbf{zero} & \textbf{total} & \textbf{domains} \\
        \midrule
        Bible & 25 & 71 & 172 & 1289 & 4 & 1561 & 1 \\
        \floresplus & 24 & 82 & 73 & 31 & 3 & 213 & 3 \\
        \bouquet & 23 & 72 & 79 & 48 & 53 & 275 & 8 \\
        \midrule 
        All benchmarks & 26 & 89 & 222 & 1318 & 59 & 1714 \\
        All benchmarks, coarse grouping & 25 & 83 & 206 & 1296 & 56 & 1666 \\
        \bottomrule
    \end{tabular}
    \caption{Number of language varieties, grouped by resource level (as per \Cref{sec:resourcelevels}) in each of the benchmark datasets, and the number of domains covered by them. The line ``All benchmarks`` counts the union of all language codes in the three individual benchmarks, and the following line counts only unique ISO 639-3 codes, ignoring the variation in scripts, locales, or dialects.}
    \label{tab:eval_datasets_language_groups_coverage}
\end{table}   

As Table \ref{tab:eval_datasets_language_groups_coverage} demonstrates, our three evaluation benchmarks collectively cover over 1,700 language varieties, or over 1,600 unique languages, if we abstract away from the more fine-grained varieties differentiated by scripts, regions, or dialects.

\section{Translation Modeling Overview}
\label{sec:tmoverview}

\subsection{Motivation and Approaches}

\paragraph{Related Work} The modeling advances that enable massively multilingual MT can be grouped into mainly encoder-decoder architectures and large‑scale decoder-only language‑model based approaches.
Early work showed that a single Transformer encoder–decoder can handle many language pairs by conditioning the decoder on a language identifier \cite{Johnson2017}. This paradigm was scaled to hundreds of languages with NLLB \citep{nllb-24}. The recent surge of large language models (LLMs) has opened a new modeling direction. \citet{zhu-etal-2024-multilingual} evaluate eight state‑of‑the‑art LLMs on a suite of 100+ languages and find that, while LLMs can acquire translation ability with few examples, they still lag behind dedicated multilingual translation systems on low‑resource pairs. Specialised LLMs to MT e.g. TowerLLM \citep{tower_llm_2024} has shown the validity of certain recipes and motivation for specialization.

\paragraph{Our approaches} In this work, we investigate how to specialize general-purpose decoder-only LLMs for the translation task, and we investigate two distinct architectural approaches. The first approach involves directly fine-tuning the LLM for translation tasks while maintaining its original decoder-only architecture. The second alternative is to build an encoder-decoder Transformer model derived from the LLM for both its encoder and its decoder. In this work, we explore both of these strategies in Sections \ref{sec:deconly} and \ref{sec:encdec}.

\subsection{Vocabulary Extension and Tokenization}

A critical prerequisite for omnilingual translation is ensuring adequate vocabulary coverage across all languages. Since Llama's original tokenizer was optimized for a limited set of languages, applying it directly to multilingual translation would result in suboptimal tokenization for many language pairs. We therefore begin by describing our approach to vocabulary extension and tokenizer adaptation.

\paragraph{Related work} Recent research has tackled the “vocabulary bottleneck” that limits the performance of large language models on low‑resource languages. One line of work introduces VEEF‑Multi‑LLM \citep{sha-etal-2025-veef}, expands the token set with Byte‑Level Byte‑Pair Encoding, then fine‑tunes only a small set of extra embeddings. Another promising direction is the Efficient and Effective Vocabulary Expansion method, which freezes most of the original embeddings and initializes new ones via subword‑level interpolation, enabling rapid adaptation to languages like Korean with just a few billion training tokens \citep{kim2024efficienteffectivevocabularyexpansion}. A broader survey of vocabulary‑centric techniques highlights adapter‑based approaches, lexical‑level curriculum learning, and even zero‑shot expansion showing that modest data can still yield noticeable gains across many typologically diverse languages \citep{spuler}. In co-occurrence to this work, \citet{sonaromni}, in the context of learning an Omnilingual multilingual embedding space, disentangle the challenge of learning a new vocabulary representation from the challenge of learning new languages. The authors minimize the MSE loss between the student and teacher \sonaromni sentence embeddings using monolingual sentences for the base languages. 

We reuse two tokenizers from \citet{sonaromni} (one for the encoder and one for the decoder side of \nllbtwo) and build a third one (for \omtllama) using the same methodology. The \nllbtwo input tokenizer is trained from scratch for over 1.5K languages, while the \nllbtwo output tokenizer extends the \llamathree tokenizer vocabulary for 200 languages. The \omtllama takes a middle ground between the two: it retains the original \llamathree tokenizer vocabulary but extends it with extra tokens for 1.5K languages. All three tokenizers have the resulting vocabulary size of 256K tokens.

\paragraph{Methodology} We chose to modify the default BPE \llamathree tokenizer to increase the granularity of its subword tokens for the long tail of languages distribution. We achieved it by two means: 
\begin{enumerate}
    \item Adjust the pre-tokenization regular expression (the rule for splitting text into ``words'') by making it more friendly to languages that use rare writing systems or a lot of diacritic characters.%
    \item Increase the vocabulary of the tokenizer from 128K to 256K tokens by continued BPE merging.
\end{enumerate}
These two measures lead to decreased fertility (number of tokens per text) of the tokenizer, especially languages with non-Latin scripts. Improved fertility always results in higher throughput of training and inference (because the same number of tokens now covers a larger length of text), and usually (but not always) results in better translation performance — because the model spends less of its capacity on reconstructing the meaning of a word from its subwords. %

To extend the tokenizer vocabulary, we implemented a byte-pair encoding ``continued training'' algorithm by sequentially merging the most frequently occurring consecutive pairs of tokens within a word. The word frequencies were computed with a balanced sample from the parallel training data in all our languages and from the \ccweb dataset of web documents (in equal proportions). As weights for balancing, we used the total number of characters in the texts, and we applied unimax sampling over the languages, squashing the proportions of the first 126 languages to uniform and upsampling the rest at most x100 (on top of this, we manually increased the weights for some languages with underrepresented scripts, such as Greek or Korean, to adjust the resulting tokenizer fertilities). For some languages, the bottleneck of tokenization fertility has been not in the vocabulary itself but in the pre-tokenization word splitting regular expression, so we extended it with additional Unicode ranges and with a pattern for matching diacritic marks within a word. As a result of these operations, the extended tokenizer achieved the average fertility of 44.8 tokens per sentence over the 212 languages in the FLORES+ dataset, as opposed to 80.7 tokens in the original \llamathree tokenizer.

When initializing representations for newly added tokens, we first tokenize them using the original tokenizer and then subsequently compute the average of the corresponding token embeddings \citep{gee2022fast, moroni2025optimizing}. 

\paragraph{Ablation} In order to measure the effects of our extended tokenizer, we perform a controlled ablation experiment. We choose the \llamathree.2 1B Instruct model as a baseline, and then extend the vocabulary from 128K to 256K tokens. Both models were continuously pre-trained for 30K steps on the same data mixture with identical hyperparameters. Results are shown in \autoref{tab:tokenizer_ablation}. Overall, we observe a relative \chrf improvement of 26\% (17.8 $\rightarrow$ 22.5) for out-of-English and 7\% (35.9 $\rightarrow$ 38.7) for into-English on \floresplus, with tangible improvements across all language resource levels. %

\begin{table}[ht]
\centering

\begin{tabular}{l|cccc|c|cccc|c}
\toprule
\textbf{Direction} & \multicolumn{5}{c|}{\textbf{\engX}} & \multicolumn{5}{c}{\textbf{\Xeng}} \\
\textbf{Resource level} & \textbf{high} & \textbf{mid} & \textbf{low} & \textbf{very low} & \textbf{all} & \textbf{high} & \textbf{mid} & \textbf{low} & \textbf{very low} & \textbf{all} \\
\midrule
Baseline model (128K) & 39.01 & 16.95 & 12.23 & 12.65 & 17.80 & 54.06 & 35.33 & 31.00 & 31.51 & 35.92 \\
+ Added tokens (256K) & 41.19 & 23.16 & 16.70 & 15.45 & 22.50 & 54.79 & 39.13 & 33.64 & 33.06 & 38.67 \\
\bottomrule
\end{tabular}

\caption{\chrf when evaluating MT systems continuously pretrained with and without our extended 256K tokenizer.\label{tab:tokenizer_ablation}}
\end{table}

\section{Decoder-only Modeling}
\label{sec:deconly}

In this section, we present the proposed translation model built on top of \llamathree. The development of this model consists of the following phases: Continual PreTraining (CPT) and Post-training. Additionally, we explore Retrieval Augmented Translation (RAG). 

\subsection{Base models}
The main \omtllama model is based on the LLaMA 3.1 8B Instruct model\footnote{\url{https://huggingface.co/meta-llama/Llama-3.1-8B-Instruct}}, inheriting its architecture and parameters. The only architectural change that it underwent was replacing its tokenizer with a more multilingual one and extending the input and output token embedding matrices accordingly, as described in the previous section. In all subsequent sections, we refer by default as ``\omtllama'' to the result of further training this 8B model.

In addition, we experiment with scaling the model size down to enable training and inference in more resource-constrained environments or simply at a lower cost. For this purpose, we create smaller models following the same recipe: 1B and 3B models. We initialize them with LLaMA 3.2 1B Instruct and LLaMA 3.2 3B Instruct, respectively, and carry out the same vocabulary extension procedure as for the main, 8B model. The smaller models also undergo the same training process as the main one, outlined in the following subsections.

\subsection{Continued Pretraining}
\label{sec:cptmodel}

Inspired by the related work of specialised MT models e.g. Tower \citep{tower_llm_2024}, we include two tasks in our Continual PreTraining (CPT): language modeling with monolingual documents, and translation with parallel documents.

In practice, our dataloader samples batches from multiple sources. Long monolingual documents are wrapped; short documents are packed together to fill the maximum sequence length. We sample from the streams of tokens from different sources proportionally to their weights described in Table \ref{tab:CPTtrainingdetails}. %

Before each monolingual document, we insert the name of the language, to teach the model to associate languages with names. For each translation pair, we use a simplified translation prompt indicating the source and target languages as follows:

\colorbox{gray!20}{%
  \begin{minipage}{0.95\textwidth}
      \itshape
      Translate {\color{blue} source-sentence} from {\color{blue} source-language} into {\color{blue}target-language}: {\color{blue} target-sentence}
  \end{minipage}}

\paragraph{Training Configuration} We continuously pretrain our base models for up to 50,000 steps, distributed across a cluster of 256 NVIDIA A100 GPUs. For model variants necessitating vocabulary adaptation, we precede the main pretraining phase with a dedicated warmup stage. This warmup consists of 10,000 steps executed on 80 NVIDIA A100 GPUs, during which all model parameters are held fixed except for the token embedding matrix and the output projection layer, which remain trainable to facilitate efficient vocabulary integration. All training procedures utilize the AdamW optimizer, configured with a base learning rate of $\eta = 5 \times 10^{-5}$, $\beta_1 = 0.9$, $\beta_2 = 0.95$, and weight decay $\lambda = 0.1$. The maximum input sequence length is set to 8,192 tokens for all training runs. During the vocabulary adaptation warmup phase, we employ an elevated learning rate of $\eta = 2 \times 10^{-4}$ to accelerate convergence of the newly introduced parameters.

\subsection{Post-training}
\label{sec:post_training}

Post-training is used to recover and enhance instruction-following behavior after continued pretraining (CPT), while further specializing the model for high-quality machine translation. We apply supervised fine-tuning (SFT) and reinforcement learning (RL), and analyze their respective contributions relative to the CPT model.

\subsubsection{Supervised Fine-Tuning}
\label{sec:sft}

We fine-tune the CPT model on a mixture of instruction-following and machine translation data. The objective of supervised fine-tuning (SFT) is twofold: (i) to restore instruction-following capabilities that may be degraded during CPT, and (ii) to bias the model toward producing high-quality translations across a wide range of language pairs.

\paragraph{Training Data}
Our base fine-tuning dataset (\ourblocks) contains $\approx$ 600k multilingual instruction-tuning examples covering 10 languages (English, Portuguese, Spanish, French, German, Dutch, Italian, Korean, Chinese, Russian). The dataset covers general conversational instruction-following (42\%), machine translation (25\%), machine translation evaluation (22\%), automatic post-editing (6\%), and other language-related tasks such as named entity recognition and paraphrasing. %
The data is predominantly English (53\%), with substantial mixed-language content (18\%, largely English combined with code). %
All examples are formatted as instruction–response pairs.

To extend the language coverage of the base fine-tuning dataset, we format the SMOL and \medley translation datasets with diverse translation prompts. In the training mix, we weigh the three datasets in the 3:1:1 proportion.

\paragraph{Training Configuration}
We optimize using AdamW~\citep{loshchilov2019decoupledweightdecayregularization} with learning rate $\eta = 1 \times 10^{-6}$, $\beta_1 = 0.9$, $\beta_2 = 0.95$, and weight decay $\lambda = 0.1$. We use a cosine annealing schedule with 1,000 warmup steps and a final learning rate scale of 0.2. Training runs for 10,000 steps with a maximum sequence length of 8,192 tokens, validating every 100 steps.

Training employs Fully Sharded Data Parallel (FSDP) with FP32 gradient reduction and layer-wise activation checkpointing. All examples are formatted using the  \llamathree chat template~\citep{llama3modelcard}. %

\subsubsection{Reinforcement Learning}
\label{sec:rl}

We further apply reinforcement learning (RL) to improve translation quality beyond SFT. Initial experiments using Group Relative Policy Optimization (GRPO) with lexical rewards such as \chrf and BLEU revealed that dataset curation was critical: narrowly templated instruction data led to in-distribution improvements but poor generalization. Using the \ourblocks subset, which exhibits substantial instruction diversity, enabled stable and generalizable gains.

Consistent with MT-R1-Zero~\citep{feng2025mtr1zeroadvancingllmbasedmachine}, we use a reward that averages normalized \chrf and BLEU scores and adopt a direct translation setup without explicit reasoning tokens. While reasoning-based approaches such as DeepTrans~\citep{wang2025deeptransdeepreasoningtranslation} show strong results for literary translation, reliably eliciting such behavior remains an open challenge.

Applying RL on top of SFT checkpoints introduces optimization difficulties due to low entropy and vanishing gradients under standard GRPO~\citep{yu2025dapoopensourcellmreinforcement}. To address this, we adopt Decoupled Clip and Dynamic Sampling Policy Optimization (DAPO) with larger group sizes ($N=64$). DAPO preserves exploration through asymmetric clipping and ensures non-zero gradient signals via dynamic sampling. We reintroduce KL regularization to constrain deviation from the SFT checkpoint and do not apply overlong reward shaping. The final reward objective is a balanced 50/50 combination of \chrf and MetricX.

\subsubsection{Results}
\label{sec:cpt_ft_effects}

For some datasets, we report results both on the full evaluation set and on a subset of language directions corresponding to those explicitly used during supervised fine-tuning and reinforcement learning. We refer to this subset as \emph{SFT langs}. Results are summarized in Table~\ref{tab:post_training_mt}.

\begin{table}[t]
\centering
\scriptsize
\setlength{\tabcolsep}{2.5pt}
\renewcommand{\arraystretch}{1.05}

\begin{tabular}{lcccccccccccc}
\toprule
\textbf{Model}
& \multicolumn{4}{c}{\textbf{\bouquet}}
& \multicolumn{4}{c}{\textbf{\floresplus}}
& \multicolumn{4}{c}{\textbf{\bible}} \\
\cmidrule(lr){2-5} \cmidrule(lr){6-9}
\cmidrule(lr){10-13} 
& \textbf{$\rightarrow$ en$_{(s.)}$} & \textbf{en $\rightarrow$$_{(s.)}$} & \textbf{$\rightarrow$ en$_{(p.)}$} & \textbf{en $\rightarrow$$_{(p.)}$}
& \textbf{$\rightarrow$ en} & \textbf{en $\rightarrow$} & \textbf{$\rightarrow$ en$_{(h)}$} & \textbf{en $\rightarrow$$_{(h)}$}
& \textbf{$\rightarrow$ en} & \textbf{en $\rightarrow$}
& \textbf{$\rightarrow$ en {\tiny\textsc{[sft langs]}}} & \textbf{en $\rightarrow$ {\tiny\textsc{[sft langs]}}} \\
\midrule
CP4
& .725 & .621 & .527 & .404 
& .735 & .523 & .376 & .231 
& .683 & .609 & .703 & .641 \\

 $\hookrightarrow$ TB (SFT)
& .732 & .611 & .549 & .404 
& .735 & .539 & .382 & .233 
& .685 & .656 & .706 & .685 \\

 \phantom{a}  $\hookrightarrow$ RL (DAPO)
& .741 & .616 & .555 & .403 
& .747 & .541 & .393 & .233 
& .689 & .661 & .708 & .685 \\
\bottomrule
\end{tabular}

\caption{\label{tab:post_training_mt}
Machine translation performance (xCOMET) after post-training. We compare the CPT model (CP4), supervised fine-tuning (SFT), and further reinforcement learning with DAPO. \bouquet is evaluated at the sentence (s) and paragraph (p) level. We evaluate both in \floresplus and \floreshard (h). %
}
\end{table}

\paragraph{Effects of Supervised Fine-Tuning}
Supervised fine-tuning yields consistent improvements over CPT across all evaluated benchmarks. On \bouquet, SFT improves most directions, particularly paragraph-level translation into English (from 0.527 to 0.549) and sentence-level translation into English (from 0.725 to 0.732), while remaining largely neutral for English-to-other directions.

On \floresplus, SFT produces small but consistent improvements on the hardest subsets, increasing $\rightarrow$ en$_{(hard)}$ from 0.376 to 0.382. %
On the full evaluation set, SFT improves all directions, with especially large gains for English-to-other languages (from 0.609 to 0.656). On the SFT language subset, SFT further improves translation quality (e.g., $\rightarrow$ en from 0.703 to 0.706), reflecting targeted specialization on languages seen during post-training.

\paragraph{Effects of Reinforcement Learning}
Reinforcement learning provides consistent additional improvements over SFT, though with smaller magnitude. On \bouquet, RL further improves most directions, notably sentence-level translation into English (from 0.732 to 0.741) and paragraph-level translation into English (from 0.549 to 0.555).

On \floresplus, RL yields clear gains on harder subsets, improving $\rightarrow$ en$_{(hard)}$ from 0.382 to 0.393. %
Gains are observed both on the full evaluation set (e.g., $\rightarrow$ en from 0.685 to 0.689) and on the SFT language subset (e.g., $\rightarrow$ en from 0.706 to 0.708), indicating that RL refines translation quality without overfitting to the languages used during post-training. Importantly, RL does not degrade performance in any evaluated direction.

\subsection{Retrieval-Augmented Translation}
\label{sec:retrievalaugmentedtranslation}

\paragraph{Motivation, related work and use cases}
Retrieval-augmented LLM systems become more and more popular, 
and using them for translation enables adaptation to new languages and domains without retraining.
RAG \citep{lewis2020retrieval} has been successfully extended to MT appending retrieved source‑target pairs to the input on low‑resource language pairs \citep{Vardhan2022}.
RAG is specially relevant for faster quality assessment of collected or generated translation data; continuous integration of new curated and domain specific translation examples into a retrieval database; and allowing the adaptation of closed LLM systems that cannot be finetuned.

\subsubsection{Algorithm Overview}

For retrieval-augmented translation, we query a database of parallel texts for the sources similar to the current source text to be translated, and insert the retrieved source-translation pairs as few-shot examples into the translation prompt. 

Our database for the RAG translation system consists of all parallel data sources from different translation directions and domains, as described in Section \ref{sec:CPTtrainingdata}. 
The source texts are indexed for both full text search (FTS) and vectorial search (VS).
We also index several of scalar bi-text quality signals to allow a fast filtering during the retrieval.
For the vectorial search we are using \sonaromni text embeddings, which exist for all considered languages.

To maximize good matching chances and to generate more diverse retrieved examples, we use not only an entire input text but also split it into smaller text chunks. More concretely, we first apply sentence-based level segmentation, and then we split sentences into ngrams of words of a certain size so that the total number of text chunks remain reasonable (usually between 5 and 30).
Then, for each text chunk we query similar bi-text examples based on cosine similarity ($cossim$) and BM25-based\footnote{\url{https://en.wikipedia.org/wiki/Okapi_BM25}} score similarity (up 64 examples at most for each strategy). For each query above, we also apply some filtering based on the quality signals that can remove up $30\%$ of the original samples depending on the translation direction. Technically speaking, we are using Lance binary format\footnote{\url{https://github.com/lance-format/lance}} with all indexing functionalities and all queries are executed in parallel.

After the retrieval phase, all candidate examples are merged together.
They are next deduplicated and reranked based on a linear mixture of $cossim$, BM25 and quality scores.
We additionally optimize for the word level recall (so that the union of words from top candidates covers the maximum of words in the original input text). We keep up to 80 examples (going beyond that showed only negligible improvement).

If the number of samples in RAG database is small or zero for a given direction, we can use an extra candidate generation strategy. Since \sonaromni representations are language-agnostic, we use source text embedding to find the most similar examples directly among all target examples (this subset can be large especially for higher resource languages). We keep only the examples where $cossim > 0.7$ and we say that these matching examples are actual translations (on-the-fly mining).

\subsubsection{Experiments and Results}

\paragraph{Experimental framework.} To understand the effect of retrieval, 
we added RAG examples in the prompts of 3 baseline models: \llamathree.1-8B, \llamathree.3-70B and \omtllama-8B.
We run an evaluation on a subset of 56 directions from \bouquet for which we have some data to build RAG system. In particular, among these directions, 31 directions have more than 30K RAG samples and 25 directions have less than 30k of them. Note that for the directions where there are no available samples we rely purely on-the-fly mining strategy.

\begin{table}[htbp]
\centering
\caption{Average performance metrics by system and level over 56 directions from the \bouquet dataset. In parentheses, differences from applying RAG with the same system \label{tab:ragresultsaverage}}
\begin{tabular}{lllll}
\toprule
\textbf{System} & \textbf{RAG samples} & \textbf{\chrf} & \textbf{\xcomet\_both} & \textbf{\metricx\_ref} \\
\midrule
\textit{sentence} &  &  &  &  \\
\midrule
\omtllama 8B          & <30K & 31.04 & 0.46 & 10.72 \\
 $\hookrightarrow$RAG    & <30K & 31.56 (0.52) & 0.47 (0.01) & 10.74 (0.02) \\
\omtllama 8B          & >=30K & 39.83 & 0.57 & 7.24 \\
   $\hookrightarrow$RAG   & >=30K & 42.13 (2.30) & 0.58 (0.01) & 6.70 (-0.54) \\
\midrule   
\llamathree 8B              & <30K & 21.64 & 0.41 & 13.58 \\
 $\hookrightarrow$RAG    & <30K & 23.28 (1.64) & 0.41 (0.00) & 12.86 (-0.72) \\
\llamathree 8B              & >=30K & 24.29 & 0.49 & 12.89 \\
 $\hookrightarrow$RAG       & >=30K & 28.21 (3.92) & 0.50 (0.01) & 11.05 (-1.84) \\
\midrule
\llamathree 70B             & <30K & 27.70 & 0.45 & 11.96 \\
 $\hookrightarrow$RAG      & <30K & 28.45 (0.75) & 0.45 (0.00) & 11.40 (-0.56) \\
\llamathree 70B             & >=30K & 32.67 & 0.54 & 9.95 \\
 $\hookrightarrow$RAG        & >=30K & 36.18 (3.51) & 0.54 (0.00) & 8.54 (-1.41) \\
\midrule
\textit{paragraph} &  &  &  &  \\
\midrule
\omtllama 8B          & <30K & 32.36 & 0.28 & 10.49 \\
 $\hookrightarrow$RAG    & <30K & 32.96 (0.60) & 0.28 (0.00) & 10.42 (-0.07) \\
\omtllama 8B          & >=30K & 44.29 & 0.35 & 7.00 \\
 $\hookrightarrow$RAG     & >=30K & 45.16 (0.87) & 0.35 (0.00) & 6.79 (-0.21) \\
\midrule
\llamathree 8B              & <30K & 26.66 & 0.24 & 13.28 \\
 $\hookrightarrow$RAG      & <30K & 27.50 (0.84) & 0.24 (0.00) & 13.10 (-0.18) \\
\llamathree 8B              & >=30K & 28.46 & 0.27 & 12.31 \\
 $\hookrightarrow$RAG       & >=30K & 30.28 (1.82) & 0.27 (0.00) & 11.60 (-0.71) \\
\midrule
\llamathree 70B             & <30K & 33.65 & 0.27 & 11.82 \\
 $\hookrightarrow$RAG     & <30K & 34.24 (0.59) & 0.27 (0.00) & 11.23 (-0.59) \\
\llamathree 70B             & >=30K & 38.08 & 0.32 & 10.01 \\
 $\hookrightarrow$RAG      & >=30K & 40.35 (2.27) & 0.32 (0.00) & 8.97 (-1.04) \\
\bottomrule
\end{tabular}
\end{table}

\paragraph{Results.} Table \ref{tab:ragresultsaverage} presents the evaluation metrics averaged over directions with a breakdown by evaluation level (sentence or paragraph) and number of available RAG examples.

As for averaged results, we note that RAG enabled models consistently improve over baseline model in all automatic metrics.
We see that the absolute gains are stronger if RAG systems have a large number of available examples. From the same perspective, the gain on sentence level translations is stronger than on paragraph (specially for smaller models) probably because most of our database example data is at sentence (or even word) level and finding good matches for the paragraph is more difficult.
Note that all 3 models manifest a similar tendency in the gain for different metrics.

\newpage 
\section{Encoder-Decoder Modeling}
\label{sec:encdec}

In this section, we focus on an alternative translation modeling architecture to the one presented in previous section. We use the well-known encoder–decoder Transformer architecture that has been classically used for the MT task \citep{nllb-24}. Concretely, we train a compact 3B-parameter Transformer model, namely \textsc{OMT - No Language Left Behind} (\nllbtwo), that can translate from \TOTALlanguages source languages into 250 target languages. 

\begin{figure*}
    \centering
    \includegraphics[width=\linewidth]{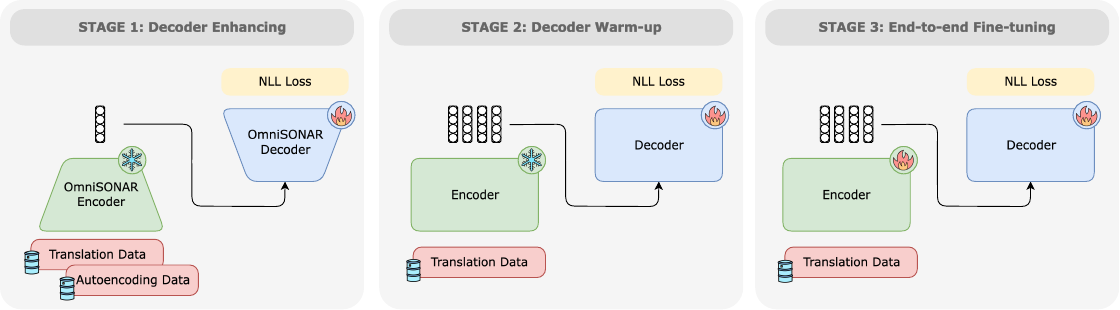}
    \caption{Overview of the proposed algorithm to train \nllbtwo model.}
    \label{fig:nllbtwo}
\end{figure*}

A known limitation of standard sequence-to-sequence training for MT is the need for parallel data, to train the system in a supervised way. To bypass this limitation, we propose a new training strategy, described in \Cref{fig:nllbtwo}.

Our method builds on top of \sonaromni, a multilingual model that maps sentences in 1600 languages into a shared representation space. In this space, equivalent sentences in different languages are mapped to the same (or very similar) sentence embeddings. \sonaromni is paired with a multilingual decoder that can decode text from the representation space in 200 target languages, via cross-attention on pooled representations. Taken together, the \sonaromni encoder and decoder already define a translation system operating through a single cross-lingual sentence embedding.

We leverage the cross-lingual alignment property to exploit both parallel and non-parallel data. In a first stage, we keep the cross-lingually aligned encoder fixed and train a decoder on a mixture of: (1) standard parallel MT data, and (2) monolingual data via an auto-encoding objective. For the auto-encoding task, the encoder maps a monolingual sentence into the shared \sonaromni space, and the decoder is trained to reconstruct the same sentence in the same language. This allows us to substantially increase the amount of training data, especially for low-resource languages where parallel corpora are scarce but monolingual text is widely available. As a result, the decoder becomes stronger and more robust across languages, since it is exposed to much more linguistic variation than it would see from parallel data alone.

A key limitation of this setup, however, is the bottleneck representation between encoder and decoder. The original \sonaromni architecture relies on a pooled sentence-level representation: the encoder compresses the input sequence into a single vector, which is then fed to the decoder. While this design is well-suited to building a shared cross-lingual space, it restricts the amount of fine-grained information that can be passed from the encoder to the decoder, and it prevents the model from fully exploiting token-level cross-attention as in standard Transformer-based MT.

To address this, the second key idea of our method is to remove this bottleneck and move from cross-attention on sentence-level pooled representations to token-level encoder-decoder attention. After the initial stage that exploits the aligned \sonaromni space, we connect the encoder and decoder through standard cross-attention over the full sequence of encoder hidden states. Since the removal of the bottleneck breaks the original alignment in the shared embedding space, the following training stages operate only on parallel translation data. We then apply a two-stage training procedure in this non-pooled setup. 

First, we perform a decoder warm-up phase, where only the decoder parameters (including cross-attention layers) are updated, while the encoder remains frozen. This step allows the decoder to adapt its cross-attention to reading full token sequences instead of a single pooled vector, and stabilizes training when transitioning away from the sentence-level bottleneck. 

In the second phase, we fine-tune the entire model end-to-end on parallel data, jointly updating encoder and decoder. This final stage enables the model to fully exploit token-level interactions while retaining the benefits of the initial training on large amounts of monolingual data through the cross-lingual encoder.

Overall, our approach combines the strengths of a cross-lingually aligned encoder with the flexibility of a standard encoder-decoder Transformer. It allows us to train with non-parallel data via auto-encoding in the initial stage, and then to recover a powerful sequence-to-sequence MT model without the representational bottleneck imposed by a single sentence embedding. Furthermore, the resulting system remains compact, with only 3B parameters, since it preserves the original size of the \sonaromni encoder–decoder architecture.

\subsection{Leveraging an Aligned Encoder for Enhanced Decoder Training}
\paragraph{Training Data.} For the first stage of our experiments, we use the data described in \Cref{tab:CPTtrainingdetails}, which was curated for continual pretraining of our decoder-only model. We leverage the bilingual parallel data from this collection to train our model for translation using traditional supervised learning. Additionally, we incorporate the monolingual non-parallel data to train the model with an autoencoding objective.

\paragraph{Experimental Framework.} For our first experiments we employ the \sonaromni codebase, keeping its original architecture but keeping the encoder frozen. We initialize our encoder and decoder using the ones from the \sonaromni model. 
In our experiments, we set the training objective to Negative Log Likelihood in the Machine Translation and Autoencoding tasks. The batch size per GPU is set to 6k tokens per batch, and the models were trained on 16 nodes, of 8 A100 GPUs each. We utilize the AdamW optimizer~\citep{loshchilov2019decoupledweightdecayregularization}. The learning rate is set to 0.001, and this first stage takes a total of 40k steps, including a learning rate warmup in the first 200.

\paragraph{Results.} We evaluate our approach both with and without the autoencoding (AE) objective to assess its contribution. As shown in \Cref{tab:encdec_results}, our first-stage training shows that leveraging the cross-lingually aligned encoder for enhanced decoder training is effective. 

The \textit{Decoder Enhancing MT} variant, trained exclusively on parallel translation data while keeping the encoder frozen, shows improvements over the baseline \sonaromni model. Furthermore, the \textit{Decoder Enhancing MT+AE} experiment demonstrates larger performance improvements. By incorporating monolingual data through the autoencoding objective, the decoder is exposed to more linguistic diversity and becomes more robust across languages.  These results confirm that exploiting monolingual data via the cross-lingually aligned \sonaromni encoder provides an enhancement to decoder quality.

\begin{figure*}
    \centering
    \includegraphics[width=0.5\linewidth]{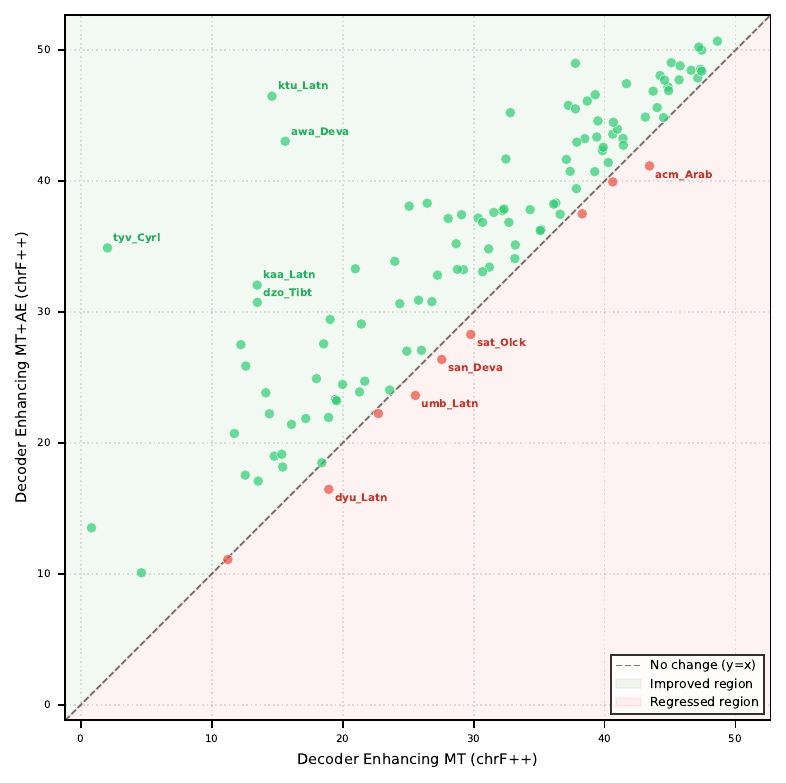}
    \caption{ChrF++ improvements (English to target translation) in languages where autoencoding data is used. Points above the diagonal indicate languages in which autoencoding data helped to improve the performance of the model.}
    \label{fig:ae_by_language}
\end{figure*}

To understand if these gains are consistent, we further analyze performance specifically on languages for which we added substantial amounts of autoencoding data in Stage 1. \Cref{fig:ae_by_language} shows the performance of both \textit{Decoder Enhancing MT+AE} and \textit{Decoder Enhancing MT} models by language. We can see that the trend clearly shows that AE data improved the performance of those languages where it was added. In particular, the performance increased in 105 out of the 114 languages where AE data was added, with an average improvement of 5.20 chrF++ points. This confirms our hypothesis that autoencoding data is valuable for low-resource languages where parallel corpora are limited. However, we observe that the performance improvement of the overall model increases less than the performance of these specific languages, showing signs of the well-known \textit{Curse of Mulilinguality} \citep{massively_multilingual_nmt,lifting_the_curse_of_multilinguality,alastruey2025interferencematrixquantifyingcrosslingual}. 

\subsection{Decoder Warm-up for Token-Level Cross-Attention}
\paragraph{Training Data.} In the second stage of our experiments, we train the model exclusively on the bilingual parallel data described in \Cref{tab:CPTtrainingdetails}, removing the monolingual non-parallel data and its associated autoencoding component. 

\paragraph{Experimental Framework.}
For the second stage of our experiments we employ the \sonaromni codebase, still keeping the encoder frozen but adding a custom adaptation to remove the original bottleneck. We initialize our model using the original \sonaromni encoder and the enhanced decoder we obtained in the previous stage training with both translation and autoencoding data. We set the training objective to Negative Log Likelihood in only the Machine Translation task. The rest of the training setup remains the same as the one used during the Decoder Enhancing step.
\paragraph{Results.}
The decoder warm-up stage, which removes the sentence-level pooling bottleneck and introduces token-level cross-attention, yields further improvements across all benchmarks, as shown in model \textit{Decoder Warm-up} in \Cref{tab:encdec_results}. By allowing the decoder to attend to the full sequence of encoder hidden states rather than a single pooled vector, the model can access fine-grained, token-level information that was previously lost in the bottleneck. Importantly, this warm-up phase, where only decoder parameters are updated, allows the model to adapt its cross-attention mechanism to reading full sequences without destabilizing the encoder representations. We observe that skipping this step causes training instabilities (exploding gradients) in the third training stage. Therefore, this warmup proves to be helpful in transitioning from the pooled to the non-pooled architecture, obtaining consistent gains over the \textit{Decoder Enhancing MT + AE} baseline and making it possible to perform the final fine-tuning. 

\subsection{End-to-End Parallel Fine-Tuning}
\paragraph{Training Data.} In the final stage of our experiments, we maintain the same training configuration as in the previous step, continuing to use only the bilingual parallel data described in \Cref{tab:CPTtrainingdetails}
\paragraph{Experimental Framework.} 
For the last step of our training we employ our adapted \sonaromni codebase to remove the bottleneck, but we unfreeze the whole system. We initialize the model using the original \sonaromni encoder and the warmed-up enhanced decoder obtained in the end of the second stage. We set the training objective to Negative Log Likelihood in only the Machine Translation task, and we utilize the AdamW optimizer setting the learning rate is set to 0.0004. This last stage takes a total of 100k steps, including a learning rate warmup in the first 200. The rest of the training setup remains the same as the one used during the previous training stages.
\paragraph{Final Results.} 
The final \nllbtwo model, after the end-to-end fine-tuning stage where both encoder and decoder are jointly updated on parallel data, produces the overall strongest results, as reported in \Cref{tab:encdec_results}. By unfreezing the encoder, the model can now fully optimize the token-level interactions between encoder and decoder, adapting the encoder representations specifically for the translation task rather than relying solely on the pre-trained \sonaromni alignment. Compared to the original \sonaromni baseline and the following proposed training steps, our final model achieves notable improvements across all evaluation sets. 

Overall, these results demonstrate that our three-stage training strategy successfully combines the benefits of cross-lingual alignment, monolingual data exploitation through autoencoding, and token-level encoder-decoder attention. The final \textsc{\nllbtwo} model achieves substantial improvements over the \sonaromni baseline while maintaining the same compact 3B parameter size, making it both effective and efficient for multilingual machine translation.

\begin{table}[H]
\setlength{\tabcolsep}{3pt}
    \centering
    \begin{tabular}{lcccccc}
        \toprule
        &  \multicolumn{3}{c}{\textbf{\floresplus}} & \multicolumn{3}{c}{\textbf{\bible}} \\
        \cmidrule(lr){2-4} \cmidrule(lr){5-7} 
        \textbf{System} & \textbf{\chrf $\uparrow$} & \textbf{\xcomet $\uparrow$} & \textbf{\metricX $\downarrow$} & \textbf{\chrf $\uparrow$} & \textbf{\xcomet $\uparrow$} & \textbf{\metricx $\downarrow$}
        \\
        \midrule
        \sonaromni & 47.54 & 0.65 & 5.66 & 40.61 & 0.52 & 8.79 \\
        \midrule
        Decoder Enhancing MT & 49.58 & 0.65 & 5.51  & 44.85	& 0.52 & 8.61 \\
        Decoder Enhancing MT+AE & 49.74 & 0.65 & 5.48 & 45.93 & 0.53 & 8.59\\
        \midrule
        Decoder Warm-up  & 50.12 & 0.65 & 5.36 & 46.40 & 0.51 & 8.58 \\
        \midrule
        \textbf{\nllbtwo}  & 50.92 & 0.66 & 5.32  & 46.28  & 0.50 & 8.61 \\
        \bottomrule
    \end{tabular}
    \caption{Results of our proposed model on each training step of the proposed methodology compared to the original \sonaromni}
    \label{tab:encdec_results}
\end{table}

\section{Proposed Evaluation Metrics and Dataset}%
\label{sec:mtmetrics}

Model evaluation constitutes a key contribution of the \OmniMT effort. The primary challenge we encounter is the limited reliability of current automatic metrics for long-tail languages, compounded by the lack of a robust methodology to assess metric quality in this context. In this section, we describe our contributions toward advancing \OmniMT evaluation. We present a variation of the previously established human evaluation protocol XSTS \citep{licht-etal-2022-consistent}; the human annotations collected using this protocol, which constitute the largest dataset of human annotations in terms of language coverage, Met-\bouquet; and the largest multilingually trained MT metric, \blaser. Finally, this section includes a comprehensive benchmarking of MT metrics using Met-\bouquet, including our proposed \blaser, enabling us to quantify the reliability of several MT metrics across a broad representation of languages and language pairs.

\subsection{Human Evaluation Protocol: XSTS+R+P}
\label{subsec:xstsrp}

MT human evaluation is the most relevant way of comparing system performance but
it is not free of challenges. Human evaluation is expensive, slow (or even unfeasible if annotators are not available)
and its quality is highly dependent on the human evaluation protocol. In our case, we want a human evaluation that is easy enough to scale to a large number of languages, but still compatible with omnilinguality, by for example taking aspects as register, which is highly relevant across cultures, and context, which can reduce relevant ambiguities.

\paragraph{Related Work} Existing protocols include Direct Assessment (DA) \citep{graham-etal-2013-continuous}, one of the simplest protocols that simply uses a single continuous rating scale of quality. Then, XSTS \citep{licht-etal-2022-consistent} which assesses the faithfulness of translations in a 5-likert scale and focuses on semantic similarity. Multidimensional Quality Metric (MQM) \citep{lommel2024multirangetheorytranslationquality} which is one of the most complex ones because it annotates each error span with its severity level and the error type selected from seven high-level error type dimensions. And most recently, Error Span Annotation (ESA) \citep{kocmi-etal-2024-error} combines the continuous rating of DA with the high-level error severity span marking of MQM, without specifying error types. While MQM is by far the most complete and specific, which may cover most aspects of language varieties, it is quite complex and expensive. However, since we are aiming at omnilinguality, it is highly relevant that the protocol does not sacrifice sensitivity to omnilingual aspects of translation, avoiding covering only English-centric errors.%

\paragraph{XSTS+R+P} Among the existing human evaluation protocols--DA, MQM, ESA, XSTS---we prioritized a well-documented protocol with an associated calibration method, which focuses on semantic equivalence rather than on error analysis. The XSTS protocol best meets these initial requirements. It offers higher inter-annotator agreement than DA, as shown in \citep{licht-etal-2022-consistent}, and is easier to implement at omnilingual scale, where finding the expertise necessary to master MQM can prove challenging. 
However, using XSTS in its original formulation would preclude taking full advantage of two central aspects of the central evaluation dataset of our work (\bouquet, section \ref{sec:bouquet}): its language register diversity, and its paragraph-based design. To meet our full requirements, while building on XSTS, our proposed protocol, XSTS+R+P, specifies scoring criteria for two additional situations. First, it takes into account elements of pragmatics by indicating how annotators should rate register (R) discrepancies at the sentence level. Second, it provides annotators with a means to downgrade the rating of a translation that is semantically equivalent at the sentence level but causes confusion when considered at the paragraph (P) level (e.g., that is inconsistent in degree of formality, verb conjugation, or grammatical gender attribution with other sentences of the same paragraph). Detailed guidelines with examples are available in Appendix \ref{app:metbouquet}.

\paragraph{Annotation and Calibrations Process}  We commissioned XSTS+R+P language annotations across different vendors and checked all the deliveries we received. Each delivery contained three parts: the calibration file which had twenty annotations, the file with source-to-target annotations and the file with the opposite language direction. To ensure the quality of all those deliveries, we established a validation process which had three main steps. During the first step, we checked the calibration: we compared how well the ratings we received were aligned with the references we had prepared. We also studied all the comments left by the raters to ensure they were following the guidelines. For the most part, the comments proved to be relevant and showcased good understanding of the guidelines. However, in some cases we had to emphasize that the goal of the protocol was to only evaluate semantic similarity, not translation style and quality.
During the second step, we studied the main part of the delivery (by bidirectional pairs) focusing on the paragraphs where the raters showed the most misalignment. We automatically highlighted the paragraphs where the deviation between the scores of different raters was equal or more than 2 (for example, one rater gave a sentence a score of 5, whereas another one gave it a score of 3). We looked at how many such rows there were: most deliveries had around 2\% of rows showing annotation misalignment, and if this parameter was significantly higher than 3\%, we sent the delivery back for rework. The common reasons which caused misalignment included, above all, different interpretation of the guidelines and edge cases, such as code-mixing and the presence of loan words in the target sentence.
Finally, we spot-checked some of the rows manually to the best of our ability, especially if the misalignment for the row was apparent. We utilized machine translation where possible and checked if the rating was compliant with the guidelines we provided. We checked around 10\% of all items that way. 

To complement the manual validation process, we implemented automated checks across all language directions. For each direction, we computed pairwise exact match rates between annotators to detect potential duplication of annotations, and analyzed per-annotator score distributions using divergence metrics (Jensen-Shannon Divergence, Wasserstein Distance) and item-level statistics (Mean Absolute Error, bias, Spearman correlation) to identify outliers. Annotators flagged by these checks were manually reviewed, and deliveries with confirmed issues were returned to the vendor for rework.

\paragraph{Final score.} %
The sentence‑level consensus is the median of the three annotator ratings. Paragraph‑level scores are computed as the harmonic mean of the sentence‑level consensus values, chosen for its greater sensitivity to low‑scoring sentences than the arithmetic mean.

\paragraph{Protocol comparison} For protocol comparison we choose 7 language pairs and annotated 1358 \bouquet sentences (dev and test partitions) on each pair (English-Russian, English-Spanish, English-Korean, English-Romanized Hindi, Hungarian-Czech, German-Croatian, French-Kinhasa Lingala). Linguistic considerations for pairs included: dominant word order, use of registers, number of grammatical genders and number of pairs with English. We controlled our protocol with the initial XSTS, for similarity, and RSQM, a simplified version of the MQM protocol.

\begin{figure}
    \centering
    \includegraphics[width=1\linewidth]{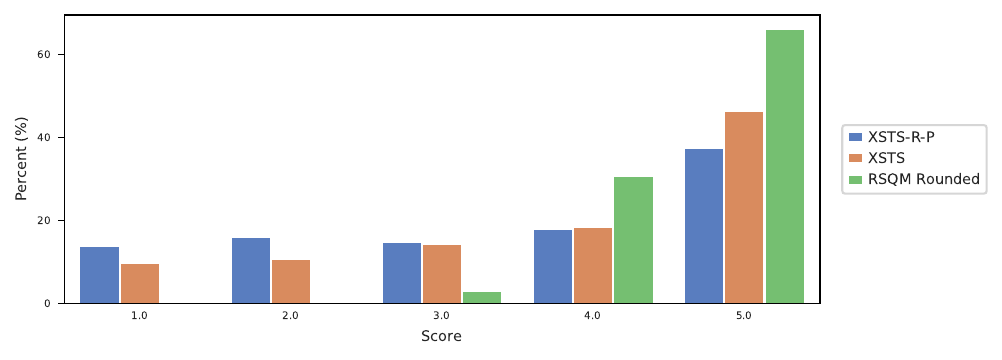}
    \caption{Score Distribution of XSTS+R+P, XSTS and RSQM$_{rnd}$.}
    \label{fig:comp_score_dist}
\end{figure}

\subparagraph{Score Distribution}
Figure \ref{fig:comp_score_dist} shows the distributions of scores for the three protocols. We notice that the distribution of XSTS+R+P is similar to XSTS' but slightly less concentrated on the upper end. This hints at XSTS+R+P's improved ability to provide additional nuance in the evaluation of translations.
We also show the distribution of RSQM scores, projected within the same range as XSTS+R+P and XSTS. RSQM exhibits a distinctively skewed distribution, with a significant concentration on higher scores.
To compare RSQM scores (range $[0, 100]$) with XSTS+R+P scores (range $[1, 5]$), we define $RSQM_{\text{rnd}}$ as a linear projection of RSQM onto the $[1, 5]$ scale, rounded to the nearest integer:
\begin{equation}
    \text{RSQM}_{\text{rnd}} = \text{round}\left( 0.04 \times \text{RSQM} + 1 \right)
\end{equation}

Computing the correlation, we see that XSTS+R+P is strongly and positively correlated with XSTS (0.65) and RSQM (0.62) on terms of Kendall's Tau.

\subparagraph{Inter-Annotator Agreement}
Table \ref{tab:iaa} shows the inter-annotator agreement (IAA) calculated with Krippendorf α using the squared distance penalty.
Our inter-annotator agreement results are broadly consistent with, and in several cases exceed, values reported in the literature for comparable translation evaluation frameworks \cite{song-etal-2025-enhancing} and \cite{licht-etal-2022-consistent}.  Crucially, our proposed XSTS+R+P protocol achieves a substantially higher mean Krippendorff's α of 0.80, representing a marked improvement over both our baseline protocols and the agreement levels typically reported in the translation evaluation literature. These findings indicate that XSTS+R+P provides a more reliable and reproducible framework for cross-lingual translation quality assessment.

\begin{table}[h!]
\centering
\begin{tabular}{lccc}
\toprule
 \textbf{Direction}  &   \textbf{XSTS+R+P}  & \textbf{XSTS} & \textbf{RSQM}\\
\midrule
 Czech$\rightarrow$Hungarian & \textbf{0.86} & 0.52 & 0.40 \\
 German$\rightarrow$Croatian & \textbf{0.89} & 0.72 & 0.49 \\
 English$\rightarrow$Korean & \textbf{0.67} & 0.54 & 0.52 \\
 English$\rightarrow$Russian & \textbf{0.60} & 0.51 & 0.43 \\
 English$\rightarrow$Spanish & \textbf{0.71} & 0.53 & 0.47 \\
 French$\rightarrow$Kinshasa Lingala & \textbf{0.90} & 0.77 & 0.51 \\
 French$\rightarrow$Swahili & \textbf{0.84} & 0.36 & 0.22 \\
 Croatian$\rightarrow$German & \textbf{0.90} & 0.67 & 0.49 \\
 Hungarian$\rightarrow$Czech & \textbf{0.91} & 0.53 & 0.27 \\
 Korean$\rightarrow$English & \textbf{0.70} & 0.60 & 0.59 \\
 Kinshasa Lingala$\rightarrow$French & \textbf{0.88} & 0.68 & 0.57 \\
 Russian$\rightarrow$English & 0.58 & \textbf{0.65} & 0.49 \\
 Spanish$\rightarrow$English & \textbf{0.82} & 0.34 & 0.48 \\
 Swahili$\rightarrow$French & \textbf{0.91} & 0.51 & 0.26 \\
\midrule
 Mean & \textbf{0.80} & 0.57 & 0.44\\
 \bottomrule
\end{tabular}
\caption{Inter-annotator agreement (IAA) for different language directions, calculated with Krippendorf α using the squared distance penalty. Best results in bold.}
\label{tab:iaa}
\end{table}

\subparagraph{Differences between Protocols}

Using \bouquet sentences gives us the unique opportunity to compare the differences between XSTS+R+P and XSTS protocols across domains but also across three functional areas of register: connectedness, preparedness, and social differential. See Figure 5 in \cite{bouquet} for deeper explanation of register.
We observe that, for the same translations, the average absolute difference in scores between protocols varies according to the social differential present in the sentence. As expected, translations where the speaker addresses someone of higher social status (lower-to-higher social differential) are penalized most by XSTS+R+P compared to XSTS, while those where the speaker addresses someone of lower status (higher-to-lower social differential) receive the least penalty. Additionally, we find a clear trend: as the length of the source segment increases, the penalty imposed by XSTS+R+P also grows. Overall, XSTS+R+P scores tend to diverge (decrease relative to XSTS) as both the context size and the social differential increase.

\subsection{Met-\bouquet}
\label{sec:metbouquet}

\paragraph{Motivation} While making Machine Translation (MT) more and more massively multilingual, MT metrics have to continue evolving to meet the needs of the field. Met-\bouquet is contributing towards this end by presenting a highly multilingual, multi-way parallel annotation dataset and benchmark for MT evaluation metrics and quality estimation. This dataset is designed in two different rounds with complimentary rationales. The first round rationale was to collect a variety of systems outputs to optimise for a variety of translation errors and diverse quality annotations. The second round was designed to a selection of our best decoder-only systems (\omtllama) to strongest external baseline systems (following automatic evaluation). The language selection optimised for language and language direction coverage. Annotations were done based on XSTS+R+P (Section \ref{subsec:xstsrp}).

\paragraph{Related work}  While WMT competitions \citep{kocmi-etal-2025-findings} provide a powerful arena to evaluate MT metrics, the dataset developed there has its own challenges. Originally, the metric evaluation benchmark was designed to evaluate MT systems and not metrics themselves. This means that covered languages are not necessarily representative enough to cover linguistic families. Additionally, competitions' rules vary year to year, meaning that collections that are derived from them have some mismatches such as several protocols and close to random collection of languages.   %
However, there are indeed other collections that have been created and designed for the purpose of metric evaluation which do not have these challenges but are more limited in size. This includes  MLQE-DA-PE \citep{fomicheva-etal-2022-mlqe} and the Indic collection \citep{sai-b-etal-2023-indicmt}. Rarely, the existing datasets share the same source sentences across several source languages which some exceptions e.g. NLLB \citep{nllb-24}, with the aim of making the dataset the closest to multi-way parallel that it can be. Note that we cannot aim at having fully multi-way parallel dataset because the MT outputs differ for each language pair translation. The motivation for close to-multi-way parallel, hereinafter we will avoid ``close'' for simplicity, is the same as in for MT evaluation which is comparing the performance across languages.

\paragraph{\bouquet Dataset selection Round 1}
Additional motivations to construct an MT metric evaluation data set is the need to do it in one of the latest MT evaluation sets \bouquet \cite{bouquet}, described in section \ref{sec:bouquet}. %
\bouquet in addition to being highly multilingual and constantly expanding through its online contribution tool\footnote{https://bouquet.metademolab.com/}, it has other interesting characteristics, which are relevant for building a MT metrics dataset including diversity of domains and registers and non-English-centric data, among others. For this round, we use the dev (564 sentences) and test (854 sentences) partitions of \bouquet.%

\paragraph{\bouquet Dataset selection Round 2} Again, we evaluate the outputs on the \bouquet dataset.%
Differently from Round 1, we prioritise the test partition (854 sentences), to optimize the annotations budget that we have with more language pairs evaluated and two outputs per source sentence. The second round was intended as a part of evaluation study of the \omtllama models (see Section \ref{sec:humaneval}), so for each translation, we annotated the translations from one OMT-based system (either \omtllama or a system based on \OmniMT retrieval-augmented translation) and from one external baseline system.

\paragraph{Language selection Round 1 and 2} 
Met-\bouquet covers a diversity in language directions. 
The criteria to choose those language directions are mainly guided by the languages available in \bouquet  (see Table 6 in \citep{bouquet}) that cover a wide range of high and extremely low resource languages representing a wide variety of language families and geographical regions. %
To choose language pairs, and the complete list of language pairs is reported in Table \ref{tab:bouquetlangs}, we are motivated by:

\begin{itemize}
\item Language pairs need to have a source language available in Bouquet. 
\item Optimize for a large number of languages evaluated, do not limit to having bidirectional pairs so that we can include languages that are not in \bouquet as target languages.
\item Optimize for non-English pairs and use pivot languages instead, to follow the \bouquet non-English-centric criteria
\item Include a variability of likely-zero-shot languages so that we are able to study what should be the lowest performing languages. 
\item Include a variability of internal priority languages.
\item Include languages available in \medley, so that we can explore its effectiveness in likely-zero-shot languages and low-resource languages.
\item Do final selection to optimize for a diverse range of language families and language scripts.
\end{itemize}

Specifically, we end up covering 104 language directions in Round 1 and  57 language directions, mostly complimentary to Round 1, in Round 2. In total, \metbouquet currently covers 161 language directions and 119 unique language varieties. %

Pairs were formed based on known or likely patterns of bilingualism between source and target languages due to geographical adjacency (including pairings of subnational languages with a regional lingua franca or a national language of wider/official communication).

\paragraph{MT Systems and Outputs Considered Round 1}
 We aim to cover a variety of errors and also provide a diversity of scores. The WMT data provide a large number of systems, but this is not the case for other benchmarks MLQE-PE \citep{fomicheva-etal-2022-mlqe}, IndicMT \citep{sai-b-etal-2023-indicmt} and NLLB data \citep{nllb-24}. In our case, we prioritize a variety of open systems which cover a wide range of languages. We include a variation of open systems and early variations of \omtllama, all of them reported in Table \ref{tab:mt_systems_comprehensive}.
For each source sentence and translation direction, we sampled exactly one output translation, trying to balance the representation of various levels of translation quality and the diversity of systems. Exact details on the selection of MT outputs can be found on appendix \ref{app:metbouquet}.

\paragraph{MT Systems and Outputs Considered Round 2}
For each direction, we evaluate two main systems of interest to compare: the strongest external baseline (one of Gemma-3 27B, MADLAD-400-MT-10B, Aya-101 13B, Aya-Expanse 8B, EuroLLM 9B) and the strongest of several \omtllama-based systems (\omtllama 8B with or without RAG examples, and in a few cases, vanilla \llamathree 70B with the \OmniMT RAG examples). To select the two candidate systems for each direction, we use the dev split of \bouquet and a combination of automated metrics: the average of normalized \blaser (reported in Section \ref{sec:blaser} and \metricx scores and, when references are available, \chrf scores. See table \ref{tab:mt_systems_comprehensive} for a summary of system selection and outputs considered as well as the goal and setup of each Round. 

\begin{table}[ht!]
\centering
\small
\begin{tabular}{l l l l}
\hline
\textbf{Category} & \textbf{Model Version} & \textbf{Params} & \textbf{Citation Reference} \\ \hline
\multirow{6}{*}{\textbf{Open Systems}} & NLLB-200 & 3B & \citep{nllb-24} \\
 & \llamathree & 3B, 8B, 70B & \citep{llama3modelcard} \\
 & TowerInstruct v02 & 7B & \citep{tower_llm_2024} \\
 & Aya-101 & 13B & \citep{ustun2024aya} \\
 & Aya Expanse & 8B & \citep{dang2024ayaexpansecombiningresearch} \\
 & Cohere R & 7B & \citep{cohere2025commandaenterprisereadylarge} \\
 & Qwen 2.5 & 7B & \citep{qwen2,qwen2.5} \\
 & EuroLLM & 9B & \citep{martins2024eurollmmultilinguallanguagemodels} \\
 & Babel-9B-chat & 9B & \citep{zhao2025babelopenmultilinguallarge} \\
 & MADLAD-400-MT & 3B, 10B & \citep{kudugunta2023madlad} \\ \hline
\multirow{4}{*}{\textbf{\omtllama Systems}} & \omtllama (FT) & 3B, 8B & Section \ref{sec:post_training} \\
 & Step-by-Step & 8B, 70B & \citep{briakou-etal-2024-translating} \\
 & OMT RAG (Low-res) & 70B & Section \ref{sec:retrievalaugmentedtranslation} \\
 & OmniSONAR & -- & \citep{sonaromni} \\ \hline
\textbf{Round 1 Goal} & \multicolumn{3}{l}{Maximize diversity of systems, error types, and score ranges.} \\
\textbf{Round 1 Setup} & \multicolumn{3}{l}{Sampled exactly one output translation per source sentence/direction.} \\ \hline
\textbf{Round 2 Goal} & \multicolumn{3}{l}{Comparison between the strongest baseline and strongest OMT system.} \\
\textbf{Round 2 Baselines} & \multicolumn{3}{l}{Gemma-3 27B, MADLAD 10B, Aya-101/Expanse, EuroLLM 9B.} \\
\textbf{Round 2 Metrics} & \multicolumn{3}{l}{Avg. of normalized \textsc{Blaser} and MetricX; ChrF++ (if refs available).} \\ \hline
\end{tabular}
\caption{Summary of MT Systems (not including small variants of each), Selection Methodology and setup and goals of Round 1 and Round 2}
\label{tab:mt_systems_comprehensive}
\end{table}

\paragraph{Statistics Round 1}
 Following the XSTS+R+P protocol, we collect 1358 sentence annotations (318 paragraphs) which correspond to the \bouquet dev and test partitions\footnote{Note that the \cite{bouquet} paper reports incorrectly the number of test sentences (10 sentences more than it has)} for 53 languages and 104 language directions from 31 different MT systems.

\paragraph{Statistics Round 2} In this round, we collect annotations of 854 sentences translated in 57 directions between 80 unique languages, with each sentence translated by two systems.
We use stronger MT systems and do not perform adversarial sampling, unlike in the first round, but we also choose more challenging translation directions.\footnote{For several translation directions with extremely low-resourced source or target languages, we had to cancel the annotation after receiving over 80\% of lowest score in the first annotation batch, indicating that both systems fail to produce even minimally meaningful translations. They were therefore excluded from Round 2.} As a result, the distribution of XSTS+R+P scores in the two rounds is roughly the same, with the average score of 3.0 and about 30\% of the scores in the ``very low'' and ``very high'' areas each. 

\paragraph{Preference annotations as a by-product} Because each sentence in Round 2 has been translated in each target language twice, the resulting annotated dataset can be viewed as a dataset of human preferences (induced from the human labels indirectly, because the alternative translations were not shown to the annotators side-by-side). Out of the $\approx$49K annotated translation pairs, ~57\% have different consensus scores and therefore express a preference.

\paragraph{Available annotations.} The complete \metbouquet is available as part of \bouquet effort.\footnote{See \url{https://huggingface.co/datasets/facebook/bouquet}. Note that \metbouquet similarly to \bouquet, is dynamic and we are constantly extending it new languages.} The data includes both rounds of XSTS+R+P annotations, as well as the experimental annotations of a subset of Round 1 data with XSTS and RSQM protocols that were used in Section \ref{subsec:xstsrp}. Comparison to other datasets as well as details on score distribution are reported on appendix \ref{app:metbouquet} showing that Met-\bouquet uniquely contains 73\% of directions without English (118 directions).

\subsection{\blaser}
\label{sec:blaser}

State-of-the-art reference-free quality estimation (QE) metrics, like \xcomet~\citep{guerreiro-etal-2024-xcomet} and MetricX-24~\citep{juraska-etal-2024-metricx}, despite being powerful, have several limitations. Specifically: (1) they are not multilingual enough, being trained on a handful of directions; (2) they have limited zero-shot cross-lingual generation power to unseen languages, since the base encoders XLM-R~\citep{conneau2020unsupervised} or mT5~\citep{xue-etal-2021-mt5} are fully finetuned; and (3) they can only handle a single evaluation protocol,\footnote{DA scores are used during pre-training, or converted to the MQM scale during finetuning.} MQM~\citep{burchardt-2013-multidimensional}, restricting generalization and limiting training resources. Given these constraints, they are not ideal candidates for evaluating omnilingual translation. Thus, we take a first step towards omnilingual QE, by proposing \blaser, a highly multilingual and multi-protocol metric build. Like its predecessor~\citep{dale-costa-jussa-2024-blaser}, it is build on top of cross-lingual embeddings from SONAR~\citep{duquenne2023sonarsentencelevelmultimodallanguageagnostic}, where now we use \sonaromni~\citep{sonaromni}, unlocking the potential to generalize to thousands of languages. To further push the multilinguality capacity of our proposed QE model, we train it with multi-task learning on several evaluation protocols, and on a mix of real and synthetic data, covering hundreds of directions.

\subsubsection{Related Work}

Reference‑free quality estimation (QE) aims to predict translation quality without relying on a human reference.  Over the past decade the field has moved from simple lexical proxies to deep contextual models that can be deployed at runtime.  Learnable Surface‑Form Metrics train a model on human‑rated QE data using shallow features (e.g. QuEst \citep{specia2013quest} and QT21 \citep{rei2021qt21}). Neural Sentence‑Embedding metrics embed source and hypothesis sentences in a shared semantic space and compute a similarity score e.g. COMET \citep{rei2020comet}, SentSim \citep{rei2022sentsim}, PRISM \citep{thompson2020prism}. Zero‑Shot and Prompt‑Based Methods leverage large language models (LLMs) without any task‑specific fine‑tuning e.g. InstructScore \citep{lu2023instructscore}, GPT‑QE \citep{kumar2023gptqe}. Finally, Hybrid and Ensemble metrics combine the strengths of different reference‑free signals e.g. QE‑Ensemble \citep{fernandes2023ensemble}, e.g. OpenKi \citep{liu2021openki}. Overall, reference‑free metrics have progressed from hand‑crafted feature sets to end‑to‑end neural regressors and, most recently, to zero‑shot LLM prompts.

\subsubsection{Methodology}

\begin{figure}[H]
    \centering
    \begin{subfigure}[b]{0.49\textwidth}
        \centering
        \resizebox{\textwidth}{!}{\begin{tikzpicture}[
  font=\fontsize{7}{8.4}\sffamily\selectfont,
  >=stealth,
  thick,
]

\definecolor{orangefill}{RGB}{255,229,168}
\definecolor{orangeborder}{RGB}{222,143,5}
\definecolor{bluefill}{RGB}{196,228,246}
\definecolor{blueborder}{RGB}{86,180,233}
\definecolor{darkblue}{RGB}{1,115,178}

\node (source)     at (-1.1, 0.0) {source};
\node (hypothesis) at ( 1.1, 0.0) {hypothesis};

\fill[bluefill, draw=blueborder, line width=1pt]
  (-1.6, 0.55) -- (1.6, 0.55) -- (1.05, 2.05) -- (-1.05, 2.05) -- cycle;
\node at (0, 1.3) {\textsc{OmniSONAR}};

\draw[->] (source.north)     -- (-1.1, 0.55);
\draw[->] (hypothesis.north) -- ( 1.1, 0.55);

\node[draw=darkblue, fill=bluefill, rounded corners=5pt,
      inner xsep=7pt, inner ysep=3pt] (esrc) at (-0.75, 2.7) {src};
\node[draw=darkblue, fill=bluefill, rounded corners=5pt,
      inner xsep=7pt, inner ysep=3pt] (emt)  at ( 0.75, 2.7) {mt};
\node[font=\fontsize{7}{8.4}\sffamily\selectfont] at (1.55, 2.7) {$\times d$};

\draw[->] (-0.75, 2.05) -- (esrc.south);
\draw[->] ( 0.75, 2.05) -- (emt.south);

\node[draw=darkblue, fill=bluefill, rounded corners=5pt,
      inner xsep=8pt, inner ysep=4pt,
      minimum width=5.0cm] (concat) at (0, 3.5)
      {src\,;\,mt\,;\,$|$src$-$mt$|$\,;\,src$\odot$mt};
\node[font=\fontsize{7}{8.4}\sffamily\selectfont] at (3, 3.5) {$\times 4d$};

\draw[->] (esrc.north) -- (esrc.north |- concat.south);
\draw[->] (emt.north)  -- (emt.north  |- concat.south);

\node[draw=orangeborder, fill=orangefill, rounded corners=5pt,
      minimum width=5.0cm, minimum height=1.1cm] (mlp) at (0, 4.8) {MLP};

\draw[->] (concat.north) -- (mlp.south);

\node[draw=orangeborder, fill=orangefill, rounded corners=3pt,
      minimum width=0.85cm, minimum height=0.6cm] (h1) at (-1.5, 5.85) {$H_1$};
\node[draw=orangeborder, fill=orangefill, rounded corners=3pt,
      minimum width=0.85cm, minimum height=0.6cm] (h2) at ( 0.0, 5.85) {$H_2$};
\node[draw=orangeborder, fill=orangefill, rounded corners=3pt,
      minimum width=0.85cm, minimum height=0.6cm] (hn) at ( 1.5, 5.85) {$H_n$};
\node at (0.78, 5.85) {\ldots};

\draw[->] ($(mlp.north)+(-1.5, 0)$) -- (h1.south);
\draw[->] ($(mlp.north)+( 0.0, 0)$) -- (h2.south);
\draw[->] ($(mlp.north)+( 1.5, 0)$) -- (hn.south);

\node (qe) at (0, 6.65)
  {\textbf{QE score} \textit{(for protocol \#2)}};
\draw[->] (h2.north) -- (qe.south);

\node[align=center, font=\fontsize{7}{8.4}\sffamily\itshape\selectfont]
  at (-3.3, 5.85) {QE Protocol\\Regression\\Heads};

\node[align=center, font=\fontsize{7}{8.4}\sffamily\itshape\selectfont]
  at (-3.3, 4.80) {(trainable)};

\node[align=center, font=\fontsize{7}{8.4}\sffamily\itshape\selectfont]
  at (-3.3, 3.50) {Concatenated\\Input};

\node[align=center, font=\fontsize{7}{8.4}\sffamily\itshape\selectfont]
  at (-3.3, 2.70) {Cross-lingual\\Embeddings};

\node[align=center, font=\fontsize{7}{8.4}\sffamily\itshape\selectfont]
  at (-3.3, 1.30) {(frozen)};

\end{tikzpicture}}
        \caption{Architecture of \blaser.}
        \label{fig:blaser3_meth}
    \end{subfigure}
    \hfill
    \begin{subfigure}[b]{0.49\textwidth}
        \centering
        \resizebox{\textwidth}{!}{\begin{tikzpicture}[
  font=\fontsize{7}{8.4}\sffamily\selectfont,
  >=stealth,
  thick,
]

\definecolor{purplefill}{RGB}{237,212,237}
\definecolor{purpleborder}{RGB}{170,100,160}
\definecolor{bluefill}{RGB}{210,232,250}
\definecolor{blueborder}{RGB}{86,160,220}
\definecolor{greenfill}{RGB}{178,235,215}
\definecolor{greenborder}{RGB}{2,158,115}
\definecolor{yellowfill}{RGB}{252,248,185}
\definecolor{yellowborder}{RGB}{150,145,0}

\node[draw=blueborder, fill=bluefill, line width=0.8pt, rounded corners=5pt,
      minimum width=2.1cm, minimum height=2.4cm]
  (bluebox) at (1.1, 1.80) {};

\node[draw=purpleborder, fill=purplefill, line width=0.8pt, rounded corners=5pt,
      minimum width=6.5cm, minimum height=0.75cm]
  (purplebar) at (3.55, 3.85) {};

\node[font=\fontsize{7}{8.4}\sffamily\selectfont]
  (src_real) at (1.1, 2.50) {source};
\node[font=\fontsize{7}{8.4}\sffamily\selectfont]
  (ref_real) at (1.1, 1.1) {reference};

\node[font=\fontsize{7}{8.4}\sffamily\selectfont]
  (src_synth)   at (1.10, 3.85) {source};
\node[font=\fontsize{7}{8.4}\sffamily\selectfont]
  (hyp_synth)   at (3.30, 3.85) {hypothesis};
\node[font=\fontsize{7}{8.4}\sffamily\selectfont]
  (score_synth) at (6, 3.85) {score};

\node[draw=greenborder, fill=greenfill, rounded corners=8pt,
      inner xsep=12pt, inner ysep=6pt]
  (mt) at (3.30, 2.50) {MT Model};

\node[draw=yellowborder, fill=yellowfill, rounded corners=8pt,
      inner xsep=9pt, inner ysep=6pt, align=center, yshift=0.2cm, xshift=-0.2cm]
  (qe) at (5.10, 1.60) {ref-based\\QE Model};

\node[font=\fontsize{7}{8.4}\sffamily\itshape\selectfont]
  at ($(purplebar.north)+(0, 0.35)$) {Synthetic QE example};

\node[align=center, font=\fontsize{7}{8.4}\sffamily\itshape\selectfont]
  at ($(bluebox.south)+(0, -0.50)$) {Real\\translation\\example};

\draw[->] (src_real.east) -- (mt.west);

\draw[->] (mt.north) -- (hyp_synth.south);

\draw[->, dashed] (src_real.north) -- (src_synth.south);

\coordinate (rightOfBlue) at ($(bluebox.east)+(0.20, 0)$);
\draw[->, dashed] (src_real.south) |- (qe.west);

\draw[->] (hyp_synth.east) -| (qe.north);

\draw[->] (ref_real.east) -| (qe.south);

\draw[->] (qe.east) -| (score_synth.south);

\end{tikzpicture}}
        \caption{Approach for synthetic QE example generation.}
        \label{fig:blaser3_synth}
    \end{subfigure}
    \caption{\blaser Methodology}
    \label{fig:blaser3}
\end{figure}

\paragraph{Task} QE data are triplets of ($\text{src}^{(x)}$, $\text{mt}^{(y)}$, $s^{(i)}$), where $\text{src}^{(x)}$ is the source text in language $x$, $\text{mt}^{(y)}$ is the translation hypothesis text in language $y$, and $s^{(i)} \in \mathbb{R}$ is the human annotation score for the pair ($\text{src}^{(x)}$, $\text{mt}^{(y)}$), according to an evaluation protocol $i$ (e.g. MQM, XSTS). The task of reference-free QE aims to learn a model to predict $s^{(i)}$ given the pair ($\text{src}^{(x)}$, $\text{mt}^{(y)}$).

\paragraph{Architecture} The architecture of \blaser is illustrated in \Cref{fig:blaser3_meth}. We use (frozen) \sonaromni to extract cross-lingual embeddings $\mathbf{e}^\text{src},\mathbf{e}^\text{mt} \in \mathbb{R}^{d_s}$ for the source and hypothesis, where $d_s$ is the dimensionality of the \sonaromni embedding space. The two individual embeddings, together with two element-wise interaction embeddings, are concatenated to obtain obtain $\mathbf{e}^\text{input} \in \mathbb{R}^{4d_s}$ as in:
\begin{equation}
    \mathbf{e}^\text{input} = \mathbf{e}^\text{src}; \mathbf{e}^\text{mt}; |\mathbf{e}^\text{src} - \mathbf{e}^\text{mt}|; \mathbf{e}^\text{src} \odot \mathbf{e}^\text{mt},
\end{equation}
where $;$ denotes horizontal concatenation, and $\odot$ denotes element-wise multiplication. The input $\mathbf{e}^\text{input}$ is passed through a two-layer MLP $[4d_s\rightarrow d_h \rightarrow d_o]$ to obtain an output embedding $\mathbf{c} \in \mathbb{R}^{d_o}$. According to the protocol of each example, the output embedding is routed through the corresponding regression head $\text{Head}_i$, which is a linear layer $d_o \rightarrow 1$ that predicts the QE score $\hat{s}^{(i)} \in \mathbb{R}$. The MLP and protocol head parameters are optimized with an MSE loss: $\mathcal{L}_\text{mse} = ||s^{(i)} - \hat{s}^{(i)}||^2$.

\paragraph{Synthetic Examples} QE datasets are not very multilingual nor diverse, usually covering a couple of directions and domains~\citep{blain-etal-2023-findings,zerva-etal-2024-findings}, with some exceptions specifically for the XSTS protocol~\cite{nllb-24}. Although our model can take advantage of multiple data sources due to its multi-tasking nature, ideally we would like to cover more languages, for which there is no QE data availability. Thus, we propose a synthetic data generation pipeline using MT models and reference-based QE (\Cref{fig:blaser3_synth}). We use a large corpus of translation data, which are tuples of ($\text{src}^{(x)}$, $\text{ref}^{(y)}$). For each example, we translate the source with $K$ MT models, to obtain a set of hypotheses $\{\text{mt}^{(y)}_k\}_{k=1}^K$ in language $y$. We then use a reference-based QE model, that takes as input a triplet of ($\text{src}^{(x)}$, $\text{ref}^{(y)}$, $\text{mt}^{(y)}_k$) to label it with a score $\bar{s}_k$. Finally, we disregard the reference that was used to label this example, and use the triples of $\{\text{src}^{(x)}, \text{mt}^{(y)}_k, \bar{s}_k\}_{k=1}^K$ as training examples for \blaser. The motivation is that we can obtain examples of diverse quality due to the multiple MT models used, and can be labeled reliably using the reference of the example, which we then discard. Our methodology here is akin to distilling a reference-based QE model into a reference-free one.

\subsubsection{Experimental Setup}

\paragraph{Data} Our training data is comprised from several different datasets, covering in total 6 protocols (DA, ESA, MQM, SQM, XSTS, XSTS+R+P) and 204 unique directions, amounting to 1.6M examples. We use a variety of data 
including, but not limited to, IndicMT-Eval~\citep{sai-b-etal-2023-indicmt}, DA data from MLQE-PQ~\citep{fomicheva-etal-2022-mlqe}%
and XSTS data from BLASER 2~\citep{dale-costa-jussa-2024-blaser}. Finally, we allocate 15 paragraphs of the development set of the XSTS+R+P data of \metbouquet\footnote{Using 77 annotated directions from Round 1 that were already available at the moment of \blaser experiments; the other 25 directions of Round 1 \metbouquet are not included in this section.} for training (around 80$\%$), while the rest is used for validation during training. We evaluate our models primarily on the test set of \metbouquet.%
The statistics of the training and test data are available on Table~\ref{tab:qe_training_data}.

For each protocol we aggregate duplicate examples (same source-hypothesis) by averaging their scores. To minimize cross-contamination, we explicitly remove all training examples where the tuple source-hypothesis also appears in our validation/test sets. All scores are normalized to 0-1 scale, by applying min-max normalization with according to the ranges for each protocol (e.g. 0-100 for DA, 1-5 for XSTS). MQM is scaled with a different formula to make the normalized score more uniform ($1-(\text{score}/25)^{0.5}$).

\begin{table}[H]
    \centering
    \resizebox{0.6\linewidth}{!}{%
    \begin{tabular}{lrr}
    \toprule
    \textbf{Protocol} & \textbf{Unique Directions} & \textbf{Examples (K)} \\
    \midrule
    \midrule
    \emph{Training}  &  & \\
    DA & 46 & 745 \\
    ESA & 9 & 72 \\
    MQM & 12 & 187 \\
    SQM & 15 & 150 \\
    XSTS & 121 & 422 \\
    XSTS+R+P & 77 & 30 \\
    \midrule
    \textbf{Combined} & \textbf{204} & \textbf{1,605} \\
    \midrule
    \midrule
    \emph{Test}  &  & \\
    XSTS+R+P (\metbouquet) & 77 & 63 \\
    \bottomrule
    \end{tabular}
    }
    \caption{Training data statistics by protocol, and combined. Examples are in thousands.}
    \label{tab:qe_training_data}
\end{table}

\paragraph{Synthetic Data} We sample 2M translation examples from our MT training data covering aRound 100 directions, tailored around the Met-BOUQuET directions.\footnote{In the future we plan to extend to all possible directions for which we have translation data.} We translate the source text with 5 different MT models: 2 variants of \sonaromni~\citep{sonaromni}, NLLB-3B~\citep{nllb-24}, Madlad-10B~\citep{kudugunta2023madlad}, and Gemma3-27B~\citep{gemma3}. We use the reference-based MetricX-24 to label the examples, thus resulting in 10M synthetic QE examples.

\paragraph{Architecture \& Training} The embeddings from \sonaromni have dimensionality $d_s=1024$, and thus the input to the MLP has dimensionality $4d=4096$. For the MLP we use $d_h=2048$ and $d_o=256$, with GeLU activations~\citep{hendrycks2023gaussianerrorlinearunits}. We have in total 7 QE regression heads, one for each of the 6 protocols, plus a separate one for the synthetic data. Since scores are normalized to 0-1 scale, we apply a sigmoid function to the regression logits. The total parameters of the model are 9M. We train with AdamW~(0.9, 0.98)~\citep{loshchilov2019decoupledweightdecayregularization} using a base learning rate of 1e-3, and a cosine annealing scheduler. We use a batch size of 1024 examples, and train for 40k steps. Dropout is set to 0.1 in input/output and 0.3 in the MLP. Training takes only 90 minutes on a single A100 GPU (using pre-extracted \sonaromni embeddings).

\paragraph{Evaluation} We use Spearman's rank correlation coefficient $\rho$ as our main evaluation metric. We pick the best checkpoint according to \metbouquet validation performance using the XSTS+R+P head and report results in \metbouquet test, using their corresponding regression heads.

\subsubsection{Results}

On Table~\ref{tab:blaser_results}, we compare our proposed method \blaser, with its predecessor BLASER 2~\citep{dale-costa-jussa-2024-blaser}, and two strong QE models from the literature, xCOMET-XL\footnote{\url{https://huggingface.co/Unbabel/XCOMET-XL}}~\cite{guerreiro-etal-2024-xcomet} and MetricX-24\footnote{\url{https://huggingface.co/google/metricx-24-hybrid-xl-v2p6}}~\cite{juraska-etal-2024-metricx} on the test sets of \metbouquet. For \metbouquet we report results on all the directions, and additionally on three subsets, depending on the use of English in source or target.

Our results show that \blaser surpasses previous models on multilingual QE on \metbouquet, on average achieving gains of +0.08 compared to MetricX-24, and +0.12 compared to xCOMET-XL. By specifically looking into the direction with non-English source (X$\rightarrow$Eng) and non-English target (Eng$\rightarrow$Y), we see that the improvements of \blaser can be attributed to better performance on the source-side. We hypothesize that cross-lingual generalization from the omnilingual embedding space is particularly strong on the source-side of QE, since it contains proper sentences. Contrary, the target-side of QE, naturally contains errors, making generalization from \sonaromni embeddings more difficult. %
Finally, our ablations show that using synthetic data is helpful, thus indicating that scaling-up domain/language coverage through synthetic data can lead to further improvements.

\begin{table}[H]
    \centering
    \begin{tabular}{lcccc}
        \toprule
         {\textbf{Model}} & \textbf{\Xeng} & \textbf{\engX} & \textbf{\XX} & \textbf{All}  \\
         & (17) & (19) & (41) & (77)  \\
        \midrule
        \emph{Previous Works}  & & & & \\
        BLASER 2 & 0.47 & 0.38 & 0.45 & 0.44  \\
        xCOMET-XL & 0.51 & 0.45 & 0.40 & 0.43 \\
        MetricX-24 & 0.52 & \textbf{0.47} & 0.45 & 0.47 \\
        \midrule
        \emph{Proposed}  & & & & \\
        \blaser & \textbf{0.65} & 0.46 & \textbf{0.55} & \textbf{0.55} \\
        \midrule
        \emph{Ablations} & & & & \\
        $\hookrightarrow$ w/o Error Pred. & 0.65 & 0.47 & 0.55 & 0.55\\
        \phantom{a} $\hookrightarrow$ w/o Synthetic Data & 0.62 & 0.46 & 0.53 & 0.53  \\
        \phantom{aa} $\hookrightarrow$ w/o Multi-tasking & 0.59 & 0.44 & 0.51 & 0.51 \\
        \bottomrule
    \end{tabular}
    \caption{Spearman's $\rho$ ($\uparrow$) on the \metbouquet (XSTS+R+P) test set. X indicates non-English source languages, and Y indicates non-English target languages. In \textbf{bold} is the best among the proposed \blaser and the three baselines. In parenthesis is the number of pairs in each group (note that we use a subset of \metbouquet from r1 annotations, available at the time of experimentation).}
    \label{tab:blaser_results}
\end{table}

\subsection{Metrics Benchmarking}
\label{sec:mtbenchmark}

\subsubsection{Automatic analysis}

Besides training and evaluating \blaser, we take advantage of the \metbouquet dataset to benchmark a wider set of popular and traditional MT metrics\footnote{We are not adding in our benchmarking LLM-based metrics since there is not an standard metric of this type and it would require an extensive amount of work on experimenting with prompt and models that we leave for further work.}. Consistently with section \ref{sec:blaser}, we evaluate each metric by the average of its Spearman correlations with human XSTS+R+P scores in each translation direction of the \metbouquet test set\footnote{note that for BLEU and \chrf, we are using their sentence-level versions}. 
We evaluate a set of lexical-based metrics (BLEU \citep{bleu}, \chrf \citep{chrf}, METEOR \citep{meteor}), a set of model-based (BLASER 2 \citep{dale-costa-jussa-2024-blaser}, BLEURT \citep{bleurt}, COMETKiWi \citep{rei-etal-2022-cometkiwi}, MetricX \citep{juraska-etal-2024-metricx}, SONAR \citep{duquenne2023sonarsentencelevelmultimodallanguageagnostic} and \sonaromni \citep{sonaromni}, xCOMET \citep{guerreiro-etal-2024-xcomet}) available and the newly proposed \blaser in previous section.\footnote{We did not include any LLM-based metrics, because the space of the models, prompt templates, and decision strategies is too large to explore in this small section. We defer this to future work.} 
For SONAR and \sonaromni, we are using cosine similarity of the translation embedding and source/reference embedding. For consistency between the reference-based and reference-free metrics, we drop the translation directions for which the reference translations are not yet available. Results are presented in Table \ref{tab:metrics_benchmark}. 

\begin{table}[H]
    \centering
    \begin{tabular}{l|ccc|ccc}
    \toprule
     & \multicolumn{3}{c|}{\textbf{Standalone}} & \multicolumn{3}{c}{\textbf{LID-adjusted}} \\
     \midrule
     & \textbf{src} & \textbf{ref} & \textbf{both} & \textbf{src} & \textbf{ref} & \textbf{both} \\
    BLEU & - & 0.318 & - & - & 0.328 & - \\
    ChrF++ & - & 0.414 & - & - & 0.404 & - \\
    METEOR & - & 0.332 & - & - & 0.336 & - \\
    \midrule
    BLASER2 & 0.368 & \textbf{0.492} & - & 0.431 & 0.477 & - \\
    BLEURT & - & 0.486 & - & - & 0.481 & - \\
    COMETKiWi & 0.349 & - & - & 0.399 & - & - \\
    MetricX & 0.370 & 0.453 & \textbf{0.481} & 0.419 & 0.485 & \textbf{0.505} \\
    SONAR & 0.359 & 0.485 & - & 0.429 & 0.471 & - \\
    \sonaromni & 0.431 & 0.465 & - & 0.486 & 0.475 & - \\
    xCOMET & 0.316 & 0.466 & 0.450 & 0.391 & 0.478 & 0.478 \\
    \midrule
    BLASER3 & \textbf{0.474} & 0.483 & - & \textbf{0.523} & \textbf{0.516} & - \\
    
    \bottomrule
    \end{tabular}
    \caption{Mean per-direction Spearman's $\rho$ on the test set of \metbouquet (XSTS+R+P) for several popular metrics, depending on whether they compare the translation with source, reference, or both, and whether they are adjusted with the GlotLID score. The best metric for each setting is in boldface.}
    \label{tab:metrics_benchmark}
\end{table}

One question addressed by this benchmark is about the role of source and reference information for translation evaluation. Consistently with the results of the WMT25 evaluation campaign \citep{lavie-etal-2025-findings}, we find that for metrics capable of using the source and the reference simultaneously, their combination typically works better than the source or the reference alone (except for the case of xCOMET without LID correction, where adding the sources actually hurts the performance).

Another issue, also reported by \citet{lavie-etal-2025-findings}, is that automatic metrics often cannot detect translations into a wrong target language (which is a catastrophic error), even despite seeing a reference in the correct language. We addressed this issue by multiplying each metric by the confidence of a GlotLID v3 model \citep{kargaran-etal-2023-glotlid} that the translation is in the target language.\footnote{To make it work, the base metric has to be put on a scale where 0 corresponds to the worst quality and some positive number corresponds to the best quality. MetricX violates this requirement, so before multiplying we rescaled it with a formula $MetricX_{adjusted}=1-(MetricX/25)^{0.5}$ which makes its scale comparable to the one of xCOMET.}
The right half of Table \ref{tab:metrics_benchmark} shows that this adjustment is highly beneficial for all reference-free metrics, as well as for the majority of model-based metrics that use reference.

Based on the above comparison, we recommend three strongest model-based metrics for evaluating the quality of highly multilingual translation: reference-free \blaser with LID adjustment, and, in case translation references are available, the versions of MetricX and xCOMET that use both source and reference and are also adjusted with LID. %

To understand how these comparisons depend on the difficulty of the source and target languages, we grouped all \metbouquet languages in two buckets: ``high'' (all ``truly high resource'' languages with at least 50M primary parallel sentences, as per our definition in \Cref{sec:resourcelevels}) and ``low'' (all other languages), and grouped our 144 directions accordingly, with 4 groups by source and target resource levels of roughly similar sizes. Table \ref{tab:metrics_benchmark_by_level} reports the results grouped by the combination of translation directions. Low-resourced languages make the automatic evaluation more difficult on the source and especially on the target side. \blaser turns out to be competitive for each group of directions, outperforming, on average, every other metric in each group. Comparing the metrics' signatures makes intuitive sense: LID adjustment improves the evaluation for translation into lower-resourced languages (where out-of-target translations usually occur), and reference-based metrics clearly outperform reference-free ones either when translating form a low-resourced languages to a high-resourced one (so that comparison to the high-resourced reference is easier than to the low-resourced source) or when evaluating translation into low-resourced languages without LID adjustment (when comparison with references can partially compensate for the lack of off-target detection). These results confirm the need to invest in improving metrics, hinting at more urgency to evaluate probably fluency of low-resource languages.

\begin{table}[H]
    \small
    \centering
    \begin{tabular}{ll|cccc}
    \toprule
    \multicolumn{2}{c|}{Best metrics per directions group} & high-high & low-high & high-low & low-low \\
    \midrule
    \multicolumn{2}{c|}{Number of directions} & 34 & 36 & 32 & 42 \\
    \midrule
    \multirow[p]{11}{*}{Best metric signature} 
     & BLEU & 0.368 & 0.349 & 0.346 & 0.262 \\
     & ChrF++ & 0.418 & 0.430 & 0.424 & 0.389 \\
     & METEOR & 0.382 & 0.371 & 0.340 & 0.265 \\
     & BLASER2 & 0.544 & 0.545 & 0.446 & 0.453 \\
     & BLEURT & 0.588 & 0.571 & 0.417 & 0.390 \\
     & COMETKiWi & 0.541 & 0.353 & 0.395 & 0.325 \\
     & MetricX & 0.599 & 0.549 & 0.466 & 0.435 \\
     & SONAR & 0.566 & 0.533 & 0.443 & 0.446 \\
     & \sonaromni & 0.576 & 0.544 & 0.435 & 0.453 \\
     & xCOMET & 0.601 & 0.559 & 0.447 & 0.385 \\
     & BLASER3 & \textbf{0.619} & \textbf{0.575} & \textbf{0.470} & \textbf{0.489} \\
     \midrule
    
    \multirow[c]{3}{*}{Best standalone metric}
        & src & \textbf{0.619} & 0.517 & 0.320 & 0.436 \\
        & ref & 0.588 & \textbf{0.575} & 0.446 & 0.453 \\
        & both & 0.601 & 0.549 & 0.380 & 0.403 \\
    \midrule
    \multirow[c]{3}{*}{Best LID-adjusted metric}
        & src & 0.615 & 0.520 & \textbf{0.470} & \textbf{0.489} \\
        & ref & 0.569 & 0.573 & 0.453 & 0.473 \\
        & both & 0.587 & 0.541 & 0.466 & 0.435 \\
    \bottomrule
    \end{tabular}
    \caption{Mean per-direction Spearman's $\rho$ on the test set of \metbouquet, aggregated per group of translation directions. Top part: for each metric, we report the best  correlation over the metric signatures (whether to use source, reference, and LID adjustment). Bottom part: for each signature, we report the best score over different metrics.}
    \label{tab:metrics_benchmark_by_level}
\end{table}

Examples of translation directions with the lowest correlations include French to Zarma, English to Plains Cree, and English to Egyptian Arabic. The first two simply contain very few good translations, and the third direction, while containing a substantial proportion of semantically similar translations, often includes penalties for translating into Modern Standard Arabic instead of Egyptian Arabic and for problems with fluency, which all automatic metrics fail to reflect. For a future generation of automatic quality metrics for translation, it would probably make sense to build some language identification capabilities directly into them.

\subsubsection{Manual analysis: automatic metrics vs human judgment}
\label{sec:manualanalysis}
Using a subset of Round 1 annotations, we performed a side-by-side analysis of translation quality, where we looked at how the automatic metrics matched human judgment (or not). Overall, automatic metrics correlate rather well with human judgment, but they are not good at capturing: paragraph-level discrepancies; the relative importance of salient words; language register discrepancies.

We want to find out where the biggest translation errors come from and how these evaluations compare to automatic metrics. We chose four language pairs (Swedish to English, English to Swedish, Italian to Romanian, Romanian to Italian) according to our internal capabilities. 

In most cases, the automatic metrics align with human judgment. It is especially apparent in such cases as obvious hallucinations and the opposite meaning of the target text. LID metrics successfully judge the presence of the wrong language in the translation. However,  there are additional factors that automatic tools cannot take into account. Primarily, these tools cannot work on the paragraph level, which results in inaccurate judgment when the context is needed for correct translation.

Secondly, automatic metrics do not always judge what words are key words in a sentence in the same way as a human does. Often, when only the key word is mistranslated (which leads to a completely different meaning and low human score) automatic metrics score the sentence significantly higher than expected, since “almost everything” in the sentence is correct.

Interestingly, when translating out of English, register problems are much more obvious.

\begin{table}[ht]
\centering
\small
\setlength{\tabcolsep}{5pt}
\begin{tabularx}{\linewidth}{@{}>{\raggedright\arraybackslash}p{0.27\linewidth} >{\raggedright\arraybackslash}p{0.27\linewidth} >{\raggedright\arraybackslash}X@{}}
\toprule
\textbf{Source text} & \textbf{Translated text} & \textbf{Comment} \\
\midrule
Åh! Ursäkta, men ta med den snälla. &
Oh! Excuse me, but take it kindly. &
This is the last sentence of a dialogue where one person forgot their purse at work and said to the other person “Oh! Sorry, but please take it with you” meaning “please take my purse with you”. Not knowing this is about a purse, even a human can misunderstand the sentence, and the literal translation is very close to what the MT here produced. \\
\addlinespace
En av gurkorna växte sig större. &
One of the tomatoes grew bigger. &
In this Swedish original, we see cucumbers, not tomatoes. According to our XSTS+R+P protocol guidelines, this should score 2 (crucial information is missing/mistranslated in the target), but the automatic metrics do not penalize in the same manner. \\
\addlinespace
Att köpa en kaffe för 5 euro på Starbucks varje dag kan kosta dig 1 300 euro per år. &
Buying a \$5 coffee at Starbucks every day can cost you \$1,300 a year. &
Though the meaning is clear, the localization is redundant and unnecessary, and the details are changed. \\
\addlinespace
Sorry man, I am going to Shirdi with my family. &
Ursäkta, jag ska till Shirdi med min familj. &
In this example, “man” is omitted completely and nothing conveys the informal address. That’s why this translation scored very low in human evaluation. However, the automatic metrics give a score much higher than expected. \\
\bottomrule
\end{tabularx}
\caption{Examples where automated metrics and human judgment do not align}
\label{tab:csv_examples}
\end{table}

All in all, it is clear that register and paragraph coherence represent challenges to automatic metrics. Besides that, these metrics tend to give higher scores to literal translation, which is not necessarily representative of how the translation should work.

\subsection{\omnitox}
\label{sec:omnitox}

To define toxicity we refer to previous works \citep{nllb-24,seamless2025} that consider toxicity as instances of profanity or language that may incite hate, violence or abuse against an individual or a group (such as a religion, race or gender). When detecting toxicity in MT, it is important to report toxicity unbalances between the source (or input) and the MT output. Toxicity unbalances can be of two kinds: deleted toxicity, where the output contains fewer toxic items than the source, and added toxicity, where the output contains more toxic items than the source. While both cases are critical, we focus here on added toxicity, following prior work by \cite{nllb-24} and \cite{seamless2025}. It has been our experience that consumers of MT regard added toxicity as more problematic than deleted toxicity, as exemplified in real situations \citep{garcia-gilabert-etal-2024-resetox}.%

There are many works in multilingual toxicity detection in NLP---e.g., \cite{kivlichan2020jigsaw}---and even going beyond multilingual toxicity detection in explainability and interpretability \citep{dementieva-etal-2025-multilingual}. However, there is a limited number of toxicity detectors that scale to the long-tail of languages; e.g., ETOX \citep{costa-jussa-etal-2023-toxicity} and MuTox \citep{costa-jussa-etal-2024-mutox}, both covering 200 languages.

In this section we describe \omnitox, a new toxicity classifier serving \TOTALlanguages languages. \omnitox achieves a mean per-language ROC AUC of 0.86, outperforming the previous state-of-the-art MuTox (0.80) by +0.06 points.
In particular, \omnitox shows strong zero-shot capabilities, when trained exclusively on English 
and Spanish, it achieves 0.82 mean per-language ROC AUC across 30 evaluation languages.

\subsubsection{Methodology}

\paragraph{Overview} \omnitox is a direct successor to MuTox~\citep{costa-jussa-etal-2024-mutox}, designed to extend multilingual toxicity detection from 200 to 1600+ languages. Rather than introducing architectural complexity, we focus on upgrading the underlying representation space: replacing SONAR embeddings with \sonaromni.

\paragraph{Task} Toxicity detection data are tuples of ($\text{sent}^{(x)}$, $y$), where $\text{sent}^{(x)}$ is a sentence in language $x$, and $y \in \{0, 1\}$ is the binary toxicity label (0 for non-toxic, 1 for toxic). The task of toxicity detection aims to learn a classifier $f$ that predicts $\hat{y} = f(\text{sent}^{(x)})$.

\paragraph{Architecture} %
Following MuTox~\citep{costa-jussa-etal-2024-mutox}, we employ a simple MLP classifier on top of  cross-lingual sentence embeddings. The key difference is the underlying encoder: we replace SONAR (200 languages) with \sonaromni (1600+ languages)~\citep{sonaromni}, which provides stronger multilingual representations and broader language coverage.
We use (frozen) \sonaromni to extract cross-lingual embeddings $e \in \mathbb{R}^{d_s}$ for each input sentence, where $d_s$ is the dimensionality of the \sonaromni embedding space. The embedding $e$ is passed through a two-layer MLP $[d_s \rightarrow d_1 \rightarrow d_2 \rightarrow 1]$ to obtain the toxicity prediction:
\begin{equation}
    h_1 = \text{ReLU}(W_1 e + b_1), \quad h_2 = \text{ReLU}(W_2 h_1 + b_2), \quad \hat{y} = \sigma(W_3 h_2 + b_3),
\end{equation}
where $W_1 \in \mathbb{R}^{d_1 \times d_s}$, $W_2 \in \mathbb{R}^{d_2 \times d_1}$, $W_3 \in \mathbb{R}^{1 \times d_2}$ are learnable weight matrices, $b_1, b_2, b_3$ are bias terms, and $\sigma$ denotes the sigmoid function. During training, we  apply dropout ($p=0.4$) after each hidden layer for regularization.

The MLP parameters are optimized with binary cross-entropy loss:
\begin{equation}
    \mathcal{L}_{\text{BCE}} = -\left[y \log \hat{y} + (1-y) \log(1-\hat{y})\right].
\end{equation}
Using a simple classifier on top of powerful cross-lingual embeddings is a deliberate design choice that facilitates zero-shot cross-lingual transfer, allowing the model to generalize to languages unseen during training.

\subsubsection{Experimental Setup}

\paragraph{Data} Our training data is comprised of a subset of the MuTox dataset~\citep{costa-jussa-etal-2024-mutox}, covering 30 languages with varying resource levels. We maintain the original data partitions: 'train' for model training, 'dev' for validation and hyperparameter tuning, and 'devtest' for final evaluation. To ensure data integrity, we exclude instances without explicit partition assignments. The dataset statistics are presented in Table~\ref{tab:mutox_ds_subset}.
English and Spanish serve as high-resource anchor languages, comprising approximately 40\% of the training data (12,906 and 10,716 examples, respectively). The remaining 28 languages contain between 887--1,476 training examples each. This imbalanced distribution allows us to evaluate both supervised performance on well-resourced languages and cross-lingual transfer to lower-resource languages. The full list of languages is provided in the MuTox paper \citep{costa-jussa-etal-2024-mutox}.

\begin{table}[H]
    \centering
    \begin{tabular}{@{}llrr@{}}
        \toprule
        \textbf{Partition} & \textbf{Languages} & \textbf{Size Range} & \textbf{Total} \\
        \midrule
        \multirow{3}{*}{Train} 
            & English & 12,906 & \multirow{3}{*}{60,194} \\
            & Spanish & 10,716 & \\
            & Other (28 langs) & 887--1,476 & \\
        \midrule
        \multirow{3}{*}{Dev}  
            & English & 894 & \multirow{3}{*}{19,488} \\
            & Spanish & 758 & \\
            & Other (28 langs) & 414--738 & \\
        \midrule
        \multirow{3}{*}{DevTest}  
            & English & 1,743 & \multirow{3}{*}{9,259} \\
            & Spanish & 1,546 & \\
            & Other (28 langs) & 137--245 & \\
        \bottomrule
    \end{tabular}
    \caption{Training and evaluation data statistics for \omnitox. We use a subset of the MuTox dataset covering 30 languages across three partitions.}
    \label{tab:mutox_ds_subset}
\end{table}

\paragraph{Architecture \& Training} The embeddings from \sonaromni have dimensionality $d_s = 1024$, serving as input to the MLP. For the MLP, we use $d_1 = 512$ and $d_2 = 128$, with ReLU activations. The total number of trainable parameters is approximately 590K. We train with Adam using a base learning rate of 1e-3, weight decay of 1e-3, and a CosineAnnealingLR scheduler. We use a batch size of 32 and train for 30 epochs. Dropout is set to 0.4 for regularization, and gradient clipping is applied at 5. We select the best checkpoint according to development set performance. Training takes approximately 5 minutes on an NVIDIA Quadro GV100 GPU (using pre-extracted \sonaromni embeddings).

\paragraph{Baselines} Our experimental design isolates the contribution of \sonaromni embeddings through controlled comparisons. We compare against MuTox~\citep{costa-jussa-etal-2024-mutox}, the previous state-of-the-art multilingual toxicity detector built on SONAR embeddings covering 200 languages. To disentangle the effects of the embedding space from classifier optimization,  we train two additional models:

\begin{itemize}
    \item Baseline: Uses \sonaromni embeddings with MuTox's original classifier architecture and hyperparameters. %
    Comparing MuTox vs Baseline isolates the impact of upgrading from SONAR to \sonaromni, holding the classifier constant.
    
    \item Baseline ZS: Identical to Baseline but trained exclusively on English and Spanish data, then evaluated on all 30 languages. This tests the zero-shot cross-lingual transfer capabilities enabled by \sonaromni embeddings.
 Comparison of Baseline ZS versus \omnitox shows the differences from classifier optimization (architecture, dropout, weight decay, learning rate tuning).
\end{itemize}

\paragraph{Evaluation Metrics} We use the Receiver Operating Characteristic Area Under the Curve (ROC AUC) as our primary evaluation metric. ROC AUC quantifies the classifier's ability to distinguish between toxic and non-toxic classes across all possible classification thresholds, providing a threshold-agnostic measure of ranking quality. We report: (1) overall ROC AUC computed on the aggregated DevTest set with 95\% confidence intervals via bootstrap resampling (1000 iterations); (2) mean ROC AUC averaged across individual languages; and (3) per-language ROC AUC range to assess cross-lingual consistency. This evaluation framework allows for subsequent threshold tuning tailored to specific use cases, potentially at the language level.

\subsubsection{Results}

\begin{figure}[H]
    \centering
    \includegraphics[width=1\linewidth]{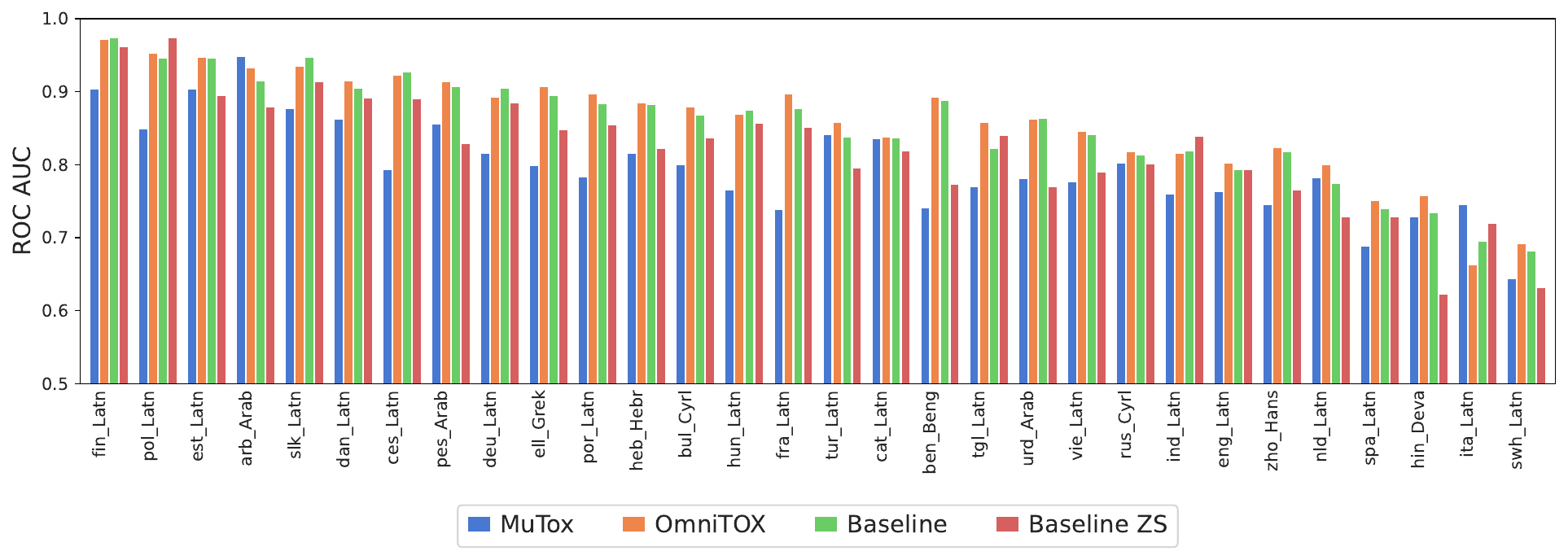}
    \caption{ROC AUC per language for MuTox and \omnitox. \omnitox outperforms MuTox on 28 of 30 languages.}
    \label{fig:auc_mutox_omnitox}
\end{figure}

Table~\ref{tab:omnitox_perf} presents the performance comparison between \omnitox, our baseline variants, and MuTox on the DevTest partition across 30 languages. We report overall ROC AUC on the aggregated test set, mean ROC AUC averaged across individual languages, and the per-language performance range. Figure~\ref{fig:auc_mutox_omnitox} provides a detailed per-language breakdown.

Our results show that \omnitox achieves an overall ROC AUC of 0.845, outperforming MuTox by +0.058 points. More importantly, our controlled experiments reveal that the majority of this improvement stems from upgrading the embedding space rather than classifier optimization.

\paragraph{Impact of Embedding Upgrade} Comparing MuTox to Baseline, which differ only in the underlying encoder (SONAR vs \sonaromni) while using identical classifier architecture and hyperparameters, we observe a gain of +0.052 ROC AUC. This accounts for 90\% of the total improvement over MuTox, validating our hypothesis that representation quality is the primary driver of cross-lingual toxicity detection performance. We attribute this to \sonaromni's improved cross-lingual alignment and broader language coverage, which yields more semantically coherent embeddings across diverse languages.

\paragraph{Impact of Classifier Optimization} Comparing Baseline to \omnitox, which differ only in classifier architecture and hyperparameters, we observe a modest additional gain of +0.006 ROC AUC. This confirms our design philosophy: when embeddings are sufficiently powerful, a simple classifier suffices, and architectural complexity yields diminishing returns.

\paragraph{Zero-Shot Cross-Lingual Transfer} Remarkably, Baseline ZS, trained exclusively on English and Spanish, achieves 0.793 overall ROC AUC, outperforming MuTox (0.787) despite using 28 fewer training languages. This demonstrates the exceptional zero-shot transfer capabilities of \sonaromni embeddings. The mean per-language ROC AUC of 0.821 for Baseline ZS compared to 0.798 for MuTox further confirms that \sonaromni's cross-lingual alignment enables effective generalization to unseen languages without explicit supervision.

\paragraph{Cross-Lingual Consistency} As shown in Figure~\ref{fig:auc_mutox_omnitox}, \omnitox achieves consistent improvements across the language spectrum. The per-language ROC AUC ranges from 0.664 to 0.972, with \omnitox outperforming MuTox on 28 of 30 languages. The mean per-language ROC AUC improves from 0.798 (MuTox) to 0.860 (\omnitox), a gain of +0.062 points.

\begin{table}[H]
    \centering
    \begin{tabular}{@{}lccc@{}}
        \toprule
        \textbf{Model} & \textbf{Overall} & \textbf{Mean per Lang.} & \textbf{Range per Lang.} \\
        & \textbf{ROC AUC} & \textbf{ROC AUC} & \textbf{ROC AUC} \\
        \midrule
        MuTox & 0.787$_{[0.775, 0.799]}$ & 0.798 & 0.645--0.949 \\
        \midrule
        \omnitox & \textbf{0.845}$_{[0.835, 0.856]}$ & \textbf{0.860} & 0.664--0.972 \\
        \midrule
        \multicolumn{4}{@{}l}{\textit{Ablations}} \\
        $\hookrightarrow$ w/o Classifier Tuning (Baseline) & 0.839$_{[0.829, 0.850]}$ & 0.854 & 0.682--0.975 \\
        $\hookrightarrow$ w/o Multilingual Training (Baseline ZS) & 0.793$_{[0.781, 0.807]}$ & 0.821 & 0.623--0.974 \\
        \bottomrule
    \end{tabular}
    \caption{Performance comparison on the DevTest partition (30 languages). Overall ROC AUC: computed on the aggregated test set with 95\% confidence intervals via bootstrap resampling (1000 iterations). Mean per Lang.: average of individual language ROC AUCs. Range per Lang.: minimum and maximum ROC AUC across languages. Baseline uses \sonaromni embeddings with MuTox hyperparameters; Baseline ZS is trained only on English and Spanish. Best results in bold.}
    \label{tab:omnitox_perf}
\end{table}

\paragraph{Summary} Our experiments confirm that \omnitox's improvements over MuTox are primarily driven by the upgrade from SONAR to \sonaromni embeddings. The strong zero-shot performance of Baseline ZS suggests that \sonaromni's cross-lingual representations can generalize effectively to the 1600+ languages beyond our 30 training languages, enabling truly omnilingual toxicity detection.

\paragraph{Limitations}

Our work has several limitations that suggest directions for future research. 
\subparagraph{Embedding Space Dependence} \omnitox inherits both the strengths and weaknesses of \sonaromni. While \sonaromni provides broad language coverage, its  representations may be weaker for languages with limited pre-training data. 
Additionally, \sonaromni was not explicitly trained to preserve toxicity-related semantic distinctions, which may limit fine-grained toxicity detection in some languages.

\subparagraph{Limited Language Evaluation} Although \sonaromni supports 1600+ languages, we evaluated \omnitox on only 30 languages due to the availability of annotated data. 
Performance on the remaining 1570+ languages relies entirely on zero-shot transfer, which remains unvalidated. Furthermore, our training data is imbalanced: English and Spanish comprise approximately 40\% of examples, potentially biasing the model toward Western toxicity norms.

\subparagraph{Annotation Subjectivity} Toxicity perception is inherently subjective and culturally dependent~\citep{costa-jussa-etal-2024-mutox}. Our training data may reflect annotator biases toward certain toxicity patterns, and culturally-specific 
forms of toxicity may be underrepresented.

\subparagraph{Task Scope} \omnitox performs sentence-level binary classification, which has two implications: (1) context-dependent toxicity spanning multiple sentences may be missed, and (2) the binary output does not capture toxicity severity or type (e.g., hate speech vs.\ profanity). These limitations may affect downstream applications requiring fine-grained toxicity analysis.

\section{MT Results}
\label{sec:evaluations}

This section reports evaluations of general translation quality and added toxicity for our final decoder-only and encoder-decoder models presented in previous sections.

\subsection{Automatic evaluation of translation quality}
\label{sec:quality}

\subsubsection{Evaluation framework}

We report results on evaluation datasets presented in section \ref{sec:MTevaldata} which includes Bible, \bouquet, and \floresplus.%
Throughout the paper, we have been presenting MT results with metrics that are reported in Appendix \ref{app:cards}. In particular, these metrics include lexical-based (BLEU \cite{bleu}, \chrf \cite{chrf}) and model-based (XCOMET \cite{guerreiro-etal-2024-xcomet}, MetricX \citep{juraska-etal-2024-metricx}). In this section, and based on findings from section \ref{sec:mtbenchmark}, we present results with a subset of them,  and a newly proposed reference-free and model-based metric, \blaser (Section \ref{sec:blaser}). We complement MetricX, xCOMET, and \blaser with LID to compensate for off-target mistakes of the metrics (as motivated in \Cref{sec:mtbenchmark}). As baseline systems, we report a variety of models of different sizes reported in Table \ref{tab:MTsystemsevaluated}.

\begin{table}[H]
    \centering
    \begin{tabular}{lccc}
        \toprule
        \textbf{System} & \textbf{Size(s)} & \textbf{Reference} & \textbf{Comment} \\
        \midrule
        \omtllama & 8B, 3B, 1B & Section \ref{sec:deconly} & Specialised\\
        \nllbtwo & 3B & Section \ref{sec:encdec} & Specialised \\
        \midrule
        Aya-101 & 13B & \citep{ustun2024aya}  & General\\
                Aya Expanse & 8B & \citep{dang2024ayaexpansecombiningresearch} & General  \\
     EuroLLM & 9B & \citep{martins2024eurollmmultilinguallanguagemodels} & General\\
        Gemma3 & 27B & \citep{gemma3} & General\\ 
        GPT-OSS\tablefootnote{We use these models in the ``low reasoning effort'' mode, to avoid the infinitely looped chains of thought that they otherwise sometimes produce, as well as to make the evaluation more comparable to other models that produce translations directly, without intermediate reasoning.} & 20B, 120B & \citep{openai2025gptoss120bgptoss20bmodel} & General \\
        Hunyuan-MT\tablefootnote{For fair comparison with other systems, we are using only the base model, without applying the ensemble.} & 7B & \citep{zheng2025hunyuanmttechnicalreport} & Specialised\\ 
       
        MADLAD & 10B & \citep{kudugunta2023madlad} & Specialised \\ 
         NLLB-200 & 3B & \citep{nllb-24} & Specialised\\
         \llama 3& 8B, 70B & \citep{grattafiori2024llama}  & General\\
        Tiny Aya Global & 3B & \citep{tiny_aya_unofficial} & General \\
        Tower+ & 72B & \citep{tower_llm_2024} & Specialised \\
        TranslateGemma & 27B & \citep{finkelstein2026translategemma} & Specialised \\
            \bottomrule
    \end{tabular}
    \caption{Proposed and external MT systems evaluated throughout this section and classified as General or Specialised (on MT).}
\label{tab:MTsystemsevaluated}
\end{table}

Some of these translation models come with a pre-defined translation prompt template (e.g. NLLB, MADLAD, TranslateGemma) which we had only to tweak to either extend the set of supported languages or to replace an unsupported language code with the ``most similar'' supported language.\footnote{In many cases, this amounts to simply replacing language tags with nearly-equivalent ones, e.g. for NLLB-200, we replaced \texttt{cmn\_Hans} with \texttt{zho\_Hans}, where the former means ``Mandarin Chinese'' and the latter simply ``Chinese'' (still implying Mandarin). However, for very low-resourced languages not supported for \nllbtwo, NLLB-200, or MADLAD, we had to substitute them with one language from the model's supported set, selected by genealogical proximity or, in its absence, by geographical one.} With most of the other models, we use the same minimalist prompt template (with the language names as described in Section \ref{sec:language_names}):

\noindent\begin{minipage}{\textwidth}
\small
\begin{tcolorbox}[]
\begin{verbatim}
  Translate the following text from {source language} into {target language}.
  Please write only its translation to {target language}, without any additional comments.
  Make sure that your response is a translation to {target language} and not the original text.
  {source language}: {source text}
  {target language}: 
\end{verbatim}
\end{tcolorbox}
\captionof{boxes}{Prompt template for translation with instruction-following models}
\label{eval:box:translation_prompt:generic}
\end{minipage}

For \omtllama models, we did not include the additional instructions (lines 2 and 3) into the prompt because these models are already trained to produce concise translations, unless explicitly requested otherwise.

\subsubsection{Evaluating with standard benchmarks}

\paragraph{Performance on \bouquet by the language resource level}
Table \ref{tab:result_bouquet_xx2en_by_resource} reports \chrf results on \bouquet dataset by language resource level defined as in Section \ref{sec:languages} and \ref{sec:benchmarkspe}. OMT models (both \omtllama and \nllbtwo) are very competitive for into-English translation from languages of any resource group. 
For translation into high- and mid-resourced languages, \nllbtwo is preferable, on average, while \omtllama shines for translation into low- and very-low-resourced languages (note that some of the lower-resourced \bouquet languages are not officially supported by \nllbtwo on the output side).

\begin{table}[H]
    \centering
    \footnotesize

\begin{tabular}{l|rrrrrr|rrrrrr|r}
\toprule
 & \multicolumn{6}{c|}{\textbf{\Xeng}} & \multicolumn{6}{c|}{\textbf{\engX} } & \multicolumn{1}{c}{\textbf{All} } \\
\textbf{ Resource level} & \textbf{high} & \textbf{mid} & \textbf{low} & \textbf{v. low} & \textbf{zero} & \textbf{total}& \textbf{high} & \textbf{mid} & \textbf{low} & \textbf{v. low} & \textbf{zero} & \textbf{total} & \textbf{total} \\
\midrule

OMT-LLaMA 8B & {\cellcolor[HTML]{023E62}} \color[HTML]{F1F1F1} 65.1 & {\cellcolor[HTML]{056BA9}} \color[HTML]{F1F1F1} 53.5 & {\cellcolor[HTML]{69A5CC}} \color[HTML]{F1F1F1} 38.7 & {\cellcolor[HTML]{B4C4DF}} \color[HTML]{000000} 27.6 & {\cellcolor[HTML]{D5D5E8}} \color[HTML]{000000} 21.6 & {\cellcolor[HTML]{589EC8}} \color[HTML]{F1F1F1} 40.6 & {\cellcolor[HTML]{045382}} \color[HTML]{F1F1F1} 60.7 & {\cellcolor[HTML]{2F8BBE}} \color[HTML]{F1F1F1} 45.8 & {\cellcolor[HTML]{A1BBDA}} \color[HTML]{000000} 30.8 & {\cellcolor[HTML]{E0DDED}} \color[HTML]{000000} 18.8 & {\cellcolor[HTML]{F4EEF6}} \color[HTML]{000000} 12.6 & {\cellcolor[HTML]{94B6D7}} \color[HTML]{000000} 32.8 & {\cellcolor[HTML]{78ABD0}} \color[HTML]{F1F1F1} 36.7 \\
OMT-NLLB & {\cellcolor[HTML]{03446A}} \color[HTML]{F1F1F1} 64.0 & {\cellcolor[HTML]{04639B}} \color[HTML]{F1F1F1} 56.1 & {\cellcolor[HTML]{73A9CF}} \color[HTML]{F1F1F1} 37.4 & {\cellcolor[HTML]{B0C2DE}} \color[HTML]{000000} 28.4 & {\cellcolor[HTML]{CED0E6}} \color[HTML]{000000} 23.2 & {\cellcolor[HTML]{529BC7}} \color[HTML]{F1F1F1} 41.3 & {\cellcolor[HTML]{03517E}} \color[HTML]{F1F1F1} 61.1 & {\cellcolor[HTML]{2383BA}} \color[HTML]{F1F1F1} 47.6 & {\cellcolor[HTML]{B7C5DF}} \color[HTML]{000000} 27.3 & {\cellcolor[HTML]{E5E1EF}} \color[HTML]{000000} 17.5 & {\cellcolor[HTML]{F7F0F7}} \color[HTML]{000000} 11.5 & {\cellcolor[HTML]{9AB8D8}} \color[HTML]{000000} 31.9 & {\cellcolor[HTML]{79ABD0}} \color[HTML]{F1F1F1} 36.6 \\
OMT-LLaMA 3B & {\cellcolor[HTML]{034267}} \color[HTML]{F1F1F1} 64.4 & {\cellcolor[HTML]{0568A3}} \color[HTML]{F1F1F1} 54.6 & {\cellcolor[HTML]{69A5CC}} \color[HTML]{F1F1F1} 38.8 & {\cellcolor[HTML]{B3C3DE}} \color[HTML]{000000} 28.0 & {\cellcolor[HTML]{D2D3E7}} \color[HTML]{000000} 22.3 & {\cellcolor[HTML]{549CC7}} \color[HTML]{F1F1F1} 41.0 & {\cellcolor[HTML]{04598C}} \color[HTML]{F1F1F1} 59.4 & {\cellcolor[HTML]{348EBF}} \color[HTML]{F1F1F1} 45.0 & {\cellcolor[HTML]{AFC1DD}} \color[HTML]{000000} 28.7 & {\cellcolor[HTML]{E7E3F0}} \color[HTML]{000000} 16.8 & {\cellcolor[HTML]{F7F0F7}} \color[HTML]{000000} 11.3 & {\cellcolor[HTML]{9EBAD9}} \color[HTML]{000000} 31.3 & {\cellcolor[HTML]{7DACD1}} \color[HTML]{F1F1F1} 36.2 \\
NLLB-200 & {\cellcolor[HTML]{034E7B}} \color[HTML]{F1F1F1} 61.5 & {\cellcolor[HTML]{056FAE}} \color[HTML]{F1F1F1} 52.4 & {\cellcolor[HTML]{99B8D8}} \color[HTML]{000000} 32.1 & {\cellcolor[HTML]{C1CAE2}} \color[HTML]{000000} 25.5 & {\cellcolor[HTML]{D1D2E6}} \color[HTML]{000000} 22.7 & {\cellcolor[HTML]{6FA7CE}} \color[HTML]{F1F1F1} 37.9 & {\cellcolor[HTML]{034973}} \color[HTML]{F1F1F1} 62.9 & {\cellcolor[HTML]{2786BB}} \color[HTML]{F1F1F1} 46.8 & {\cellcolor[HTML]{C6CCE3}} \color[HTML]{000000} 24.6 & {\cellcolor[HTML]{E5E1EF}} \color[HTML]{000000} 17.5 & {\cellcolor[HTML]{F2ECF5}} \color[HTML]{000000} 13.2 & {\cellcolor[HTML]{9EBAD9}} \color[HTML]{000000} 31.3 & {\cellcolor[HTML]{88B1D4}} \color[HTML]{000000} 34.6 \\
GPT-OSS 120B & {\cellcolor[HTML]{034165}} \color[HTML]{F1F1F1} 64.6 & {\cellcolor[HTML]{1278B4}} \color[HTML]{F1F1F1} 49.9 & {\cellcolor[HTML]{8EB3D5}} \color[HTML]{000000} 33.6 & {\cellcolor[HTML]{C1CAE2}} \color[HTML]{000000} 25.4 & {\cellcolor[HTML]{C8CDE4}} \color[HTML]{000000} 24.3 & {\cellcolor[HTML]{6DA6CD}} \color[HTML]{F1F1F1} 38.3 & {\cellcolor[HTML]{03466E}} \color[HTML]{F1F1F1} 63.6 & {\cellcolor[HTML]{3F93C2}} \color[HTML]{F1F1F1} 43.7 & {\cellcolor[HTML]{BDC8E1}} \color[HTML]{000000} 26.0 & {\cellcolor[HTML]{EBE6F2}} \color[HTML]{000000} 16.0 & {\cellcolor[HTML]{F2ECF5}} \color[HTML]{000000} 13.4 & {\cellcolor[HTML]{A1BBDA}} \color[HTML]{000000} 30.8 & {\cellcolor[HTML]{88B1D4}} \color[HTML]{000000} 34.5 \\
Gemma 3 & {\cellcolor[HTML]{023D60}} \color[HTML]{F1F1F1} 65.3 & {\cellcolor[HTML]{0872B1}} \color[HTML]{F1F1F1} 51.3 & {\cellcolor[HTML]{86B0D3}} \color[HTML]{000000} 34.8 & {\cellcolor[HTML]{B5C4DF}} \color[HTML]{000000} 27.4 & {\cellcolor[HTML]{C6CCE3}} \color[HTML]{000000} 24.6 & {\cellcolor[HTML]{63A2CB}} \color[HTML]{F1F1F1} 39.4 & {\cellcolor[HTML]{034A74}} \color[HTML]{F1F1F1} 62.6 & {\cellcolor[HTML]{569DC8}} \color[HTML]{F1F1F1} 40.9 & {\cellcolor[HTML]{C9CEE4}} \color[HTML]{000000} 24.0 & {\cellcolor[HTML]{EEE8F3}} \color[HTML]{000000} 14.8 & {\cellcolor[HTML]{F7F0F7}} \color[HTML]{000000} 11.6 & {\cellcolor[HTML]{ADC1DD}} \color[HTML]{000000} 28.9 & {\cellcolor[HTML]{8BB2D4}} \color[HTML]{000000} 34.1 \\
TranslateGemma & {\cellcolor[HTML]{03456C}} \color[HTML]{F1F1F1} 63.6 & {\cellcolor[HTML]{0C74B2}} \color[HTML]{F1F1F1} 50.9 & {\cellcolor[HTML]{84B0D3}} \color[HTML]{F1F1F1} 35.0 & {\cellcolor[HTML]{B4C4DF}} \color[HTML]{000000} 27.7 & {\cellcolor[HTML]{C4CBE3}} \color[HTML]{000000} 24.9 & {\cellcolor[HTML]{63A2CB}} \color[HTML]{F1F1F1} 39.3 & {\cellcolor[HTML]{04588A}} \color[HTML]{F1F1F1} 59.5 & {\cellcolor[HTML]{4C99C5}} \color[HTML]{F1F1F1} 42.0 & {\cellcolor[HTML]{D2D3E7}} \color[HTML]{000000} 22.1 & {\cellcolor[HTML]{EDE7F2}} \color[HTML]{000000} 15.5 & {\cellcolor[HTML]{F4EEF6}} \color[HTML]{000000} 12.4 & {\cellcolor[HTML]{AFC1DD}} \color[HTML]{000000} 28.5 & {\cellcolor[HTML]{8CB3D5}} \color[HTML]{000000} 33.9 \\
OMT-LLaMA 1B & {\cellcolor[HTML]{034871}} \color[HTML]{F1F1F1} 63.0 & {\cellcolor[HTML]{0872B1}} \color[HTML]{F1F1F1} 51.5 & {\cellcolor[HTML]{86B0D3}} \color[HTML]{000000} 34.7 & {\cellcolor[HTML]{C9CEE4}} \color[HTML]{000000} 24.1 & {\cellcolor[HTML]{DBDAEB}} \color[HTML]{000000} 19.9 & {\cellcolor[HTML]{6FA7CE}} \color[HTML]{F1F1F1} 37.9 & {\cellcolor[HTML]{04629A}} \color[HTML]{F1F1F1} 56.5 & {\cellcolor[HTML]{4A98C5}} \color[HTML]{F1F1F1} 42.2 & {\cellcolor[HTML]{C4CBE3}} \color[HTML]{000000} 25.1 & {\cellcolor[HTML]{EEE8F3}} \color[HTML]{000000} 14.9 & {\cellcolor[HTML]{F5EEF6}} \color[HTML]{000000} 12.3 & {\cellcolor[HTML]{ACC0DD}} \color[HTML]{000000} 29.1 & {\cellcolor[HTML]{8FB4D6}} \color[HTML]{000000} 33.5 \\
LLaMA 3 70B & {\cellcolor[HTML]{023E62}} \color[HTML]{F1F1F1} 65.0 & {\cellcolor[HTML]{2383BA}} \color[HTML]{F1F1F1} 47.6 & {\cellcolor[HTML]{8FB4D6}} \color[HTML]{000000} 33.3 & {\cellcolor[HTML]{BDC8E1}} \color[HTML]{000000} 26.2 & {\cellcolor[HTML]{C9CEE4}} \color[HTML]{000000} 24.0 & {\cellcolor[HTML]{73A9CF}} \color[HTML]{F1F1F1} 37.6 & {\cellcolor[HTML]{045585}} \color[HTML]{F1F1F1} 60.2 & {\cellcolor[HTML]{75A9CF}} \color[HTML]{F1F1F1} 37.2 & {\cellcolor[HTML]{CCCFE5}} \color[HTML]{000000} 23.7 & {\cellcolor[HTML]{EAE6F1}} \color[HTML]{000000} 16.2 & {\cellcolor[HTML]{F0EAF4}} \color[HTML]{000000} 14.3 & {\cellcolor[HTML]{B1C2DE}} \color[HTML]{000000} 28.2 & {\cellcolor[HTML]{93B5D6}} \color[HTML]{000000} 32.9 \\
MADLAD & {\cellcolor[HTML]{023F64}} \color[HTML]{F1F1F1} 64.7 & {\cellcolor[HTML]{2182B9}} \color[HTML]{F1F1F1} 47.8 & {\cellcolor[HTML]{A1BBDA}} \color[HTML]{000000} 31.0 & {\cellcolor[HTML]{C8CDE4}} \color[HTML]{000000} 24.3 & {\cellcolor[HTML]{D9D8EA}} \color[HTML]{000000} 20.5 & {\cellcolor[HTML]{7DACD1}} \color[HTML]{F1F1F1} 36.0 & {\cellcolor[HTML]{034B76}} \color[HTML]{F1F1F1} 62.2 & {\cellcolor[HTML]{81AED2}} \color[HTML]{F1F1F1} 35.5 & {\cellcolor[HTML]{D5D5E8}} \color[HTML]{000000} 21.5 & {\cellcolor[HTML]{EDE8F3}} \color[HTML]{000000} 15.3 & {\cellcolor[HTML]{F7F0F7}} \color[HTML]{000000} 11.5 & {\cellcolor[HTML]{BBC7E0}} \color[HTML]{000000} 26.6 & {\cellcolor[HTML]{9EBAD9}} \color[HTML]{000000} 31.3 \\
GPT-OSS 20B & {\cellcolor[HTML]{034A74}} \color[HTML]{F1F1F1} 62.5 & {\cellcolor[HTML]{2685BB}} \color[HTML]{F1F1F1} 47.0 & {\cellcolor[HTML]{A2BCDA}} \color[HTML]{000000} 30.7 & {\cellcolor[HTML]{CED0E6}} \color[HTML]{000000} 23.2 & {\cellcolor[HTML]{D2D2E7}} \color[HTML]{000000} 22.4 & {\cellcolor[HTML]{80AED2}} \color[HTML]{F1F1F1} 35.7 & {\cellcolor[HTML]{045A8D}} \color[HTML]{F1F1F1} 59.2 & {\cellcolor[HTML]{78ABD0}} \color[HTML]{F1F1F1} 36.7 & {\cellcolor[HTML]{DAD9EA}} \color[HTML]{000000} 20.2 & {\cellcolor[HTML]{F2ECF5}} \color[HTML]{000000} 13.1 & {\cellcolor[HTML]{F5EEF6}} \color[HTML]{000000} 12.1 & {\cellcolor[HTML]{BDC8E1}} \color[HTML]{000000} 26.1 & {\cellcolor[HTML]{A1BBDA}} \color[HTML]{000000} 30.9 \\
Aya-101 & {\cellcolor[HTML]{04588A}} \color[HTML]{F1F1F1} 59.6 & {\cellcolor[HTML]{1C7FB8}} \color[HTML]{F1F1F1} 48.3 & {\cellcolor[HTML]{A1BBDA}} \color[HTML]{000000} 30.9 & {\cellcolor[HTML]{C1CAE2}} \color[HTML]{000000} 25.5 & {\cellcolor[HTML]{D0D1E6}} \color[HTML]{000000} 23.0 & {\cellcolor[HTML]{7BACD1}} \color[HTML]{F1F1F1} 36.3 & {\cellcolor[HTML]{056DAC}} \color[HTML]{F1F1F1} 52.8 & {\cellcolor[HTML]{8EB3D5}} \color[HTML]{000000} 33.7 & {\cellcolor[HTML]{E8E4F0}} \color[HTML]{000000} 16.6 & {\cellcolor[HTML]{F7F0F7}} \color[HTML]{000000} 11.4 & {\cellcolor[HTML]{F8F1F8}} \color[HTML]{000000} 10.9 & {\cellcolor[HTML]{CED0E6}} \color[HTML]{000000} 23.1 & {\cellcolor[HTML]{A8BEDC}} \color[HTML]{000000} 29.7 \\
Tower+ & {\cellcolor[HTML]{023858}} \color[HTML]{F1F1F1} 66.5 & {\cellcolor[HTML]{4295C3}} \color[HTML]{F1F1F1} 43.3 & {\cellcolor[HTML]{9CB9D9}} \color[HTML]{000000} 31.7 & {\cellcolor[HTML]{C2CBE2}} \color[HTML]{000000} 25.2 & {\cellcolor[HTML]{C6CCE3}} \color[HTML]{000000} 24.4 & {\cellcolor[HTML]{7DACD1}} \color[HTML]{F1F1F1} 36.0 & {\cellcolor[HTML]{045B8E}} \color[HTML]{F1F1F1} 58.9 & {\cellcolor[HTML]{B9C6E0}} \color[HTML]{000000} 26.9 & {\cellcolor[HTML]{EDE8F3}} \color[HTML]{000000} 15.2 & {\cellcolor[HTML]{F7F0F7}} \color[HTML]{000000} 11.3 & {\cellcolor[HTML]{FAF2F8}} \color[HTML]{000000} 10.3 & {\cellcolor[HTML]{D6D6E9}} \color[HTML]{000000} 21.3 & {\cellcolor[HTML]{AFC1DD}} \color[HTML]{000000} 28.7 \\
Tiny Aya Global & {\cellcolor[HTML]{034973}} \color[HTML]{F1F1F1} 62.7 & {\cellcolor[HTML]{69A5CC}} \color[HTML]{F1F1F1} 38.1 & {\cellcolor[HTML]{C1CAE2}} \color[HTML]{000000} 24.6 & {\cellcolor[HTML]{DCDAEB}} \color[HTML]{000000} 18.8 & {\cellcolor[HTML]{D5D5E8}} \color[HTML]{000000} 20.6 & {\cellcolor[HTML]{9EBAD9}} \color[HTML]{000000} 30.5 & {\cellcolor[HTML]{045B8F}} \color[HTML]{F1F1F1} 58.5 & {\cellcolor[HTML]{A5BDDB}} \color[HTML]{000000} 29.5 & {\cellcolor[HTML]{EFE9F3}} \color[HTML]{000000} 13.3 & {\cellcolor[HTML]{FAF3F9}} \color[HTML]{000000} 9.0 & {\cellcolor[HTML]{F7F0F7}} \color[HTML]{000000} 10.3 & {\cellcolor[HTML]{D3D4E7}} \color[HTML]{000000} 21.0 & {\cellcolor[HTML]{BBC7E0}} \color[HTML]{000000} 25.8 \\
LLaMA 3 8B & {\cellcolor[HTML]{045280}} \color[HTML]{F1F1F1} 61.0 & {\cellcolor[HTML]{60A1CA}} \color[HTML]{F1F1F1} 39.8 & {\cellcolor[HTML]{B5C4DF}} \color[HTML]{000000} 27.5 & {\cellcolor[HTML]{D5D5E8}} \color[HTML]{000000} 21.6 & {\cellcolor[HTML]{D9D8EA}} \color[HTML]{000000} 20.6 & {\cellcolor[HTML]{99B8D8}} \color[HTML]{000000} 32.1 & {\cellcolor[HTML]{056DAC}} \color[HTML]{F1F1F1} 52.7 & {\cellcolor[HTML]{BFC9E1}} \color[HTML]{000000} 26.0 & {\cellcolor[HTML]{F2ECF5}} \color[HTML]{000000} 13.1 & {\cellcolor[HTML]{FCF4FA}} \color[HTML]{000000} 9.5 & {\cellcolor[HTML]{FFF7FB}} \color[HTML]{000000} 8.2 & {\cellcolor[HTML]{DFDDEC}} \color[HTML]{000000} 19.1 & {\cellcolor[HTML]{C0C9E2}} \color[HTML]{000000} 25.6 \\
Hunyuan-MT & {\cellcolor[HTML]{046299}} \color[HTML]{F1F1F1} 56.6 & {\cellcolor[HTML]{8CB3D5}} \color[HTML]{000000} 33.9 & {\cellcolor[HTML]{C6CCE3}} \color[HTML]{000000} 24.4 & {\cellcolor[HTML]{DAD9EA}} \color[HTML]{000000} 20.3 & {\cellcolor[HTML]{D8D7E9}} \color[HTML]{000000} 21.0 & {\cellcolor[HTML]{ACC0DD}} \color[HTML]{000000} 29.0 & {\cellcolor[HTML]{2182B9}} \color[HTML]{F1F1F1} 47.8 & {\cellcolor[HTML]{D5D5E8}} \color[HTML]{000000} 21.5 & {\cellcolor[HTML]{F1EBF4}} \color[HTML]{000000} 13.9 & {\cellcolor[HTML]{F4EEF6}} \color[HTML]{000000} 12.4 & {\cellcolor[HTML]{F8F1F8}} \color[HTML]{000000} 10.8 & {\cellcolor[HTML]{E0DEED}} \color[HTML]{000000} 18.5 & {\cellcolor[HTML]{CACEE5}} \color[HTML]{000000} 23.8 \\
\bottomrule
\end{tabular}

    \caption{Translation performance on \bouquet (\chrf) by the non-English language resource level.}
    \label{tab:result_bouquet_xx2en_by_resource}
\end{table}

\paragraph{Performance on \floresplus} To corroborate our \bouquet evaluation results, we report similar evaluation numbers based on \floresplus in \Cref{tab:result_flores_xx2en_by_resource}. Similarly to \bouquet, on \floresplus, OMT systems perform comparably to strong baselines like Tower+ for translation between high-resource languages and English, and outperform them by a significant margin when it comes to mid- and low-resourced languages.

\begin{table}[H]
    \centering
    \footnotesize

\begin{tabular}{l|rrrrr|rrrrr|r}
\toprule
 & \multicolumn{5}{c}{\textbf{\Xeng}} & \multicolumn{5}{c}{\textbf{\engX}} & \multicolumn{1}{c}{\textbf{All}} \\
\textbf{ Resource level} & \textbf{high} & \textbf{mid} & \textbf{low} & \textbf{v. low} & \textbf{total}& \textbf{high} & \textbf{mid} & \textbf{low} & \textbf{v. low}  & \textbf{total} & \textbf{total} \\
\midrule

OMT-NLLB & {\cellcolor[HTML]{023858}} \color[HTML]{F1F1F1} 65.0 & {\cellcolor[HTML]{045C90}} \color[HTML]{F1F1F1} 58.3 & {\cellcolor[HTML]{056AA6}} \color[HTML]{F1F1F1} 54.6 & {\cellcolor[HTML]{0A73B2}} \color[HTML]{F1F1F1} 52.3 & {\cellcolor[HTML]{046097}} \color[HTML]{F1F1F1} 57.3 & {\cellcolor[HTML]{045F95}} \color[HTML]{F1F1F1} 57.6 & {\cellcolor[HTML]{3F93C2}} \color[HTML]{F1F1F1} 46.0 & {\cellcolor[HTML]{71A8CE}} \color[HTML]{F1F1F1} 41.0 & {\cellcolor[HTML]{A7BDDB}} \color[HTML]{000000} 34.6 & {\cellcolor[HTML]{4E9AC6}} \color[HTML]{F1F1F1} 44.6 & {\cellcolor[HTML]{157AB5}} \color[HTML]{F1F1F1} 50.9 \\
OMT-LLaMA 8B & {\cellcolor[HTML]{023A5B}} \color[HTML]{F1F1F1} 64.6 & {\cellcolor[HTML]{04639B}} \color[HTML]{F1F1F1} 56.4 & {\cellcolor[HTML]{187CB6}} \color[HTML]{F1F1F1} 50.6 & {\cellcolor[HTML]{2A88BC}} \color[HTML]{F1F1F1} 48.3 & {\cellcolor[HTML]{0569A5}} \color[HTML]{F1F1F1} 54.6 & {\cellcolor[HTML]{046096}} \color[HTML]{F1F1F1} 57.4 & {\cellcolor[HTML]{3B92C1}} \color[HTML]{F1F1F1} 46.5 & {\cellcolor[HTML]{7DACD1}} \color[HTML]{F1F1F1} 39.8 & {\cellcolor[HTML]{ACC0DD}} \color[HTML]{000000} 33.8 & {\cellcolor[HTML]{509AC6}} \color[HTML]{F1F1F1} 44.2 & {\cellcolor[HTML]{2182B9}} \color[HTML]{F1F1F1} 49.4 \\
NLLB-200 & {\cellcolor[HTML]{034165}} \color[HTML]{F1F1F1} 63.5 & {\cellcolor[HTML]{0567A2}} \color[HTML]{F1F1F1} 55.3 & {\cellcolor[HTML]{157AB5}} \color[HTML]{F1F1F1} 50.9 & {\cellcolor[HTML]{2383BA}} \color[HTML]{F1F1F1} 49.3 & {\cellcolor[HTML]{056BA9}} \color[HTML]{F1F1F1} 54.2 & {\cellcolor[HTML]{046097}} \color[HTML]{F1F1F1} 57.2 & {\cellcolor[HTML]{509AC6}} \color[HTML]{F1F1F1} 44.3 & {\cellcolor[HTML]{80AED2}} \color[HTML]{F1F1F1} 39.4 & {\cellcolor[HTML]{B5C4DF}} \color[HTML]{000000} 32.5 & {\cellcolor[HTML]{5EA0CA}} \color[HTML]{F1F1F1} 43.0 & {\cellcolor[HTML]{2786BB}} \color[HTML]{F1F1F1} 48.6 \\
Gemma 3 & {\cellcolor[HTML]{023C5F}} \color[HTML]{F1F1F1} 64.2 & {\cellcolor[HTML]{0A73B2}} \color[HTML]{F1F1F1} 52.3 & {\cellcolor[HTML]{3B92C1}} \color[HTML]{F1F1F1} 46.4 & {\cellcolor[HTML]{2C89BD}} \color[HTML]{F1F1F1} 48.1 & {\cellcolor[HTML]{1077B4}} \color[HTML]{F1F1F1} 51.4 & {\cellcolor[HTML]{045B8E}} \color[HTML]{F1F1F1} 58.6 & {\cellcolor[HTML]{79ABD0}} \color[HTML]{F1F1F1} 40.1 & {\cellcolor[HTML]{BDC8E1}} \color[HTML]{000000} 31.5 & {\cellcolor[HTML]{DBDAEB}} \color[HTML]{000000} 26.4 & {\cellcolor[HTML]{8BB2D4}} \color[HTML]{000000} 38.1 & {\cellcolor[HTML]{4C99C5}} \color[HTML]{F1F1F1} 44.8 \\
TranslateGemma & {\cellcolor[HTML]{03466E}} \color[HTML]{F1F1F1} 62.5 & {\cellcolor[HTML]{1379B5}} \color[HTML]{F1F1F1} 51.2 & {\cellcolor[HTML]{4496C3}} \color[HTML]{F1F1F1} 45.5 & {\cellcolor[HTML]{308CBE}} \color[HTML]{F1F1F1} 47.5 & {\cellcolor[HTML]{197DB7}} \color[HTML]{F1F1F1} 50.4 & {\cellcolor[HTML]{0567A1}} \color[HTML]{F1F1F1} 55.4 & {\cellcolor[HTML]{80AED2}} \color[HTML]{F1F1F1} 39.5 & {\cellcolor[HTML]{C4CBE3}} \color[HTML]{000000} 30.5 & {\cellcolor[HTML]{D6D6E9}} \color[HTML]{000000} 27.5 & {\cellcolor[HTML]{93B5D6}} \color[HTML]{000000} 37.2 & {\cellcolor[HTML]{569DC8}} \color[HTML]{F1F1F1} 43.8 \\
GPT-OSS 120B & {\cellcolor[HTML]{03466E}} \color[HTML]{F1F1F1} 62.5 & {\cellcolor[HTML]{2081B9}} \color[HTML]{F1F1F1} 49.5 & {\cellcolor[HTML]{5EA0CA}} \color[HTML]{F1F1F1} 42.9 & {\cellcolor[HTML]{4C99C5}} \color[HTML]{F1F1F1} 44.7 & {\cellcolor[HTML]{2987BC}} \color[HTML]{F1F1F1} 48.5 & {\cellcolor[HTML]{045D92}} \color[HTML]{F1F1F1} 58.2 & {\cellcolor[HTML]{73A9CF}} \color[HTML]{F1F1F1} 40.9 & {\cellcolor[HTML]{BCC7E1}} \color[HTML]{000000} 31.7 & {\cellcolor[HTML]{CDD0E5}} \color[HTML]{000000} 29.2 & {\cellcolor[HTML]{84B0D3}} \color[HTML]{F1F1F1} 38.8 & {\cellcolor[HTML]{589EC8}} \color[HTML]{F1F1F1} 43.6 \\
LLaMA 3 70B & {\cellcolor[HTML]{023C5F}} \color[HTML]{F1F1F1} 64.1 & {\cellcolor[HTML]{1E80B8}} \color[HTML]{F1F1F1} 49.8 & {\cellcolor[HTML]{4C99C5}} \color[HTML]{F1F1F1} 44.8 & {\cellcolor[HTML]{4094C3}} \color[HTML]{F1F1F1} 45.7 & {\cellcolor[HTML]{2081B9}} \color[HTML]{F1F1F1} 49.5 & {\cellcolor[HTML]{046299}} \color[HTML]{F1F1F1} 56.7 & {\cellcolor[HTML]{8CB3D5}} \color[HTML]{000000} 37.9 & {\cellcolor[HTML]{C0C9E2}} \color[HTML]{000000} 31.1 & {\cellcolor[HTML]{DBDAEB}} \color[HTML]{000000} 26.3 & {\cellcolor[HTML]{96B6D7}} \color[HTML]{000000} 36.8 & {\cellcolor[HTML]{5C9FC9}} \color[HTML]{F1F1F1} 43.2 \\
MADLAD & {\cellcolor[HTML]{034369}} \color[HTML]{F1F1F1} 63.1 & {\cellcolor[HTML]{4496C3}} \color[HTML]{F1F1F1} 45.4 & {\cellcolor[HTML]{569DC8}} \color[HTML]{F1F1F1} 43.8 & {\cellcolor[HTML]{509AC6}} \color[HTML]{F1F1F1} 44.3 & {\cellcolor[HTML]{348EBF}} \color[HTML]{F1F1F1} 47.1 & {\cellcolor[HTML]{046097}} \color[HTML]{F1F1F1} 57.2 & {\cellcolor[HTML]{ABBFDC}} \color[HTML]{000000} 34.1 & {\cellcolor[HTML]{BDC8E1}} \color[HTML]{000000} 31.4 & {\cellcolor[HTML]{CCCFE5}} \color[HTML]{000000} 29.3 & {\cellcolor[HTML]{9FBAD9}} \color[HTML]{000000} 35.7 & {\cellcolor[HTML]{6DA6CD}} \color[HTML]{F1F1F1} 41.4 \\
GPT-OSS 20B & {\cellcolor[HTML]{034D79}} \color[HTML]{F1F1F1} 61.2 & {\cellcolor[HTML]{348EBF}} \color[HTML]{F1F1F1} 47.1 & {\cellcolor[HTML]{7BACD1}} \color[HTML]{F1F1F1} 39.9 & {\cellcolor[HTML]{75A9CF}} \color[HTML]{F1F1F1} 40.7 & {\cellcolor[HTML]{4094C3}} \color[HTML]{F1F1F1} 45.8 & {\cellcolor[HTML]{0569A5}} \color[HTML]{F1F1F1} 54.8 & {\cellcolor[HTML]{ACC0DD}} \color[HTML]{000000} 33.9 & {\cellcolor[HTML]{E1DFED}} \color[HTML]{000000} 25.0 & {\cellcolor[HTML]{E6E2EF}} \color[HTML]{000000} 24.1 & {\cellcolor[HTML]{B5C4DF}} \color[HTML]{000000} 32.6 & {\cellcolor[HTML]{81AED2}} \color[HTML]{F1F1F1} 39.2 \\
Aya-101 & {\cellcolor[HTML]{045C90}} \color[HTML]{F1F1F1} 58.2 & {\cellcolor[HTML]{2987BC}} \color[HTML]{F1F1F1} 48.5 & {\cellcolor[HTML]{67A4CC}} \color[HTML]{F1F1F1} 42.1 & {\cellcolor[HTML]{589EC8}} \color[HTML]{F1F1F1} 43.5 & {\cellcolor[HTML]{348EBF}} \color[HTML]{F1F1F1} 47.1 & {\cellcolor[HTML]{2987BC}} \color[HTML]{F1F1F1} 48.5 & {\cellcolor[HTML]{B8C6E0}} \color[HTML]{000000} 32.3 & {\cellcolor[HTML]{EBE6F2}} \color[HTML]{000000} 22.9 & {\cellcolor[HTML]{F4EDF6}} \color[HTML]{000000} 20.3 & {\cellcolor[HTML]{C8CDE4}} \color[HTML]{000000} 29.9 & {\cellcolor[HTML]{88B1D4}} \color[HTML]{000000} 38.5 \\
Tower+ & {\cellcolor[HTML]{023858}} \color[HTML]{F1F1F1} 65.0 & {\cellcolor[HTML]{4094C3}} \color[HTML]{F1F1F1} 45.8 & {\cellcolor[HTML]{5EA0CA}} \color[HTML]{F1F1F1} 43.1 & {\cellcolor[HTML]{4094C3}} \color[HTML]{F1F1F1} 45.8 & {\cellcolor[HTML]{308CBE}} \color[HTML]{F1F1F1} 47.4 & {\cellcolor[HTML]{056AA6}} \color[HTML]{F1F1F1} 54.5 & {\cellcolor[HTML]{D7D6E9}} \color[HTML]{000000} 27.3 & {\cellcolor[HTML]{EDE7F2}} \color[HTML]{000000} 22.6 & {\cellcolor[HTML]{EBE6F2}} \color[HTML]{000000} 22.9 & {\cellcolor[HTML]{D0D1E6}} \color[HTML]{000000} 28.8 & {\cellcolor[HTML]{8BB2D4}} \color[HTML]{000000} 38.1 \\
LLaMA 3 8B & {\cellcolor[HTML]{034E7B}} \color[HTML]{F1F1F1} 61.0 & {\cellcolor[HTML]{529BC7}} \color[HTML]{F1F1F1} 44.1 & {\cellcolor[HTML]{84B0D3}} \color[HTML]{F1F1F1} 38.9 & {\cellcolor[HTML]{7DACD1}} \color[HTML]{F1F1F1} 39.7 & {\cellcolor[HTML]{529BC7}} \color[HTML]{F1F1F1} 44.1 & {\cellcolor[HTML]{1077B4}} \color[HTML]{F1F1F1} 51.4 & {\cellcolor[HTML]{D6D6E9}} \color[HTML]{000000} 27.5 & {\cellcolor[HTML]{F0EAF4}} \color[HTML]{000000} 21.7 & {\cellcolor[HTML]{F2ECF5}} \color[HTML]{000000} 20.6 & {\cellcolor[HTML]{D4D4E8}} \color[HTML]{000000} 27.9 & {\cellcolor[HTML]{9CB9D9}} \color[HTML]{000000} 36.0 \\
Tiny Aya Global & {\cellcolor[HTML]{034C78}} \color[HTML]{F1F1F1} 61.3 & {\cellcolor[HTML]{7DACD1}} \color[HTML]{F1F1F1} 39.9 & {\cellcolor[HTML]{A8BEDC}} \color[HTML]{000000} 34.6 & {\cellcolor[HTML]{9FBAD9}} \color[HTML]{000000} 35.6 & {\cellcolor[HTML]{78ABD0}} \color[HTML]{F1F1F1} 40.4 & {\cellcolor[HTML]{056DAC}} \color[HTML]{F1F1F1} 53.6 & {\cellcolor[HTML]{D5D5E8}} \color[HTML]{000000} 27.6 & {\cellcolor[HTML]{F6EFF7}} \color[HTML]{000000} 19.5 & {\cellcolor[HTML]{F5EFF6}} \color[HTML]{000000} 19.7 & {\cellcolor[HTML]{D6D6E9}} \color[HTML]{000000} 27.4 & {\cellcolor[HTML]{ACC0DD}} \color[HTML]{000000} 33.9 \\
Hunyuan-MT & {\cellcolor[HTML]{046097}} \color[HTML]{F1F1F1} 57.3 & {\cellcolor[HTML]{8BB2D4}} \color[HTML]{000000} 38.1 & {\cellcolor[HTML]{9FBAD9}} \color[HTML]{000000} 35.5 & {\cellcolor[HTML]{8EB3D5}} \color[HTML]{000000} 37.7 & {\cellcolor[HTML]{7DACD1}} \color[HTML]{F1F1F1} 39.7 & {\cellcolor[HTML]{4295C3}} \color[HTML]{F1F1F1} 45.6 & {\cellcolor[HTML]{EEE8F3}} \color[HTML]{000000} 22.2 & {\cellcolor[HTML]{F3EDF5}} \color[HTML]{000000} 20.5 & {\cellcolor[HTML]{E8E4F0}} \color[HTML]{000000} 23.4 & {\cellcolor[HTML]{E2DFEE}} \color[HTML]{000000} 24.9 & {\cellcolor[HTML]{B8C6E0}} \color[HTML]{000000} 32.3 \\
Aya Expanse & {\cellcolor[HTML]{04598C}} \color[HTML]{F1F1F1} 59.0 & {\cellcolor[HTML]{A5BDDB}} \color[HTML]{000000} 34.8 & {\cellcolor[HTML]{A8BEDC}} \color[HTML]{000000} 34.5 & {\cellcolor[HTML]{89B1D4}} \color[HTML]{000000} 38.2 & {\cellcolor[HTML]{89B1D4}} \color[HTML]{000000} 38.3 & {\cellcolor[HTML]{2C89BD}} \color[HTML]{F1F1F1} 48.0 & {\cellcolor[HTML]{EEE8F3}} \color[HTML]{000000} 22.3 & {\cellcolor[HTML]{F4EEF6}} \color[HTML]{000000} 20.1 & {\cellcolor[HTML]{EFE9F3}} \color[HTML]{000000} 21.8 & {\cellcolor[HTML]{E2DFEE}} \color[HTML]{000000} 24.9 & {\cellcolor[HTML]{BCC7E1}} \color[HTML]{000000} 31.6 \\
EuroLLM & {\cellcolor[HTML]{03466E}} \color[HTML]{F1F1F1} 62.5 & {\cellcolor[HTML]{ADC1DD}} \color[HTML]{000000} 33.7 & {\cellcolor[HTML]{ABBFDC}} \color[HTML]{000000} 34.2 & {\cellcolor[HTML]{89B1D4}} \color[HTML]{000000} 38.2 & {\cellcolor[HTML]{89B1D4}} \color[HTML]{000000} 38.2 & {\cellcolor[HTML]{0566A0}} \color[HTML]{F1F1F1} 55.7 & {\cellcolor[HTML]{FFF7FB}} \color[HTML]{000000} 16.6 & {\cellcolor[HTML]{F8F1F8}} \color[HTML]{000000} 18.8 & {\cellcolor[HTML]{EDE8F3}} \color[HTML]{000000} 22.3 & {\cellcolor[HTML]{EAE6F1}} \color[HTML]{000000} 23.2 & {\cellcolor[HTML]{C2CBE2}} \color[HTML]{000000} 30.7 \\

\bottomrule
\end{tabular}

    \caption{Translation performance on \floresplus (\chrf) by the non-English language resource level.}
    \label{tab:result_flores_xx2en_by_resource}
\end{table}

\paragraph{Relative comparison of non-English centric performance.} 
For each \bouquet non-English language, we select at least one other high- or mid-resource non-English ``proxy language'' based on their geographical and cultural proximity (see Section \ref{sec:languages}). Most usually, it amounts to pairing a lower-resourced language with a high-resource language spoken in the same country (majority language or a local lingua franca): for example, Spanish gets paired with Catalan and Basque languages spoken in Spain, as well as with many native American languages from Spanish-speaking countries such as Mexico. Each pair is evaluated in both directions.\footnote{This results in 514 distinct non-English-centric directions: less than two times 274 (the number of non-English languages in \bouquet), because some high- or mid-resourced languages are paired to each other symmetrically, reducing the total number of unordered pairs.} 
We evaluate how different systems compare on this set of directions, as well as how the \omtllama 8B performance for each language depends on whether it is paired with English or not.

We compare the systems ranking in different types of directions in \ref{tab:nonenglishcentric}. The two metrics we report (reference-based ChrF++ and reference-free \blaser + glotlid combination) mostly agree with each other, and most systems are ranked similarly regardless of the direction. One interesting outlier is NLLB-200 which ranks relatively high in \engX directions compared to other systems, which is probably a sign that all other systems (including the OMT ones) are still underinvesting into generation of diverse languages.

\begin{table}[H]
    \centering

\footnotesize

\begin{tabular}{l|cccc|cccc}
\toprule
Metric & \multicolumn{4}{c|}{\textbf{ChrF++}} & \multicolumn{4}{c}{\textbf{BLASER3+GlotLID}} \\
\textbf{Direction} & \textbf{XX-En} & \textbf{En-YY} & \textbf{XX-Proxy} & \textbf{Proxy-YY} & \textbf{XX-En} & \textbf{En-YY} & \textbf{XX-Proxy} & \textbf{Proxy-YY} \\
\midrule
OMT-LLaMA 8B & {\cellcolor[HTML]{03466E}} \color[HTML]{F1F1F1} 40.6 & {\cellcolor[HTML]{023858}} \color[HTML]{F1F1F1} 32.8 & {\cellcolor[HTML]{023858}} \color[HTML]{F1F1F1} 35.8 & {\cellcolor[HTML]{023858}} \color[HTML]{F1F1F1} 30.0 & {\cellcolor[HTML]{045E93}} \color[HTML]{F1F1F1} 0.47 & {\cellcolor[HTML]{023858}} \color[HTML]{F1F1F1} 0.32 & {\cellcolor[HTML]{034F7D}} \color[HTML]{F1F1F1} 0.40 & {\cellcolor[HTML]{023B5D}} \color[HTML]{F1F1F1} 0.32 \\
OMT-NLLB & {\cellcolor[HTML]{023858}} \color[HTML]{F1F1F1} 41.3 & {\cellcolor[HTML]{034973}} \color[HTML]{F1F1F1} 31.9 & {\cellcolor[HTML]{023F64}} \color[HTML]{F1F1F1} 35.5 & {\cellcolor[HTML]{023E62}} \color[HTML]{F1F1F1} 29.7 & {\cellcolor[HTML]{023858}} \color[HTML]{F1F1F1} 0.49 & {\cellcolor[HTML]{023858}} \color[HTML]{F1F1F1} 0.32 & {\cellcolor[HTML]{023858}} \color[HTML]{F1F1F1} 0.42 & {\cellcolor[HTML]{023858}} \color[HTML]{F1F1F1} 0.32 \\
OMT-LLaMA 3B & {\cellcolor[HTML]{023C5F}} \color[HTML]{F1F1F1} 41.0 & {\cellcolor[HTML]{045483}} \color[HTML]{F1F1F1} 31.3 & {\cellcolor[HTML]{034C78}} \color[HTML]{F1F1F1} 34.8 & {\cellcolor[HTML]{04598C}} \color[HTML]{F1F1F1} 28.4 & {\cellcolor[HTML]{046198}} \color[HTML]{F1F1F1} 0.46 & {\cellcolor[HTML]{034267}} \color[HTML]{F1F1F1} 0.31 & {\cellcolor[HTML]{045E93}} \color[HTML]{F1F1F1} 0.39 & {\cellcolor[HTML]{034973}} \color[HTML]{F1F1F1} 0.31 \\
GPT-OSS 120B & {\cellcolor[HTML]{056FAE}} \color[HTML]{F1F1F1} 38.3 & {\cellcolor[HTML]{045C90}} \color[HTML]{F1F1F1} 30.8 & {\cellcolor[HTML]{045E94}} \color[HTML]{F1F1F1} 33.8 & {\cellcolor[HTML]{03517E}} \color[HTML]{F1F1F1} 28.8 & {\cellcolor[HTML]{2484BA}} \color[HTML]{F1F1F1} 0.43 & {\cellcolor[HTML]{045788}} \color[HTML]{F1F1F1} 0.29 & {\cellcolor[HTML]{046096}} \color[HTML]{F1F1F1} 0.38 & {\cellcolor[HTML]{034F7D}} \color[HTML]{F1F1F1} 0.30 \\
Gemma 3 & {\cellcolor[HTML]{045F95}} \color[HTML]{F1F1F1} 39.4 & {\cellcolor[HTML]{0F76B3}} \color[HTML]{F1F1F1} 28.9 & {\cellcolor[HTML]{046097}} \color[HTML]{F1F1F1} 33.7 & {\cellcolor[HTML]{0570B0}} \color[HTML]{F1F1F1} 26.7 & {\cellcolor[HTML]{0D75B3}} \color[HTML]{F1F1F1} 0.44 & {\cellcolor[HTML]{0771B1}} \color[HTML]{F1F1F1} 0.26 & {\cellcolor[HTML]{0567A1}} \color[HTML]{F1F1F1} 0.37 & {\cellcolor[HTML]{056BA9}} \color[HTML]{F1F1F1} 0.27 \\
NLLB-200 & {\cellcolor[HTML]{0F76B3}} \color[HTML]{F1F1F1} 37.9 & {\cellcolor[HTML]{045483}} \color[HTML]{F1F1F1} 31.3 & {\cellcolor[HTML]{1278B4}} \color[HTML]{F1F1F1} 32.0 & {\cellcolor[HTML]{0566A0}} \color[HTML]{F1F1F1} 27.5 & {\cellcolor[HTML]{0771B1}} \color[HTML]{F1F1F1} 0.44 & {\cellcolor[HTML]{034871}} \color[HTML]{F1F1F1} 0.30 & {\cellcolor[HTML]{056FAE}} \color[HTML]{F1F1F1} 0.36 & {\cellcolor[HTML]{04588A}} \color[HTML]{F1F1F1} 0.29 \\
LLaMA 3 70B & {\cellcolor[HTML]{187CB6}} \color[HTML]{F1F1F1} 37.6 & {\cellcolor[HTML]{2182B9}} \color[HTML]{F1F1F1} 28.2 & {\cellcolor[HTML]{187CB6}} \color[HTML]{F1F1F1} 31.8 & {\cellcolor[HTML]{1379B5}} \color[HTML]{F1F1F1} 26.2 & {\cellcolor[HTML]{3D93C2}} \color[HTML]{F1F1F1} 0.42 & {\cellcolor[HTML]{2C89BD}} \color[HTML]{F1F1F1} 0.24 & {\cellcolor[HTML]{1C7FB8}} \color[HTML]{F1F1F1} 0.35 & {\cellcolor[HTML]{2987BC}} \color[HTML]{F1F1F1} 0.24 \\
OMT-LLaMA 1B & {\cellcolor[HTML]{0F76B3}} \color[HTML]{F1F1F1} 37.9 & {\cellcolor[HTML]{0872B1}} \color[HTML]{F1F1F1} 29.1 & {\cellcolor[HTML]{2D8ABD}} \color[HTML]{F1F1F1} 31.1 & {\cellcolor[HTML]{358FC0}} \color[HTML]{F1F1F1} 25.1 & {\cellcolor[HTML]{2A88BC}} \color[HTML]{F1F1F1} 0.43 & {\cellcolor[HTML]{045C90}} \color[HTML]{F1F1F1} 0.29 & {\cellcolor[HTML]{1C7FB8}} \color[HTML]{F1F1F1} 0.35 & {\cellcolor[HTML]{056AA6}} \color[HTML]{F1F1F1} 0.27 \\
GPT-OSS 20B & {\cellcolor[HTML]{5C9FC9}} \color[HTML]{F1F1F1} 35.7 & {\cellcolor[HTML]{65A3CB}} \color[HTML]{F1F1F1} 26.1 & {\cellcolor[HTML]{589EC8}} \color[HTML]{F1F1F1} 29.8 & {\cellcolor[HTML]{589EC8}} \color[HTML]{F1F1F1} 24.2 & {\cellcolor[HTML]{7DACD1}} \color[HTML]{F1F1F1} 0.39 & {\cellcolor[HTML]{328DBF}} \color[HTML]{F1F1F1} 0.24 & {\cellcolor[HTML]{4496C3}} \color[HTML]{F1F1F1} 0.32 & {\cellcolor[HTML]{2D8ABD}} \color[HTML]{F1F1F1} 0.24 \\
Aya-101 & {\cellcolor[HTML]{4697C4}} \color[HTML]{F1F1F1} 36.3 & {\cellcolor[HTML]{B9C6E0}} \color[HTML]{000000} 23.1 & {\cellcolor[HTML]{5C9FC9}} \color[HTML]{F1F1F1} 29.8 & {\cellcolor[HTML]{9EBAD9}} \color[HTML]{000000} 22.0 & {\cellcolor[HTML]{4697C4}} \color[HTML]{F1F1F1} 0.41 & {\cellcolor[HTML]{8BB2D4}} \color[HTML]{000000} 0.20 & {\cellcolor[HTML]{4094C3}} \color[HTML]{F1F1F1} 0.32 & {\cellcolor[HTML]{79ABD0}} \color[HTML]{F1F1F1} 0.20 \\
MADLAD & {\cellcolor[HTML]{4E9AC6}} \color[HTML]{F1F1F1} 36.0 & {\cellcolor[HTML]{529BC7}} \color[HTML]{F1F1F1} 26.6 & {\cellcolor[HTML]{83AFD3}} \color[HTML]{F1F1F1} 28.6 & {\cellcolor[HTML]{8BB2D4}} \color[HTML]{000000} 22.7 & {\cellcolor[HTML]{81AED2}} \color[HTML]{F1F1F1} 0.39 & {\cellcolor[HTML]{2C89BD}} \color[HTML]{F1F1F1} 0.24 & {\cellcolor[HTML]{76AAD0}} \color[HTML]{F1F1F1} 0.30 & {\cellcolor[HTML]{5A9EC9}} \color[HTML]{F1F1F1} 0.21 \\
Tower+ & {\cellcolor[HTML]{4E9AC6}} \color[HTML]{F1F1F1} 36.0 & {\cellcolor[HTML]{DEDCEC}} \color[HTML]{000000} 21.3 & {\cellcolor[HTML]{79ABD0}} \color[HTML]{F1F1F1} 28.9 & {\cellcolor[HTML]{C2CBE2}} \color[HTML]{000000} 20.6 & {\cellcolor[HTML]{8FB4D6}} \color[HTML]{000000} 0.38 & {\cellcolor[HTML]{C4CBE3}} \color[HTML]{000000} 0.16 & {\cellcolor[HTML]{86B0D3}} \color[HTML]{000000} 0.29 & {\cellcolor[HTML]{B7C5DF}} \color[HTML]{000000} 0.16 \\
LLaMA 3 8B & {\cellcolor[HTML]{D0D1E6}} \color[HTML]{000000} 32.1 & {\cellcolor[HTML]{F9F2F8}} \color[HTML]{000000} 19.1 & {\cellcolor[HTML]{D2D3E7}} \color[HTML]{000000} 25.6 & {\cellcolor[HTML]{E5E1EF}} \color[HTML]{000000} 18.9 & {\cellcolor[HTML]{EAE6F1}} \color[HTML]{000000} 0.33 & {\cellcolor[HTML]{C2CBE2}} \color[HTML]{000000} 0.16 & {\cellcolor[HTML]{D1D2E6}} \color[HTML]{000000} 0.23 & {\cellcolor[HTML]{C6CCE3}} \color[HTML]{000000} 0.15 \\
Tiny Aya Global & {\cellcolor[HTML]{ECE7F2}} \color[HTML]{000000} 30.5 & {\cellcolor[HTML]{E1DFED}} \color[HTML]{000000} 21.0 & {\cellcolor[HTML]{DFDDEC}} \color[HTML]{000000} 24.9 & {\cellcolor[HTML]{E2DFEE}} \color[HTML]{000000} 19.0 & {\cellcolor[HTML]{EDE8F3}} \color[HTML]{000000} 0.32 & {\cellcolor[HTML]{B5C4DF}} \color[HTML]{000000} 0.17 & {\cellcolor[HTML]{CDD0E5}} \color[HTML]{000000} 0.24 & {\cellcolor[HTML]{C6CCE3}} \color[HTML]{000000} 0.15 \\
Hunyuan-MT & {\cellcolor[HTML]{FFF7FB}} \color[HTML]{000000} 29.0 & {\cellcolor[HTML]{FFF7FB}} \color[HTML]{000000} 18.5 & {\cellcolor[HTML]{FFF7FB}} \color[HTML]{000000} 22.4 & {\cellcolor[HTML]{FFF7FB}} \color[HTML]{000000} 16.8 & {\cellcolor[HTML]{FFF7FB}} \color[HTML]{000000} 0.30 & {\cellcolor[HTML]{FFF7FB}} \color[HTML]{000000} 0.10 & {\cellcolor[HTML]{FFF7FB}} \color[HTML]{000000} 0.17 & {\cellcolor[HTML]{FFF7FB}} \color[HTML]{000000} 0.08 \\
\bottomrule
\end{tabular}

    \caption{Systems ranking on \bouquet depending on the direction type.\label{tab:nonenglishcentric}}
\end{table}

\begin{figure}[H]
    \centering
    \includegraphics[scale=1]{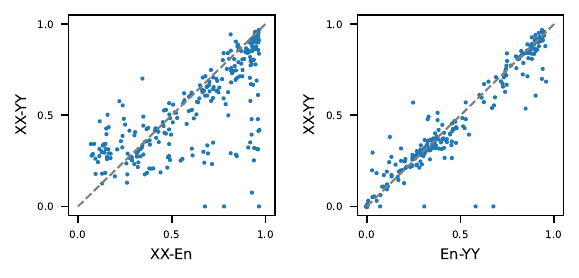}
    \caption{Translation quality (LID-normalized xCOMET) of the \omtllama model out of (left) and into (right) each language paired with English (horizontal axis) and non-English proxy languages (vertical axis). \label{fig:non_eng_eval}}

\end{figure}

To evaluate the role of the proxy languages, for each non-English source language, we compare the quality of translating it with \omtllama 8B into English and into its non-English proxy language. We do similarly for the target languages paired with either English or non-English sources. The results are presented in Figure \ref{fig:non_eng_eval}. For out-of-a-language translation into either English or a non-English language (left pane), a large part of the distribution is below the diagonal: translation into a non-English language is often harder than into English.\footnote{There is a cluster of exceptions, though: some directions like Amis-Chinese, Aguaruna-Spanish, Baatonum-French look better than their into-English equivalents, either because these language pairs have more training data, or simply because of the bias in the evaluation metrics, which are generally not intended for comparisons between different target languages and might be less exigent for non-English target languages.}. But for into-a-language translation out of either English or non-English languages (right pane), the distribution is mostly diagonal, indicating that the source language does not affect difficulty as much. In other words, the difficulty of non-English-centric pairs for \omtllama is more often driven by a non-English target than by a non-English source language.

For a non-negligible proportion of languages, quality from and into non-English language is quite high, which back-up the idea that is worth to explore non-English translations. Further research is needed on how to pair languages and experimenting with them.

\subsubsection{Evaluating long-tail understanding and generation}

\begin{figure}[H]
    \centering
    \includegraphics[width=1\linewidth]{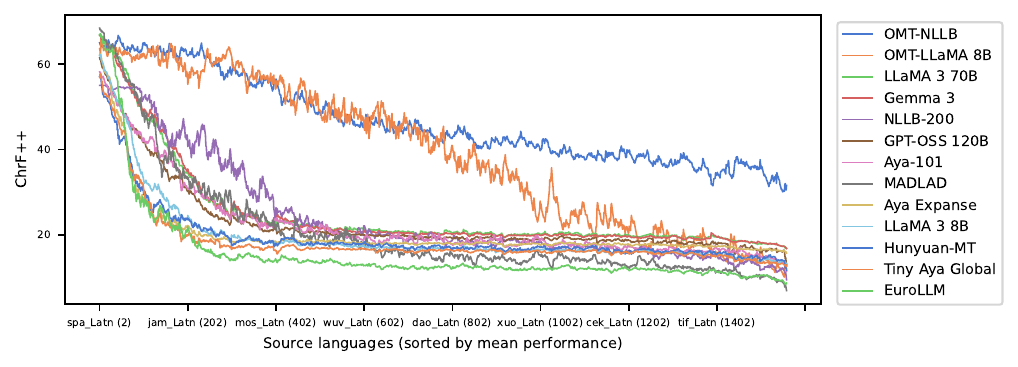}
    \includegraphics[width=1\linewidth]{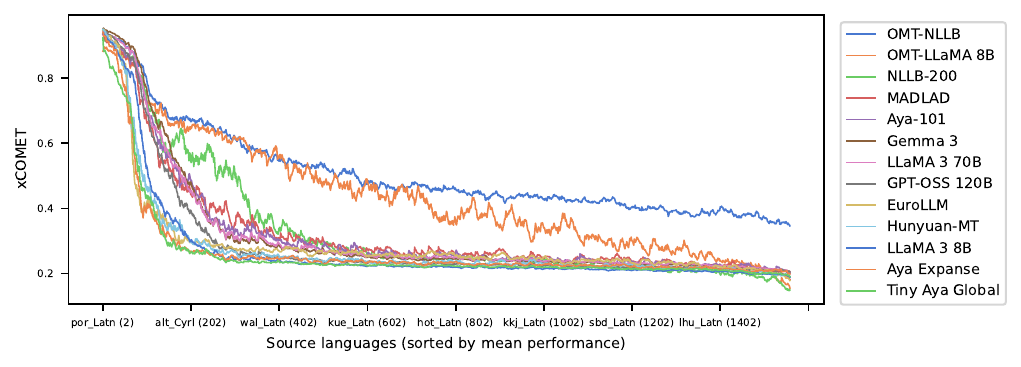}
    \caption{Relative performance in terms of \chrf (top) and XCOMET (bottom) of OMT models in understanding the longtail in the Bible domain compared to external baselines. Languages sorted by average performance of the models; curves smoothed with exponential moving average. %
    \label{fig:xx2englongtail}}

\end{figure}

\paragraph{Relative Performance on language understanding in the long tail} 
We compare our models to open models included in Table \ref{tab:MTsystemsevaluated}. 
Figure \ref{fig:xx2englongtail} shows how the OMT models understand the longtail of languages compared to baseline systems on the Bible evaluation benchmark (test split) in terms of \chrf and \xcomet for \Xeng. The number of languages where \omtllama 8B strictly outperforms all the baselines in the Bible is 1045 (which is approximately 2/3 of the languages). Furthermore, our 3B model \nllbtwo consistently outperforms all baselines in the \TOTALlanguages evaluated languages. When testing on MetricX and \blaser, we obtain similar results.

\begin{figure}[H]
    \centering
\includegraphics[scale=1]{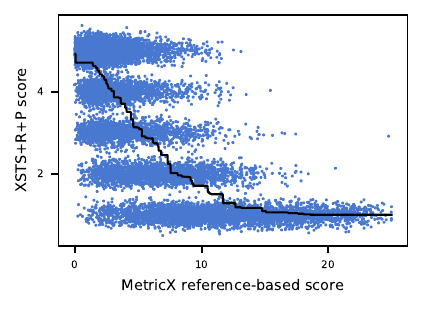}
    \caption{A curve to predict XSTS+R+P scores from reference-based MetricX (on \metbouquet, \Xeng subset;
    to visualize XSTS+R+P distribution, a random jitter has been added to the Y axis).
    \label{fig:xstsrp_curve_fit}}
\end{figure}

\paragraph{Languages passing the quality bar on understanding the long tail} Beyond the relative performance of models in understanding hundreds of languages, we seek to determine how many languages the OMT models understand "well enough" in absolute terms. We define a passing quality threshold as an average XSTS+R+P score above 2.5. Based on the definition of XSTS+R+P scores (see Section \ref{subsec:xstsrp}), this roughly means that the system is capable of conveying the core meaning of a sentence in the majority of cases. We rely on MetricX (reference-based) to estimate whether a translation meets this criterion. The primary motivation for using MetricX is that it is a well-established external metric that effectively leverages both target and reference translations. Furthermore, unlike \blaser, it is not based on \sonaromni, thereby avoiding potential bias toward the \nllbtwo model, which shares the same encoder. We apply a monotonic regression over Met-\bouquet (restricted to the \Xeng directions for direct compatibility with the Bible evaluation setup) to map between MetricX and XSTS+R+P scores (see Figure \ref{fig:xstsrp_curve_fit}).

\begin{figure}[H]
    \centering
\includegraphics[scale=1]{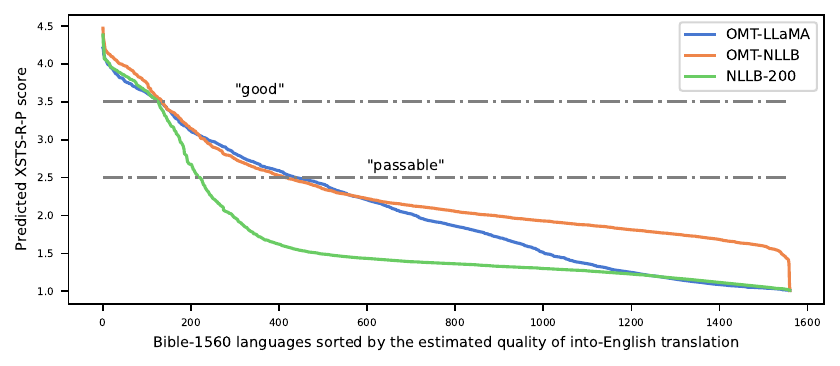}
    \caption{
    Predicted XSTS+R+P machine translation quality score for the Bible languages into English using MetricX extrapolation on the test set (languages sorted individually for each of the evaluated models) \label{fig:qualitybar}.}
\end{figure}

We use the Bible benchmark (comprising \biblelanguages languages) to estimate the average XSTS+R+P score using the MetricX proxy for each source language translated into English. The right panel of Figure \ref{fig:qualitybar} presents the results of this extrapolation for \omtllama, \nllbtwo, and NLLB-200 as a baseline. All three models cover approximately the same number of Bible languages (around 130) at the "good" quality threshold of at least 3.5 extrapolated XSTS+R+P points on average. However, the number of Bible languages for which the "passable" quality threshold of 2.5 points is exceeded is substantially larger for the OMT models: 440 languages for the \omtllama 8B model and 416 languages for \nllbtwo—nearly double the 221 varieties for which NLLB-200 surpasses this threshold. We therefore conclude that, while we remain far from completely "solving" machine translation for the long tail of languages, the OMT models double the number of reasonably well-understood source languages compared to previous massively multilingual models.

\begin{figure}[H]
    \centering
    \includegraphics[width=1\linewidth]{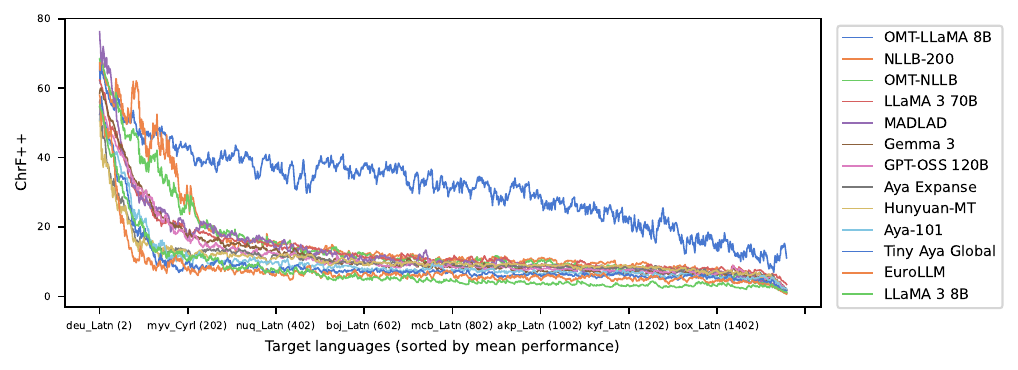}
    \includegraphics[width=1\linewidth]{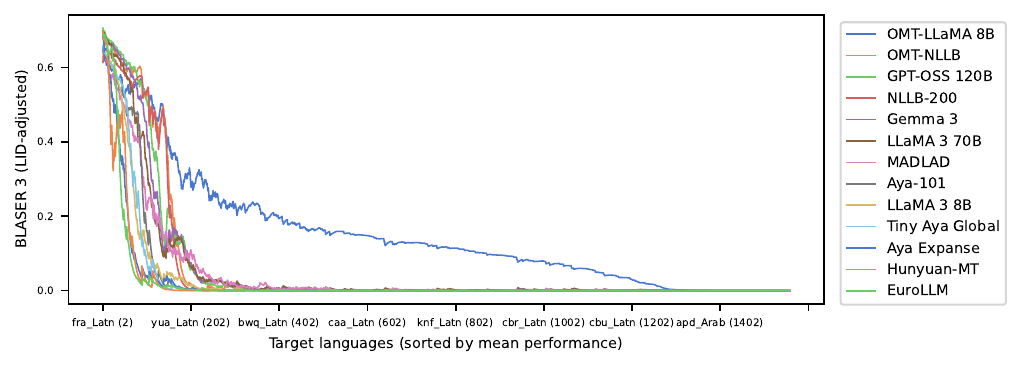}
    \caption{Relative performance in terms of \chrf (top) and LID-adjusted \blaser (bottom) of OMT models in generating the longtail in the Bible domain compared to external baselines.  \label{fig:eng2xxlongtail} 
    }

\end{figure}

\paragraph{Relative Performance on generating the long tail of languages} 
For reference evaluation, we use the Bible and translate the English part into \biblelanguages languages; report ChrF++ and LID-adjusted \blaser, MetricX (rescaled to 0-1), and xCOMET scores. Results are shown in Figure \ref{fig:eng2xxlongtail}. 

For all baseline models and all scores, the quality becomes near-random-level at about 300-400 languages, while for OMT models, it holds for about $\approx$ 1,200 languages.
On the first 150 languages (the intersection of the NLLB-200 set with Bible), the \omtllama model sometimes underperforms compared to \nllbtwo and NLLB-200.
However, \nllbtwo nearly always outperforms NLLB-200.

Some limitations of this experiment include that none of our automatic metrics are capable of adequately evaluating the grammaticality/fluency/naturalness of translations into long-tail languages, so we cannot say for sure, for how many languages our translations are “good enough”, without extensive human evaluation results (which are reported for Round 2 of \metbouquet in Section \ref{sec:humaneval}). Also, we had to adjust all scores (except ChrF++) by LID, because all models tend to generate outputs in the wrong language. Finally, we admit that we did not do thorough prompt-engineering to ensure that all the long-tail target languages are correctly identified by each of the baseline models, and there might be a room for improvement by using better instructions and/or few-shot examples with the baseline models.

\subsubsection{Comparing across model sizes and architectures}
\label{sec:familyofmodels}

\begin{figure}[H]
    \centering
    \includegraphics[scale=1]{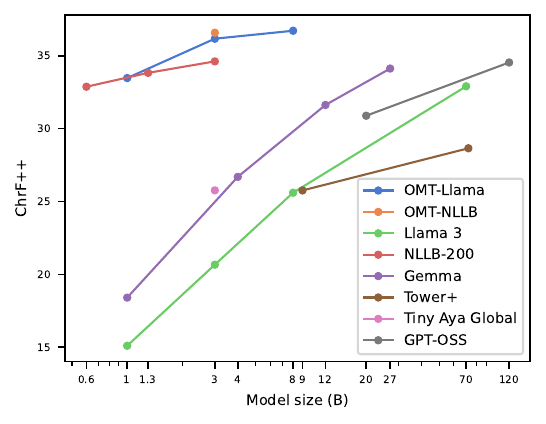}
    \caption{Interaction of model size and translation quality for the OMT models and several other open model families on the \bouquet dataset (from and into English).
    \label{fig:params_scaling_bouquet}
    }

\end{figure}

\paragraph{Size vs performance}
Given that \OmniMT models come in different sizes, it is interesting to explore the size-performance tradeoff.
Figure \ref{fig:params_scaling_bouquet} displays the interaction of model size (in billions of parameters) and translation quality (\chrf) on the \bouquet dataset (all resource levels, translation into and from Englihs) for \omtllama and for several other open model families. %
Across all size categories, the OMT models are outperforming the baseline models of the corresponding size.

\begin{figure}[H]
    \centering
    \includegraphics[scale=1]{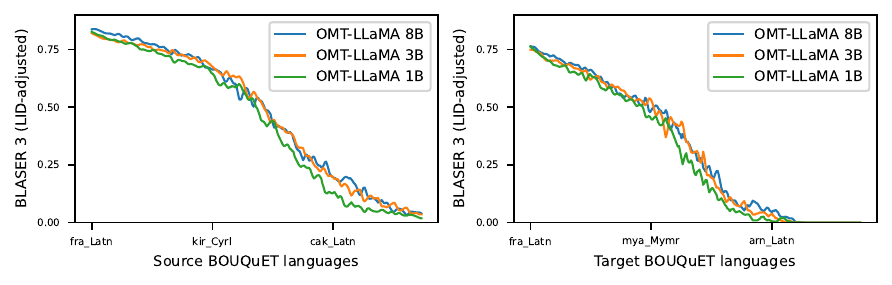}
    \caption{Performance of \omtllama models on \bouquet (test set, sentence-level) by language for \Xeng translation (left) and \engX translation (right). Languages sorted by the decreasing performance of the average of the three models; curved smoothed with sliding window.
    \label{fig:params_scaling_bouquet_languages}
    }

\end{figure}

\paragraph{\omtllama downscaling effect on languages} To see which languages contribute most to the differences in \omtllama generation and understanding at three different model scales, we plot the per-language LID-adjusted \blaser performance in Figure \ref{fig:params_scaling_bouquet_languages}. The results look qualitatively similar for \Xeng and \engX directions. The 1B model slightly underperforms compared to the larger variants for most languages, but especially for the hardest ones, perhaps because the smaller number of parameters prevents it from learning as efficiently from very low-resourced data or from generalizing to languages never seen in the traning data.

\paragraph{\omtllama vs \nllbtwo} These models share several key characteristics, including language coverage, massively multilingual tokenization, training data, extensibility, and a common LLaMA 3 backbone. However, they differ in certain respects, as detailed in Table \ref{tab:nllbtwoomtllama}. Most notably, \omtllama models have been instruction-finetuned and therefore are compatible with diverse translation instructions, such as using few-shot prompts, either static or retrieved from a database based on the source text. The \nllbtwo model has not been trained to support any instructions apart from basic translation, although in principle it can be fine-tuned this way.

\begin{table}[H]
    \centering
    \begin{tabular}{lp{5cm}p{5cm}}
        \toprule
        \multirow{1}{*}{\textbf{Feature}}           & \textbf{\omtllama} & \textbf{\nllbtwo} \\ \midrule
        Sizes & 1B, 3B, 8B & 3B \\
        Architecture & decoder-only & encoder-decoder \\
        Understanding Languages & around 1000 & almost any language  \\
        Generating Languages & around 1000 & around 250 \\
        Zero/few-shot & few-shot & 0-shot (source side) \\
        \bottomrule
    \end{tabular}
    \caption{List of features comparing \omtllama vs \nllbtwo models presented in this paper.  }
    \label{tab:nllbtwoomtllama}
\end{table}

As demonstrated in the previous subsections, the \OmniMT models also exhibit variation in relative performance across the translation directions they support. %
On the input side, \nllbtwo is the best model when it comes to translating from low- and zero-resourced languages, whereas \omtllama seems to be competitive for translation from more high-resourced ones. On the output side, the trend is opposite: \nllbtwo is capable of generating ``only'' 250 languages, compared to over 1000 languages with \omtllama, but for many high- and mid-resourced languages, its generation quality is superior.

Further experimental evidence would be required to identify the main factors driving the difference in performance between \omtllama and \nllbtwo on the set of translation directions that they both support. Those differences might stem from the architecture (separating the encoder and decoder modules seems to benefit cross-lingual generalization on the input side), the training tasks (with language modeling as a secondary task for \omtllama and reconstruction, for \nllbtwo, inducing different competences), the composition of the training data (more imbalanced across languages for \omtllama than for \nllbtwo), or even simply the ``intensity'' of training (how much the language competences are retained from the base model or acquired during continual training) — or a combination of these factors. In future research, a more principled set of experiments could shed more light on the effects of each of these choices.

\subsection{Human Evaluation}
\label{sec:humaneval}

Our human evaluation analysis corresponds to the \metbouquet Round 2 annotations. Details of the data and systems evaluated are in section \ref{sec:metbouquet} when describing Round 2. In short, we compare a variety of \omtllama systems to the strongest baseline in terms of automatic evaluation in the test partition of \bouquet. The set of languages is given in Table \ref{tab:bouquetlangs} specified as \metbouquet r2 and the annotation protocol is XSTS+R+P described in Section \ref{subsec:xstsrp}. 

\begin{table}[H]
    \centering
    \begin{tabular}{lrrrr}
        \toprule
        Direction group &  \omtllama & \omtllama & Baseline & Num \\
        &Win rate & Mean score & Mean score & Directions \\
        \midrule
        high-high & 0.75 & 4.03 & 3.45 & 16 \\
        high-low & 0.83 & 2.99 & 1.61 & 18 \\
        low-high & 0.80 & 3.26 & 2.69 & 15 \\
        low-low & 0.62 & 2.96 & 2.66 & 8 \\
        \midrule
        total & 0.77 & 3.38 & 2.67 & 57 \\
        \bottomrule
        \end{tabular}
    \caption{Average results of \metbouquet Round 2 annotations: \omtllama win rate, \omtllama and baseline mean score and number of directions for each. ``Win rate'' is defined as the proportion of directions with the average scores for \omtllama higher than for the baseline system.}
    \label{tab:metbouquet_round2_sbs_agg}
\end{table}

\begin{figure}[H]
    \centering
    \includegraphics[scale=1.0]{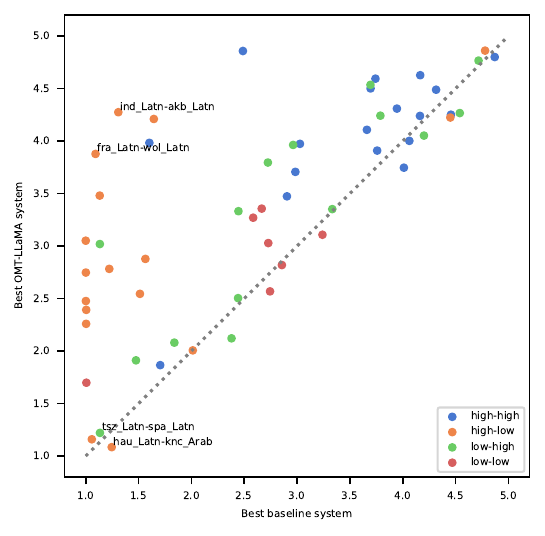}
    \caption{Mean XSTS+R+P scores for each of the 57 directions of Met-Bouquet Round 2 for the OMT system (vertical axis) and the baseline system (horizontal axis). 
    \label{fig:metbouquet_round2_sbs}}
\end{figure}

The currently available annotation results cover 57 directions composed of 80 unique language varieties. Of these 57 directions, in 44 (77\%), the \omtllama system outperforms the baseline, according to the mean XSTS+R+P score. Figure \ref{fig:metbouquet_round2_sbs} and Table \ref{tab:metbouquet_round2_sbs_agg} report these results, aggregated by translation direction resource levels (with ``high'' standing for high- and mid-resource languages, i.e. the ones with at least 1M of primary parallel sentences, and ``low'' standing for low- and very-low-resource languages). 

The largest improvements of \omtllama over the baseline systems, in agreement with the automatic evaluation results, are observed when translating from higher-resourced to lower-resourced languages: while the baseline systems often struggle to produce meaningful translations, OMT models generate a significant proportion of least moderate-quality translations. But overall, \omtllama is outperforming the baselines in each group of directions, and in no directions it lags behind the baseline by more than 0.3 XSTS+R+P points on average.

The three directions where both \omtllama and the baseline systems produce the majority of minimal scores are Hausa to Central Kanuri (Arabic script), Spanish to Tzotzil, and Purepecha to Spanish. Some directions with the largest \OmniMT gains include Indonesian to Batak Angkola, between French and Wolof, and Spanish to Alacatlatzala Mixtec. 

Overall, this manual annotation campaign demonstrates the significant progress that \OmniMT made in challenging language pairs but shows that much more progress is yet to follow: the evolution from the average score from 2.67 (baseline) to 3.38 (OMT) represents a qualitative jump from ``useless'' to ``useful'' for many of the directions, but it is only halfway to the ``really good'' score of 4 and the ``perfect'' score of 5.

\subsection{Added Toxicity Automatic Evaluation}
\label{sec:toxicity}

\paragraph{Added toxicity definition and experimental framework} Using \omnitox, we benchmark \omtllama and \nllbtwo for added toxicity comparing against Gemma-3-27B and NLLB-200-3B as baselines. We define added toxicity as the increase in toxicity between the source text and its translation, quantified through the difference in \omnitox logits.

\paragraph{Language debiasing} To mitigate inherent language-level biases in the classifier, we establish per-language baselines using the \bouquet dataset, which contains professionally crafted, non-toxic sentences across a comprehensive set of languages. For each language $l$, we compute the mean logit $\mu_l = \mathbb{E}[z \mid \text{language} = l]$ from both source and target texts in \bouquet. Each raw logit $z$ is then debiased as $z_{\text{debiased}} = z - \mu_l$.

\paragraph{Added toxicity computation} For a translation pair, added toxicity ($\Delta$) is computed as the difference between debiased logits:
\begin{equation}
    \Delta = z_{\text{tgt}}^{\text{debiased}} - z_{\text{src}}^{\text{debiased}}
\end{equation}
where $z_{\text{src}}^{\text{debiased}}$ and $z_{\text{tgt}}^{\text{debiased}}$ are the debiased logits for the source and target texts, respectively. When either language lacks a baseline (i.e., is not represented in \bouquet), we use raw logits for both source and target to ensure consistent comparisons.

\paragraph{Threshold calibration} We calibrate flagging thresholds using \bouquet to achieve an approximate 5\% False Positive Rate (FPR). Specifically, we compute the 95th percentile of the $\Delta$ distribution. To account for variation across translation directions, we compute \emph{per-direction thresholds} $\tau_{l_s \rightarrow l_t}$ for each direction available in \bouquet. For the other directions, we fall back to a \emph{global threshold} $\tau_{\text{global}}$ computed across all \bouquet translation pairs.

\paragraph{Flagging criterion} A translation is flagged for added toxicity if it satisfies two conditions:
\begin{equation}
    \Delta > \tau \quad \text{and} \quad p_{\text{tgt}} > 0.20
\end{equation}
where $\tau$ is the per-direction threshold $\tau_{l_s \rightarrow l_t}$ when available, or the global threshold $\tau_{\text{global}}$ otherwise. The probability constraint ($p_{\text{tgt}} = \sigma(z_{\text{tgt}}) > 0.20$) filters out false positives arising from minor fluctuations at low toxicity levels, where small logit changes can produce disproportionately large relative differences.

\paragraph{Results} 
Table \ref{tab:addedtoxicity} shows the average added toxicity and flagging rates across 207 languages available in \bouquet at the time of this experiment. Overall, none of the evaluated systems exhibit meaningful added toxicity, with flagging rates remaining below 1.5\% across all configurations. This suggests that both our systems and the external baselines reliably preserve the toxicity profile of source texts during translation. Note that similarly to \floresplus, \bouquet may be limited for triggering added toxicity, and in the future, we can consider using more adequate datasets e.g. \citep{tan-etal-2025-towards}.

We observe a directional asymmetry that correlates with model architecture. Encoder-decoder models (NLLB-200-3B and \nllbtwo) show slightly higher added toxicity for X$\rightarrow$Eng translations, whereas decoder-only LLMs (Gemma-3-27B and \omtllama) exhibit the opposite pattern, with marginally higher values for Eng$\rightarrow$X. While the absolute differences are small, this consistency across architectures suggests a systematic effect that may warrant further investigation.

\begin{table}[H]
    \centering
    \begin{tabular}{l|cc|cc}
        \toprule
        \textbf{System} & \multicolumn{2}{c|}{\textbf{Average Added Toxicity}} & \multicolumn{2}{c}{\textbf{Flagging Rate}}\\
        \midrule
        & \textbf{\Xeng} & \textbf{\engX} & \textbf{\Xeng} & \textbf{\engX} \\
        \midrule
        NLLB-200-3B & 0.19 & -0.20 & 1.10\% & 0.07\% \\
        Gemma-3-27b & -0.20 & 0.07 & 0.25\% & 0.06\%\\
        \midrule
        \nllbtwo & 0.02 & -0.17 & 0.69\% & 0.04\%\\
        \omtllama & -0.07 & 0.17 & 0.52\% & 0.06\%\\
        \bottomrule
    \end{tabular}
    \caption{Average Added Toxicity and Flagging Rate (\%) on \bouquet dataset for 207 languages.}
    \label{tab:addedtoxicity}
\end{table}

\section{Extensibility of \omtllama}
\label{sec:extensibility}

\paragraph{\textit{Motivation}}
An omnilingual model, by definition, intends to provide support for any language. In many practical use cases, however, MT models are applied to a limited subset of language pairs and optimized for improved translation quality for that subset. In this section, we select several difficult languages and explore how the \omtllama models could be extended for their improved support using additional data. We consider two approaches of feeding this focused data to models: via fine-tuning (already described in Section \ref{sec:sft}) and through retrieval-augmented translation (explored in Section \ref{sec:retrievalaugmentedtranslation}).

\paragraph{\textit{Languages}}
For extension experiments, we selected the languages among those for which the \omtllama model demonstrated low performance on \bouquet (mostly low- and very-low-resourced ones), either in understanding or in generation, with the additional criterion of being included in the version of \medley available at the time of the experiment.\footnote{These languages are \texttt{azb\_Arab, bam\_Latn, dik\_Latn, fuv\_Latn, kam\_Latn, kmb\_Latn, lug\_Latn, mam\_Latn, miq\_Latn, mos\_Latn, pcm\_Latn, sba\_Latn, shn\_Mymr, tsz\_Latn, tzh\_Latn, umb\_Latn, vmw\_Latn, wol\_Latn, yor\_Latn, yua\_Latn}.} In addition, we selected 5 mid-resourced moderately difficult \bouquet languages\footnote{\texttt{gaz\_Latn, lin\_Latn, mya\_Mymr, swh\_Latn, tir\_Ethi}}. We tune the systems of this section for translation of these 25 languages into and out of English.

\paragraph{\textit{Data}} For the selected 25 languages, for both fine-tuning and retrieval, we use a subset of primary (non-synthetic) parallel data already described in Section \ref{sec:CPTtrainingdata}. The number of parallel examples (mostly sentences, but also words in case of Panlex and paragraphs in case of \medley) per language ranges from 300K (Swahili) to 11K (Ngambay). Note that because a large part of the data comes from massively parallel sources (such as the Bible or \medley), the same English texts often appear multiple times in it, paired with different languages.

\paragraph{Experimental setup} We use the above data for addapting models to translation of specific languages in two ways: via fine-tuning and via retrieval-augmented translation. We compare two models as the base model for fine-tuning and direct translation:
\begin{itemize}
    \item LLaMA-base: the original LLaMA 3.1 8B Instruct model without any modifications;
    \item \omtllama: the 8B version of \omtllama that underwent the standard \OmniMT continual pretraining (as described in Section \ref{sec:cptmodel}) and then was fine-tuned only with the \ourblocks dataset (which does not include the low-resourced languages) to isolate the effect of massively multilingual fine-tuning data.
\end{itemize}

For all the baselines and the models fine-tuned in this section, we evaluate retrieval-augmented translation as well as standard translation with the minimalistic prompt. We report \chrf and LID-augmented \blaser scores on the \bouquet dataset (sentence-level part).

\begin{table}
\scriptsize
\centering
\begin{tabular}{l|rrrr|rrrr}
\toprule
Metric & \multicolumn{4}{c|}{ChrF++} & \multicolumn{4}{c}{\blaser + LID} \\
Languages & \multicolumn{2}{c}{20 hardest} & \multicolumn{2}{c|}{5 hard} & \multicolumn{2}{c}{20 hardest} & \multicolumn{2}{c}{5 hard} \\
Direction & \engX & \Xeng &  \engX & \Xeng & \engX  & \Xeng & \engX & \Xeng \\
\midrule

OMT-LLaMA + FT + RAG & {\cellcolor[HTML]{023858}} \color[HTML]{F1F1F1} 29.25 & {\cellcolor[HTML]{023858}} \color[HTML]{F1F1F1} 36.67 & {\cellcolor[HTML]{023E62}} \color[HTML]{F1F1F1} 46.74 & {\cellcolor[HTML]{023F64}} \color[HTML]{F1F1F1} 53.22 & {\cellcolor[HTML]{023858}} \color[HTML]{F1F1F1} 0.2577 & {\cellcolor[HTML]{023A5B}} \color[HTML]{F1F1F1} 0.3427 & {\cellcolor[HTML]{023C5F}} \color[HTML]{F1F1F1} 0.4682 & {\cellcolor[HTML]{023C5F}} \color[HTML]{F1F1F1} 0.6349 \\
OMT-LLaMA + FT & {\cellcolor[HTML]{023C5F}} \color[HTML]{F1F1F1} 28.91 & {\cellcolor[HTML]{69A5CC}} \color[HTML]{F1F1F1} 29.03 & {\cellcolor[HTML]{023858}} \color[HTML]{F1F1F1} 47.49 & {\cellcolor[HTML]{5A9EC9}} \color[HTML]{F1F1F1} 42.69 & {\cellcolor[HTML]{023B5D}} \color[HTML]{F1F1F1} 0.2547 & {\cellcolor[HTML]{0567A2}} \color[HTML]{F1F1F1} 0.3010 & {\cellcolor[HTML]{023858}} \color[HTML]{F1F1F1} 0.4747 & {\cellcolor[HTML]{1C7FB8}} \color[HTML]{F1F1F1} 0.5285 \\
OMT-LLaMA + RAG & {\cellcolor[HTML]{04598C}} \color[HTML]{F1F1F1} 26.79 & {\cellcolor[HTML]{023F64}} \color[HTML]{F1F1F1} 36.18 & {\cellcolor[HTML]{034E7B}} \color[HTML]{F1F1F1} 44.84 & {\cellcolor[HTML]{023858}} \color[HTML]{F1F1F1} 53.96 & {\cellcolor[HTML]{034267}} \color[HTML]{F1F1F1} 0.2486 & {\cellcolor[HTML]{023858}} \color[HTML]{F1F1F1} 0.3445 & {\cellcolor[HTML]{034369}} \color[HTML]{F1F1F1} 0.4595 & {\cellcolor[HTML]{023858}} \color[HTML]{F1F1F1} 0.6420 \\
OMT-LLaMA & {\cellcolor[HTML]{0872B1}} \color[HTML]{F1F1F1} 24.12 & {\cellcolor[HTML]{4E9AC6}} \color[HTML]{F1F1F1} 29.92 & {\cellcolor[HTML]{045D92}} \color[HTML]{F1F1F1} 43.01 & {\cellcolor[HTML]{056AA6}} \color[HTML]{F1F1F1} 48.43 & {\cellcolor[HTML]{046198}} \color[HTML]{F1F1F1} 0.2178 & {\cellcolor[HTML]{045D92}} \color[HTML]{F1F1F1} 0.3135 & {\cellcolor[HTML]{034E7B}} \color[HTML]{F1F1F1} 0.4422 & {\cellcolor[HTML]{034D79}} \color[HTML]{F1F1F1} 0.6130 \\
LLaMA-base + FT + RAG & {\cellcolor[HTML]{60A1CA}} \color[HTML]{F1F1F1} 20.11 & {\cellcolor[HTML]{88B1D4}} \color[HTML]{000000} 27.94 & {\cellcolor[HTML]{5EA0CA}} \color[HTML]{F1F1F1} 33.16 & {\cellcolor[HTML]{79ABD0}} \color[HTML]{F1F1F1} 40.93 & {\cellcolor[HTML]{589EC8}} \color[HTML]{F1F1F1} 0.1514 & {\cellcolor[HTML]{3790C0}} \color[HTML]{F1F1F1} 0.2630 & {\cellcolor[HTML]{509AC6}} \color[HTML]{F1F1F1} 0.3109 & {\cellcolor[HTML]{4496C3}} \color[HTML]{F1F1F1} 0.4950 \\
LLaMA-base + FT & {\cellcolor[HTML]{73A9CF}} \color[HTML]{F1F1F1} 19.37 & {\cellcolor[HTML]{ACC0DD}} \color[HTML]{000000} 26.39 & {\cellcolor[HTML]{7BACD1}} \color[HTML]{F1F1F1} 31.17 & {\cellcolor[HTML]{9CB9D9}} \color[HTML]{000000} 38.82 & {\cellcolor[HTML]{71A8CE}} \color[HTML]{F1F1F1} 0.1399 & {\cellcolor[HTML]{549CC7}} \color[HTML]{F1F1F1} 0.2504 & {\cellcolor[HTML]{76AAD0}} \color[HTML]{F1F1F1} 0.2804 & {\cellcolor[HTML]{7BACD1}} \color[HTML]{F1F1F1} 0.4530 \\
LLaMA-base + RAG & {\cellcolor[HTML]{99B8D8}} \color[HTML]{000000} 17.50 & {\cellcolor[HTML]{E9E5F1}} \color[HTML]{000000} 22.93 & {\cellcolor[HTML]{C5CCE3}} \color[HTML]{000000} 25.01 & {\cellcolor[HTML]{F1EBF4}} \color[HTML]{000000} 31.07 & {\cellcolor[HTML]{ABBFDC}} \color[HTML]{000000} 0.1049 & {\cellcolor[HTML]{ABBFDC}} \color[HTML]{000000} 0.2064 & {\cellcolor[HTML]{D3D4E7}} \color[HTML]{000000} 0.1824 & {\cellcolor[HTML]{D7D6E9}} \color[HTML]{000000} 0.3584 \\
LLaMA-base & {\cellcolor[HTML]{FFF7FB}} \color[HTML]{000000} 9.35 & {\cellcolor[HTML]{FFF7FB}} \color[HTML]{000000} 20.71 & {\cellcolor[HTML]{FFF7FB}} \color[HTML]{000000} 16.07 & {\cellcolor[HTML]{FFF7FB}} \color[HTML]{000000} 28.66 & {\cellcolor[HTML]{FFF7FB}} \color[HTML]{000000} 0.0184 & {\cellcolor[HTML]{FFF7FB}} \color[HTML]{000000} 0.1285 & {\cellcolor[HTML]{FFF7FB}} \color[HTML]{000000} 0.0917 & {\cellcolor[HTML]{FFF7FB}} \color[HTML]{000000} 0.2779 \\

\bottomrule
\end{tabular}
\caption{Results of extending the \omtllama and LLaMA-base models via focused fine-tuning, retrieval augmentation, or both. Reported on the dev split of \bouquet, for 20 hard languages and 5 mid-resourced languages.\label{tab:extensibility}}
\end{table}

\paragraph{\textit{Fine-tuning}} 
As a base fine-tuning experiment, we mix the parallel data described above (all 25 languages from and into English) with \ourblocks (in equal proportion) and fine-tune each of the three baseline models with the same hyperparameters as in Section \ref{sec:sft}. As Table \ref{tab:extensibility} shows, fine-tuning on average improves out-of-English results, but is detrimental for into-English translation. We hypothesize that it is the same effect as observed in the SMOL paper \citep{caswell2025smol}, with the model degenerating after training on repetitive English outputs.

\paragraph{\textit{RAG}}
We use a simplified version of the retrieval algorithm described in Section \ref{sec:retrievalaugmentedtranslation}, with TF-IDF matching of words only (without extra retrieval with embeddings or on-the-fly mining). As Table \ref{tab:extensibility} shows, RAG almost always leads to improvement both of the baseline model and of the fine-tuned models (despite the latter have been exposed to the same data during fine-tuning), with the only exception of translation into the 5 mid-resourced languages. 

\paragraph{\textit{Comparison with LLaMA-base}} The results of applying focused fine-tuning and RAG to the base LLaMA 3 model are mostly qualitatively similar to those obtained for \omtllama: finetuning is crucial for reaching good \engX translation, whereas RAG is sufficient for reaping a large part of the improvements in \Xeng results, and the positive effects of these two techniques add up. One difference between LLaMA-base and \omtllama is that the former doesn't exhibit the negative effects of fine-tuning on \Xeng translation, maybe simply because of the low base effect. But, crucially, even after applying a combination of instruction fine-tuning and RAG translation, the base LLaMA model does not outperform the unadapted \omtllama model, highlighting that \omtllama is a strongly preferable base model for adaptation into a translation model specialized on certain language pairs.

\paragraph{\textit{Conclusions}}
Both fine-tuning and RAG, when applied to a focused set of languages, yield improved translation performance, enabling targeted customization of the \omtllama model. The two techniques are complementary: fine-tuning is particularly effective for translation into challenging languages, while RAG is essential for enhancing translation quality from these languages.

\section{Conclusion}
\label{sec:conclusions}

\OmniMT demonstrates that scaling multilingual translation is not simply a matter of increasing the number of supported languages, but of rethinking how MT systems are built, trained, and evaluated. Expanding from 200 to more than one thousand languages required coordinated advances across every layer of the pipeline. Our data strategy — combining massive public corpora with newly created resources such as \medley and \bouquet bitext, deliberate data creation targeting linguistic varieties, domains, and registers that existing corpora overlook—allows us to improve translation generation quality and more meaningfully evaluate long-tail coverage. Our modeling strategy — based on pretrained language models with extended tokenizer vocabulary — encompassed 2 distinct architectures. Our decoder-only models (\omtllama) introduce only minimal changes to the architecture and training scheme of standard language models that are necessary to support over a thousand languages. Our encoder-decoder (\nllbtwo) model introduces a novel three-stage training strategy that effectively exploits non-parallel data to achieve substantial quality improvements.

\OmniMT models deliver strong, consistent gains for broad-coverage languages, and provides the first non-trivial MT for hundreds of emerging-support languages where no usable systems previously existed. %
Our evaluation efforts further show that our 1B to 8B parameter specialized MT models can match or exceed the MT performance of a 70B LLM, offering a clear Pareto improvement and enabling high-quality translation in low-compute real-world settings. Our English-to-\TOTALlanguages evaluation additionally reveals a consistent failure mode in existing systems: while many models can interpret undersupported languages, they frequently cannot generate them with meaningful fidelity. \OmniMT dramatically improves in cross-lingual transfer, being close to solving the ``understanding'' part of the puzzle in MT; and it substantially expands the set of languages for which coherent generation is feasible, underscoring the central bottleneck for genuine large-scale language coverage: robust generation in under-resourced languages. We also show that targeted techniques such as finetuning and retrieval-augmented generation can yield further quality improvements in our models when additional data in the languages and domains of interest is available. %

Our experiments show that post-training techniques targeting multilingual extension of models do not compensate the lack of large improvements obtained by embracing massively multilingual training data and vocabulary in the earlier stages of training. Therefore, our findings should motivate model researchers and developers interested in boosting multilingual tasks performance to gather high-quality massively multilingual data and train models that are highly multilingual by design and are well equipped to extend their support to any additional language, if necessary. %

Taken together, \OmniMT positions large-scale inclusion as an ongoing technical and scientific challenge — one that demands continued investment in data creation, architecture design, evaluation methodology, and under-resourced-language generation capabilities. %
By pairing broad coverage with efficient specialization and flexible avenues for improvement, \OmniMT provides inspiration for future research and applications across translation, multilingual LLM development, and speech-to-text systems. Ultimately, setting a baseline for  \TOTALlanguages languages is therefore not a terminus but an encouragement to sustained innovation in pursuit of genuinely inclusive language technology.

Our key dataset for massively multilingual evaluation of machine translation, \bouquet (including its Met-\bouquet extension with human judgments of translation quality) is publicly available, enabling researchers from anywhere to reproduce our evaluation results. Moreover, we hope that our \seed detailed guidelines can be used to create more high-quality and diverse data. Finally, we encourage the scientific community to use our \omtllama, \nllbtwo, \blaser and \omnitox recipes to develop and release yet other radically inclusive foundational models for machine translation, evaluation, and general-purpose massively multilingual language processing.

\newpage

\section{Contribution Statements}

\begin{spacing}{0.9}

We outline the contributions of each member of \OmniMT. However, there are no possible words to describe the emotional dedication stemming from the multilingual passion that characterizes this team.

\textbf{Data}

Niyati Bafna - designed and led efforts on \medley

Andrea Caciolai - led experiments on backtranslation, contributed to the creation of \medley and led its experiments, supported \omtnllb data curation

Jean Maillard - managed linguistic partnerships and community engagement, coordinating with language communities and external organizations

Holger Schwenk - inspired large scale data mining

\textbf{Modeling}

Belen Alastruey - led, designed and drove efforts for \nllbtwo 

Pere Lluís Huguet Cabot, João Maria Janeiro - contributed to the training of \nllbtwo

Paul-Ambroise Duquenne - contributed to the technical supervision of \nllbtwo

Kevin Heffernan - drove CPT experiments, parallel mining, contributed to model scaling, vocabulary extension

Artyom Kozhevnikov - drove retrieval-augmented translation experiments

Eduardo S\'anchez - designed and drove efforts on \omtllama post-training

Edan Toledo - contributed to post-training experiments, implementing the RL stage

Ioannis Tsiamas - spearheaded the development of \blaser and engineered data pipelines for CPT

\textbf{Evaluation}

Chierh Cheng, Joe Chuang, Gabriel Mejia Gonzalez - insured the quality of manual translations and annotations

Mark Duppenthaler - developed the online \bouquet collection tool

Nate Ekberg, Cynthia Gao - drove relationships with language service providers

Christophe Ropers - led the linguistic team and data creation efforts across \medley, \bouquet, Met-\bouquet and XSTS+R+P

Charles-Eric Saint-James - developed \omnitox and added toxicity analysis, built data pipelines and ran data validation for Met-BOUQuET %

Arina Turkatenko - led extensive analysis on the quality of manual translations and annotations

Albert Ventayol-Boada - led linguistic efforts and feature retention analysis in \medley, led language selection across projects, and contributed to translation quality and error analyses

\textbf{Project management}

Rashel Moritz - Technical Program Manager, coordinated the Language Technology Partnership Program

Alexandre Mourachko - Research Manager, helped with the overall direction, strategy and resourcing plan and supported data efforts

Surya Parimi - Technical Program Manager, supported data efforts

Shireen Yates - Product Manager, helped with the overall direction and strategy

Mary Williamson - Research director, helped with overall direction and strategy

\textbf{Technical leadership}

David Dale - co-technical lead,  devised the continual pretraining framework, led the direction of the family of models strategy, coordinated engineering efforts across the team

Marta R. Costa-juss\`a - co-technical lead, led the overall direction for evaluation, main driver on \bouquet, Met-\bouquet, \blaser, \omnitox, XSTS+R+P

\section*{Acknowledgements}

We extend our thanks to Mikel Artetxe for his continued support and partnering while brainstorming modeling directions. We thank Sebastian Ruder for his feedback on early drafts of the paper. We thank Anaelia Ovalle for the discussions and her involvement on exploration experiments on agentic extensions of the model. We thank Luke Zettlemoyer for his guidance and feedback on the project strategy. Finally, we thank all the participants of the Language Technology Partnership Program, for their contributions and keen interest in our workshops as well as the contributors to the \bouquet open-initiative.

\end{spacing}

\clearpage
\newpage
\bibliographystyle{assets/plainnat}
\bibliography{bibliography/paper,bibliography/bib-bouquet,bibliography/bib-backtranslation,bibliography/bib-languages,bibliography/general,bibliography/bib-blaser3,bibliography/bib-medley}

\clearpage
\newpage

\beginappendix

\addcontentsline{toc}{section}{Appendices}

\section{MeDLEy details}
\label{app:medley}

\subsection{More details on the approach}
\label{medley:sec:approach_details}

The goal of \datasetname is to provide a bitext corpus that is domain-diverse and grammatically diverse in a large number of included languages.
Further, we would like for lay people of various language communities to be able to extend the dataset to their native language in the future via simple translation, while maintaining this property. 
Our approach therefore does not rely on linguistic expertise in specific target languages.
Instead, we formulate a framework of grammatical diversity which is transferrable via translation (cf. \cref{medley:sec:gram_div_corpus}), and craft \datasetnamesource with  domain and grammatical diversity in mind.
Here are the steps involved in creating \datasetname, as depicted in \cref{medley:fig:medley_creation}.

\begin{enumerate}
    \item[1.] \textbf{Feature enumeration} We curate a list of broad grammatical categories of interest, with associated features per category. Features are chosen to be representative of known cross-linguistic grammatical phenomena. See the list of features in \cref{medley:app:features}.
    \item[2a.] \textbf{Domain selection} We choose the following 5 domains: \emph{informative}, \emph{dialogue}, \emph{casual}, \emph{narrative}, and \emph{instruction-response}. Notably, we include data in the style of user instructions and large language model (LLM) responses, given the increasing practice of and need for translating instruction fine-tuning datasets into LRLs in the era of LLMs \citep{upadhayay2023taco,singh-etal-2024-aya}. %
    \item[2b.] \textbf{Source language selection} We choose $5$ \emph{source languages}, in which paragraphs are crafted: English, Mandarin, Russian, Spanish, and German, based on team's linguistic proficiency. 
    \item[2c.] \textbf{Template generation} We create linguistic templates, consisting of constrained combinations of grammatical features and a domain for each paragraph. We assign each template to a source language uniformly at random.
    \item[3a.] \textbf{Creation of grammatically-diverse, domain-diverse source paragraphs, with accompanying context} Expert native speaker linguists craft source paragraphs given the set of templates assigned to each source language, within the associated domain for a template, and exhibiting the listed grammatical features. We prioritize naturalness, and avoid highly specialized or technical jargon for the sake of accessibility. This results in a set of multi-centric, domain-diverse, easy-to-translate, and grammatically diverse source paragraphs. Each source paragraph is accompanied by notes regarding its context, which may provide additional relevant information. Notes may specify the gender or age of the referents involved, or the surrounding context of a conversation, since these may become relevant for translations into some languages.
    \item[3b.] \textbf{Quality checks and iteration} We check the created paragraphs for naturalness and the feasibility of including various features, and iterate on the feature list and annotation instructions. The process is repeated with the refined guidelines.
    \item[4a.] \textbf{Pivot selection} We select $8$ \emph{pivot languages}: English, Mandarin, Hindi, Indonesian, Modern Standard Arabic, Swahili, Spanish, and French. This selection was done with the goal of covering common L2 languages spoken by LRL communities around the world, and as per the availability of professional translators in these languages.
    \item[4b.] \textbf{N-way parallelization} The source paragraphs, including context notes, created in the $5$ source languages are then manually n-way parallellized across the $8$ pivot languages. This is done in two stages: firstly, all source paragraphs are translated into English. The resulting English dataset is then translated into all the pivot languages. Additionally, the contextual information is transcreated into all the pivot languages (e.g., information about grammatical gender of participants is added or removed based how readily available it is in the pivot translation). This forms \datasetnamesource. 
    \item[5a.] \textbf{Selection of LRL target languages} We then select \numberofLRLs~ LRL target languages, based on availability of professional translators, previous coverage in open source initiatives, and language family representativeness. See the list of languages pertaining to \medley in \cref{tab:languages}.
    \item[5b.] \textbf{Translation into LRLs} Finally, we commissioned professional translations of \datasetnamesource into the above LRLs. Translators worked out of the pivot language of their choice. This results in grammatical diverse bitext in our target languages. See guidelines and annotator details for creation and translation in \cref{medley:app:guidelines_creation} and \cref{medley:app:annotator_details}.
\end{enumerate}

\subsection{More details on data creation}
\label{medley:sec:data_creation}

Data creation was done in two major batches of $254$ and $352$ paragraphs respectively, with procedural refinements in Batch 2 based on feedback from linguists from Batch 1 creation process.

\paragraph{Feature enumeration}

As per \cref{medley:sec:gram_div_corpus}, we are interested in common cross-linguistic grammatical features.
Each feature is associated with a meaning that can be cued in any language, regardless of its typology.
We chose 18 grammatical categories and 61 features across them. Of these, 2 features were dropped in Batch 2 due to lack of generalized transfer in translation. See~\cref{medley:tab:gram_features} for a list of features and associated functions.

\paragraph{Template generation}

Each template consists of a random combination of $k$ features such that each feature occurs at least $N=45$ times over all templates. Each feature category may be represented at most twice in a single template.
Each template is then assigned a domain, a source language, and a number of sentences between 2-5, uniformly at random. 
The number of sentences per paragraph is to ensure variation in the length of paragraphs and avoid length artifacts.

We received feedback about the low compatibility of some features with certain domains for Batch 1, so we added constraints to disallow such combinations for Batch 2. E.g., we disallowed dialogue-relevant features such as \ttt{Inclusive/exclusive distinction} for templates with \ttt{narrative, informative, or literary} domain.
We also decreased the maximum number of times a feature could appear in a template from 2 to 1.
Finally, while $k$ was set to $5$ for Batch 1, we reduced it to $4$ for Batch 2, to increase the ease of creation of naturalistic paragraphs for the linguists.

\paragraph{Sentence-level alignments}
We did not require the translators to provide one-to-one sentence translations in order to preserve naturalness of the document-level translations.
However, some translation models may require aligned sentence pairs to train. Therefore, we segment the paragraphs into sentences and align them across languages automatically after dataset creation, and provide this annotation alongside with other metadata for optional use.
See~\cref{medley:app:sentence_alignment} for more details.

\subsection{Features}
\label{medley:app:features}
We list all the grammatical features that we use in corpus creation as described in \cref{medley:sec:approach_details} in \cref{medley:tab:gram_features}. Of these, \ttt{middle voice} and \ttt{suppletion} were dropped in Batch 2.

\setlength{\LTpre}{0pt}
\setlength{\LTpost}{0pt}
\renewcommand{\arraystretch}{1.15}

{\small
\begin{longtable}{@{}p{0.2\textwidth} p{0.3\textwidth} p{0.4\textwidth}@{}}

\caption{Feature Inventory}\\
\toprule
\textbf{Category} & \textbf{Feature} & \textbf{Function} \\
\midrule
\endfirsthead

\toprule
\textbf{Category} & \textbf{Features} & \textbf{Function} \\
\midrule
\endhead

\midrule
\multicolumn{3}{r}{\textit{Continued on next page}}\\
\midrule
\endfoot

\bottomrule
\endlastfoot

Case marking & Nominative case & Marks subject of a clause \\
             & Accusative case & Marks patient or theme \\
             & Genitive case & Marks possession \\
             & Dative case & Marks recipient or experiencer \\
             & Locative or spatial case & Marks location \\
             & Instrumental or comitative case & Marks means, tool, or companion \\
\midrule
Number marking & Singular & Marks one entity \\
               & Plural & Marks more than one entity \\
               & Dual & Marks two entities \\
\midrule
Tense marking & Present tense & Marks current time relative to moment of speaking \\
              & Past tense & Marks previous time relative to moment of speaking \\
              & Future tense & Marks propsective time relative to moment of speaking \\
\midrule
Aspect marking & Perfective aspect & Marks completed event \\
               & Imperfective or progressive aspect & Marks ongoing or incomplete event \\
               & Habitual aspect & Marks repeated or customary events \\
               & Perfect aspect & Marks event as complete at the time of reference \\
\midrule
Mood marking & Indicative mood & Marks statements \\
             & Imperative mood & Marks commands or requests \\
             & Conditional or subjunctive mood & Expresses hypotheticals or counterfactual events \\
\midrule
Evidentiality marking & Evidential marker (direct, reported, inferred) & Marks source of information \\
\midrule
Politeness \& honorifics & Formal or polite form & Marks respect or social distance \\
                          & Informal or casual form & Marks familiarity or solidarity \\
                          & Honorifics or self-humbling used & Marks status of others or self-lowering status \\
\midrule
Voice marking & Active voice & Subject is the agent, doer or experiencer \\
              & Passive voice & Subject is the patient, theme or recipient \\
              & Middle voice & Subject is both the agent and the patient \\
              & Causative construction & Marks a causer acting on a causee to do something \\
\midrule
Valency & Impersonal & No explicit participants \\
                             & Intransitive & One-participant event \\
                             & Monotransitive used & Two-participant event \\
                             & Ditransitive & Three-participant event \\
                             & Intransitive + transitive sequence & Sequence of events involving differing valencies \\
\midrule
Negation marking & Clause-level negation & Marks a negated proposition \\
                 & Negative polarity item & Marks affirmation or negation in a licensing environment \\
                 & Double or emphatic negation present & Reinforces or intensifies negation \\
\midrule
Questions & Polar question & Elicits yes/no answers \\
                   & Wh-question & Elicits specific information \\
                   & Tag, echo or rhetorical question & Elicits agreement, clarification or does not elicit a response \\
\midrule
Subordination & Relative clause & Modifies a nominal referent \\
                             & Complement clause present & Modifies a verb phrase \\
                             & Adverbial clause present & Adds information about time, reason, condition, etc. \\
\midrule
Information structure & Topic marking present & Marks what the utterance is about \\
                      & Focus marking present & Marks new or contrastive information \\
\midrule
Anaphora \& coreference & Personal pronoun & Refers to an aforementioned participants in discourse \\
                         & Reflexive/Reciprocal pronoun & Refers to a participant acting upon itself \\
                         & Null subject or argument & Referent is understood by not overt \\
\midrule
Pronouns \& persons & Inclusive/exclusive distinction & Marks inclusion/exclusion of addressee in first person plural forms \\
                    & Deictic pronoun & Marks space and time relative to the context of the utterance and the speaker \\
                    & Placeholder & Syntactically-integrated filler word to denote a forgotten word or one that the speaker is unsure about \\
\midrule
Coordination & Conjunction & Joins clauses or phrases (e.g., and) \\
                            & Disjunction & Marks alternatives in a clause or phrase (e.g., or) \\
\midrule
Morphosyntactic constructions & Serial verb construction & Single event encoded with 2+ verbs \\
                              & Productive compound & Word with more than one stem which follows most common patterns of word formation \\
                              & Suppletion & Displays distinct roots in different grammatical environments \\
\midrule
Emphasis & Lexical intensifier (e.g., ``very'') & Marks additional emotional context to a modified entity \\
                             & Focus particle (e.g., ``only'', ``even'') & Marks arrowed or restricted scope \\
                             & Emphatic pronoun & Reinforces referent identity \\
                             & Cleft and pseudo-cleft & Emphasizes focus of one or more constitutents with subordination \\
                             & Exclamative construction & Expresses heightened emotion \\
                             & Repetition for emphasis & Reinforces meaning through duplication \\
                             & Marked word order & Highlights information structure or emphasis \\
\label{medley:tab:gram_features}
\end{longtable}
}

\subsection{Guidelines for linguists and translators}
\label{medley:app:guidelines_creation}

We crafted two sets of guidelines: one for writing the source paragraphs in various languages for \datasetnamesource, and one for the commissioned translations of \datasetnamesource into various target languages.

\paragraph{Guidelines for source paragraph creation}

We held a session with the linguists to explain our goals and expectations for the source paragraph creation. We also provided a document explaining the same. In particular, this contained:

\begin{itemize}
    \item Basic instructions explaining the domains, templates, and features: for each template, we asked linguists to craft a paragraph in the assigned domain that contained at least one example of each of the template features. We asked that the paragraph roughly containing the suggested number of sentences.
    \item We emphasized naturalness as a first priority. The linguists were allowed to drop features when including them led to unnaturalness. Similarly, it was acceptable to add sentences if required to accommodate the listed features in a natural way. This resulted in 21 dropped feature instances over all templates (of a total of 2500+ instances). 50 features are covered 45 times over the dataset, 9 features are covered between 38-44 times, and 2 features were dropped after the first batch due to poor generalizability (\cref{medley:sec:approach_details}).
    \item Linguists were also requested to provide any additional context (in English), including any relevant details about the text that would not be readily obvious from the content of the text itself, such as the broader context of the utterance or the genders of the mentioned human referents. This information was collected to inform  text translations and consistency across several language translations.
    \item We also provided a checklist for additional phenomena to include across all translations. For example, we asked linguists to include at least 5 examples of lexical phenomena such as slang, acronyms, and filler words. We also asked them to include examples with human referents of various genders to avoid a gender-biased corpus. For the full checklist see \cref{medley:tab:annotator_checklist}. %

\end{itemize}

\begin{table}[!htp]
\centering
\small
\begin{tabular}{>{\raggedright\arraybackslash}p{0.9\textwidth}}
\toprule
\textbf{Number} \\
\midrule
    - At least 5 quantified expressions (e.g., “some”, “many”, “all”, etc.) \\
    - At least 5 numbers (including digits and spelled out) \\
\midrule
\midrule
\textbf{Gender / Noun classes} \\
\midrule
    - Nouns in all genders of the language \\
    - Personal pronouns in all genders (e.g., Spanish pronouns “nosotras/vosotras”); it includes things like past tense verbs in the feminine in Russian for first and second persons \\
    - Animate and inanimate forms in various semantic/syntactic roles (i.e., subject, object, beneficiary, etc.) \\
\midrule
\midrule
\textbf{Subordination} \\
\midrule
    - Relativized noun in various semantic/syntactic roles (i.e., subject, object, locative, etc.) \\
    - Relative pronoun in various semantic/syntactic roles (i.e., subject, object, possessor, etc.) \\
\midrule
\midrule
\textbf{Morphology} \\
\midrule
    - At least 5 diminutives/augmentatives if available in the language (e.g., piglet, booklet) \\
    - At least 10 examples of productive derivation (e.g., symmetric > asymmetric; inform > information) AND/OR zero conversion / flexible POS use (e.g., “I'm just gonna earthquake everyone out of their misery now”) \\
    - At least 5 examples of opaque/non-predictable compounding (e.g., “redneck”) \\
    - Any examples of reduplication (e.g., “easy-peasy”, “Did you TALK-ABOUT-IT-talk-about-it, or did you just mention it?”) \\
    - Any morphological quirks in the language (e.g., English applicatives like “outrun”) \\
\midrule
\midrule
\textbf{Modifying expressions} \\
\midrule
    - At least 5 modifiers (e.g., adjectives, adverbs, prepositional clauses) \\
    - At least 2 sequences of 2+ modifiers (e.g., adverb + adjective, adjective + adjective + adjective) \\
\midrule
\midrule
\textbf{Lexical variation} \\
\midrule
    - At least 1 idiom \\
    - At least 1 slang term (e.g., “yo!”, “vibe”, “tea” for gossip, “like” for say, etc.) \\
    - At least 1 acronym (e.g., “ETA”) or abbreviation (e.g., “lit.”) \\
    - At least 1 named entity (e.g., “Los Angeles”) \\
    - At least 1 filler (e.g., “uh/uhm”, “like”) \\
    - At least 1 emoji \\
\bottomrule
\end{tabular}
\caption{A global checklist for linguists to include over all source paragraphs for a single source language.}
\label{medley:tab:annotator_checklist}
\end{table}

\paragraph{Guidelines for post-editors and translators}
The above source paragraphs were manually translated into English by the same person that created them in the original pivot language.
We then used these English translations to prepare automatic translations of all source paragraphs into each target pivot language.
These translations were manually post-edited by professional translators, who also transcreated the contextual information. 
By ``transcreated" we mean that the information was not merely translated into the target, but also adapted.
The process of adaptation involves removing information that might no longer apply to the target language, as well as adding information that might be lost in the translation process.

For example, English ``we” is ambiguous in English, as it can have two different readings: ``you and I” (inclusive) and ``I and somebody else but not you” (exclusive). 
Where English has one form with two meanings, Indonesian has two different forms: \textit{kita} for inclusive, and \textit{kami} for exclusive. 
If the translation makes clear what the reading is, translators were informed to delete information about inclusivity in the transcreation process.
Conversely, where English has male and female third person singular pronouns (\textit{he/she}), Indonesian only has one (\textit{dia}). 
If the contextual information does not readily indicate the gender of a participant because it is obvious in the source but that information does copy over into the target, then translators were instructed to add that information in the transcreation process (e.g., ``the third person singular is male”; ``Sam is female”; ``the character is female”). 
Thus, that process ensures that any translations out of target match the original text regardless of the language it was crafted in. 
The result is \datasetnamesource.

\subsection{Annotator details}
\label{medley:app:annotator_details}

\datasetnamesource was translated into \numberofLRLs target languages by professional translators.
We commissioned the translations through third-party vendors, which resourced native-level speakers in our list of target low-resource languages who also had a level of proficiency in the source language equivalent to CEFR C2. 
Translators were able to choose the pivot language to translate from (i.e., English, French, Hindi, Indonesian, Mandarin, Russian, Spanish, and Swahili).  
Translators were provided instructions to pay heed to the contextual information of each paragraph, ensuring that translations were semantically adequate as well as contextually appropriate. 
We expressly forbade the use of AI or any automatic machine translation tools in the translation process.

Translations were checked for format and quality both on the vendor's side and by us. 
Checks included preservation of new lines and paragraph boundaries, digits (where applicable and sensible), emojis, quotes in reported speech, and mark-up style tags in angular brackets. 
Vendors were compensated at market rates.

\subsection{Sentence-level annotations}
\label{medley:app:sentence_alignment}
By construction, \datasetname is multiway parallel at the paragraph level, but we provide additional sentence-level segmentation aligned across all languages, so that the dataset can be viewed as parallel sentences for any pair of languages. The process of extracting this segmentation is described below.

Given a pair of aligned source (in a pivot language) and target paragraphs, we use a neural sentence boundary detector, SaT~\citep{frohmann-etal-2024-segment}, to get character-level sentence boundary probabilities for both. To align the sentence boundaries across languages, we use SONAR-based multilingual text encoder~\citep{Duquenne:2023:sonar_arxiv} to extract contextualized cross-lingual representations of subword tokens. The words are not aligned across languages in a one-to-one or monotonic way, but we expect this to be the case for sentences (or sentence-like structures), so we compute a forced monotonic alignment path across the token representations of two languages using a dynamic time warping algorithm and use this alignment to compare potential locations of sentence boundaries in two languages. The token alignment algorithm chooses a strictly monotonic path (with each token aligned to at most one other token) that maximizes the sum of adjusted cosine similarities of token representations.

Concretely, let $\mathbf{X} \in \mathbb{R}^{m \times d}$ and $\mathbf{Y} \in \mathbb{R}^{n \times d}$ be the SONAR token embeddings for the source and target paragraphs, respectively, and let
\begin{equation}
S_{ij} \;=\; \cos(\mathbf{X}_i, \mathbf{Y}_j)
\end{equation}
be the pairwise cosine similarity matrix. We then compute a dynamic-programming table $\mathbf{C} \in \mathbb{R}^{m \times n}$ of cumulative scores
\begin{equation}
C_{ij} \;=\; \max\Bigl\{
  S_{ij} + C_{i-1,j-1},\;
  C_{i-1,j},\;
  C_{i,j-1}
\Bigr\},
\end{equation}
with appropriate boundary conditions, and backtrack from $(m,n)$ to $(0,0)$ to obtain an optimal monotonic alignment path
\begin{equation}
\mathcal{A}^{\text{src}\rightarrow\text{tgt}} \;=\; \bigl\{(i_k,j_k)\bigr\}_{k=1}^{K}.
\end{equation}
We perform the same procedure in the reverse direction, using $S^\top$, to obtain
\begin{equation}
\mathcal{A}^{\text{tgt}\rightarrow\text{src}} \;=\; \bigl\{(i'_k,j'_k)\bigr\}_{k=1}^{K'},
\end{equation}
where $K$ is the number of tokens of the source paragraph and $K'$ of the target paragraph. Then, we take the intersection $\mathcal{A} = \mathcal{A}^{\text{src}\rightarrow\text{tgt}} \cap \mathcal{A}^{\text{tgt}\rightarrow\text{src}}$ as a set of high-confidence alignment links. The resulting sparse mapping is then densified by linear interpolation to define a total, approximately monotonic mapping from source token indices to target token indices and vice versa.

On the source side, we treat SaT’s character-level probabilities as primary, with the intuition that the SaT probabilities will be more reliable for higher resource pivot languages, and detect peaks to define a set of source character boundaries, which we map to source tokens to obtain token boundaries that we project to the target side via the dense token alignment $\mathcal{A}$, yielding candidate target token boundaries.

Finally, we refine the candidate target sentence boundaries at the character level. For each projected target token boundary, we consider a small character window around the token span and generate a set of candidate split positions with high SaT probability. For each candidate, we compute a combined score
\begin{equation}
\text{score} \;=\; \lambda_{\text{prob}} \cdot p_{\text{tgt}} \;+\; \lambda_{\text{sim}} \cdot s_{\text{bi}},
\end{equation}
where $p_{\text{tgt}}$ is the normalized SaT boundary probability (with an additional bonus for candidates following sentence-final punctuation), and $s_{\text{bi}}$ is a bidirectional semantic similarity term that compares (i) the current source sentence to the candidate left target segment, and (ii) the remaining source text to the candidate right target segment using SONAR sentence embeddings and cosine similarity. The best-scoring candidate in each window is selected as the final target character boundary. This procedure produces a 1:1 sequence of aligned source–target sentence segments with boundaries that are jointly supported by monolingual boundary probabilities, cross-lingual token alignment, and sentence-level semantic similarity.

In post-processing, we optionally enforce length constraints by merging very short aligned sentence pairs and splitting very long ones. When the number of detected boundaries differs between source and target, we either (i) add missing boundaries on the side with fewer sentences by projecting and refining peaks from the other side, or (ii) remove the lowest-confidence boundaries on the side with more sentences, ensuring that the final alignment consists of well-formed, semantically corresponding sentence pairs.

Due to imperfections in sentence boundary detection, cross-lingual token alignment, and the intrinsic variability of human translation (e.g., sentence splits/merges or reorderings), the resulting automatic sentence alignment is not guaranteed to be perfect. To help users identify potentially mismatched or noisy pairs, we provide, for each aligned sentence pair, automatically computed diagnostic scores.
\begin{itemize}
    \item \textbf{Split confidence.} For each language side, we evaluate how confident the SaT model is at the chosen sentence boundaries. For every split position, we look up the SaT boundary probability at the corresponding character and aggregate these values, reporting the mean and minimum confidence as well as the list of per-split confidences. 
    \item \textbf{Length ratios.} To detect structurally implausible alignments, we compute, for each aligned sentence pair, the ratio between the longer and the shorter sentence length (in characters). From these ratios we report the mean, maximum, and the full list of per-pair ratios. Very large ratios may indicate over- or under-segmentation on one side or missing content in the translation.
    \item \textbf{Semantic similarity.} Finally, we measure semantic adequacy of each sentence pair using the same SONAR encoder. For every aligned source–target sentence pair, we compute cosine similarity between their sentence embeddings and report the list of per-pair similarities, their mean and minimum values, and a count of pairs whose similarity falls below a certain threshold (by default, 0.7). Low similarity scores highlight pairs that are likely mistranslated, misaligned, or otherwise noisy.
\end{itemize}
These validation scores are not used to modify the alignment itself, but they provide a convenient way to automatically flag questionable sentence pairs. In downstream applications, they can be used to filter out low-quality alignments (e.g., by discarding pairs with low boundary confidence, extreme length ratios, or low semantic similarity) or to prioritize them for manual inspection.

\subsection{Examples from our dataset}
\label{medley:app:examples_dataset}
In~\cref{medley:tab:app:examples-from-dataset} we report some examples from our dataset, detailing the template used to craft the original text, the context surrounding the text, useful for accurate professional translations, the English translation, the text of the dataset entry in the translated language, and the \emph{route} taken to obtain the translation, i.e. the sequence of translation steps from the original text language to the final text language.

\newcolumntype{Y}{>{\raggedright\arraybackslash}X}

\begin{sidewaystable}[h!]
  \centering
  \scriptsize
  \setlength{\tabcolsep}{3pt}
  \renewcommand{\arraystretch}{1.2}

  \begin{tabularx}{\textwidth}{%
    >{\raggedright\arraybackslash}p{2.5cm}  %
    p{1.1cm}                                 %
    p{1.3cm}                                 %
    >{\raggedright\arraybackslash}p{2.2cm}  %
    Y                                        %
    Y                                        %
    p{1.1cm}                                 %
    Y                                        %
    >{\raggedright\arraybackslash}p{1.5cm}  %
  }
  \toprule
  \textbf{Template} &
  \textbf{Original Language} &
  \textbf{Domain} &
  \textbf{Context} &
  \textbf{Original Text} &
  \textbf{English Translation} &
  \textbf{Language} &
  \textbf{Text} &
  \textbf{Route} \\
  \midrule

  \makecell[tl]{%
    \textbullet~Emphasis and\\
    \quad intensification:\\
    \quad Focus particle\\
    \textbullet~Tense marking:\\
    \quad Past tense verb\\
    \textbullet~Negation marking:\\
    \quad Clause-level\\
    \textbullet~Number marking:\\
    \quad Singular\\
    \textbullet~Valency:\\
    \quad Impersonal used
  } &
  deu\_Latn &
  casual &
  A comment online &
  Ich muss sagen, dass ich überhaupt nicht verstehe, warum sich hier jemand beschwert! Es wird nur gelacht! Und ich finde es sogar irgendwie lustig. So etwas war jedenfalls noch nie da. &
  I have to say, I really don't understand why anyone is complaining here! Everyone's just laughing! And I actually find it kind of funny. Nothing like this has ever happened here before. &
  wol\_Latn &
  Lii mom fokma wahko, wah degue meunuma ndande loutax kounek rek di niahtou fii! Kounek rek di retane! Mandé lii dafa moujeh retanlou sakh ak mane. Lou melni meusoul fi ame. &
  \makecell[tl]{deu\_Latn\\$\downarrow$\\eng\_Latn\\$\downarrow$\\wol\_Latn} \\
  \midrule

  \makecell[tl]{%
    \textbullet~Conjunction and\\
    \quad disjunction:\\
    \quad Conjunction used\\
    \textbullet~Morphosyntactic:\\
    \quad Serial verb constr.\\
    \textbullet~Valency:\\
    \quad Impersonal used\\
    \textbullet~Anaphora:\\
    \quad Reflexive pronoun\\
    \textbullet~Valency:\\
    \quad Intrans. + trans.
  } &
  spa\_Latn &
  instruction-response &
  Interaction between a user (male) and a chatbot to brainstorm ideas for a meeting. The user's partner is referred to with a gender-neutral word---if gender needs to be specified it should be male. &
  ---~Mañana voy a cenar a casa de los padres de mi pareja por primera vez. ¿Qué debería llevar para hacer una buena impresión?\par
  ---~Para causar una buena impresión, puedes llevar postre o una botella de vino si beben alcohol, pero confírmalo con tu pareja primero.\par
  ---~Una botella de vino, ¡qué buena idea! ¿Y qué me pongo? Parece que va a nevar todo el día.\par
  ---~Te recomiendo que vayas arreglado con unos pantalones de vestir y una camisa. Ponte un buen abrigo de invierno para protegerte del frío si va a nevar. &
  ---~Tomorrow I'm going to have dinner at my partner's parents' place for the first time. What should I bring to make a good impression?\par
  ---~To make a good impression, you can bring dessert or a bottle of wine if they drink alcohol, but check with your partner first.\par
  ---~A bottle of wine, what a great idea! And what should I wear? It looks like it's going to snow all day.\par
  ---~I recommend you dress up with dress pants and a shirt. Wear a good winter coat to protect yourself from the cold if it's going to snow. &
  heh\_Latn &
  \_Milau tubita kulya ichakulya cha pamiyee kwa mara ya kwanza ukwao ku vazazi na muyangu. Mele uwushauri.\par
  \_ili ulekesee ipicha nofu, wivesa kusa ni kitindamlo awu ichupa ni divai nda winywa pombe, lavilise ta kwa muyago.\par
  \_Ichupa yinya kinywaj, iliwaso linofu!, nvwale kiki? Yiwoneka witya yitonya theluj isiku mbeyeli.\par
  \_Ngukupendekesa ufwale isuruvali ya heshma ni gwanda, fwale na likoti linofu lya ng'ala ili kuhesa ng'ala kama yiva yitonya theluji. &
  \makecell[tl]{spa\_Latn\\$\downarrow$\\eng\_Latn\\$\downarrow$\\swa\_Latn\\$\downarrow$\\heh\_Latn} \\
  \midrule

  \makecell[tl]{%
    \textbullet~Case marking:\\
    \quad Locative case\\
    \textbullet~Tense marking:\\
    \quad Past tense verb\\
    \textbullet~Mood marking:\\
    \quad Imperative mood\\
    \textbullet~Valency:\\
    \quad Intransitive used\\
    \textbullet~Valency:\\
    \quad Intransitive used
  } &
  spa\_Latn &
  narrative &
  Beginning of a coming-of-age novel that introduces the main character, Marcos (male). The public servant mentioned is also male. &
  Llevaba una hora en secretaría. Marcos se había tenido que tomar la mañana libre para ir hasta la facultad y el funcionario de turno era más lento que una tortuga. Quería acercarse a la ventanilla y gritarle «¡Venga, va, apresúrate, que voy a llegar tarde al trabajo!», pero sabía que esto solo lo enfurecería y lo haría trabajar todavía más lento. Inhaló profundamente y se convenció de que llegaría a tiempo. &
  He had been at the administration office for an hour. Marcos had had to take the morning off to go to the faculty, and your typical civil servant was as slow as a snail. He wanted to approach the counter window and scream "Come on, hurry up, I'm going to be late for work!", but he knew this would only make him angry and make him work even more slowly. He took a deep breath and convinced himself that he would arrive on time. &
  akb\_Latn &
  Madung di kantor pangurusan ma ia sa jom on. Marcos ingkon mangido izin tarlambat tu ganan parharejoan niai dungi songon na biaso, pagawe negeri karejo ama na santing lamabt songon dalkop-dalkop. Giot i padonok ia tu tingkap ni konter sambil marungut "Apebo, paipas, tarlambat ma au harejo!", Tai binoto ia do on um na mambahen ia mangamuk dot martamba lolot harejo nia. i sirup ia hosa na nia bagas dohot mayakinkon iba nia bahaso na angkan sampe ia pas di waktu nai. &
  \makecell[tl]{spa\_Latn\\$\downarrow$\\eng\_Latn\\$\downarrow$\\ind\_Latn\\$\downarrow$\\akb\_Latn} \\
  \bottomrule
  \end{tabularx}

  \caption{Examples from our dataset. We can see the feature template used to craft the original text, the original text itself and its language, the domain it belongs to, the english translation of the original text, the text of the dataset entry, the language it has been translated to, and the route used to obtain this translation. For instance \emph{spa\_Latn $\rightarrow$ eng\_Latn $\rightarrow$ ind\_Latn $\rightarrow$ akb\_Latn} means that the text was originally created in Spanish, then translated into English, then into Indonesian, and finally into Batak.}
  \label{medley:tab:app:examples-from-dataset}
\end{sidewaystable}

\clearpage
\subsection{Grammatical feature analyses}
\label{medley:app:grammatical_feature_analyses}

\subsubsection{Feature distribution analysis}
\begin{table}[!htb]
\centering
\small
\begin{tabular}{lcccc}
\toprule
\textbf{Paradigm} & \textsc{SmolSent} & \textsc{SmolDoc} & \textsc{\nllbseed} & \datasetname~ \\
\midrule
Case     & \textbf{1.01} & 0.92 & 1.00 & 0.92 \\
Number   & 0.52 & \textbf{0.59} & 0.58 & 0.54 \\
Person   & 0.86 & 0.81 & 0.13 & \textbf{0.97} \\
Tense    & 0.80 & 0.79 & 0.69 & \textbf{0.83} \\
Aspect   & 0.93 & 1.16 & 0.98 & \textbf{1.27} \\
Mood     & \textbf{0.43} & 0.14 & 0.02 & 0.29 \\
Voice    & 0.45 & 0.37 & \textbf{0.53} & 0.35 \\
Valency  & 1.29 & 1.29 & 1.22 & \textbf{1.33} \\
Negation & 0.32 & 0.38 & 0.37 & \textbf{0.62} \\
\bottomrule
\end{tabular}
\caption{Entropy over paradigm distributions over features for different datasets (bold indicates highest value per paradigm).}
\label{tab:feature_entropy_eng}
\end{table}

\begin{table}[!htb]
\centering
\small
\begin{tabular}{llcccc}
\toprule
\textbf{Paradigm} & \textbf{Feature} & \textsc{SmolSent} & \textsc{SmolDoc} & \textsc{\nllbseed} & \textsc{\datasetname} \\
\midrule
case & Accusative & 0.16 & 0.16 & 0.15 & 0.23 \\
 & Genitive & 0.35 & 0.21 & 0.34 & 0.15 \\
 & Nominative & 0.49 & 0.63 & 0.51 & 0.62 \\
number & Dual & 0.00 & 0.00 & 0.00 & 0.00 \\
 & Plural & 0.21 & 0.27 & 0.25 & 0.21 \\
 & Singular & 0.79 & 0.73 & 0.75 & 0.79 \\
person & P1 & 0.18 & 0.21 & 0.03 & 0.32 \\
 & P2 & 0.15 & 0.10 & 0.00 & 0.14 \\
 & P3 & 0.67 & 0.69 & 0.97 & 0.54 \\
tense & Future & 0.03 & 0.03 & 0.00 & 0.04 \\
 & Past & 0.42 & 0.47 & 0.59 & 0.40 \\
 & Present & 0.55 & 0.51 & 0.40 & 0.55 \\
aspect & Imperfective & 0.45 & 0.32 & 0.30 & 0.34 \\
 & Perfect & 0.04 & 0.08 & 0.10 & 0.14 \\
 & Perfective & 0.49 & 0.49 & 0.58 & 0.39 \\
 & Progressive & 0.03 & 0.11 & 0.02 & 0.13 \\
mood & Conditional & 0.00 & 0.00 & 0.00 & 0.00 \\
 & Imperative & 0.15 & 0.03 & 0.00 & 0.09 \\
 & Indicative & 0.85 & 0.97 & 1.00 & 0.91 \\
 & Subjunctive & 0.00 & 0.00 & 0.00 & 0.00 \\
voice & Active & 0.87 & 0.90 & 0.80 & 0.91 \\
 & Causative & 0.02 & 0.02 & 0.01 & 0.03 \\
 & Passive & 0.11 & 0.08 & 0.20 & 0.06 \\
valency & Ditransitive & 0.11 & 0.15 & 0.25 & 0.13 \\
 & Impersonal & 0.31 & 0.14 & 0.09 & 0.23 \\
 & Intransitive & 0.20 & 0.36 & 0.17 & 0.30 \\
 & Monotransitive & 0.38 & 0.35 & 0.49 & 0.35 \\
 & Other & 0.00 & 0.00 & 0.00 & 0.00 \\
negation & ClauseNeg & 0.07 & 0.08 & 0.07 & 0.14 \\
 & DoubleNeg & 0.01 & 0.01 & 0.01 & 0.02 \\
 & NPI & 0.00 & 0.01 & 0.01 & 0.02 \\
 & None & 0.92 & 0.90 & 0.91 & 0.81 \\
\bottomrule
\end{tabular}
\caption{Feature distributions per paradigm and dataset. Cell values are proportions of that feature given the paradigm.}
\label{tab:feature_distributions_full}
\end{table}

Given the aim of covering naturally rarer features in our dataset, we compare grammatical feature distributions in our dataset versus others. 
We look at several paradigms such as tense, aspect, formality, among others, and look at the entropy of the distribution over various features in each paradigm in each dataset for English (see \cref{tab:feature_entropy_eng}. See \cref{tab:feature_distributions_full} for the full distributions over features.). We label features automatically using a Stanza parser \citep{qi-etal-2020-stanza} as well as some heuristics in cases where Stanza annotation was not rich enough for the target feature.\footnote{We are limited to languages that lie in the intersection of our considered datasets, for which we also have parsers and linguistic expertise. English was the only such language.} A higher entropy signals a more balanced distribution over the paradigm.

We find that \nllbseed, which is Wikipedia domain text, unsurprisingly shows high concentrations of past tense, indicative mood, third person features (low entropies for tense, mood, and person). The other datasets are more balanced, with \datasetname showing the highest entropy in 5 of 9 categories. \datasetname often has higher proportions of rarer features in a paradigm, such as first person text, the perfect aspect, or clausal negation.

We also automatically translated \textsc{SmolSent} and \nllbseed to Hindi and repeated this analysis. Note that the translations produced as a result may affect our findings.
We find that for Hindi, as for English, \nllbseed shows distributions representative of Wikipedia, e.g. almost no second person pronouns, no informal pronouns, only indicative mood.
\textsc{SmolSent} and \datasetname are more diverse, with the latter often showing the highest distribution entropy (i.e. most diversity). For example, it has a more balanced distribution over tense, verbal valencies, as well as negation types.
See~\cref{medley:tab:paradigm-features} for a list of studied features for Hindi, and \cref{medley:tab:feature_entropy_hin} for the entropies of the paradigm distributions. 

\begin{table}[!htb]
\centering
\small
\begin{tabular}{l p{6cm}}
\toprule
\textbf{Paradigm} & \textbf{Features} \\
\midrule

Number    & Singular, Plural \\
Formality & Formal, Informal \\
Tense     & Present, Past, Future \\
Aspect    & Habitual, Imperfective, Perfect, Perfective \\
Mood      & Conditional, Imperative, Indicative, Subjunctive \\
Valency   & Impersonal, Intransitive, Monotransitive, Ditransitive \\
Negation  & Clausal Negation, Double Negation, Negative Polarity Item, No Negation \\

\bottomrule
\end{tabular}
\caption{Feature types contained in each paradigm for Hindi case study.}
\label{medley:tab:paradigm-features}
\end{table}

\begin{table}[!htb]
\centering
\small
\begin{tabular}{lccc}
\toprule
\textbf{Paradigm} & \datasetname & \textsc{SmolSent} & \textsc{\nllbseed} \\
\midrule

Formality & \textbf{0.60} & 0.07 & 0.00 \\
Tense     & \textbf{0.79} & 0.67 & 0.64 \\
Aspect    & \textbf{1.22} & 1.19 & 1.13 \\
Valency   & \textbf{1.28} & 1.17 & 1.18 \\
Negation  & \textbf{0.78} & 0.60 & 0.53 \\
Number    & 0.43 & 0.41 & \textbf{0.46} \\
Mood      & 0.33 & \textbf{0.47} & 0.09 \\

\bottomrule
\end{tabular}
\caption{Entropy of feature paradigms for Hindi across datasets (bold indicates highest entropy per paradigm).}
\label{medley:tab:feature_entropy_hin}
\end{table}

\subsubsection{Feature retention study}

\begin{table}[!htb]
\centering
\small
\begin{tabular}{lcccc}
\toprule
\textbf{Feature} & \textbf{rus-eng} & \textbf{rus-eng-spa} & \textbf{spa-eng} & \textbf{spa-eng-rus} \\
\midrule
Dative & 46.7 & 60.0 & 55.6 & 22.2 \\
Instrumental & 50.0 & 50.0 & 92.3 & 69.2 \\
Past tense & 96.6 & 89.7 & 94.1 & 91.2 \\
Passive & 88.9 & 100.0 & 81.8 & 72.7 \\
Imperative & 100.0 & 100.0 & 100.0 & 100.0 \\
Ditransitive & 91.7 & 91.7 & 68.8 & 68.8 \\
Marked word order & 17.6 & 23.5 & 0.0 & 33.3 \\
Compound & 88.9 & 66.7 & 85.7 & 88.9 \\
Lex. intensifier & 56.2 & 56.2 & 87.5 & 87.5 \\
Evidentiality & 80.0 & 80.0 & 100.0 & 100.0 \\
\bottomrule
\end{tabular}
\caption{Percentage of transferred feature for 10 features, starting with Spanish or Russian as source languages, and for 1 or 2 hops. }
\label{tab:preservation-directions}
\end{table}

We are also interested in measuring the extent of feature transfer with our approach: i.e., the extent to which we can cue a feature in a source language (regardless of its typology) via its underlying function and have the target form in an arbitrary target surface when the text is translated to that language. To study the retention of features across translation, we analyzed 10 features in the 120 paragraphs originally created in Russian and Spanish, their translations into English, and their 2-hop translations into Spanish and Russian (i.e., the Russian-to-English-to-Spanish and Spanish-to-English-to-Russian translations). 
The features included: dative and instrumental case, past tense, passive voice, imperative mood, ditransitive constructions, marked word order, compound words, lexical intensifiers, and evidentiality.   
These features were chosen to target different levels of linguistic structure, from morphology to morphosyntax, to information structure. 
Evidentiality was included to investigate a known feature that is not overtly marked morphosyntactically in either pivot language but rather through various lexical choices. 

We are also interested in whether these features are preserved across translation hops; thus, we also look at feature transfer for the same paragraphs created via 2-hop translation in our pipeline (Spanish-English-Russian and vice versa).

\paragraph{Results and summary of findings} See \cref{tab:preservation-directions} for this analysis.
Broadly, we find that most morphosyntactic features have decent transfer rates ($>50\%$), with the exception of marked word order.
Crucially, we find that forms that are ``lost'' (i.e. do not surface in a target translation) can resurface in next hop from that language.
For example, although marked word order is lost in the Russian-to-English translation, some instances of marked word order resurface in Spanish with the 2-hop Russian-English-Spanish translation. 
Such resurfacing likely occurs because both Spanish and Russian share pragmatic uses of word order that are utterance dependent and, thus, are not dependent on the form not appearing in English.\footnote{This loss and resurfacing of marked word order is not entirely surprising, since English has 1) more rigid word order and 2) fewer available patterns of marked word order than both Russian and Spanish. However, our approach shows that even when English lacks some forms, they can still be cued into a target translation of out English.}

\paragraph{Analysis in more detail}
The results show that the most robust feature is the imperative, which is always copied both in 1- and 2-hop translations. 
All three languages have mechanisms to convey commands and requests (as most, if not all, languages), therefore the close mapping is to be expected. 
Overall, features that are prototypically marked as verbal morphology (i.e., past tense, passive voice, imperative mood) have the highest retention rates. 
The slightly lower retention for passives out of Spanish are due to two factors: the so-called Spanish ``pasiva refleja'' does not have a clear equivalent in English and is hard to recover in Russian (\ref{pasiva-refleja}), and some passive readings in Spanish and English can be expressed with marked word order in Russian (\ref{passive-to-MWO}).

{
\small
\begin{exe}
    \ex \label{pasiva-refleja} \begin{xlist}
        \ex \gll Spanish: En esta empresa \textbf{se} \textbf{trabaja} muy bien. \\
        {} in \Dem{} company \Pass{} work.\Prs.\Tsg{} very well \\
        \ex \gll Russian: \rus{Работать} \rus{в} \rus{этой} \rus{компании} \rus{очень} \rus{приятно}. \\
        {} work.\Inf{} in \Dem.\Loc{} company.\Loc{} very nice \\ 
        \ex English: It's very nice working in this company. \\
    \end{xlist}

    \ex \label{passive-to-MWO} \begin{xlist}
        \ex \gll Spanish: La comida \textbf{fue} \textbf{preparada} por nuestra chef invitada, la señora Gómez. \\
        {} the food be.\Pst.\Tsg{} prepare.\Ptcp.\F{} by our chef guest the Mrs. Gómez \\
        \ex \gll Russian: \rus{Еду} \rus{готовила} \rus{наша} \rus{гостья} \rus{шеф-повар}, \rus{миссис} \rus{Гомес}. \\
        {} food.\Acc{} prepare.\Pst.\F{} our guest chef Mrs. Gómez \\ 
        \ex English: The food \textbf{was prepared} by our guest chef, Mrs. Gómez. \\
    \end{xlist} 
\end{exe}
}

The preservation of case shows many idiosyncrasies, which is to be expected considering that Russian does have morphological case, while English and Spanish only have remnants of a case system.
For Spanish and English, the notion of ``instrumental'' case is restricted to uses of the preposition ``con'' and ``with'', respectively; but in Russian this case can also mark agents in passive constructions or attributes in copular or copular-like constructions, which do not elicit these prepositions in English or Spanish. 
Similarly, dative case in Spanish in pronominal verbs or the so-called ``ethic dative'' constructions may not always have clear equivalents in English or Russian. 

{
\small
\begin{exe}
    \ex \label{rus-ins} \begin{xlist}
        \ex \gll Russian: \rus{Тогда} \rus{ей} \rus{подойдет} \rus{человек}, \rus{который} \rus{уже} \rus{преподавал} \rus{в} \rus{нескольких} \rus{школах} \rus{или} \rus{работал} \textbf{\rus{завучем}}.\\
        {} then \Tsg.\Dat{} suit.\Prs.\Tsg{} person who.\Nom{} already teach.\Pst.\M{} in few.\Pl.\Loc{} school.\Pl.\Loc{} or work.\Pst.\M{} deputy.\Ins{} \\ 
        \ex \gll Spanish: Entonces, le gustaría alguien que ya haya enseñado en varias escuelas o haya sido subdirector. \\
        {} then \Tsg.\Dat{} like.\Cond.\Tsg{} someoone who already has.\Pst.\Sjv.\Tsg{} teach.\Ptcp{} in several.\Pl{} school.\Pl{} or has.\Pst.\Sjv.\Tsg{} be.\Ptcp{} deputy \\
        \ex English: Then someone who has taught in several schools or worked as a deputy principal would suit her. \\
    \end{xlist}

    \ex \label{spa-dat} \begin{xlist}
        \ex \gll Spanish: [..] ahora \textbf{te} los tienes que fumar igual aunque pagues. \\
        {} {} now \Ssg.\Dat{} \Tpl.\Acc{} have.\Prs.\Ssg{} to smoke.\Inf{} equally even pay.\Prs.\Sjv.\Ssg \\
        \ex \gll Russian: [...] \rus{теперь} \rus{приходится} \rus{мириться} \rus{с} \rus{ней}, \rus{даже} \rus{если} \rus{платишь} \\
        {} {} now have\_to.\Prs.\Tsg{} reconcile.\Inf{} with \Tsg.\Ins{} even if pay.\Prs.\Ssg \\ 
        \ex English: [...] now you have to put up with them even if you pay. \\
    \end{xlist} 
\end{exe}
}

The examples above show that features may be lost in the first hop, as in (\ref{pasiva-refleja}) and (\ref{rus-ins}). 
However, they can also redistribute to other rare features of interest, like marked word order in (\ref{passive-to-MWO}) and instrumental case in (\ref{spa-dat}). 
Additionally, features that are lost in the first hop can reappear in the second, as shown in (\ref{rus-MWO}-\ref{spa-MWO}), in which post-verbal subjects (i.e., ``a car'', ``two dogs'', ``several rumors'') are attested in the 2-hop in Spanish and Russian but not in the 1-hop in English.

{
\small
\begin{exe}
    \ex \label{rus-MWO} \begin{xlist}
        \ex \gll Russian: \rus{Солнечные} \rus{лучи} \rus{только} \rus{осветили} \rus{комнату}, \rus{как} \rus{по} \rus{улице} \rus{проехала} \rus{машина}; \rus{за} \rus{ней} \rus{бежали} \rus{две} \rus{собаки}.\\
        {} solar.\Pl.\Nom{} ray.\Pl.\Nom{} only light\_up.\Pst.\Pl{} bedroom.\Acc{} as along street.\Dat{} pass.\Pst.\F{} car.\Nom{} behind \Tsg.\Ins{} run.\Pst.\Pl{} two.\Nom{} dog.\Gen{} \\ 
        \ex \gll Spanish: Los rayos del sol acababan de iluminar la habitación cuando pasó un coche por la calle; detrás de él corrían dos perros. \\
        {} The ray.\Pl{} of.the sun finish.\Pst.\Tpl{} of light\_up.\Inf{} the bedroom when pass.\Pst.\Tsg{} a car on the street behind of \Tsg{} run.\Pst.\Tpl two dog.\Pl{} \\
        \ex English: The sun's rays had just lit up the room when a car drove down the street; two dogs ran after it.  \\
    \end{xlist}

    \ex \label{spa-MWO} \begin{xlist}
        \ex \gll Spanish: Ayer aparecieron varios rumores que La Oreja de Van Gogh podría disolverse. \\
        {} yesterday appear.\Pst.\Tpl{} several.\Pl{} rumors.\Pl{} that La Oreja de Van Gogh might.\Cond.\Tsg{} dissolve.\Inf{}\\
        \ex \gll Russian: \rus{Вчера} \rus{появилось} \rus{несколько} \rus{слухов} \rus{о} \rus{том}, \rus{что} La Oreja de Van Gogh \rus{может} \rus{распасться}. \\
        {} yesterday appear.\Pst.\N{} few rumors.\Gen{} about \Dem.\Loc{} that La Oreja de Van Gogh might.\Prs.\Tsg{} break\_up.\Inf{}  \\ 
        \ex English: Yesterday, several rumors appeared that La Oreja de Van Gogh might disband. \\
    \end{xlist} 
\end{exe}
}

Taken together, the results from the qualitative analysis suggests that 1) some features will naturally be lost in the translation process into LRLs, since languages differ in terms of the forms and functions they codify in their grammars; 2) some features that are lost in the translation process can cue other, equally-rare linguistic features; and 3) some features that do not appear in a translation hop can resurface in a subsequent hop, especially if the two languages share a similar use of that feature (even if the intermediary language does not).

\clearpage
\subsection{More details on datasets and language statistics}
\label{medley:app:experiment_dataset_details}

\begin{figure}[h]
    \centering
    \includegraphics[width=\linewidth]{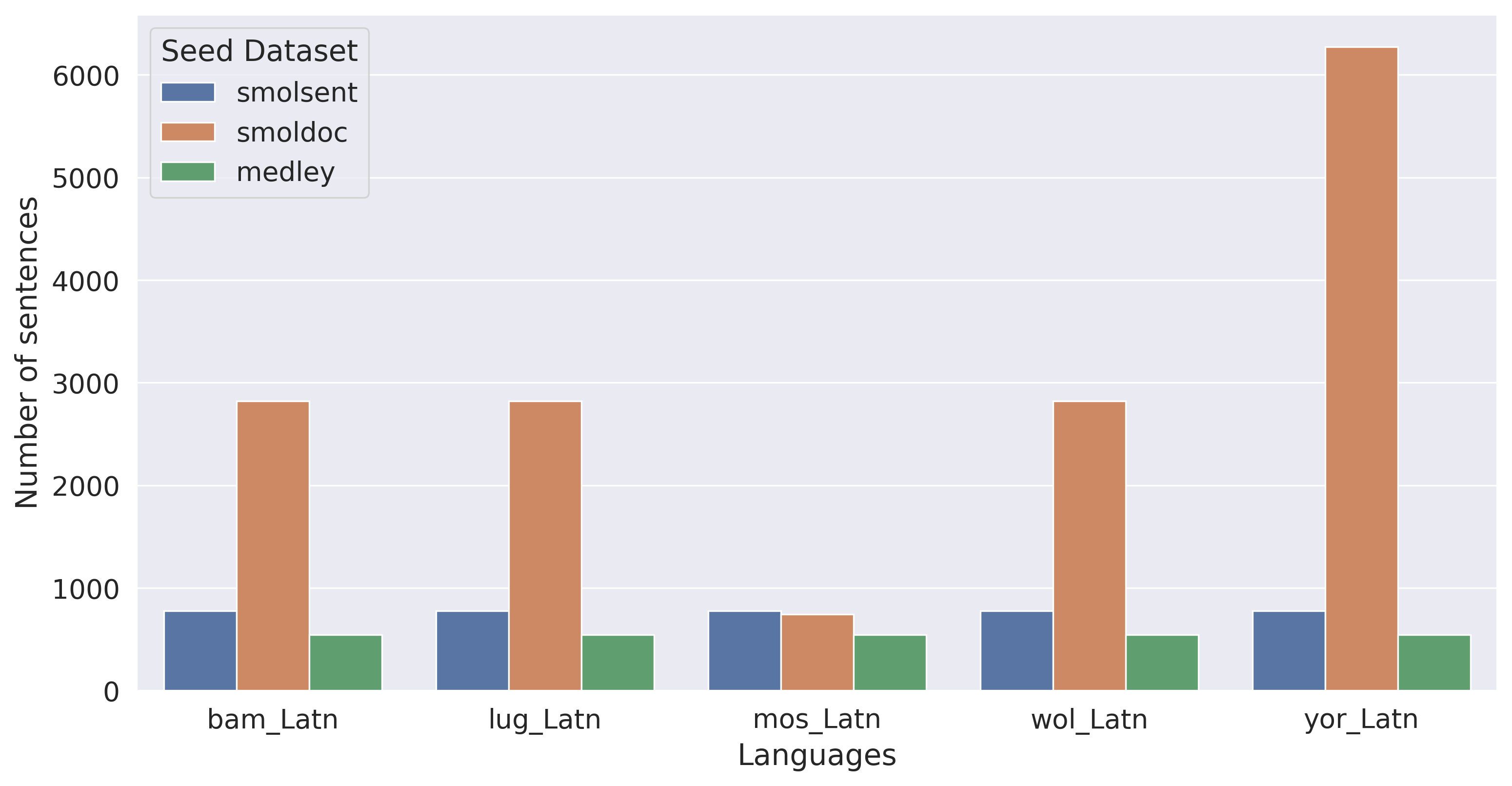}
    \caption{Number of examples per language across seed datasets.}
    \label{fig:app:experiment_dataset_details:num_examples}
\end{figure}

\begin{figure}[h]
    \centering
    \includegraphics[width=\linewidth]{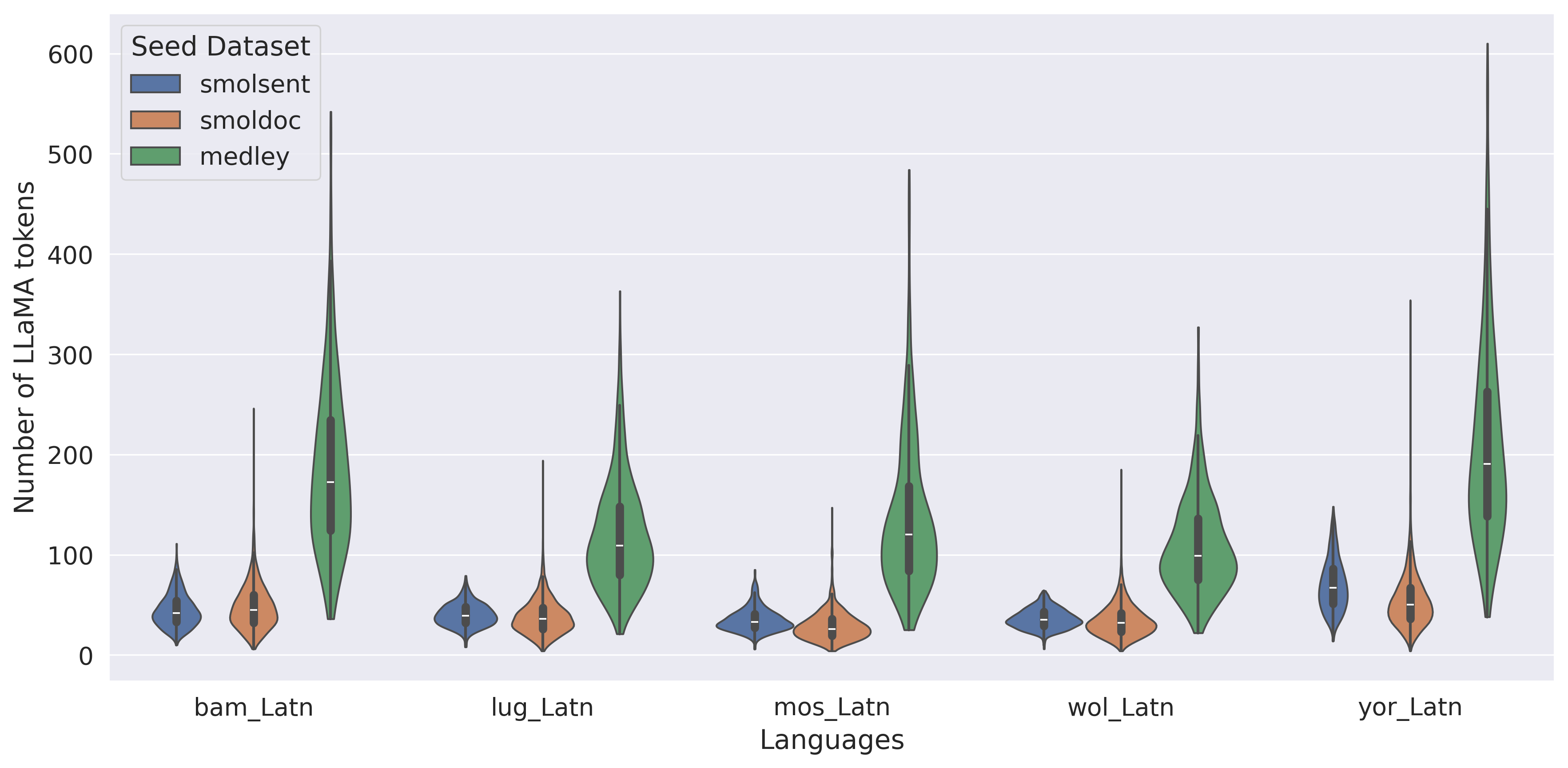}
    \caption{Token length distribution per language across seed datasets.}
    \label{fig:app:experiment_dataset_details:num_tokens_before}
\end{figure}

\begin{figure}[h]
    \centering
    \includegraphics[width=\linewidth]{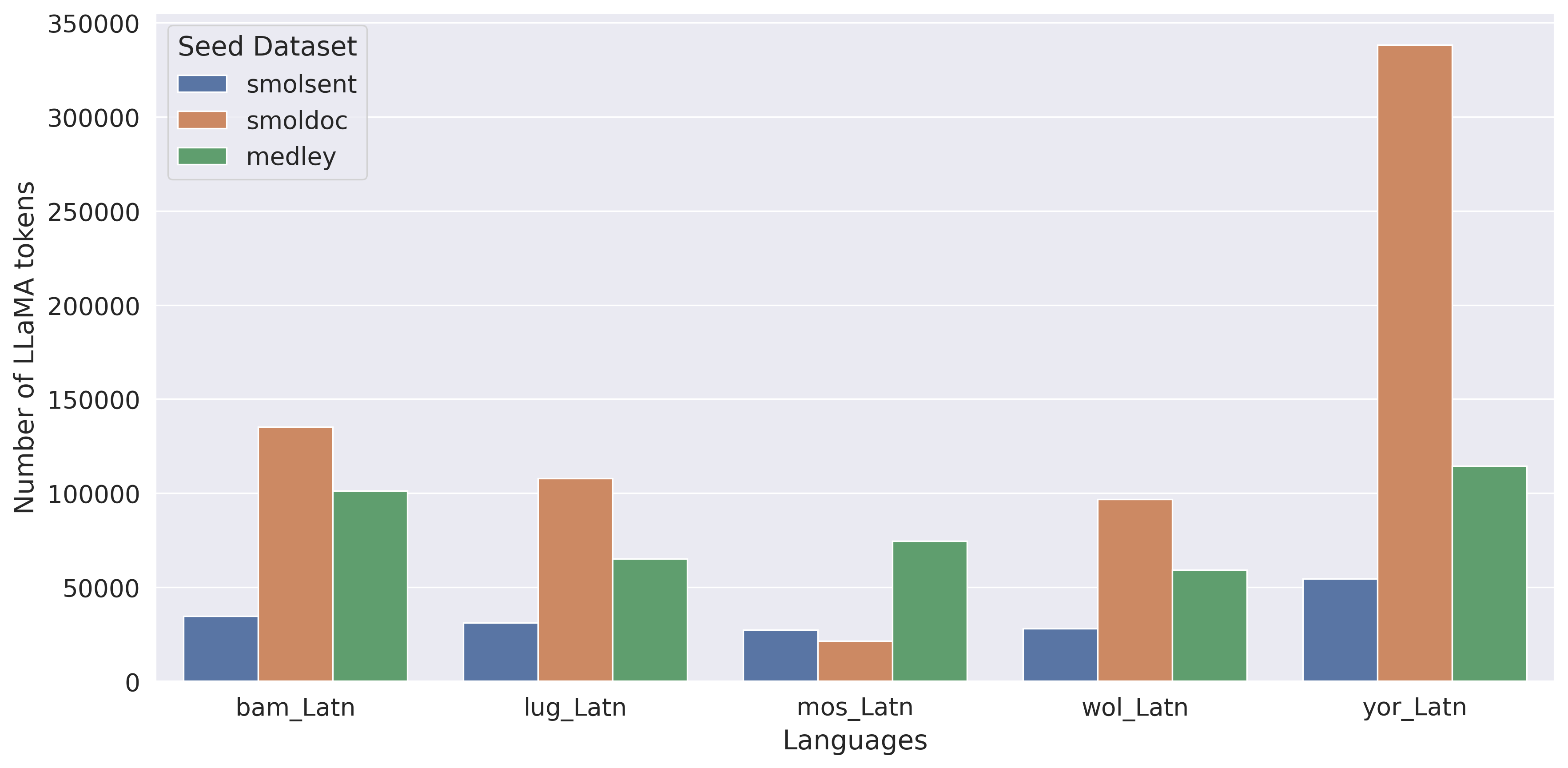}
    \caption{Total number of tokens per language across seed datasets.}
    \label{fig:app:experiment_dataset_details:tot_num_tokens_before}
\end{figure}

Here we provide statistics about the datasets and language we leverage to conduct our experiments. We experiment on into English and out of English directions, considering five low resource languages: Bambara, Ganda, Mossi, Wolof, Yoruba\footnote{\href{https://glottolog.org/resource/languoid/id/bamb1269}{bam\_Latn}, \href{https://glottolog.org/resource/languoid/id/gand1255}{lug\_Latn}, \href{https://glottolog.org/resource/languoid/id/moss1236}{mos\_Latn}, \href{https://glottolog.org/resource/languoid/id/nucl1347}{wol\_Latn}, \href{https://glottolog.org/resource/languoid/id/yoru1245}{yor\_Latn}}. These are the languages for which all the seed datasets we evaluate, and the evaluation datasets, contain examples. Note that given the different nature of the three seed datasets, number of examples and number of tokens vary across them, as can be seen in~\cref{fig:app:experiment_dataset_details:num_examples}. 
In particular, \datasetname tends to have longer examples, while SmolSent and SmolDoc tend to have shorter sequences, in terms of number of tokens, see~\cref{fig:app:experiment_dataset_details:num_tokens_before}. 
Furthermore, SmolDoc is much larger than the other two as can be seen in~\cref{fig:app:experiment_dataset_details:tot_num_tokens_before}. These discrepancies motivated as to perform the token-controlled experiments, sampling  from the three sources while sticking to a per-language shared fixed budget of tokens, determined by the seed dataset containing the least number of tokens for that language.

\clearpage
\subsection{More details on the experimental setup}
\label{medley:app:experiment_setup_details}

We experiment with two popular MT paradigms: \textsc{\nllb-200-3.3B}\footnote{\href{facebook/nllb-200-3.3B}{https://huggingface.co/facebook/nllb-200-3.3B}} as a representative of sequence-to-sequence (seq2seq) models~\citep{nllb-24}, and \textsc{\llama-3.1-8B-Instruct}\footnote{\href{meta-llama/Llama-3.1-8B-Instruct}{https://huggingface.co/meta-llama/Llama-3.1-8B-Instruct}} representing LLM-based MT, or autoregressive decoder-only instruction-following models~\citep{grattafiori2024llama}. For \nllb, tokenized source text is fed to the model, with appropriate delimiting language tags, and translations are obtained via beam search decoding, whereas for \llama, we use a minimal prompt template instructing the model to translate from source to target taken from~\citet{alves2024tower}.

We consider into-English and out-of-English directions per language, and fine-tune and evaluate models for each language pair and direction separately.
We train both models in a supervised way by maximizing the log-likelihood of the target sequence given the source, with teacher forcing. We train for a fixed number of epochs, monitoring the model performance on a held-out validation set to pick the best checkpoint. For training \nllb, we use the same hyperparameters reported in~\citet{nllb-24} for various fine-tuning experiments, while for \llama~we adopt a similar setup to~\citet{caswell2025smol}. For \nllb, we dynamically pad tokenized source sequences to the longest in the batch. For \llama, we apply packing, merging tokenized prompts together when possible to minimize padding.

We experiment with 5 languages that are covered by these baseline datasets, \datasetname-\numberofLRLs, as well as our evaluation datasets: Bambara, Mossi, Wolof, Yoruba, and Ganda.
Since \textsc{SmolDoc} samples can range to over several thousand tokens, we break it into sentence-level chunks with the provided sentence alignments in accordance with \citet{caswell2025smol}. See~\cref{medley:app:experiment_dataset_details} for training dataset statistics.

We perform supervised fine-tuning leveraging ``labelled'' datasets, i.e. bitext corpora (the seed datasets), in which each example in these datasets is a piece of text (source text), of known language (source language), along with its translation (target text) in a give language (target language). We then work with datasets $\{(sl_i, tl_i, st_i, tt_i)_{i=1}^N\}$ where $sl_i$ is a representation of the source language, $tl_i$ is a representation of target language, $st_i$ is a representation of the source text, and $tt_i$ is a representation of the target text. These representations differ slighly across the two models used to solve the task. For NLLB, we encode them as can be seen in~\boxref{box:experiment_setup_details:nllb}, since the NLLB tokenizer reserves special tokens for representing language codes of the supported languages. Note that all of the languages we evaluate on are (technically) supported by NLLB, meaning that the tokenizer has a reserved language code token for them. For LLaMa, given prior work~\citep{grattafiori2024llama} highlighting the good performance of instruction-following LLMs at translation tasks, we perform instruction fine-tuning, with a very simple prompt template and task description, as can be seen in~\boxref{box:experiment_setup_details:nllb}.

\noindent\begin{minipage}{\textwidth}
\begin{tcolorbox}[]
\begin{verbatim}
  Source: <{source language code}> {source text} </s> <pad> ... <pad>
  Target: <{target language code}> {target text} </s> <pad> ... <pad>
\end{verbatim}
\end{tcolorbox}
\captionof{boxes}{Example encoding of a translation pair for fine-tuning NLLB }
\label{box:experiment_setup_details:nllb}
\end{minipage}

\noindent\begin{minipage}{\textwidth}
\begin{tcolorbox}[]
\begin{verbatim}
  Translate the following text from {source language} into {target language}.
  {source language}: {source text}
  {target language}: 
\end{verbatim}
\end{tcolorbox}
\captionof{boxes}{Prompt template for fine-tuning LLaMa}
\label{medley:box:experiment_setup_details:llama}
\end{minipage}

All fine-tuning experiments and subsequent evaluations are conducted on A100 GPUs\footnote{\url{https://www.nvidia.com/en-gb/data-center/a100/}}, using the HuggingFace transformers~\citep{wolf2020huggingfacestransformersstateoftheartnatural} implementations. The models are efficiently fine-tuned using mixed precision training~\citep{micikevicius2017mixed}, with dynamic padding for NLLB (padding to the longest sequence in the batch) for NLLB, and using packing\footnote{\url{https://huggingface.co/docs/trl/main/en/sft_trainer\#packing}}, FlashAttention-2~\citep{dao2023flashattention} and FSDP\footnote{\href{https://engineering.fb.com/2021/07/15/open-source/fsdp/}{Fully Sharded Data Parallel}} for LLaMa, using the AdamW optimizer~\citep{loshchilov2019decoupledweightdecayregularization}. Sequences longer than the model max length are truncated from the left (keeping the source language code for conditioning) for NLLB, while from the right for LLaMa. All training hyperparameters are detailed in~\cref{table:experiment_setup_details:hparams:train}.

\begin{table}[h]
\centering
\small
\begin{tabular}{@{}llll@{}}
\toprule
Hyperparameter    & NLLB         & LLaMa            & Notes              \\ \midrule
Epochs            & 10           & 10               &                    \\
Batch Size        & 8            & 1                & \llama uses packing \\
Max Num Tokens    & 512          & 1024             &                    \\
Learning Rate     & 5e-5         & 1e-5             &                    \\
Optimizer         & AdamW        & AdamW            &                    \\
Learning Rate Schedule      & Inverse Square Root & Cosine Annealing &  With 10\% of warmup steps                  \\
Gradient Clipping & 1.0          & 1.0              &                    \\
Weight Decay      & 1e-2         & 1e-2             &                    \\
AdamW Betas       & (0.9, 0.95)  & (0.9, 0.95)      &                    \\ \bottomrule
\end{tabular}
\caption{Training Hyperparameters}
\label{table:experiment_setup_details:hparams:train}
\end{table}

Then, to efficiently produce inference results from the resulting checkpoints for evaluation , we leverage ctranslate2~\citep{klein2020efficient} for NLLB and vLLM~\citep{kwon2023efficient} for LLaMa. All inference hyperparameters are listed in~\cref{table:experiment_setup_details:hparams:eval}.

\begin{table}[]
\centering
\small
\begin{tabular}{@{}llll@{}}
\toprule
Hyperparameter     & NLLB        & LLaMa  & Notes       \\ \midrule
Decoding           & Beam Search & Greedy &             \\
Beam Width         & 5           & N/A    &             \\
Max Num Tokens     & 512         & 1024   &             \\
Temperature        & 0.0         & 0.0    &             \\
Top-P              & 1.0         & 1.0    & No Sampling \\
Top-K              & N/A         & N/A    & No Sampling \\
Repetition Penalty & 1.0         & 1.0    &             \\ \bottomrule
\end{tabular}
\caption{Inference hyperparameters}
\label{table:experiment_setup_details:hparams:eval}
\end{table}

\subsection{More details on the experiment results}
\label{app:experiment_results_details}

We observe the benefits of any seed data over the no-seed baseline, especially for \llama. This is true to a lesser extent for \textsc{\nllb}, which has already been fine-tuned for these languages.\footnote{Note that \datasetname covers 79 languages that are not supported by \nllb, for which we can expect higher gains over the baseline. However, we lack the evaluation resources in these languages to demonstrate this.} \textsc{NLLB} shows higher baseline and fine-tuned performance across the board, highlighting that smaller sequence-to-sequence models are still state-of-the-art for low-resource MT as compared to LLM-based MT in accordance with previous findings \citep{stap-araabi-2023-chatgpt,scalvini-etal-2025-rethinking}.

We see that \datasetname  matches or outperforms baseline datasets in the token-controlled, and shows gains in the into-English direction. We also show similar findings on a comparison on \nllbseed on a separate set of intersection languages in \cref{app:experiment_results_details}.\footnote{In addition, we also show that \nllbseed contains a high proportion of difficult-to-translate texts potentially due to technical or obscure terminology (54\% as compared to 10.41\%), which may hinder lay community translators.} We confirm these trends over various other MT evaluation metrics such as \xcomet and \metricx~\citep{guerreiro2024xcomet,juraska-etal-2024-metricx}, among others, reported in \cref{app:experiment_results_details}.
This supports a major application of seed datasets, i.e., synthetic data generation from monolingual LRL data via better \ttt{xx-en} systems as discussed in \cref{medley:sec:intro}.
 
We observe some improvements from adding \datasetname to existing seed datasets.
However, these are generally small, indicating the challenges of making significant improvements for LRLs via manual collection of data at the scale of a few thousands of sentences. Note that \datasetname contains 92 languages not covered by \textsc{SMOL}.

We note that regardless of the seed dataset used, scores remain low in general, especially in the \texttt{en-xx} direction. 
Given that scaling up data collection is infeasible for this range of languages, one possible takeaway from this is that standard supervised fine-tuning-based approaches may not be sufficient for language learning in this data regime. 
Current literature already investigates various creative ways of using structured information about a language to boost LLM performance for unseen or very low-resource language \citep{tanzer2023benchmark,kornilov2024mteb,zhang2024hire,hus2024back,hus2025machine}; future work in this direction may further investigate the most efficient usage of a seed dataset to augment these methods.
Our dataset, in focusing on the controlled and principled coverage of grammatical features, opens up possibilities for future research in more efficient language learning.

\paragraph{Summary results}

\begin{figure}
    \centering
    \includegraphics[width=0.7\linewidth]{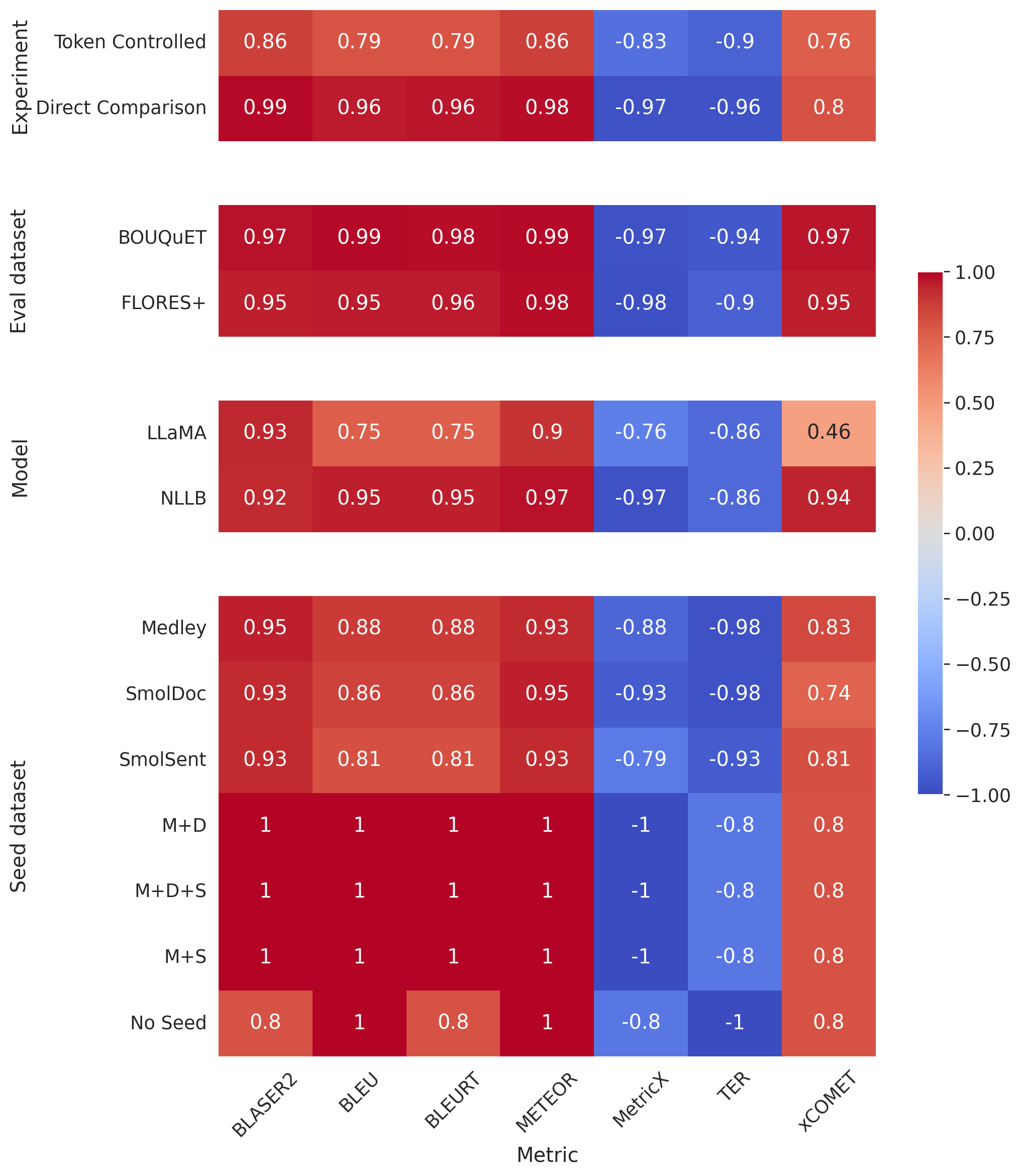}
    \caption{Spearman rank correlation of chrF++ with other evaluation metrics, across different dimensions.}
    \label{fig:medley:results:metrics-correlations}
\end{figure}

Beside chrF++ score we also report xCOMET~\citep{guerreiro2024xcomet} and MetricX~\citep{juraska-etal-2024-metricx} as they are usually better correlated with human judgement and thus can provide a more practical measure of translation quality, see~\cref{tab:app:results:token-controlled:avg} for token-controlled and~\cref{tab:app:results:token-uncontrolled:avg} for direct comparison results. Furthermore, we report Spearman rank correlation between the metric we report in the main manuscript, namely chrF++, and a selection of other widely used evaluation metrics, namely BLASER~\citep{dale2024blaser}, BLEU~\citep{bleu}, BLEURT~\citep{bleurt}, METEOR~\citep{banerjee2005meteor}, MetricX~\citep{juraska-etal-2024-metricx}, TER~\citep{snover2006study}, xCOMET~\citep{guerreiro2024xcomet}, see~\cref{fig:medley:results:metrics-correlations}.

\begin{table}[htbp]
\centering
\resizebox{\linewidth}{!}{%
\begin{tabular}{@{}llcccccccccccc@{}}
\toprule
& & \multicolumn{6}{c}{\textbf{BOUQuET}} & \multicolumn{6}{c}{\textbf{FLORES+}} \\
\cmidrule(lr){3-8} \cmidrule(lr){9-14}
& & \multicolumn{3}{c}{\textbf{en-xx}} & \multicolumn{3}{c}{\textbf{xx-en}} & \multicolumn{3}{c}{\textbf{en-xx}} & \multicolumn{3}{c}{\textbf{xx-en}} \\
\cmidrule(lr){3-5} \cmidrule(lr){6-8} \cmidrule(lr){9-11} \cmidrule(lr){12-14}
\textbf{Model} & \textbf{Seed Dataset} & \textbf{chrF++} & \textbf{MetricX} & \textbf{XCOMET} & \textbf{chrF++} & \textbf{MetricX} & \textbf{XCOMET} & \textbf{chrF++} & \textbf{MetricX} & \textbf{XCOMET} & \textbf{chrF++} & \textbf{MetricX} & \textbf{XCOMET} \\
\midrule

\multirow{4}{*}{LLaMA} 
& No Seed  & 8.49  & 16.02 & 0.22 & 14.45 & \textbf{10.44} & 0.24 & 10.07 & 17.16 & 0.18 & 16.81 & \textbf{10.86} & 0.20 \\
& SmolDoc  & \textbf{20.08} & \textbf{13.89} & \textbf{0.34} & 18.74 & 12.66 & \textbf{0.35} & \textbf{21.53} & \textbf{14.91} & \textbf{0.24} & 22.25 & 12.96 & 0.27 \\
& SmolSent & 18.16 & 15.04 & 0.33 & 18.74 & 13.00 & 0.34 & 18.46 & 16.36 & 0.23 & 22.42 & 13.67 & \textbf{0.28} \\
& Medley   & 19.60 & 13.97 & 0.33 & \textbf{20.39} & 11.94 & 0.33 & 20.69 & 15.48 & \textbf{0.24} & \textbf{23.73} & 12.63 & 0.27 \\

\midrule

\multirow{4}{*}{NLLB} 
& No Seed  & 31.75 & \textbf{9.76}  & 0.36 & 39.43 & 8.97  & 0.54 & 29.01 & \textbf{10.74} & \textbf{0.24} & 39.75 & 9.14  & 0.46 \\
& SmolDoc  & 31.70 & 10.06 & \textbf{0.37} & 40.88 & 8.23  & \textbf{0.57} & 29.85 & 10.92 & 0.23 & 39.34 & 8.89  & 0.46 \\
& SmolSent & \textbf{32.54} & 9.86  & 0.36 & 40.88 & 8.30  & 0.56 & \textbf{30.27} & 10.79 & 0.23 & 39.31 & 8.97  & 0.46 \\
& Medley   & 30.58 & 10.08 & 0.36 & \textbf{43.05} & \textbf{7.89}  & \textbf{0.57} & 29.35 & 11.02 & 0.23 & \textbf{40.72} & \textbf{8.49}  & \textbf{0.47} \\

\bottomrule
\end{tabular}
}
\caption{Average translation performance when fine-tuning on token-controlled seed datasets.}
\label{tab:app:results:token-controlled:avg}
\end{table}

\begin{table}[htbp]
\resizebox{\linewidth}{!}{%
\begin{tabular}{@{}rrrrrrrrrrrrrr@{}}
\toprule
    \multirow{3}{*}{\textbf{Model}} &
    \multirow{3}{*}{\textbf{Seed Dataset}} & 
    \multicolumn{6}{c}{\textbf{BOUQuET}} & 
    \multicolumn{6}{c}{\textbf{FLORES+}} \\
    \cmidrule(lr){3-8} \cmidrule(lr){9-14}
    & & 
    \multicolumn{3}{c}{\textbf{en-xx}} &
    \multicolumn{3}{c}{\textbf{xx-en}} &
    \multicolumn{3}{c}{\textbf{en-xx}} &
    \multicolumn{3}{c}{\textbf{xx-en}} \\
    \cmidrule(lr){3-5} \cmidrule(lr){6-8} \cmidrule(lr){9-11} \cmidrule(lr){12-14}
    & & 
    \textbf{chrF++} & \textbf{MetricX} & \textbf{XCOMET} &
    \textbf{chrF++} & \textbf{MetricX} & \textbf{XCOMET} &
    \textbf{chrF++} & \textbf{MetricX} & \textbf{XCOMET} &
    \textbf{chrF++} & \textbf{MetricX} & \textbf{XCOMET} \\
    \midrule
\multirow{7}{*}{LLaMA} 
    & No Seed      & 8.49  & 16.02 & 0.22 & 14.45 & 10.44 & 0.24 & 10.07 & 17.16 & 0.18 & 16.81 & \textbf{10.86} & 0.20 \\
    & SmolDoc (D)  & 25.07 & 11.99 & 0.35 & 23.06 & 11.41 & 0.40 & 25.03 & 12.94 & \textbf{0.24} & 25.09 & 11.85 & 0.31 \\
    & SmolSent (S) & 18.69 & 14.69 & 0.33 & 19.47 & 12.85 & 0.34 & 19.70 & 15.98 & \textbf{0.24} & 23.87 & 12.99 & 0.29 \\
    & Medley (M)   & 22.30 & 12.66 & 0.34 & 23.18 & 11.35 & 0.37 & 22.43 & 14.33 & \textbf{0.24} & 25.71 & 12.00 & 0.29 \\
    \cline{2-14}
    & M+D          & 26.98 & 11.40 & \textbf{0.36} & 27.44 & 10.12 & 0.44 & 26.10 & 12.58 & \textbf{0.24} & 26.94 & 11.17 & 0.34 \\
    & M+S          & 24.33 & 12.25 & 0.34 & 25.98 & 10.68 & 0.41 & 23.27 & 13.81 & \textbf{0.24} & 26.43 & 11.88 & 0.32 \\
    & M+D+S        & \textbf{28.12} & \textbf{11.25} & \textbf{0.36} & \textbf{29.04} & \textbf{9.98}  & \textbf{0.46} & \textbf{26.79} & \textbf{12.36} & \textbf{0.24} & \textbf{27.90} & 11.22 & \textbf{0.35} \\
\midrule
\multirow{7}{*}{NLLB}  
    & No Seed      & 31.75 & 9.76  & 0.36 & 39.43 & 8.97  & 0.54 & 29.01 & 10.74 & \textbf{0.24} & 39.75 & 9.14  & 0.46 \\
    & SmolDoc (D)  & 32.92 & 9.98  & \textbf{0.37} & 41.24 & 8.19  & 0.57 & 30.05 & 10.88 & 0.23 & 39.36 & 8.95  & 0.46 \\
    & SmolSent (S) & 32.29 & 9.91  & \textbf{0.37} & 41.77 & 8.24  & 0.57 & 30.10 & 10.86 & \textbf{0.24} & 39.94 & 8.91  & 0.47 \\
    & Medley (M)   & 30.60 & 10.04 & \textbf{0.37} & \textbf{43.40} & \textbf{7.85}  & \textbf{0.58} & 29.81 & 10.91 & \textbf{0.24} & \textbf{40.94} & \textbf{8.46}  & \textbf{0.48} \\
    \cline{2-14}
    & M+D          & 33.43 & 9.84  & \textbf{0.37} & 42.67 & 7.94  & \textbf{0.58} & 30.69 & 10.77 & 0.23 & 40.17 & 8.61  & \textbf{0.48} \\
    & M+S          & 32.92 & 9.78  & \textbf{0.37} & 41.98 & 8.00  & 0.57 & 31.02 & 10.68 & 0.23 & 39.83 & 8.73  & 0.47 \\
    & M+D+S        & \textbf{34.73} & \textbf{9.63}  & \textbf{0.37} & 42.90 & 7.89  & \textbf{0.58} & \textbf{31.60} & \textbf{10.58} & 0.23 & 40.19 & 8.65  & 0.47 \\
\bottomrule
\end{tabular}
}
\caption{Average translation performance when fine-tuning on seed datasets and their combination.}
\label{tab:app:results:token-uncontrolled:avg}
\end{table}

\paragraph{Language-wise breakdowns}

Here we report the same as above but with aggregating per-language, see~\cref{medley:tab:app:results:token-controlled:per-lang} for token-controlled and~\cref{medley:tab:app:results:token-uncontrolled:per-lang} for direct comparison.

\begin{table}[htbp]
\centering
\small
\resizebox{\linewidth}{!}{%
\begin{tabular}{@{}llccccccccccccc@{}}
\toprule
    \multirow{3}{*}{\textbf{Model}} &
    \multirow{3}{*}{\textbf{Language}} &
    \multirow{3}{*}{\textbf{Seed Dataset}} & 
    \multicolumn{6}{c}{\textbf{BOUQuET}} & 
    \multicolumn{6}{c}{\textbf{FLORES+}} \\
    \cmidrule(lr){4-9} \cmidrule(lr){10-15}
    & & & 
    \multicolumn{3}{c}{\textbf{en-xx}} &
    \multicolumn{3}{c}{\textbf{xx-en}} &
    \multicolumn{3}{c}{\textbf{en-xx}} &
    \multicolumn{3}{c}{\textbf{xx-en}} \\
    \cmidrule(lr){4-6} \cmidrule(lr){7-9} \cmidrule(lr){10-12} \cmidrule(lr){13-15}
    & & & 
    \textbf{chrF++} & \textbf{MetricX} & \textbf{XCOMET} &
    \textbf{chrF++} & \textbf{MetricX} & \textbf{XCOMET} &
    \textbf{chrF++} & \textbf{MetricX} & \textbf{XCOMET} &
    \textbf{chrF++} & \textbf{MetricX} & \textbf{XCOMET} \\
    \midrule

\multirow{20}{*}{\textbf{LLaMA}} 
& \multirow{4}{*}{bam\_Latn} 
& No Seed  & 7.49  & 16.67 & 0.22 & 14.24 & \textbf{10.74} & 0.25 & 7.06  & 17.51 & 0.17 & 15.36 & \textbf{11.65} & 0.22 \\
& & SmolDoc  & \textbf{22.56} & \textbf{13.03} & \textbf{0.34} & 17.62 & 13.30 & \textbf{0.33} & \textbf{22.83} & \textbf{13.85} & \textbf{0.25} & 21.24 & 14.12 & \textbf{0.26} \\
& & SmolSent & 15.85 & 17.08 & 0.31 & 16.29 & 12.89 & 0.31 & 14.04 & 18.69 & 0.21 & 20.54 & 14.11 & \textbf{0.26} \\
& & Medley   & 21.34 & 13.50 & 0.32 & \textbf{18.08} & 12.24 & 0.30 & 21.05 & 14.43 & \textbf{0.25} & \textbf{22.27} & 13.66 & 0.24 \\
\cmidrule(lr){2-15}

& \multirow{4}{*}{lug\_Latn} 
& No Seed  & 11.43 & 16.37 & 0.24 & 15.02 & \textbf{9.94}  & 0.25 & 13.85 & 17.32 & 0.20 & 18.37 & \textbf{10.43} & 0.22 \\
& & SmolDoc  & \textbf{24.93} & \textbf{12.62} & \textbf{0.37} & 21.47 & 11.81 & 0.39 & \textbf{25.99} & \textbf{14.09} & 0.25 & 24.95 & 11.82 & 0.31 \\
& & SmolSent & 22.67 & 13.00 & \textbf{0.37} & 22.87 & 11.88 & \textbf{0.41} & 25.53 & 14.26 & \textbf{0.26} & \textbf{27.28} & 12.19 & \textbf{0.33} \\
& & Medley   & 23.30 & 14.28 & 0.36 & \textbf{23.73} & 10.82 & 0.37 & 24.26 & 16.32 & \textbf{0.26} & 26.44 & 11.36 & 0.28 \\
\cmidrule(lr){2-15}

& \multirow{4}{*}{mos\_Latn} 
& No Seed  & 5.95  & \textbf{14.70} & 0.16 & 12.29 & \textbf{11.31} & 0.21 & 8.01  & \textbf{16.29} & 0.16 & 14.57 & \textbf{11.42} & 0.18 \\
& & SmolDoc  & 12.29 & 17.94 & \textbf{0.31} & 13.91 & 14.48 & \textbf{0.28} & 14.76 & 18.59 & \textbf{0.23} & 18.33 & 14.21 & \textbf{0.24} \\
& & SmolSent & 13.05 & 17.59 & 0.30 & 13.46 & 15.19 & 0.25 & 14.75 & 18.76 & 0.22 & 16.48 & 15.87 & 0.22 \\
& & Medley   & \textbf{15.73} & 17.15 & \textbf{0.31} & \textbf{16.26} & 13.50 & \textbf{0.28} & \textbf{15.62} & 18.48 & \textbf{0.23} & \textbf{20.78} & 13.58 & \textbf{0.24} \\
\cmidrule(lr){2-15}

& \multirow{4}{*}{wol\_Latn} 
& No Seed  & 9.24  & 15.44 & 0.21 & 14.77 & \textbf{10.73} & 0.22 & 10.65 & 16.71 & 0.19 & 18.46 & \textbf{11.05} & 0.19 \\
& & SmolDoc  & \textbf{22.93} & \textbf{15.35} & \textbf{0.37} & \textbf{19.52} & 12.07 & \textbf{0.38} & \textbf{22.42} & \textbf{16.43} & \textbf{0.24} & 22.40 & 12.45 & 0.28 \\
& & SmolSent & 21.52 & 15.46 & 0.35 & 18.25 & 13.26 & 0.36 & 21.46 & 16.60 & \textbf{0.24} & \textbf{22.52} & 13.89 & \textbf{0.30} \\
& & Medley   & 17.19 & 15.83 & 0.34 & 19.24 & 12.31 & 0.32 & 18.42 & 17.57 & 0.23 & 21.93 & 12.81 & 0.25 \\
\cmidrule(lr){2-15}

& \multirow{4}{*}{yor\_Latn} 
& No Seed  & 8.35  & 16.94 & 0.25 & 15.95 & \textbf{9.45}  & 0.25 & 10.80 & 17.99 & 0.19 & 17.28 & \textbf{9.77}  & 0.22 \\
& & SmolDoc  & 17.71 & 10.53 & 0.31 & 21.17 & 11.66 & 0.35 & 21.65 & 11.61 & 0.22 & 24.34 & 12.19 & 0.29 \\
& & SmolSent & 17.70 & 12.07 & \textbf{0.32} & 22.86 & 11.77 & \textbf{0.39} & 16.50 & 13.48 & \textbf{0.23} & 25.28 & 12.28 & \textbf{0.31} \\
& & Medley   & \textbf{20.45} & \textbf{9.11}  & 0.31 & \textbf{24.66} & 10.84 & \textbf{0.39} & \textbf{24.11} & \textbf{10.58} & 0.22 & \textbf{27.24} & 11.74 & \textbf{0.31} \\

\midrule

\multirow{20}{*}{\textbf{NLLB}} 
& \multirow{4}{*}{bam\_Latn} 
& No Seed  & 28.61 & 9.64  & 0.36 & 31.60 & 11.05 & 0.45 & 31.04 & 9.89  & \textbf{0.25} & \textbf{38.96} & 9.56  & \textbf{0.44} \\
& & SmolDoc  & \textbf{32.05} & \textbf{9.61}  & \textbf{0.37} & 33.88 & 10.16 & \textbf{0.48} & \textbf{32.43} & \textbf{9.58}  & 0.24 & 38.31 & 9.64  & \textbf{0.44} \\
& & SmolSent & 27.91 & 10.67 & 0.35 & 31.61 & 10.31 & 0.46 & 30.09 & 10.50 & 0.24 & 36.51 & 9.69  & 0.43 \\
& & Medley   & 29.63 & 10.28 & 0.35 & \textbf{34.40} & \textbf{9.99}  & \textbf{0.48} & 30.99 & 10.18 & 0.24 & 38.85 & \textbf{9.30}  & \textbf{0.44} \\
\cmidrule(lr){2-15}

& \multirow{4}{*}{lug\_Latn} 
& No Seed  & 41.96 & \textbf{6.09}  & \textbf{0.43} & 43.22 & 8.00  & 0.60 & 38.29 & \textbf{5.86}  & 0.26 & 44.97 & 7.19  & 0.55 \\
& & SmolDoc  & \textbf{42.21} & 6.16  & 0.41 & 44.71 & 7.43  & 0.61 & \textbf{38.80} & 6.08  & 0.25 & 45.32 & 6.89  & 0.55 \\
& & SmolSent & 41.78 & 6.23  & 0.42 & 46.90 & 7.52  & 0.61 & 38.72 & 6.08  & 0.25 & 46.06 & 7.03  & 0.56 \\
& & Medley   & 40.32 & 7.01  & 0.42 & \textbf{48.25} & \textbf{6.90}  & \textbf{0.62} & 38.27 & 7.00  & \textbf{0.27} & \textbf{47.04} & \textbf{6.40}  & \textbf{0.57} \\
\cmidrule(lr){2-15}

& \multirow{4}{*}{mos\_Latn} 
& No Seed  & \textbf{34.33} & \textbf{13.11} & 0.33 & 31.86 & 10.76 & 0.48 & 23.47 & \textbf{15.50} & 0.23 & 31.76 & 11.69 & 0.36 \\
& & SmolDoc  & 17.78 & 16.28 & \textbf{0.35} & 34.71 & 9.74  & 0.51 & 18.48 & 17.53 & \textbf{0.24} & 30.84 & 11.21 & 0.36 \\
& & SmolSent & 27.95 & 13.92 & 0.33 & 34.24 & 9.67  & 0.51 & 23.54 & 15.88 & \textbf{0.24} & 31.95 & 11.10 & 0.37 \\
& & Medley   & 31.33 & 13.45 & 0.33 & \textbf{37.57} & \textbf{8.96}  & \textbf{0.55} & \textbf{23.94} & 15.66 & 0.22 & \textbf{33.74} & \textbf{10.39} & \textbf{0.38} \\
\cmidrule(lr){2-15}

& \multirow{4}{*}{wol\_Latn} 
& No Seed  & 33.56 & 13.08 & 0.37 & 41.25 & 8.97  & 0.58 & 27.65 & 14.89 & 0.23 & 39.23 & 9.95  & 0.46 \\
& & SmolDoc  & \textbf{41.03} & \textbf{12.01} & \textbf{0.39} & 41.05 & 8.25  & \textbf{0.61} & \textbf{30.12} & 14.32 & \textbf{0.24} & 38.47 & 9.72  & 0.45 \\
& & SmolSent & 39.57 & 12.11 & \textbf{0.39} & 41.36 & 8.16  & \textbf{0.61} & 29.79 & \textbf{14.25} & \textbf{0.24} & 38.46 & 9.63  & \textbf{0.47} \\
& & Medley   & 24.32 & 13.75 & 0.37 & \textbf{43.82} & \textbf{8.10}  & \textbf{0.61} & 22.50 & 15.64 & \textbf{0.24} & \textbf{39.86} & \textbf{9.51}  & 0.45 \\
\cmidrule(lr){2-15}

& \multirow{4}{*}{yor\_Latn} 
& No Seed  & 20.27 & 6.88  & 0.33 & 49.20 & 6.06  & 0.61 & 24.59 & 7.58  & \textbf{0.22} & 43.83 & 7.31  & 0.49 \\
& & SmolDoc  & 25.41 & 6.22  & \textbf{0.34} & 50.05 & 5.56  & \textbf{0.62} & 29.42 & 7.07  & 0.20 & 43.77 & 7.00  & \textbf{0.50} \\
& & SmolSent & 25.49 & 6.37  & 0.33 & 50.30 & 5.82  & 0.61 & 29.22 & 7.22  & 0.20 & 43.56 & 7.43  & 0.49 \\
& & Medley   & \textbf{27.29} & \textbf{5.89}  & \textbf{0.34} & \textbf{51.23} & \textbf{5.49}  & \textbf{0.62} & \textbf{31.03} & \textbf{6.60}  & 0.20 & \textbf{44.13} & \textbf{6.85}  & 0.49 \\
\bottomrule
\end{tabular}
}
\caption{Per-language translation performance when fine-tuning on token-controlled seed datasets.}
\label{medley:tab:app:results:token-controlled:per-lang}
\end{table}

\begin{table}[htbp]
\resizebox{\linewidth}{!}{%
\begin{tabular}{@{}llrrrrrrrrrrrrr@{}}
\toprule
    \multirow{3}{*}{\textbf{Model}} &
    \multirow{3}{*}{\textbf{Language}} &
    \multirow{3}{*}{\textbf{Seed Dataset}} & 
    \multicolumn{6}{c}{\textbf{BOUQuET}} & 
    \multicolumn{6}{c}{\textbf{FLORES+}} \\
    \cmidrule(lr){4-9} \cmidrule(lr){10-15}
    & & & 
    \multicolumn{3}{c}{\textbf{en-xx}} &
    \multicolumn{3}{c}{\textbf{xx-en}} &
    \multicolumn{3}{c}{\textbf{en-xx}} &
    \multicolumn{3}{c}{\textbf{xx-en}} \\
    \cmidrule(lr){4-6} \cmidrule(lr){7-9} \cmidrule(lr){10-12} \cmidrule(lr){13-15}
    & & & 
    \textbf{chrF++} & \textbf{MetricX} & \textbf{XCOMET} &
    \textbf{chrF++} & \textbf{MetricX} & \textbf{XCOMET} &
    \textbf{chrF++} & \textbf{MetricX} & \textbf{XCOMET} &
    \textbf{chrF++} & \textbf{MetricX} & \textbf{XCOMET}
  \\ 
\midrule
\multirow{35}{*}{LLaMA} 
& \multirow{7}{*}{bam\_Latn}  &
  No Seed &
  7.49 &
  16.67 &
  0.22 &
  14.24 &
  \textbf{10.74} &
  0.25 &
  7.06 &
  17.51 &
  0.17 &
  15.36 &
  \textbf{11.65} &
  0.22 
  \\
& &
  SmolDoc (D) &
  \textbf{27.82} &
  \textbf{10.82} &
  \textbf{0.35} &
  20.57 &
  12.16 &
  \textbf{0.38} &
  \textbf{27.32} &
  \textbf{11.42} &
  \textbf{0.24} &
  23.46 &
  12.54 &
  \textbf{0.29} \\
& &
  SmolSent (S) &
  18.6 &
  15.4 &
  0.32 &
  15.21 &
  14.51 &
  0.28 &
  19.56 &
  16.42 &
  \textbf{0.24} &
  19.35 &
  14.83 &
  0.24 \\
& &
  Medley (M) &
  24.97 &
  11.83 &
  0.32 &
  \textbf{22.15} &
  11.67 &
  \textbf{0.38} &
  23.88 &
  12.68 &
  \textbf{0.24} &
  \textbf{24.38} &
  12.69 &
  0.28 \\
\cline{3-15}
& &
  M+D &
  28.22 &
  \textbf{10.72} &
  0.34 &
  \textbf{25.75} &
  \textbf{10.64} &
  0.41 &
  \textbf{28.13} &
  \textbf{11} &
  0.24 &
  25.68 &
  \textbf{11.75} &
  \textbf{0.31} \\
& &
  M+S &
  24.96 &
  11.94 &
  0.33 &
  23.4 &
  11.44 &
  0.38 &
  24.36 &
  13.04 &
  \textbf{0.25} &
  24.12 &
  12.89 &
  0.28 \\
& &
  M+D+S &
  \textbf{28.26} &
  10.96 &
  \textbf{0.35} &
  25.36 &
  10.92 &
  \textbf{0.42} &
  27.97 &
  11.47 &
  \textbf{0.25} &
  \textbf{25.72} &
  12.19 &
  \textbf{0.31} \\
\cline{2-15}
& \multirow{7}{*}{lug\_Latn} &
  No Seed &
  11.43 &
  16.37 &
  0.24 &
  15.02 &
  \textbf{9.94} &
  0.25 &
  13.85 &
  17.32 &
  0.2 &
  18.37 &
  \textbf{10.43} &
  0.22 \\
& &
  SmolDoc (D) &
  \textbf{29.42} &
  \textbf{9.94} &
  \textbf{0.39} &
  25.64 &
  10.99 &
  \textbf{0.44} &
  \textbf{30.92} &
  \textbf{10.51} &
  0.25 &
  27.07 &
  11.48 &
  \textbf{0.34} \\
& &
  SmolSent (S) &
  22.32 &
  13.08 &
  0.37 &
  22.96 &
  12.03 &
  0.4 &
  25.57 &
  14.38 &
  \textbf{0.26} &
  26.92 &
  12.18 &
  0.33 \\
& &
  Medley (M) &
  26.61 &
  12.15 &
  0.38 &
  \textbf{26.13} &
  10.59 &
  0.41 &
  26.24 &
  14.48 &
  \textbf{0.26} &
  \textbf{27.42} &
  11.42 &
  0.32 \\
\cline{3-15}
& &
  M+D &
  33.1 &
  9.17 &
  \textbf{0.4} &
  29.72 &
  \textbf{9.58} &
  \textbf{0.49} &
  32.43 &
  9.99 &
  \textbf{0.26} &
  29.03 &
  10.43 &
  \textbf{0.39} \\
& &
  M+S &
  29.5 &
  10.62 &
  0.38 &
  \textbf{30.63} &
  9.61 &
  0.48 &
  29.47 &
  11.97 &
  \textbf{0.26} &
  \textbf{31.04} &
  \textbf{10.22} &
  0.37 \\
& &
  M+D+S &
  \textbf{33.53} &
  \textbf{9.04} &
  0.39 &
  30.62 &
  9.62 &
  \textbf{0.49} &
  \textbf{33.17} &
  \textbf{9.55} &
  0.25 &
  29.24 &
  10.51 &
  \textbf{0.39} \\
\cline{2-15}
& \multirow{7}{*}{mos\_Latn} &
  No Seed &
  5.95 &
  \textbf{14.7} &
  0.16 &
  12.29 &
  \textbf{11.31} &
  0.21 &
  8.01 &
  \textbf{16.29} &
  0.16 &
  14.57 &
  \textbf{11.42} &
  0.18 \\
& &
  SmolDoc (D) &
  12.61 &
  17.94 &
  \textbf{0.32} &
  15.11 &
  14.25 &
  0.28 &
  15.01 &
  18.65 &
  \textbf{0.23} &
  19.16 &
  13.81 &
  0.24 \\
& &
  SmolSent (S) &
  13.74 &
  17.46 &
  \textbf{0.32} &
  14.58 &
  14.64 &
  0.28 &
  15.69 &
  18.8 &
  \textbf{0.23} &
  19.15 &
  14.47 &
  0.25 \\
& &
  Medley (M) &
  \textbf{19.19} &
  15.92 &
  0.31 &
  \textbf{19.16} &
  12.09 &
  \textbf{0.32} &
  \textbf{17.79} &
  17.78 &
  \textbf{0.23} &
  \textbf{23.03} &
  12.32 &
  \textbf{0.26} \\
\cline{3-15}
& &
  M+D &
  16.69 &
  16.69 &
  \textbf{0.33} &
  20.07 &
  12.08 &
  0.34 &
  16.38 &
  18.24 &
  \textbf{0.24} &
  22.41 &
  12.68 &
  0.27 \\
& &
  M+S &
  \textbf{20.1} &
  \textbf{15.41} &
  0.31 &
  19.49 &
  12.34 &
  0.34 &
  17.86 &
  \textbf{16.94} &
  0.23 &
  22.13 &
  13.21 &
  0.27 \\
& &
  M+D+S &
  18.61 &
  16.07 &
  \textbf{0.33} &
  \textbf{22.28} &
  \textbf{11.52} &
  \textbf{0.37} &
  \textbf{18.04} &
  17.43 &
  \textbf{0.24} &
  \textbf{23.92} &
  \textbf{12.46} &
  \textbf{0.29} \\
\cline{2-15}
& \multirow{7}{*}{wol\_Latn} &
  No Seed &
  9.24 &
  15.44 &
  0.21 &
  14.77 &
  10.73 &
  0.22 &
  10.65 &
  16.71 &
  0.19 &
  18.46 &
  11.05 &
  0.19 \\
& &
  SmolDoc (D) &
  \textbf{33.06} &
  \textbf{13.48} &
  \textbf{0.38} &
  \textbf{26.93} &
  \textbf{9.63} &
  \textbf{0.48} &
  \textbf{25.8} &
  \textbf{15.51} &
  \textbf{0.24} &
  \textbf{27.38} &
  \textbf{10.72} &
  \textbf{0.34} \\
& &
  SmolSent (S) &
  20.85 &
  15.74 &
  0.36 &
  19.83 &
  12.24 &
  0.36 &
  21.21 &
  16.86 &
  \textbf{0.24} &
  25.07 &
  12.53 &
  0.31 \\
& &
  Medley (M) &
  18.82 &
  15.17 &
  0.36 &
  21.11 &
  11.71 &
  0.33 &
  19 &
  17.1 &
  \textbf{0.24} &
  23.85 &
  12.43 &
  0.26 \\
\cline{3-15}
& &
  M+D &
  32.54 &
  13.34 &
  0.39 &
  30.33 &
  9.4 &
  0.51 &
  26.22 &
  15.49 &
  \textbf{0.25} &
  27.81 &
  11.05 &
  0.35 \\
& &
  M+S &
  24.91 &
  14.51 &
  0.36 &
  25.49 &
  10.72 &
  0.43 &
  22.89 &
  16.38 &
  0.24 &
  26.62 &
  12 &
  0.34 \\
& &
  M+D+S &
  \textbf{35.61} &
  \textbf{13.04} &
  \textbf{0.4} &
  \textbf{33.03} &
  \textbf{9.06} &
  \textbf{0.55} &
  \textbf{27.5} &
  \textbf{15.07} &
  0.24 &
  \textbf{29.81} &
  \textbf{10.52} &
  \textbf{0.38} \\
\cline{2-15}
& \multirow{7}{*}{yor\_Latn} &
  No Seed &
  8.35 &
  16.94 &
  0.25 &
  15.95 &
  \textbf{9.45} &
  0.25 &
  10.8 &
  17.99 &
  0.19 &
  17.28 &
  \textbf{9.77} &
  0.22 \\
& &
  SmolDoc (D) &
  \textbf{22.42} &
  \textbf{7.77} &
  \textbf{0.33} &
  27.07 &
  10.01 &
  \textbf{0.43} &
  \textbf{26.1} &
  \textbf{8.62} &
  0.21 &
  28.37 &
  10.69 &
  \textbf{0.35} \\
& &
  SmolSent (S) &
  17.94 &
  11.78 &
  0.32 &
  24.78 &
  10.82 &
  0.38 &
  16.47 &
  13.46 &
  \textbf{0.23} &
  28.89 &
  10.94 &
  0.32 \\
& &
  Medley (M) &
  21.93 &
  8.2 &
  0.32 &
  \textbf{27.34} &
  10.69 &
  0.41 &
  25.25 &
  9.63 &
  0.22 &
  \textbf{29.85} &
  11.14 &
  0.32 \\
\cline{3-15}
& &
  M+D &
  24.38 &
  \textbf{7.1} &
  \textbf{0.34} &
  31.34 &
  8.9 &
  0.46 &
  \textbf{27.34} &
  \textbf{8.19} &
  0.21 &
  29.8 &
  \textbf{9.92} &
  \textbf{0.36} \\
& &
  M+S &
  22.21 &
  8.76 &
  0.32 &
  30.86 &
  9.31 &
  0.44 &
  21.78 &
  10.7 &
  \textbf{0.22} &
  28.25 &
  11.06 &
  0.32 \\
& &
  M+D+S &
  \textbf{24.6} &
  7.12 &
  \textbf{0.34} &
  \textbf{33.92} &
  \textbf{8.78} &
  \textbf{0.49} &
  27.26 &
  8.25 &
  \textbf{0.22} &
  \textbf{30.81} &
  10.43 &
  \textbf{0.36} \\
\midrule
\multirow{35}{*}{NLLB} 
& \multirow{7}{*}{bam\_Latn} &
  No Seed &
  28.61 &
  9.64 &
  0.36 &
  31.6 &
  11.05 &
  0.45 &
  31.04 &
  9.89 &
  \textbf{0.25} &
  38.96 &
  9.56 &
  0.44 \\
& &
  SmolDoc (D) &
  \textbf{32.79} &
  \textbf{9.63} &
  \textbf{0.37} &
  33.45 &
  10.29 &
  \textbf{0.48} &
  \textbf{32.66} &
  \textbf{9.58} &
  0.24 &
  37.56 &
  9.9 &
  0.44 \\
& &
  SmolSent (S) &
  27.68 &
  10.76 &
  0.35 &
  31.48 &
  10.16 &
  \textbf{0.48} &
  29.71 &
  10.83 &
  0.24 &
  37.26 &
  9.54 &
  0.45 \\
& &
  Medley (M) &
  29.58 &
  10.38 &
  0.35 &
  \textbf{34.34} &
  \textbf{9.81} &
  \textbf{0.48} &
  31.47 &
  10.15 &
  0.24 &
  \textbf{39.53} &
  \textbf{9.03} &
  \textbf{0.46} \\
\cline{3-15}
& &
  M+D &
  \textbf{32.2} &
  \textbf{9.88} &
  \textbf{0.36} &
  34.35 &
  \textbf{10.12} &
  \textbf{0.48} &
  \textbf{33.08} &
  \textbf{9.63} &
  \textbf{0.24} &
  \textbf{38.84} &
  \textbf{9.53} &
  \textbf{0.45} \\
& &
  M+S &
  28.86 &
  10.64 &
  0.35 &
  32.96 &
  10.32 &
  \textbf{0.48} &
  30.56 &
  10.61 &
  \textbf{0.24} &
  37.12 &
  9.75 &
  0.44 \\
& &
  M+D+S &
  31.83 &
  9.98 &
  \textbf{0.36} &
  \textbf{34.63} &
  10.17 &
  \textbf{0.48} &
  32.95 &
  9.74 &
  \textbf{0.24} &
  38.41 &
  9.62 &
  0.44 \\
\cline{2-15}
& \multirow{7}{*}{lug\_Latn} &
  No Seed &
  41.96 &
  \textbf{6.09} &
  \textbf{0.43} &
  43.22 &
  8 &
  0.6 &
  38.29 &
  \textbf{5.86} &
  0.26 &
  44.97 &
  7.19 &
  0.55 \\
& &
  SmolDoc (D) &
  \textbf{42.17} &
  6.12 &
  0.41 &
  44.79 &
  7.55 &
  0.61 &
  38.96 &
  6.17 &
  0.24 &
  44.92 &
  7.2 &
  0.54 \\
& &
  SmolSent (S) &
  41.95 &
  6.15 &
  0.42 &
  47.27 &
  7.54 &
  0.61 &
  38.6 &
  6.13 &
  0.25 &
  46.12 &
  7.06 &
  0.55 \\
& &
  Medley (M) &
  41.57 &
  6.72 &
  \textbf{0.43} &
  \textbf{48.36} &
  \textbf{6.77} &
  \textbf{0.63} &
  \textbf{39.07} &
  6.86 &
  \textbf{0.27} &
  \textbf{47.04} &
  \textbf{6.43} &
  \textbf{0.57} \\
\cline{3-15}
& &
  M+D &
  42.28 &
  \textbf{6.28} &
  \textbf{0.42} &
  \textbf{47.15} &
  7.03 &
  \textbf{0.63} &
  39.29 &
  \textbf{5.99} &
  0.25 &
  \textbf{46.08} &
  6.74 &
  \textbf{0.56} \\
& &
  M+S &
  42.02 &
  6.41 &
  \textbf{0.42} &
  46.91 &
  \textbf{6.95} &
  0.62 &
  39.13 &
  6.19 &
  \textbf{0.26} &
  45.8 &
  \textbf{6.72} &
  0.55 \\
& &
  M+D+S &
  \textbf{43.16} &
  6.32 &
  \textbf{0.42} &
  46.66 &
  7.06 &
  0.62 &
  \textbf{39.49} &
  6.21 &
  0.25 &
  45.83 &
  6.82 &
  0.55 \\
\cline{2-15}
& \multirow{7}{*}{mos\_Latn} &
  No Seed &
  \textbf{34.33} &
  \textbf{13.11} &
  0.33 &
  31.86 &
  10.76 &
  0.48 &
  23.47 &
  \textbf{15.5} &
  0.23 &
  31.76 &
  11.69 &
  0.36 \\
& &
  SmolDoc (D) &
  17.68 &
  16.28 &
  \textbf{0.35} &
  35.86 &
  9.54 &
  0.51 &
  18.2 &
  17.47 &
  \textbf{0.25} &
  32.16 &
  10.87 &
  0.36 \\
& &
  SmolSent (S) &
  27.35 &
  14.15 &
  0.33 &
  35.12 &
  9.4 &
  0.53 &
  23.25 &
  16.03 &
  0.24 &
  32.89 &
  10.72 &
  0.39 \\
& &
  Medley (M) &
  30.84 &
  13.59 &
  0.33 &
  \textbf{39.38} &
  \textbf{8.99} &
  \textbf{0.56} &
  \textbf{24.53} &
  15.56 &
  0.23 &
  \textbf{34.56} &
  \textbf{10.34} &
  \textbf{0.41} \\
\cline{3-15}
& &
  M+D &
  23.6 &
  14.96 &
  \textbf{0.34} &
  37.21 &
  9.19 &
  0.54 &
  20.25 &
  17.03 &
  \textbf{0.25} &
  33.28 &
  \textbf{10.28} &
  0.39 \\
& &
  M+S &
  \textbf{30.39} &
  \textbf{13.59} &
  0.33 &
  37.08 &
  \textbf{9.07} &
  0.54 &
  \textbf{24.51} &
  \textbf{15.56} &
  0.23 &
  33.69 &
  10.44 &
  0.39 \\
& &
  M+D+S &
  27.31 &
  14.17 &
  \textbf{0.34} &
  \textbf{38.15} &
  9.1 &
  \textbf{0.55} &
  23.43 &
  15.92 &
  0.24 &
  \textbf{33.83} &
  10.35 &
  \textbf{0.4} \\
\cline{2-15}
& \multirow{7}{*}{wol\_Latn} &
  No Seed &
  33.56 &
  13.08 &
  0.37 &
  41.25 &
  8.97 &
  0.58 &
  27.65 &
  14.89 &
  0.23 &
  39.23 &
  9.95 &
  \textbf{0.46} \\
& &
  SmolDoc (D) &
  \textbf{45.55} &
  \textbf{11.71} &
  \textbf{0.41} &
  42.77 &
  \textbf{7.77} &
  \textbf{0.63} &
  \textbf{31.05} &
  14.15 &
  0.24 &
  39.13 &
  9.56 &
  \textbf{0.46} \\
& &
  SmolSent (S) &
  38.62 &
  12.18 &
  0.39 &
  43.78 &
  8.22 &
  0.61 &
  29.59 &
  \textbf{14.14} &
  0.24 &
  39.59 &
  9.74 &
  \textbf{0.46} \\
& &
  Medley (M) &
  23.78 &
  13.69 &
  0.38 &
  \textbf{43.94} &
  8.1 &
  0.61 &
  22.65 &
  15.49 &
  \textbf{0.25} &
  \textbf{40.05} &
  \textbf{9.54} &
  \textbf{0.46} \\
\cline{3-15}
& &
  M+D &
  42.08 &
  11.95 &
  \textbf{0.41} &
  44.24 &
  7.71 &
  0.64 &
  31.04 &
  14.2 &
  \textbf{0.24} &
  39.26 &
  9.41 &
  \textbf{0.48} \\
& &
  M+S &
  36 &
  12.4 &
  0.4 &
  42.99 &
  8 &
  0.61 &
  29.71 &
  14.28 &
  \textbf{0.24} &
  39.48 &
  9.57 &
  0.47 \\
& &
  M+D+S &
  \textbf{44.1} &
  \textbf{11.58} &
  \textbf{0.41} &
  \textbf{45.36} &
  \textbf{7.47} &
  \textbf{0.65} &
  \textbf{31.86} &
  \textbf{14.04} &
  \textbf{0.24} &
  \textbf{39.79} &
  \textbf{9.32} &
  \textbf{0.48} \\
\cline{2-15}
& \multirow{7}{*}{yor\_Latn} &
  No Seed &
  20.27 &
  6.88 &
  0.33 &
  49.2 &
  6.06 &
  \textbf{0.61} &
  24.59 &
  7.58 &
  \textbf{0.22} &
  43.83 &
  7.31 &
  \textbf{0.49} \\
& &
  SmolDoc (D) &
  26.41 &
  6.15 &
  \textbf{0.34} &
  49.34 &
  5.79 &
  \textbf{0.61} &
  29.38 &
  7.02 &
  0.2 &
  43.03 &
  7.23 &
  \textbf{0.49} \\
& &
  SmolSent (S) &
  25.85 &
  6.33 &
  0.33 &
  \textbf{51.2} &
  5.86 &
  \textbf{0.61} &
  29.35 &
  7.18 &
  0.21 &
  \textbf{43.86} &
  7.46 &
  0.48 \\
& &
  Medley (M) &
  \textbf{27.25} &
  \textbf{5.79} &
  \textbf{0.34} &
  50.99 &
  \textbf{5.61} &
  \textbf{0.61} &
  \textbf{31.31} &
  \textbf{6.5} &
  0.21 &
  43.53 &
  \textbf{6.96} &
  \textbf{0.49} \\
\cline{3-15}
& &
  M+D &
  26.98 &
  6.15 &
  \textbf{0.34} &
  \textbf{50.4} &
  5.65 &
  \textbf{0.62} &
  29.77 &
  7.01 &
  0.2 &
  \textbf{43.39} &
  \textbf{7.08} &
  \textbf{0.49} \\
& &
  M+S &
  \textbf{27.35} &
  \textbf{5.86} &
  \textbf{0.34} &
  49.96 &
  5.66 &
  0.61 &
  \textbf{31.18} &
  \textbf{6.76} &
  \textbf{0.21} &
  43.04 &
  7.19 &
  \textbf{0.49} \\
& &
  M+D+S &
  27.26 &
  6.12 &
  \textbf{0.34} &
  49.73 &
  \textbf{5.64} &
  0.61 &
  30.26 &
  7.02 &
  0.2 &
  43.06 &
  7.15 &
  \textbf{0.49} \\ 
\bottomrule
\end{tabular}
}
\caption{Per-language translation performance when fine-tuning on seed datasets and their combinations.}
\label{medley:tab:app:results:token-uncontrolled:per-lang}
\end{table}

\paragraph{Domain-wise breakdown}

We evaluate our models also at the domain-level, leveraging the domain-split provided by \bouquet, and report results in~\cref{tab:app:bouquet:domain-split:results_og}.

\begin{table}[htpb]
\resizebox{\linewidth}{!}{
\centering
\begin{tabular}{@{}rrrrrrrrrrrrrrrrrr@{}}
\toprule
  \multirow{2}{*}{\textbf{Model}} &
\multirow{2}{*}{\textbf{Seed Dataset}} &
  \multicolumn{2}{c}{\textbf{comments}} &
  \multicolumn{2}{c}{\textbf{conversation}} &
  \multicolumn{2}{c}{\textbf{instruction}} &
  \multicolumn{2}{c}{\textbf{narration}} &
  \multicolumn{2}{c}{\textbf{other misc.}} &
  \multicolumn{2}{c}{\textbf{reflection}} &
  \multicolumn{2}{c}{\textbf{social posts}} &
  \multicolumn{2}{c}{\textbf{web misc.}} \\ 
  & & \textbf{xx-en} & \textbf{en-xx} & \textbf{xx-en} & \textbf{en-xx} & \textbf{xx-en} & \textbf{en-xx} & \textbf{xx-en} & \textbf{en-xx} &
  \textbf{xx-en} & \textbf{en-xx} & \textbf{xx-en} & \textbf{en-xx} & \textbf{xx-en} & \textbf{en-xx} & \textbf{xx-en} & \textbf{en-xx}\\
\midrule
\multirow{4}{*}{LLaMA} &
  No Seed &
  7.19 &
  13.49 &
  9.27 &
  13.37 &
  6.97 &
  15.19 &
  7.87 &
  14.07 &
  6.81 &
  14.33 &
  6.42 &
  15.13 &
  7.18 &
  15.50 &
  6.50 &
  13.44 \\
 &
  SmolDoc &
  19.64 &
  14.50 &
  25.63 &
  17.35 &
  22.40 &
  17.68 &
  23.91 &
  19.47 &
  20.23 &
  15.91 &
  21.73 &
  17.61 &
  24.37 &
  20.82 &
  19.18 &
  17.04 \\
 &
  SmolSent &
  14.75 &
  13.66 &
  15.97 &
  16.00 &
  15.87 &
  16.62 &
  16.44 &
  16.55 &
  14.80 &
  15.74 &
  15.44 &
  16.31 &
  17.39 &
  18.38 &
  15.68 &
  16.55 \\
 &
  Medley &
  20.12 &
  15.61 &
  21.68 &
  16.40 &
  21.20 &
  18.30 &
  22.42 &
  19.91 &
  20.89 &
  18.42 &
  21.94 &
  18.81 &
  22.91 &
  19.80 &
  19.10 &
  18.20 \\
  \cline{2-18}
\multirow{4}{*}{NLLB} &
  No Seed &
  25.92 &
  27.85 &
  30.63 &
  34.74 &
  30.15 &
  33.18 &
  31.78 &
  32.55 &
  27.57 &
  29.49 &
  27.11 &
  27.05 &
  29.72 &
  36.22 &
  23.44 &
  27.59 \\
 &
  SmolDoc &
  29.64 &
  31.47 &
  37.27 &
  37.98 &
  33.29 &
  33.35 &
  33.69 &
  36.50 &
  29.97 &
  32.10 &
  28.80 &
  28.02 &
  32.87 &
  36.52 &
  25.48 &
  30.47 \\
 &
  SmolSent &
  24.87 &
  28.24 &
  31.30 &
  33.75 &
  29.02 &
  32.32 &
  28.52 &
  32.71 &
  25.60 &
  30.59 &
  26.53 &
  27.60 &
  29.68 &
  35.23 &
  23.93 &
  29.34 \\
 &
  Medley &
  27.78 &
  31.11 &
  32.47 &
  38.55 &
  31.06 &
  34.59 &
  31.67 &
  37.19 &
  27.20 &
  33.31 &
  27.54 &
  28.57 &
  30.80 &
  36.67 &
  25.30 &
  30.31 \\
\bottomrule
\end{tabular}
}
\caption{Per-domain \bouquet evaluation results comparing models fine-tuned on token-controlled version of the seed datasets.}
\label{tab:app:bouquet:domain-split:results_og}
\end{table}

\paragraph{Comparison with \nllbseed}

While \nllbseed does not cover 4 of 5 of our evaluation languages in \cref{sec:exp_setup}, at the time of writing it has $3$ low-resource languages languages in common with \datasetname and our evaluation datasets: Bambara, Dinka and Fulfulde (bam\_Latn, dik\_Latn, fuv\_Latn). 
We fine-tune, evaluate and compare the two with the same token-controlled methodology, and report the results of the comparison for those languages in~\cref{tab:medley:nllb-seed:results-overall} and~\cref{tab:app:nllb-seed:results-per-lang}. 
Note that there are significant differences between \nllbseed and our dataset. 
The former is single-domain English-centric data selected across a variety of topics, whereas \datasetname is multicentric and multiway, and contains human-written source sentences over 5 domains.
The Wikipedia domain has been observed to contain obscure and specialized terms that are difficult for language translators to work with \citep{taguchi2025languages,jumashev-etal-2025-kyrgyz} and may not necessarily add general utility to an MT system trained on such a corpus. 
We also perform our difficulty analysis 
, and find that \nllbseed contains a very high percentage of texts that are labeled C1 or C2 (54.1\%) as compared to \datasetname~(10.4\%) or \textsc{SMOL-sent} (9.7\%).

\begin{table}[htbp]
\centering
\resizebox{\linewidth}{!}{
\begin{tabular}{@{}llrrrrrrrrrrrr@{}}
\toprule
    \multirow{3}{*}{\textbf{Model}} &
    \multirow{3}{*}{\textbf{Seed Dataset}} & 
    \multicolumn{6}{c}{\textbf{BOUQUET}} & 
    \multicolumn{6}{c}{\textbf{FLORES+}} \\
    \cmidrule(lr){3-8} \cmidrule(lr){9-14}
    & & 
    \multicolumn{3}{c}{\textbf{en-xx}} &
    \multicolumn{3}{c}{\textbf{xx-en}} &
    \multicolumn{3}{c}{\textbf{en-xx}} &
    \multicolumn{3}{c}{\textbf{xx-en}} \\
    \cmidrule(lr){3-5} \cmidrule(lr){6-8} \cmidrule(lr){9-11} \cmidrule(lr){12-14}
    & & 
    \textbf{chrF++} & \textbf{MetricX} & \textbf{XCOMET} &
    \textbf{chrF++} & \textbf{MetricX} & \textbf{XCOMET} &
    \textbf{chrF++} & \textbf{MetricX} & \textbf{XCOMET} &
    \textbf{chrF++} & \textbf{MetricX} & \textbf{XCOMET} \\
    \midrule
\multirow{3}{*}{LLaMA} &
  No Seed &
  6.69 & 14.73 & 0.19 & 13.54 & \textbf{10.40} & 0.23 & 7.78 & 15.71 & 0.16 & 15.64 & 11.04 & 0.20 \\
 & NLLB Seed &
  18.42 & 15.05 & 0.32 & 17.43 & 14.21 & 0.29 & 21.08 & 15.23 & \textbf{0.24} & 24.03 & \textbf{13.19} & \textbf{0.29} \\
 & Medley & \textbf{24.92} & \textbf{13.20} & \textbf{0.33} & \textbf{23.17} & 11.29 & \textbf{0.37} & \textbf{22.19} & \textbf{14.40} & \textbf{0.24} & \textbf{24.35} & 12.49 & 0.28 \\
\cmidrule{2-14}
\multirow{3}{*}{NLLB} &
  No Seed &
  25.69 & 12.07 & \textbf{0.35} & 33.29 & 10.80 & 0.46 & 25.53 & 13.00 & \textbf{0.24} & 33.85 & 11.03 & 0.39 \\
 &
  NLLB Seed &
  27.39 & 12.04 & \textbf{0.35} & 34.98 & 10.32 & 0.48 & 25.97 & 13.01 & \textbf{0.24} & 34.05 & 10.83 & \textbf{0.40} \\
 &
  Medley &
  \textbf{30.51} & \textbf{11.87} & \textbf{0.35} &
  \textbf{37.97} & \textbf{9.43} & \textbf{0.51} & \textbf{26.40} & \textbf{12.88} & \textbf{0.24} & \textbf{34.80} & \textbf{10.56} & \textbf{0.40} \\
  \bottomrule
\end{tabular}
}
\caption{Average translation performance comparing \nllbseed and \datasetname~.}
\label{tab:medley:nllb-seed:results-overall}
\end{table}

\begin{table}[htbp]
\centering
\resizebox{\linewidth}{!}{
\begin{tabular}{@{}lllrrrrrrrrrrrr@{}}
\toprule
    \multirow{3}{*}{\textbf{Model}} &
    \multirow{3}{*}{\textbf{Language}} & 
    \multirow{3}{*}{\textbf{Seed Dataset}} &
    \multicolumn{6}{c}{\textbf{BOUQUET}} & 
    \multicolumn{6}{c}{\textbf{FLORES+}} \\
    \cmidrule(lr){4-9} \cmidrule(lr){10-15}
    & & &
    \multicolumn{3}{c}{\textbf{en-xx}} &
    \multicolumn{3}{c}{\textbf{xx-en}} &
    \multicolumn{3}{c}{\textbf{en-xx}} &
    \multicolumn{3}{c}{\textbf{xx-en}} \\
    \cmidrule(lr){4-6} \cmidrule(lr){7-9} \cmidrule(lr){10-12} \cmidrule(lr){13-15}
    & & &
    \textbf{chrF++} & \textbf{MetricX} & \textbf{XCOMET} &
    \textbf{chrF++} & \textbf{MetricX} & \textbf{XCOMET} &
    \textbf{chrF++} & \textbf{MetricX} & \textbf{XCOMET} &
    \textbf{chrF++} & \textbf{MetricX} & \textbf{XCOMET} \\
    \midrule
\multirow{9}{*}{LLaMA} &
  \multirow{3}{*}{bam\_Latn} &
  No Seed &
  7.55 & 16.67 & 0.22 & 14.29 & \textbf{10.75} & 0.25 & 7.13 & 17.57 & 0.18 & 15.30 & \textbf{11.52} & 0.22 \\
 & & NLLB Seed &
  19.34 & 14.54 & \textbf{0.32} & 18.17 & 13.86 & 0.30 & 22.97 & 14.32 & \textbf{0.25} & \textbf{24.29} & 13.16 & \textbf{0.27} \\
 & & Medley &
  \textbf{24.46} & \textbf{11.90} & \textbf{0.32} & \textbf{22.91} & 11.45 & \textbf{0.37} & \textbf{23.56} & \textbf{12.70} & 0.24 & 23.88 & 13.16 & \textbf{0.27} \\
  \cline{3-15}
 & \multirow{3}{*}{dik\_Latn} &
  No Seed &
  4.95 & \textbf{12.38} & 0.13 & 12.10 & \textbf{10.05} & 0.19 & 7.61 & \textbf{13.69} & 0.13 & 15.12 & \textbf{10.54} & 0.18 \\
 & & NLLB Seed &
  18.47 & 15.99 & 0.31 & 15.57 & 15.15 & 0.26 & \textbf{21.62} & 16.32 & \textbf{0.24} & 23.45 & 13.43 & \textbf{0.29} \\
 & & Medley &
  \textbf{27.08} & 14.62 & \textbf{0.33} & \textbf{23.68} & 11.32 & \textbf{0.36} & 21.51 & 16.25 & \textbf{0.24} & \textbf{25.09} & 11.78 & 0.28 \\
  \cline{3-15}
 & \multirow{3}{*}{fuv\_Latn} &
  No Seed &
  7.56 & 15.14 & 0.22 & 14.23 & \textbf{10.41} & 0.24 & 8.61 & 15.88 & 0.18 & 16.50 & \textbf{11.06} & 0.20 \\
 & & NLLB Seed &
  17.45 & 14.61 & \textbf{0.34} & 18.55 & 13.63 & 0.31 & 18.67 & 15.04 & \textbf{0.24} & \textbf{24.36} & 12.97 & \textbf{0.30} \\
 & & Medley &
  \textbf{23.24} & \textbf{13.09} & \textbf{0.34} & \textbf{22.90} & 11.10 & \textbf{0.37} & \textbf{21.49} & \textbf{14.24} & 0.23 & 24.08 & 12.54 & 0.28 \\
  \cline{2-15}
\multirow{9}{*}{NLLB} &
  \multirow{3}{*}{bam\_Latn} &
  No Seed &
  28.61 & \textbf{9.64} & \textbf{0.36} & 31.60 & 11.05 & 0.45 & 31.04 & \textbf{9.89} & \textbf{0.25} & 38.96 & 9.56 & 0.44 \\
 & & NLLB Seed &
  28.71 & 10.12 & \textbf{0.36} & 32.64 & 10.63 & 0.46 & 31.28 & 10.26 & \textbf{0.25} & 38.43 & 9.65 & \textbf{0.45} \\
 & & Medley &
  \textbf{29.37} & 10.35 & 0.35 & \textbf{34.33} & \textbf{10.01} & \textbf{0.48} & \textbf{31.34} & 10.23 & 0.24 & \textbf{39.17} & \textbf{9.25} & \textbf{0.45} \\
  \cline{3-15}
 & \multirow{3}{*}{dik\_Latn} &
  No Seed &
  23.48 & 14.65 & 0.35 & 32.37 & 11.17 & 0.45 & 22.54 & 16.08 & \textbf{0.25} & 30.42 & 12.28 & 0.36 \\
 & & NLLB Seed &
  28.20 & 14.07 & 0.34 & 35.98 & 10.63 & 0.49 & \textbf{24.26} & 15.56 & 0.24 & 31.08 & 12.03 & \textbf{0.37} \\
 & & Medley &
  \textbf{34.70} & \textbf{13.41} & \textbf{0.36} & \textbf{39.92} & \textbf{9.48} & \textbf{0.51} & 23.83 & \textbf{15.45} & \textbf{0.25} & \textbf{31.93} & \textbf{11.58} & \textbf{0.37} \\
  \cline{3-15}
 & \multirow{3}{*}{fuv\_Latn} &
  No Seed &
  24.97 & 11.90 & \textbf{0.36} & 35.91 & 10.18 & 0.48 & 23.01 & 13.04 & \textbf{0.24} & 32.16 & 11.25 & 0.37 \\
 & & NLLB Seed &
  25.25 & 11.91 & \textbf{0.36} & 36.32 & 9.69 & 0.50 & 22.39 & 13.20 & \textbf{0.24} & 32.64 & \textbf{10.80} & \textbf{0.38} \\
 & & Medley &
  \textbf{27.44} & \textbf{11.85} & 0.35 & \textbf{39.66} & \textbf{8.81} & \textbf{0.52} & \textbf{24.02} & \textbf{12.95} & 0.23 & \textbf{33.29} & 10.84 & 0.37 \\ \bottomrule
\end{tabular}
}
\caption{Per-language translation performance comparing \nllbseed and \datasetname~.}
\label{tab:app:nllb-seed:results-per-lang}
\end{table}

\clearpage
\newpage

\section{Met-\bouquet details}
\label{app:metbouquet}

\subsection{XSTS+R+P}
\label{appendix:xstsrpguidelines}
\paragraph{Goal}
The goal is to assess the degree of meaning correspondence/equivalence between a translation request and the translation of the requested paragraph. The scores will be aggregated to estimate an average semantic correspondence between languages in the dataset. 

You will need to read the source and the target, compare them sentence by sentence and assign a score to each sentence. An automatic formula will then calculate the resulting score for the whole paragraph. You will be asked not to only compare the semantic meaning, but also the register of the paragraphs.

\paragraph{Rating guidelines}
\paragraph{Notes:}
\begin{itemize}
    \item The examples you will now see are phrase or sentence-based for simplicity purposes. You will work with longer paragraphs.
    \item Please ignore minor typos and grammatical errors if they do not affect your understanding of the texts.
    \item Please ignore capitalization and punctuation differences if they do not affect your understanding.
\end{itemize}

\paragraph{[1]} The source paragraph and its translation are not semantically equivalent, share very little detail, and may be about different topics.\\
\tab \small{\textbf{Example A (different topics)}}\\
\tab \small{Text 1. (English): Train station}\\
\tab \small{Text 2. (Spanish): Restaurante vegano (Vegan restaurant)}\\
\tab \small{\textbf{Example B (false equivalents)}\\
\tab \small{Text 1. When I get home, I always lock the door. It gives me peace of mind.}\\
\tab \small{Text 2. Cuando llego a casa siempre loqueo la puerta. Me da tranquilidad.}\\
\tab \small{\textbf{Example C (very little overlap and unrelated entities)}}\\
\tab \small{Text 1. Open museums}\\
\tab \small{Text 2. Centros comerciales abiertos (Open malls)}\\
\tab \small{\textbf{Example D (indecipherable on one side or the other)}}\\
\tab \small{Text 1. lorbo lorbo lorl room}\\
\tab \small{Text 2. Habitación doble (double room)}\\
\tab \small{\textbf{Example E (untranslated text)}}\\
\tab \small{Text 1. Should you have any questions, please let me know.}\\
\tab \small{Text 2. Si tiene cualquier duda, please let me know.}\\
\tab \small{\textbf{Example F (hallucinations)}}\\
\tab \small{Text 1. Thank you for joining us today. We’re so happy you’re here.}\\
\tab \small{Text 2. Gracias por acompañarnos hoy. Nos alegra mucho que estéis aquí. Esta noche la vamos a recordar toda la vida. (We will remember this night for the rest of our lives).}

\paragraph{[2]} The source paragraph and its translation share some details, but are not equivalent. Some important information related to the primary subject/verb/object differs or is missing, which alters the intent or meaning of the paragraph. Alternatively, the register differs so much that this translation will be a big mistake. A significant change in register will always score no more than 2 points.\\
\tab \small{\textbf{Example A (opposite polarity)}}\\
\tab \small{Text 1. Flight to London}\\
\tab \small{Text 2. Vuelo desde Londres (Flight from London)}\\
\tab \small{\textbf{Example B (non-equivalent numbers)}}\\
\tab \small{Text 1. Two rooms for three people}\\
\tab \small{Text 2. Tres habitaciones para dos personas (Three rooms for two people)}\\
\tab \small{\textbf{Example C (substitution/change in named entity)}}\\
\tab \small{Text 1. Flight to Valencia}\\
\tab \small{Text 2. Vuelo a Valladolid (Flight to Valladolid)}\\
\tab \small{\textbf{Example D (different meaning due to word order)}}\\
\tab \small{Text 1. I like pizza}\\
\tab \small{Text 2. Yo gusto a la pizza (Pizza likes me)}\\
\tab \small{\textbf{Example E (equivalent constructions with different meanings)}}\\
\tab \small{Text 1. The bus is arriving at 2.}\\
\tab \small{Text 2. El autobús está llegando a las 2.\\
\tab \scriptsize{\textbf{Explanation: }Spanish present progressive cannot be used to refer to the future}}\\
\tab \small{\textbf{Example F (missing salient information)}}\\
\tab \small{Text 1. Vegan Italian restaurant}\\
\tab \small{Text 2. Restaurante italiano (Italian restaurant)}\\
\tab \small{\textbf{Example G (omitted relevant chunks)}}\\
\tab \small{Text 1. The company's new product (a flask with temperature regulation) is expected to be a success.}\\
\tab \small{Text 2. Se espera que el nuevo producto de la empresa sea un éxito.}\\
\tab \small{\textbf{Example H (register difference)}}\\
\tab \small{Text 1. What’s up, dude?}\\
\tab \small{Text 2. ¿Cómo se encuentra, señor? (How are you, sir?)}\\

\paragraph{[3]} The two paragraphs are mostly equivalent, but some unimportant details can differ.  There cannot be any significant conflicts in intent, meaning or register between the sentences, no matter how long the sentences are.\\
\tab \small{\textbf{Example A  (omitted non-critical information, but no contradictory info introduced)}}\\
\tab \small{Text 1. Table for 3 adults}\\
\tab \small{Text 2. Mesa para 3 (Table for 3)}\\
\tab \small{\textbf{Example B (unit of measurement differences)}}\\
\tab \small{Text 1. I want 2 pounds of cheese.}\\
\tab \small{Text 2. Quería 1 kg de queso. (I wanted 1kg of cheese.)}\\
\tab \small{\textbf{Example C (minor verb tense differences)}}\\
\tab \small{Text 1. When he arrived at the station, the train had already left.}\\
\tab \small{Text 2. Cuando llegue a la estación, el tren ya se habrá ido. (When he arrives at the station, the train will have already left.)}\\
\tab \small{\textbf{Example D (small, non-conflicting differences \textbf{in meaning)}}\\
\tab \small{Text 1. I love running.}\\
\tab \small{Text 2. Me gusta correr. (I like running.)}\\
\tab \small{\textbf{Example E (non-critical information added)}}\\
\tab \small{Text 1. Photos of the trip}\\
\tab \small{Text 2. Fotos de mi viaje (Photos of my trip)}\\
\tab \small{\textbf{Example F (non-equivalent constructions)}}\\
\tab \small{Text 1. The president finally signed the new education bill yesterday at noon.}\\
\tab \small{Text 2. La nueva ley de educación finalmente fue firmada por el presidente ayer al mediodía. (The new education bill was finally signed by the president yesterday at noon.)}\\
\tab \small{\textbf{Example G (inconsistent register)}}\\
\tab \small{Text 1. First, you will need to purchase all the painting supplies you need. Then, film and tape everything that is not to be painted.}\\
\tab \small{Text 2. Primero, tendrás que comprar todo lo que necesitas para pintar. Luego proteja con film y cinta de pintor todo lo que no deba ser pintado. \\
\tab \scriptsize{\textbf{Explanation: }“proteja” is a formal second person imperative form (“usted”), but the verb in the previous sentence (“tendrás”) uses an informal second person form (“tú”).}}\\
\tab \small{\textbf{Example H (inconsistent target)}}\\
\tab \small{Text 1. It’s pretty flashy, but remember that you have to wash it often.} \\
\tab \small{Text 2. Es bien chido, pero acordate que tienes que lavarlo frecuentemente. \\
\tab \scriptsize{\textbf{Explanation: }The translation mixes lexical and/or grammatical forms from different varieties/dialects (“bien chido” is broadly Mexican, “acordate” is broadly Rioplatense and “tienes” is from a non-voseo variety like European Spanish).}}\\

\paragraph{[4]} The two paragraphs are paraphrases of each other. Their meanings are near-equivalent,  with no major differences or information missing. There can only be minor differences in meaning due to differences in expression (e.g., formality level, style, emphasis, potential implication, idioms, common metaphors). For single word texts, there might be multiple meanings depending on the context they would be used in, but the one presented is still correct.\\
\tab \small{\textbf{Example A}}\\
\tab \small{Text 1. This is great}\\
\tab \small{Text 2. Esto es la leche (Lit: This is the milk)}\\
\tab \scriptsize{\textbf{Explanation:} “Esto es la leche” is an idiom, “this is great” is not.}\\
\tab \small{\textbf{Example B}}\\
\tab \small{Text 1. The day that comes after the day of today}\\
\tab \small{Text 2. Mañana (Tomorrow)}\\
\tab \scriptsize{\textbf{Explanation:} Differences in phrasing, text 1 is oddly phrased and more verbose than text2.}\\
\tab \small{\textbf{Example C}}\\
\tab \small{Text 1. Bird}\\
\tab \small{Text 2. Pajarito (birdie)\\
\tab \scriptsize{\textbf{Explanation:} Different level of formality (“Birdie” vs “bird”).}}\\

\paragraph{[5]} The two paragraphs are exactly and completely equivalent in meaning and usage expression (e.g., formality level, style, emphasis, potential implication, idioms, common metaphors). In other words, nuance is completely preserved and there is a faithful correspondence. Fidelity is also preserved.\\
\tab \small{\textbf{Example A}}\\
\tab \small{Text 1. I am so happy.}\\
\tab \small{Text 2. Estoy lleno de felicidad (I am filled with happiness).}\\
\tab \small{\textbf{Example B}}\\
\tab \small{Text 1. Hi friends}\\
\tab \small{Text 2. Hola chicos (Hello guys)}\\
\tab \small{\textbf{Example C}}\\
\tab \small{Text 1. Hello, how are you}\\
\tab \small{Text 2. Hola cómo estás (Hello how are you)}\\

\paragraph{Pilot} To validate the XSTS+R+P protocol, we designed a pilot consisting of 210 sentence pairs organized in 50 paragraphs in two high resource languages (i.e., 105 Russian-to-Spanish pairs, and 105 Spanish-to-Russian). Similar to XSTS, annotators were asked to rate each Russian-Spanish sentence pair on a scale from 1 to 5 based on how equivalent they were, with 1 being not semantically equivalent and 5 being completely equivalent in meaning and usage expression. Meaning is broadly construed here, that is, it includes both lexical and grammatical meaning (i.e., the semantics of grammatical constructions, formality levels, style, emphasis, etc.). Unlike XSTS, however, annotators were asked to consider each sentence in the context of the paragraphs they were in, as well as any additional information provided about the paragraphs’ source, genre, and communicative goal. Finally, annotators provided comments for their rating decisions.
The pilot results show that annotators were able to rate sentences considering register and paragraph information, although they tended to point out lexical mistranslations at a higher rate. We attribute this tendency to a smaller pool of examples involving grammatical non-equivalent meaning in our guidelines (e.g., identical lexical items with alternate active-passive sentences). This was corrected in the guidelines for the final evaluation.

\subsection{Detailed Selection of MT outputs Met-\bouquet Round 1} 

For the entire set of development and test sentences from \bouquet (1358), we select only one translation output among the different systems. We optimitze for a variety of quality (in order to have a wide range of scores) and a variety of systems (in order to have a wide range of error types typical for different systems). We use variations of the same selection algorithm for very strong translation directions (when we had to oversample bad translations to provide some signal for QE) and directions with very poor quality (where bad translations should be undersampled).
This resulted in the following implementation:

\emph{Step1. For every translation direction}, we have the outputs of several candidate systems (3 to 24) and for each sentence output, we have 4 translation scores (ChrF++ \cite{chrf}, Blaser 2.0-QE \cite{dale-costa-jussa-2024-blaser}, either WMT23-Cometkiwi-da-xl \cite{rei-etal-2022-cometkiwi} or \xcomet-XL \cite{guerreiro-etal-2024-xcomet}, MetricX-24-hybrid-xl-v2p6 \cite{juraska-etal-2024-metricx}). The systems are selected so that they always include a Llama-based system and NLLB (if it supports the target language), and several other systems with the best translations according to any of the scores above (at least one system per score, at most 4 systems if they are ``good'' according to this score; the systems selected by different scores may overlap).

\emph{Step2. For each translation direction} and unique source text, we select at most one translation from the ones that are “probably good enough” (MetricX$<$5, Blaser-QE$>$3, ChrF++$>$10); if several systems provide good translations for a source, the system is chosen randomly. We keep such translations for at most 50\% of all volume per direction.

\emph{Step3. 10\% translations per direction} (or more, if at the previous step less than 50\% were selected) are chosen by the best values of each metric for each system. \textit{25\% translations} are selected by the worst values for each metric for each system. 

\emph{Step4. The rest is selected randomly}, with a slight upsampling of the systems that have been underrepresented before.

\subsection{Comparison to other MT metrics evaluation datasets}

\begin{table*}[ht!]
\centering
\scriptsize
\begin{tabular}{@{}p{2cm}cccccccc@{}}
\toprule
Dataset & Protocol & Langs & Dir & Dir w/o Eng & Src Sent & Systems & Domains\\
\midrule
 \multirow{5}{*}{MLQE PE}& DA (z-scores) & 2 & 1 & 1 & 995 & NA & NA\\
 & DA & 8 & 7 & 0 & 60,788 & NA & NA \\
 & HTER & 8 & 7 & 0 & 60,757 & NA & NA \\
  & DA \& HTER & 12 & 11 & 0 & 10,970 & NA & NA \\
  & Catastrophic Errors & 5 & 4 & 0 & 15,261 & NA & NA \\
\hline
IndicMT Eval & MQM & 6 & 5 & 0 & 1,423 & 7 & NA \\
\hline
AmericasNLP25 & Semantics \& Fluency & 4 & 3 & 3 & 100 & NA & NA\\
\hline
NLLB & XSTS & 60 & 52 & 20 & 19,228 & NA & 3 \\
\hline
\multirow{3}{*}{Met-\bouquet} & R1 (XSTS+R+P, XSTS, RSQM) & 53 & 104 & 62 & 74,034 & 31 & 8\\
 & R2 (XSTS+R+P) & 80 & 57 & 56 & 38,346 & 16 & 8\\
 & R1+2 & 119 & 161 & 118 & 104,305 & 43 & 8\\
\bottomrule
\end{tabular}%
\caption{Summary of key attributes of MT metrics evaluation datasets, including language coverage, source sentences, systems evaluated, and domains. Statistics are reported only for datasets with available evaluation scores.}
\label{tabl:MTmetricsdatasets}
\end{table*}

Table \ref{tabl:MTmetricsdatasets} compares to other datasets that have been used to evaluate MT metrics. It includes %
MLQE \cite{fomicheva-etal-2022-mlqe}; IndicMT Eval \cite{sai-b-etal-2023-indicmt},  AmericasNLP (Task3) \cite{de-gibert-etal-2025-findings}, NLLB \cite{nllb-24}.

Table \ref{tabl:MTmetricsdatasets} summarizes key characteristics of each dataset, including: {Evaluation protocol} used to score translations; number of languages covered (as source or target); {Number of language pairs} (source--target combinations); {Number of source sentences} for which translations are available; {Number of translation systems} used; {Number of domains} represented in the dataset.

Most existing datasets use different sets of source sentences for different language directions. %
Notable exceptions are the {AmericasNLP 2025 Task 3} and {IndicMT Eval} datasets, which employ the same set of source sentences (in Spanish and English, respectively) in all translation directions. However, these datasets are limited in their multilingual scope as they maintain parallelism only within a single source language.

In contrast, Met-\bouquet is designed to maximize both parallelism and multilinguality. By assigning a unique identifier to each source paragraph and sentence, regardless of the source language, it enables true cross-lingual comparisons across a wide range of languages and translation directions. This approach supports parallel evaluation while also facilitating comprehensive multilingual benchmarking.%

It is also relevant noting that Met-\bouquet Round 1 uniquely contains 100\% of bidirectional pairs and 60\% of directions without English (62 directions), and Met-\bouquet Round 1 + 2 includes a total of 118 directions without English, only followed by NLLB with 20 directions without English.

Additionally, Met-\bouquet inherits advantages from \bouquet (as discussed in Section \ref{sec:bouquet}) %
by offering detailed domain, register and linguistic annotations for all sentences \cite{bouquet} and information about the specific MT system that produced the output. Although it covers a single target output, it is the only dataset reviewed that provides such comprehensive metadata, enabling more granular and fine-grained performance benchmarking.

\paragraph{Score Distribution.} Figure \ref{fig:xsts-scores-histogram} (top) reports the histogram of the XSTS+R+P consensus scores (median) across English and non-English directions for Met-\bouquet Round 1. We observe that pairs involving English tend to have higher scores than non-English pairs. Non-English pairs feature a higher number of very poor translations (XSTS+R+P of 1).
Figure \ref{fig:xsts-scores-histogram} (bottom) shows the XSTS+R+P average score across source and target languages. While for this round we aimed for a uniform score distribution, the cases further from this goal are low-resource languages such as Plains Cree (crk\_Cans) and Ngambay (sba\_Latn).

\begin{figure*}[ht!]
\centering
   \includegraphics[width=1\textwidth]{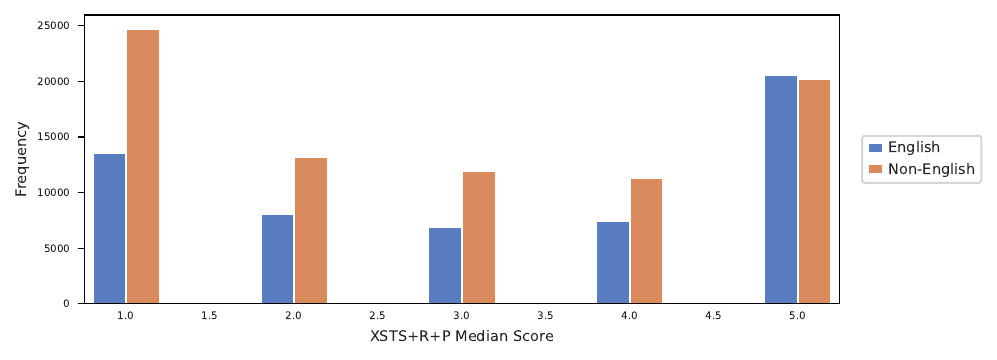} \\
      \includegraphics[width=1\textwidth]{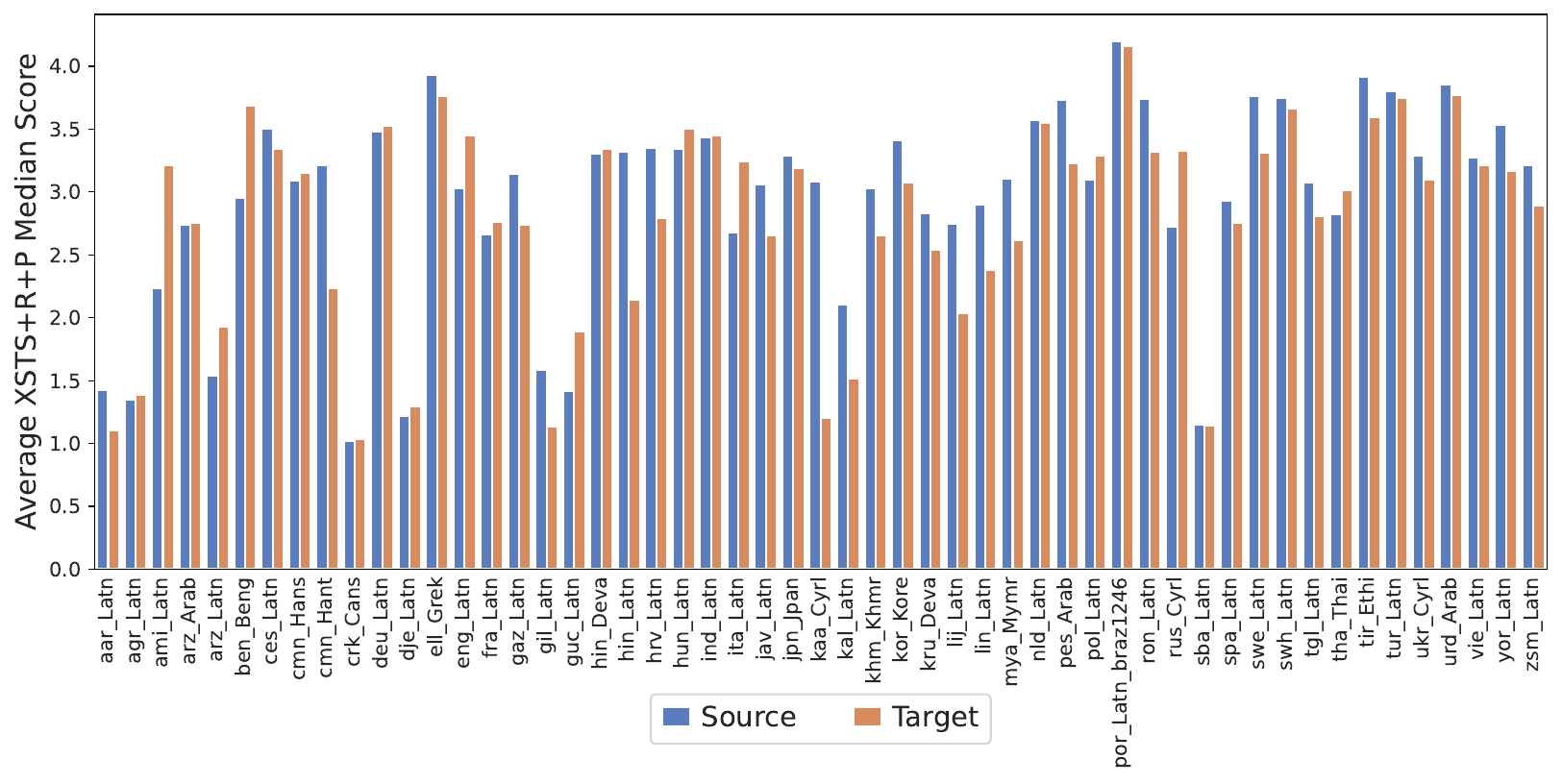} 
  \caption{\label{fig:xsts-scores-histogram} For Met-\bouquet Round 1: Top, XSTS+R+P consensus scores histogram across English and non-English directions; Bottom, XSTS+R+P consensus averages across source and target languages.
}
\end{figure*}

\begin{figure*}
\centering
   \includegraphics[width=1\textwidth]{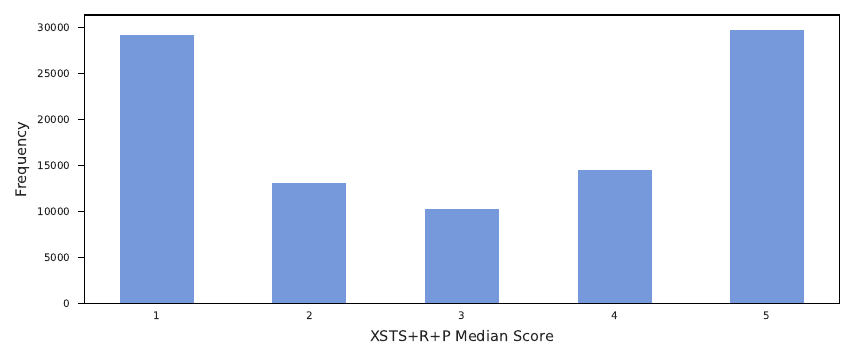} \\
      \includegraphics[width=1\textwidth]{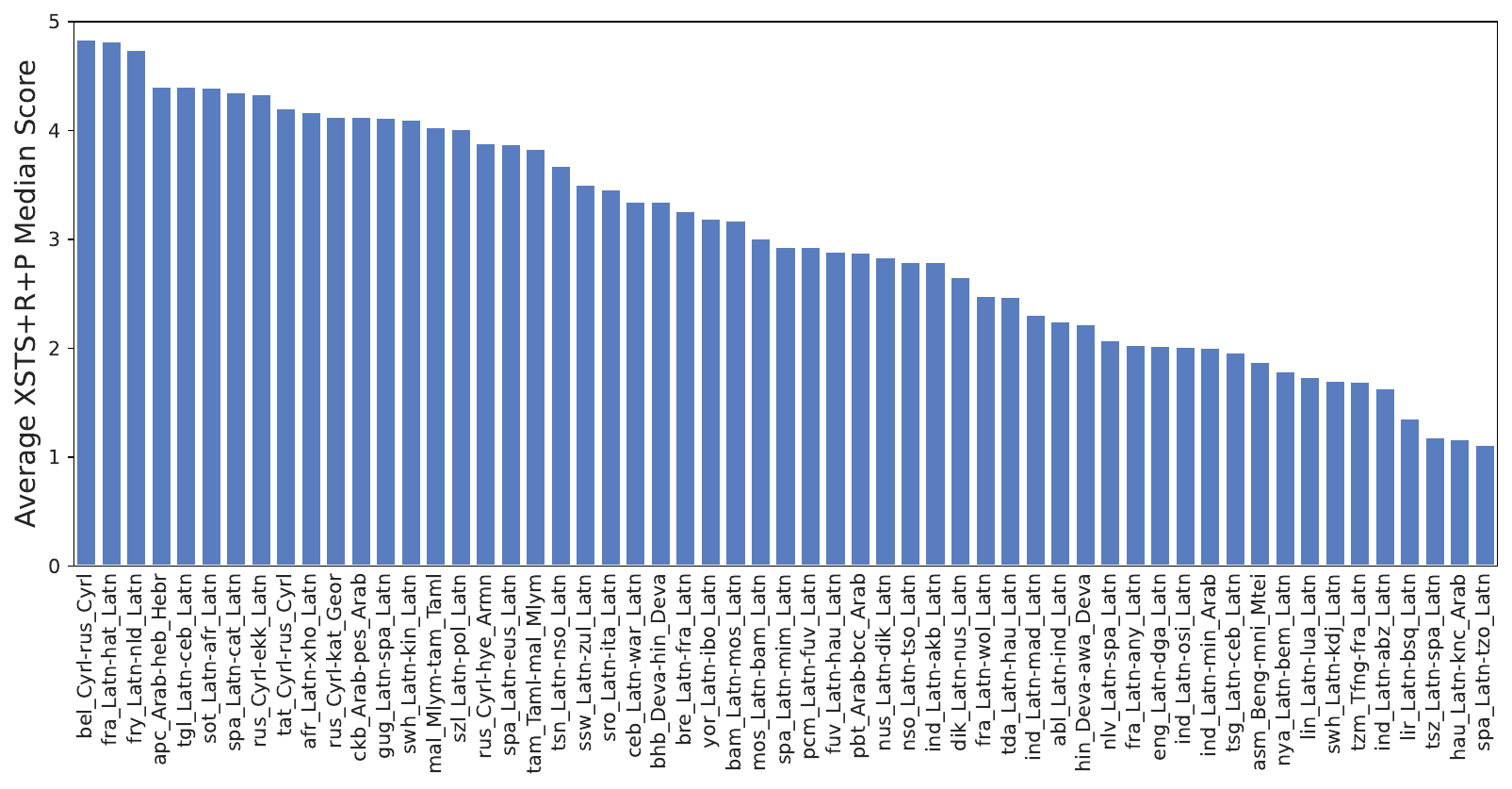} 
  \caption{\label{fig:xsts-scores-histogram_r2} For Met-\bouquet Round 2: Top, XSTS+R+P consensus scores histogram; Bottom, XSTS+R+P consensus averages across directions.
}
\end{figure*}

\clearpage
\newpage

\section{Cards}
\label{app:cards}
\begin{table}[h!]
\centering
\label{metriccard}
\scriptsize
\begin{tabular}{lp{2.5cm}ll}
\toprule
{\bf Task} & {\bf Metric} &  {\bf Type}  & {\bf Citation} \\
\midrule
\textbf{LID} & \textsc{glotlid} & Automatic & \citep{kargaran-etal-2023-glotlid} \\
\cmidrule{2-4}
{\bf MT Metrics} & Spearman's rank correlation coefficient & Automatic  \\
\cmidrule{2-4}
{\bf MT} &  {BLEU}  & Automatic &\citep{bleu} \\
&  \chrf  & Automatic & \citep{chrf}  \\
& METEOR & Automatic & \citep{meteor}  \\
&  BLASER2 & Automatic  & \citep{dale-costa-jussa-2024-blaser}\\
&  \blaser & Automatic  & Section \ref{sec:blaser}\\
& BLEURT & Automatic & \citep{bleurt}  \\
& COMETKiWi & Automatic & \citep{rei-etal-2022-cometkiwi} \\
&  MetricX & Automatic & \citep{juraska-etal-2024-metricx}\\
&  XCOMET & Automatic &\citep{guerreiro-etal-2024-xcomet}\\
&XSTS & Human & \citep{licht-etal-2022-consistent}\\
&SQM & Human & \citep{mqm} \\
&XSTS+R+P & Human & Section \ref{subsec:xstsrp}\\
\cmidrule{2-4}
{\bf Sentence Similarity} &  SONAR + Cosine Similarity & Automatic &  \citep{duquenne2023sonarsentencelevelmultimodallanguageagnostic}\\
&\sonaromni + Cosine Similarity & Automatic &  \citep{sonaromni}\\

\cmidrule{2-4}
{\bf Toxicity} & ROC AUC & Automatic  \\
& OmniTox & Automatic  & Section \ref{sec:omnitox}\\
\bottomrule
\end{tabular}
\caption{Automatic and human evaluation metrics/protocols used in this work.}
\end{table}

\clearpage
\newpage

\section{Languages at a glance}
\label{app:comparelangs}

\centering
\small
\label{tab:bouquetlangs}
\begin{longtable}{ccccccc}
\toprule
\textbf{ISO 639-3} & \textbf{ISO 15924} &\textbf{Language family} &\textbf{\bouquet} & \textbf{\medley} & \textbf{met-\bouquet}& \textbf{FLoRes+}\\
\midrule
aar & Latn & Afro-Asiatic & \checkmark & $\times$ & \checkmark (R1) & $\times$ \\
abl & Latn & Austronesian & \checkmark & $\times$ & \checkmark (R2) & $\times$ \\  %
abz & Arab & Timor-Alor-Pantar & $\times$ & $\times$ & \checkmark (R2) & $\times$ \\
ace & Arab & Austronesian & $\times$ & $\times$ & $\times$ & \checkmark \\
ace & Latn & Austronesian & $\times$ & $\times$ & $\times$ & \checkmark \\
acm & Arab & Afro-Asiatic & $\times$ & $\times$ & $\times$ & \checkmark \\
acq & Arab & Afro-Asiatic & $\times$ & $\times$ & $\times$ & \checkmark \\
aeb & Arab & Afro-Asiatic & $\times$ & $\times$ & $\times$ & \checkmark \\
afr & Latn & Indo-European & \checkmark & $\times$ & \checkmark (R2) & \checkmark \\
agr & Latn & Chicham & \checkmark & $\times$ & \checkmark (R1) & $\times$ \\
ahk & Thai & Sino-Tibetan & $\times$ & \checkmark & $\times$ & $\times$ \\
aiq & Arab & Indo-European & \checkmark & $\times$ & $\times$ & $\times$ \\
akb & Latn & Austronesian & $\times$ & \checkmark & \checkmark (R2) & $\times$ \\
als & Latn & Indo-European & \checkmark & $\times$ & $\times$ & \checkmark \\
amh & Ethi & Afro-Asiatic & \checkmark & $\times$ & $\times$ & \checkmark \\
ami & Latn & Austronesian & \checkmark & $\times$ & \checkmark (R1) & $\times$ \\
ane & Latn & Austronesian & \checkmark & $\times$ & $\times$ & $\times$ \\
any & Latn & Atlantic-Congo & $\times$ & \checkmark & \checkmark (R2) & $\times$ \\
apc & Arab & Afro-Asiatic & \checkmark & $\times$ & \checkmark (R2) & \checkmark \\  %
arb & Arab & Afro-Asiatic & $\times$ & $\checkmark$ & $\times$ & \checkmark \\
arb & Latn & Afro-Asiatic & $\times$ & $\times$ & $\times$ & \checkmark \\
arg & Latn & Indo-European & $\times$ & $\times$ & $\times$ & \checkmark \\
arh & Latn & Chibchan & \checkmark & $\times$ & $\times$ & $\times$ \\
arn & Latn & Araucanian & \checkmark & $\times$ & $\times$ & $\times$ \\
ars & Arab & Afro-Asiatic & $\times$ & $\times$ & $\times$ & \checkmark \\
ary & Arab & Afro-Asiatic & $\times$ & $\times$ & $\times$ & \checkmark \\
arz & Arab & Afro-Asiatic & \checkmark & $\times$ & \checkmark (R1) & \checkmark \\
arz & Latn & Afro-Asiatic & \checkmark & $\times$ & \checkmark (R1) & $\times$ \\
asm & Beng & Indo-European & \checkmark & $\times$ & \checkmark (R2) & \checkmark \\ %
ast & Latn & Indo-European & $\times$ & $\times$ & $\times$ & \checkmark \\
ati & Latn & Atlantic-Congo & $\times$ & \checkmark & $\times$ & $\times$ \\
awa & Deva & Indo-European & $\times$ & $\times$ & \checkmark (R2) & \checkmark \\
ayr & Latn & Aymaran & \checkmark & $\times$ & $\times$ & \checkmark \\
ayz & Latn & Maybratic & \checkmark & $\times$ & $\times$ & $\times$ \\
azb & Arab & Turkic & \checkmark & \checkmark & $\times$ & \checkmark \\
azj & Latn & Turkic & \checkmark & $\times$ & $\times$ & \checkmark \\
azm & Latn & Otomanguean & \checkmark & $\times$ & $\times$ & $\times$ \\
azz & Latn & Uto-Aztecan & \checkmark & \checkmark & $\times$ & $\times$ \\ %
bak & Cyrl & Turkic & \checkmark & $\times$ & $\times$ & \checkmark \\ %
bam & Latn & Mande & \checkmark & \checkmark & \checkmark (R2) & \checkmark \\
ban & Latn & Austronesian & $\times$ & $\times$ & $\times$ & \checkmark \\
bas & Latn & Atlantic-Congo & \checkmark & $\times$ & $\times$ & $\times$ \\   %
bba & Latn & Atlantic-Congo & \checkmark & \checkmark & $\times$ & $\times$ \\
bbc & Latn & Austronesian & $\times$ & \checkmark & $\times$ & $\times$ \\
bcc & Arab & Indo-European & $\times$ & \checkmark & \checkmark (R2) & $\times$ \\
bel & Cyrl & Indo-European & \checkmark & $\times$ & \checkmark (R2) & \checkmark \\  %
bem & Latn & Atlantic-Congo & $\times$ & $\times$ & \checkmark (R2) & \checkmark \\
ben & Beng & Indo-European & \checkmark & $\times$ & \checkmark (R1, R2) & \checkmark \\
ben & Latn & Indo-European & \checkmark & $\times$ & $\times$ & $\times$ \\
bfa & Latn & Nilotic & $\times$ & \checkmark & $\times$ & $\times$ \\
bft & Arab & Sino-Tibetan & \checkmark & $\times$ & $\times$ & $\times$ \\   %
bhb & Deva & Indo-European & \checkmark & $\times$ & \checkmark (R2) & $\times$ \\  %
bho & Deva & Indo-European & \checkmark & $\times$ & $\times$ & \checkmark \\
bib & Latn & Mande & $\times$ & \checkmark & $\times$ & $\times$ \\
bjn & Arab & Austronesian & $\times$ & $\times$ & $\times$ & \checkmark \\
bjn & Latn & Austronesian & $\times$ & $\times$ & $\times$ & \checkmark \\
blt & Latn & Tai-Kadai & $\times$ & \checkmark & $\times$ & $\times$ \\
bod & Tibt & Sino-Tibetan & \checkmark & $\times$ & $\times$ & \checkmark \\
bom & Latn & Atlantic-Congo & $\times$ & \checkmark & $\times$ & $\times$ \\
bos & Latn & Indo-European & \checkmark & $\times$ & $\times$ & \checkmark \\
bre & Latn & Indo-European & \checkmark & $\times$ & \checkmark (R2) & $\times$ \\  %
brh & Arab & Dravidian & \checkmark & $\times$ & $\times$ & $\times$ \\   %
brx & Deva & Sino-Tibetan & \checkmark & $\times$ & $\times$ & \checkmark \\
bsh & Arab & Indo-European & \checkmark & $\times$ & $\times$ & $\times$ \\   %
bsk & Arab & Burushaski & \checkmark & $\times$ & $\times$ & $\times$ \\
bsq & Latn & Kru & $\times$ & \checkmark & \checkmark (R2) & $\times$ \\
bug & Latn & Austronesian & $\times$ & $\times$ & $\times$ & \checkmark \\
bul & Cyrl & Indo-European & \checkmark & $\times$ & $\times$ & \checkmark \\
cak & Latn & Mayan & \checkmark & $\times$ & $\times$ & $\times$ \\   %
cat & Latn & Indo-European & \checkmark & $\times$ & \checkmark (R2) & \checkmark \\
ceb & Latn & Austronesian & \checkmark & $\times$ & \checkmark (R2) & \checkmark \\
ces & Latn & Indo-European & \checkmark & $\times$ & \checkmark (R1) & \checkmark \\
che & Cyrl & Nakh-Daghestanian & \checkmark & $\times$ & $\times$ & $\times$ \\ %
chr & Cher & Iroquoian & \checkmark & $\times$ & $\times$ & $\times$ \\
chv & Cyrl & Turkic & \checkmark & $\times$ & $\times$ & \checkmark \\ %
cja & Arab & Austronesian & \checkmark & $\times$ & $\times$ & $\times$ \\
cjk & Latn & Atlantic-Congo & \checkmark & \checkmark & $\times$ & \checkmark \\
ckb & Arab & Indo-European & \checkmark & $\times$ & \checkmark (R2) & \checkmark \\  %
ckl & Latn & Afro-Asiatic & \checkmark & $\times$ & $\times$ & $\times$ \\
cmn & Hans & Sino-Tibetan & \checkmark & $\times$ & \checkmark (R1) & \checkmark \\
cmn & Hant & Sino-Tibetan & \checkmark & $\times$ & \checkmark (R1) & \checkmark \\
crh & Latn & Turkic & $\times$ & $\times$ & $\times$ & \checkmark \\
crk & Cans & Algic & \checkmark & $\times$ & \checkmark (R1) & $\times$ \\
crk & Latn & Algic & \checkmark & $\times$ & $\times$ & $\times$ \\
cux & Latn & Otomanguean & \checkmark & $\times$ & $\times$ & $\times$ \\   %
cym & Latn & Indo-European & \checkmark & $\times$ & $\times$ & \checkmark \\
dan & Latn & Indo-European & \checkmark & $\times$ & $\times$ & \checkmark \\
daq & Deva & Dravidian & \checkmark & $\times$ & $\times$ & $\times$ \\
dar & Cyrl & Nakh-Daghestanian & $\times$ & $\times$ & $\times$ & \checkmark \\
deu & Latn & Indo-European & \checkmark & $\times$ & \checkmark (R1) & \checkmark \\
dga & Latn & Atlantic-Congo & $\times$ & \checkmark & \checkmark (R2) & $\times$ \\
dgo & Deva & Indo-European & \checkmark & $\times$ & $\times$ & \checkmark \\
dik & Latn & Nilotic & \checkmark & \checkmark & \checkmark (R2) & \checkmark \\
diq & Latn & Indo-European & \checkmark & $\times$ & $\times$ & $\times$ \\
div & Thaa & Indo-European & \checkmark & $\times$ & $\times$ & $\times$ \\
djc & Latn & Dajuic & \checkmark & $\times$ & $\times$ & $\times$ \\
dje & Latn & Songhay & \checkmark & $\times$ & \checkmark (R1) & $\times$ \\
dnj & Latn & Mande & $\times$ & \checkmark & $\times$ & $\times$ \\
dtm & Latn & Dogon & \checkmark & $\times$ & $\times$ & $\times$ \\
dts & Latn & Dogon & \checkmark & $\times$ & $\times$ & $\times$ \\
dua & Latn & Atlantic-Congo & \checkmark & $\times$ & $\times$ & $\times$ \\   %
dyu & Latn & Mande & $\times$ & \checkmark & $\times$ & \checkmark \\
dzo & Tibt & Sino-Tibetan & \checkmark & \checkmark & $\times$ & \checkmark \\
ekk & Latn & Uralic & \checkmark & $\times$ & \checkmark (R2) & \checkmark \\
ell & Grek & Indo-European & \checkmark & $\times$ & \checkmark (R1) & \checkmark \\
enb & Latn & Nilotic & \checkmark & $\times$ & $\times$ & $\times$ \\
eng & Latn & Indo-European & \checkmark & \checkmark & \checkmark (R1, R2) & \checkmark \\
enl & Latn & Lengua-Mascoy & \checkmark & $\times$ & $\times$ & $\times$ \\
epo & Latn & Esperanto & $\times$ & $\times$ & $\times$ & \checkmark \\
eto & Latn & Atlantic-Congo & \checkmark & $\times$ & $\times$ & $\times$ \\   %
eus & Latn & Basque & \checkmark & $\times$ & \checkmark (R2) & \checkmark \\
ewe & Latn & Atlantic-Congo & $\times$ & \checkmark & $\times$ & \checkmark \\
ewo & Latn & Atlantic-Congo & \checkmark & $\times$ & $\times$ & $\times$ \\
fao & Latn & Indo-European & \checkmark & $\times$ & $\times$ & \checkmark \\
fia & Copt & Nubian & \checkmark & $\times$ & $\times$ & $\times$ \\
fij & Latn & Austronesian & $\times$ & \checkmark & $\times$ & \checkmark \\
fil & Latn & Austronesian & $\times$ & $\times$ & $\times$ & \checkmark \\
fin & Latn & Uralic & \checkmark & $\times$ & $\times$ & \checkmark \\
fon & Latn & Atlantic-Congo & $\times$ & \checkmark & $\times$ & \checkmark \\
fra & Latn & Indo-European & \checkmark & \checkmark & \checkmark (R1, R2) & \checkmark \\
fry & Latn & Indo-European & \checkmark & $\times$ & \checkmark (R2) & $\times$ \\  %
fuc & Latn & Atlantic-Congo & \checkmark & $\times$ & $\times$ & $\times$ \\
fur & Latn & Indo-European & $\times$ & $\times$ & $\times$ & \checkmark \\
fuv & Latn & Atlantic-Congo & \checkmark & \checkmark & \checkmark (R2) & \checkmark \\
fvr & Latn & Furan & \checkmark & $\times$ & $\times$ & $\times$ \\
gax & Latn & Afro-Asiatic & \checkmark & $\times$ & $\times$ & $\times$ \\
gaz & Latn & Afro-Asiatic & \checkmark & $\times$ & \checkmark (R1) & \checkmark \\
gil & Latn & Austronesian & \checkmark & $\times$ & \checkmark (R1) & $\times$ \\
gkp & Latn & Mande & \checkmark & $\times$ & $\times$ & $\times$ \\
gla & Latn & Indo-European & \checkmark & $\times$ & $\times$ & \checkmark \\
gle & Latn & Indo-European & \checkmark & $\times$ & $\times$ & \checkmark \\
glg & Latn & Indo-European & \checkmark & $\times$ & $\times$ & \checkmark \\
gom & Deva & Indo-European & \checkmark & $\times$ & $\times$ & \checkmark \\
gor & Latn & Austronesian & $\times$ & \checkmark & $\times$ & $\times$ \\
grt & Latn & Sino-Tibetan & $\times$ & \checkmark & $\times$ & $\times$ \\
guc & Latn & Arawakan & \checkmark & $\times$ & \checkmark (R1) & $\times$ \\
gug & Latn & Tupian & \checkmark & $\times$ & \checkmark (R2) & \checkmark \\ %
guj & Gujr & Indo-European & \checkmark & $\times$ & $\times$ & \checkmark \\
guz & Latn & Atlantic-Congo & \checkmark & $\times$ & $\times$ & $\times$ \\
gxx & Latn & Kru & \checkmark & $\times$ & $\times$ & $\times$ \\
hat & Latn & Indo-European & \checkmark & $\times$ & \checkmark (R2) & \checkmark \\
hau & Latn & Afro-Asiatic & \checkmark & $\times$ & \checkmark (R2) & \checkmark \\
heb & Hebr & Afro-Asiatic & \checkmark & $\times$ & \checkmark (R2) & \checkmark \\
heh & Latn & Atlantic-Congo & \checkmark & \checkmark & $\times$ & $\times$ \\
hig & Latn & Afro-Asiatic & $\times$ & \checkmark & $\times$ & $\times$ \\
hin & Deva & Indo-European & \checkmark & \checkmark & \checkmark (R1, R2) & \checkmark \\
hin & Latn & Indo-European & \checkmark & $\times$ & \checkmark (R1) & $\times$ \\
hne & Deva & Indo-European & \checkmark & $\times$ & $\times$ & \checkmark \\
hrv & Latn & Indo-European & \checkmark & $\times$ & \checkmark (R1) & \checkmark \\
hun & Latn & Uralic & \checkmark & $\times$ & \checkmark (R1) & \checkmark \\
hve & Latn & Huavean & \checkmark & $\times$ & $\times$ & $\times$ \\
hye & Armn & Indo-European & \checkmark & $\times$ & \checkmark (R2) & \checkmark \\
ibo & Latn & Atlantic-Congo & \checkmark & $\times$ & \checkmark (R2) & \checkmark \\
ijc & Latn & Ijoid & \checkmark & $\times$ & $\times$ & $\times$ \\
ilo & Latn & Austronesian & \checkmark & $\times$ & $\times$ & \checkmark \\
ind & Latn & Austronesian & \checkmark & \checkmark & \checkmark (R1, R2) & \checkmark \\
irk & Latn & Afro-Asiatic & \checkmark & \checkmark & $\times$ & $\times$ \\
isl & Latn & Indo-European & \checkmark & $\times$ & $\times$ & \checkmark \\
ita & Latn & Indo-European & \checkmark & $\times$ & \checkmark (R1, R2) & \checkmark \\
jav & Latn & Austronesian & \checkmark & $\times$ & \checkmark (R1) & \checkmark \\
jmc & Latn & Atlantic-Congo & \checkmark & \checkmark & $\times$ & $\times$ \\
jnj & Latn & Ta-Ne-Omotic & \checkmark & $\times$ & $\times$ & $\times$ \\
jpn & Jpan & Japonic & \checkmark & $\times$ & \checkmark (R1) & \checkmark \\
kaa & Cyrl & Turkic & \checkmark & $\times$ & \checkmark (R1) & \checkmark \\
kab & Latn & Afro-Asiatic & $\times$ & $\times$ & $\times$ & \checkmark \\
kac & Latn & Sino-Tibetan & \checkmark & \checkmark & $\times$ & \checkmark \\
kai & Latn & Afro-Asiatic & \checkmark & $\times$ & $\times$ & $\times$ \\
kal & Latn & Eskimo-Aleut & \checkmark & $\times$ & \checkmark (R1) & $\times$ \\
kam & Latn & Atlantic-Congo & \checkmark & \checkmark & $\times$ & \checkmark \\
kan & Knda & Dravidian & \checkmark & $\times$ & $\times$ & \checkmark \\
kas & Arab & Indo-European & $\times$ & $\times$ & $\times$ & \checkmark \\
kas & Deva & Indo-European & $\times$ & $\times$ & $\times$ & \checkmark \\
kat & Geor & Kartvelian & \checkmark & $\times$ & \checkmark (R2) & \checkmark \\
kaz & Cyrl & Turkic & \checkmark & $\times$ & $\times$ & \checkmark \\
kbp & Latn & Atlantic-Congo & $\times$ & \checkmark & $\times$ & \checkmark \\
kde & Latn & Atlantic-Congo & $\times$ & \checkmark & $\times$ & $\times$ \\
kdj & Latn & Nilotic & $\times$ & \checkmark & \checkmark (R2) & $\times$ \\
kea & Latn & Indo-European & \checkmark & $\times$ & $\times$ & \checkmark \\
kek & Latn & Mayan & \checkmark & \checkmark & $\times$ & $\times$ \\
khk & Cyrl & Mongolic-Khitan & \checkmark & $\times$ & $\times$ & \checkmark \\
khm & Khmr & Austroasiatic & \checkmark & $\times$ & \checkmark (R1) & \checkmark \\
khq & Latn & Songhay & \checkmark & $\times$ & $\times$ & $\times$ \\
khw & Latn & Indo-European & \checkmark & $\times$ & $\times$ & $\times$ \\
kik & Latn & Atlantic-Congo & $\times$ & $\times$ & $\times$ & \checkmark \\
kin & Latn & Atlantic-Congo & \checkmark & $\times$ & \checkmark (R2) & \checkmark \\
kir & Cyrl & Turkic & \checkmark & $\times$ & $\times$ & \checkmark \\
kls & Arab & Indo-European & \checkmark & $\times$ & $\times$ & $\times$ \\   %
kmb & Latn & Atlantic-Congo & \checkmark & \checkmark & $\times$ & \checkmark \\
kmr & Latn & Indo-European & \checkmark & $\times$ & $\times$ & \checkmark \\
knc & Arab & Saharan & \checkmark & \checkmark & \checkmark (R2) & \checkmark \\
knc & Latn & Saharan & $\times$ & $\times$ & $\times$ & \checkmark \\
knw & Latn & Kxa & \checkmark & $\times$ & $\times$ & $\times$ \\
kor & Kore & Koreanic & \checkmark & $\times$ & \checkmark (R1) & \checkmark \\
krt & Latn & Saharan & \checkmark & $\times$ & $\times$ & $\times$ \\
kru & Deva & Dravidian & \checkmark & $\times$ & \checkmark (R1) & $\times$ \\
ksf & Latn & Atlantic-Congo & \checkmark & $\times$ & $\times$ & $\times$ \\   %
ktu & Latn & Atlantic-Congo & \checkmark & \checkmark & $\times$ & \checkmark \\
kuj & Latn & Atlantic-Congo & \checkmark & $\times$ & $\times$ & $\times$ \\
kus & Latn & Atlantic-Congo & $\times$ & \checkmark & $\times$ & $\times$ \\
kwy & Latn & Atlantic-congo & \checkmark & $\times$ & $\times$ & $\times$ \\
kxp & Arab & Indo-European & \checkmark & $\times$ & $\times$ & $\times$ \\   %
lao & Laoo & Tai-Kadai & \checkmark & $\times$ & $\times$ & \checkmark \\
led & Latn & Central Sudanic & \checkmark & $\times$ & $\times$ & $\times$ \\
lgg & Latn & Central Sudanic & \checkmark & \checkmark & $\times$ & $\times$ \\
lia & Latn & Atlantic-Congo & $\times$ & \checkmark & $\times$ & $\times$ \\
lij & Latn & Indo-European & \checkmark & $\times$ & \checkmark (R1) & \checkmark \\
lim & Latn & Indo-European & \checkmark & $\times$ & $\times$ & \checkmark \\
lin & Latn & Atlantic-Congo & \checkmark & $\times$ & \checkmark (R1, R2) & \checkmark \\
lir & Latn & Pidgin & \checkmark & $\times$ & \checkmark (R2) & $\times$ \\  %
lit & Latn & Indo-European & \checkmark & $\times$ & $\times$ & \checkmark \\
lld & Latn & Indo-European & $\times$ & $\times$ & $\times$ & \checkmark \\
lmo & Latn & Indo-European & $\times$ & $\times$ & $\times$ & \checkmark \\
loa & Latn & North Halmahera & \checkmark & $\times$ & $\times$ & $\times$ \\
loh & Latn & Surmic & \checkmark & $\times$ & $\times$ & $\times$ \\
lon & Latn & Atlantic-Congo & $\times$ & \checkmark & $\times$ & $\times$ \\
ltg & Latn & Indo-European & $\times$ & $\times$ & $\times$ & \checkmark \\
ltz & Latn & Indo-European & $\times$ & $\times$ & $\times$ & \checkmark \\
lua & Latn & Atlantic-Congo & $\times$ & \checkmark & \checkmark (R2) & \checkmark \\
lug & Latn & Atlantic-Congo & \checkmark & \checkmark & $\times$ & \checkmark \\
luo & Latn & Nilotic & \checkmark & \checkmark & \checkmark (R2) & \checkmark \\ %
lus & Latn & Sino-Tibetan & $\times$ & $\times$ & $\times$ & \checkmark \\
lvs & Latn & Indo-European & \checkmark & $\times$ & $\times$ & \checkmark \\
mad & Latn & Austronesian & $\times$ & \checkmark & \checkmark (R2) & $\times$ \\
maf & Latn & Afro-Asiatic & \checkmark & $\times$ & $\times$ & $\times$ \\
mag & Deva & Indo-European & $\times$ & $\times$ & $\times$ & \checkmark \\
mah & Latn & Austronesian & $\times$ & \checkmark & $\times$ & $\times$ \\
mai & Deva & Indo-European & \checkmark & $\times$ & $\times$ & \checkmark \\
mak & Latn & Austronesian & $\times$ & \checkmark & $\times$ & $\times$ \\
mal & Mlym & Dravidian & \checkmark & $\times$ & \checkmark (R2) & \checkmark \\
mam & Latn & Mayan & \checkmark & \checkmark & $\times$ & $\times$ \\
mar & Deva & Indo-European & \checkmark & $\times$ & \checkmark (R2) & \checkmark \\
mas & Latn & Nilotic & \checkmark & $\times$ & $\times$ & $\times$ \\
men & Latn & Mande & $\times$ & \checkmark & $\times$ & $\times$ \\
mey & Latn & Afro-Asiatic & \checkmark & $\times$ & $\times$ & $\times$ \\
mfe & Latn & Indo-European & $\times$ & $\times$ & $\times$ & \checkmark \\
mhr & Cyrl & Uralic & $\times$ & $\times$ & $\times$ & \checkmark \\
mie & Latn & Otomanguean & \checkmark & $\times$ & $\times$ & $\times$ \\
mim & Latn & Otomanguean & $\times$ & \checkmark & \checkmark (R2) & $\times$ \\
min & Arab & Austronesian & \checkmark & $\times$ & \checkmark (R2) & \checkmark \\
min & Latn & Austronesian & $\times$ & $\times$ & $\times$ & \checkmark \\
mio & Latn & Otomanguean & $\times$ & \checkmark & $\times$ & $\times$ \\
miq & Latn & Misumalpan & \checkmark & \checkmark & $\times$ & $\times$ \\
mkd & Cyrl & Indo-European & \checkmark & $\times$ & $\times$ & \checkmark \\
mlt & Latn & Afro-Asiatic & \checkmark & $\times$ & $\times$ & \checkmark \\
mni & Beng & Sino-Tibetan & $\times$ & $\times$ & $\times$ & \checkmark \\
mni & Mtei & Sino-Tibetan & $\times$ & \checkmark & \checkmark (R2) & \checkmark \\
mos & Latn & Atlantic-Congo & \checkmark & \checkmark & \checkmark (R2) & \checkmark \\
mri & Latn & Austronesian & \checkmark & $\times$ & $\times$ & \checkmark \\
mrw & Latn & Austronesian & $\times$ & \checkmark & $\times$ & $\times$ \\
mtq & Latn & Austroasiatic & \checkmark & $\times$ & $\times$ & $\times$ \\
mya & Mymr & Sino-Tibetan & \checkmark & $\times$ & \checkmark (R1) & \checkmark \\
myv & Cyrl & Uralic & $\times$ & $\times$ & $\times$ & \checkmark \\
myx & Latn & Atlantic-Congo & $\times$ & \checkmark & $\times$ & $\times$ \\
mzl & Latn & Mixe-Zoque & \checkmark & $\times$ & $\times$ & $\times$ \\
mzm & Latn & Atlantic-Congo & $\times$ & \checkmark & $\times$ & $\times$ \\
naq & Latn & Khoe-Kwadi & \checkmark & $\times$ & $\times$ & $\times$ \\
nga & Latn & Atlantic-Congo & $\times$ & \checkmark & $\times$ & $\times$ \\
ngl & Latn & Atlantic-Congo & $\times$ & \checkmark & $\times$ & $\times$ \\
ngu & Latn & Uto-Aztecan & $\times$ & \checkmark & $\times$ & $\times$ \\
nhe & Latn & Uto-Aztecan & \checkmark & $\times$ & $\times$ & $\times$ \\
nia & Latn & Austronesian & $\times$ & \checkmark & $\times$ & $\times$ \\
nij & Latn & Austronesian & $\times$ & \checkmark & $\times$ & $\times$ \\
nim & Latn & Atlantic-Congo & $\times$ & \checkmark & $\times$ & $\times$ \\
nld & Latn & Indo-European & \checkmark & $\times$ & \checkmark (R1, R2) & \checkmark \\
nlv & Latn & Uto-Aztecan & \checkmark & $\times$ & \checkmark (R2) & $\times$ \\  %
nno & Latn & Indo-European & \checkmark & $\times$ & $\times$ & \checkmark \\
nob & Latn & Indo-European & $\times$ & $\times$ & $\times$ & \checkmark \\
npi & Deva & Indo-European & \checkmark & $\times$ & \checkmark (R2) & \checkmark \\  %
nqo & Nkoo & N'Ko & $\times$ & $\times$ & $\times$ & \checkmark \\
nso & Latn & Atlantic-Congo & \checkmark & $\times$ & \checkmark (R2) & \checkmark \\
nuj & Latn & Atlantic-Congo & $\times$ & \checkmark & $\times$ & $\times$ \\
nus & Latn & Nilotic & \checkmark & \checkmark & \checkmark (R2) & \checkmark \\
nya & Latn & Atlantic-Congo & \checkmark & $\times$ & $\times$ & \checkmark \\
nyy & Latn & Atlantic-Congo & $\times$ & \checkmark & $\times$ & $\times$ \\
oci & Latn & Indo-European & $\times$ & $\times$ & $\times$ & \checkmark \\
ory & Orya & Indo-European & \checkmark & $\times$ & $\times$ & \checkmark \\
osi & Latn & Austronesian & $\times$ & $\times$ & \checkmark (R2) & $\times$ \\
pag & Latn & Austronesian & $\times$ & $\times$ & $\times$ & \checkmark \\
pam & Latn & Austronesian & $\times$ & \checkmark & $\times$ & $\times$ \\
pan & Guru & Indo-European & \checkmark & $\times$ & $\times$ & \checkmark \\
pap & Latn & Indo-European & $\times$ & $\times$ & $\times$ & \checkmark \\
pbs & Latn & Otomanguean & \checkmark & $\times$ & $\times$ & $\times$ \\
pbt & Arab & Indo-European & \checkmark & $\times$ & \checkmark (R2) & \checkmark \\ %
pcm & Latn & Indo-European & \checkmark & \checkmark & \checkmark (R2) & $\times$ \\  %
pes & Arab & Indo-European & \checkmark & $\times$ & \checkmark (R1, R2) & \checkmark \\
plt & Latn & Austronesian & \checkmark & $\times$ & $\times$ & \checkmark \\
pol & Latn & Indo-European & \checkmark & $\times$ & \checkmark (R1, R2) & \checkmark \\
por & Latn & Indo-European & \checkmark & $\times$ & \checkmark (R1) & \checkmark \\
prs & Arab & Indo-European & $\times$ & $\times$ & $\times$ & \checkmark \\
quc & Latn & Mayan & \checkmark & \checkmark & $\times$ & $\times$ \\
quh & Latn & Quechuan & \checkmark & $\times$ & $\times$ & $\times$ \\
quy & Latn & Quechuan & $\times$ & $\times$ & $\times$ & \checkmark \\
quz & Latn & Quechuan & \checkmark & $\times$ & $\times$ & $\times$ \\
rhg & Rohg & Indo-European & $\times$ & \checkmark & $\times$ & $\times$ \\
rim & Latn & Atlantic-Congo & $\times$ & \checkmark & $\times$ & $\times$ \\
rmy & Latn & Indo-European & $\times$ & \checkmark & $\times$ & $\times$ \\
rob & Latn & Austronesian & \checkmark & $\times$ & $\times$ & $\times$ \\
roh & Latn & Indo-European & \checkmark & $\times$ & $\times$ & $\times$ \\
ron & Latn & Indo-European & \checkmark & $\times$ & \checkmark (R1) & \checkmark \\
run & Latn & Atlantic-Congo & $\times$ & $\times$ & $\times$ & \checkmark \\
rus & Cyrl & Indo-European & \checkmark & \checkmark & \checkmark (R1, R2) & \checkmark \\
sag & Latn & Atlantic-Congo & $\times$ & $\times$ & $\times$ & \checkmark \\
san & Deva & Indo-European & $\times$ & $\times$ & $\times$ & \checkmark \\
sat & Olck & Austroasiatic & \checkmark & \checkmark & $\times$ & \checkmark \\  %
sba & Latn & Central Sudanic & \checkmark & \checkmark & \checkmark (R1) & $\times$ \\
scn & Latn & Indo-European & \checkmark & $\times$ & $\times$ & \checkmark \\
sgc & Latn & Nilotic & \checkmark & $\times$ & $\times$ & $\times$ \\
shk & Latn & Nilotic & $\times$ & \checkmark & $\times$ & $\times$ \\
shn & Mymr & Tai-Kadai & \checkmark & \checkmark & $\times$ & \checkmark \\
sif & Latn & Siamou & \checkmark & $\times$ & $\times$ & $\times$ \\
sin & Sinh & Indo-European & \checkmark & $\times$ & $\times$ & \checkmark \\
skr & Arab & Indo-European & \checkmark & $\times$ & $\times$ & $\times$ \\   %
slk & Latn & Indo-European & \checkmark & $\times$ & $\times$ & \checkmark \\
slv & Latn & Indo-European & \checkmark & $\times$ & $\times$ & \checkmark \\
sme & Latn & Uralic & \checkmark & $\times$ & $\times$ & $\times$ \\
smo & Latn & Austronesian & $\times$ & $\times$ & $\times$ & \checkmark \\
sna & Latn & Atlantic-congo & \checkmark & $\times$ & $\times$ & \checkmark \\
snd & Arab & Indo-European & \checkmark & $\times$ & $\times$ & \checkmark \\
snd & Deva & Indo-European & $\times$ & $\times$ & $\times$ & \checkmark \\
som & Latn & Afro-Asiatic & \checkmark & $\times$ & $\times$ & \checkmark \\
sot & Latn & Atlantic-Congo & \checkmark & $\times$ & \checkmark (R2) & \checkmark \\ %
spa & Latn & Indo-European & \checkmark & \checkmark & \checkmark (R1, R2) & \checkmark \\
sro & Latn & Indo-European & \checkmark & $\times$ & $\times$ & $\times$ \\  %
srd & Latn & Indo-European & $\times$ & $\times$ & $\times$ & \checkmark \\
srp & Cyrl & Indo-European & \checkmark & $\times$ & $\times$ & \checkmark \\
ssw & Latn & Atlantic-Congo & \checkmark & $\times$ & \checkmark (R2) & \checkmark \\  %
sun & Latn & Austronesian & \checkmark & $\times$ & $\times$ & \checkmark \\
swe & Latn & Indo-European & \checkmark & $\times$ & \checkmark (R1) & \checkmark \\
swh & Latn & Atlantic-Congo & \checkmark & \checkmark & \checkmark (R1, R2) & \checkmark \\
syl & Beng & Indo-European & $\times$ & \checkmark & $\times$ & $\times$ \\
szl & Latn & Indo-European & \checkmark & $\times$ & \checkmark (R2) & \checkmark \\  %
taj & Deva & Sino-Tibetan & $\times$ & \checkmark & $\times$ & $\times$ \\
tam & Latn & Dravidian & \checkmark & $\times$ & $\times$ & $\times$ \\
tam & Taml & Dravidian & \checkmark & $\times$ & \checkmark (R2) & \checkmark \\
taq & Latn & Afro-Asiatic & \checkmark & \checkmark & $\times$ & \checkmark \\
taq & Tfng & Afro-Asiatic & \checkmark & \checkmark & $\times$ & \checkmark \\
tat & Cyrl & Turkic & \checkmark & $\times$ & \checkmark (R2) & \checkmark \\  %
tda & Latn & Songhay & \checkmark & $\times$ & \checkmark (R2) & $\times$ \\  %
tel & Latn & Dravidian & \checkmark & $\times$ & $\times$ & \checkmark \\
tel & Telu & Dravidian & \checkmark & $\times$ & $\times$ & $\times$ \\
tem & Latn & Atlantic-Congo & $\times$ & \checkmark & $\times$ & $\times$ \\
teo & Latn & Nilotic & $\times$ & \checkmark & $\times$ & $\times$ \\
tgk & Cyrl & Indo-European & \checkmark & $\times$ & $\times$ & \checkmark \\
tgl & Latn & Austronesian & \checkmark & $\times$ & \checkmark (R1, R2) & $\times$ \\
tha & Thai & Tai-Kadai & \checkmark & $\times$ & \checkmark (R1) & \checkmark \\
tir & Ethi & Afro-Asiatic & \checkmark & \checkmark & \checkmark (R1) & \checkmark \\
toc & Latn & Totonacan & \checkmark & $\times$ & $\times$ & $\times$ \\
tpi & Latn & Indo-European & \checkmark & $\times$ & $\times$ & \checkmark \\
tpl & Latn & Otomanguean & \checkmark & $\times$ & $\times$ & $\times$ \\
tsg & Latn & Austronesian & \checkmark & $\times$ & \checkmark (R2) & $\times$ \\  %
tsn & Latn & Atlantic-Congo & \checkmark & \checkmark & \checkmark (R2) & \checkmark \\ %
tso & Latn & Atlantic-Congo & \checkmark & $\times$ & \checkmark (R2) & \checkmark \\
tsz & Latn & Tarascan & \checkmark & \checkmark & $\times$ & \checkmark \\ %
tui & Latn & Atlantic-Congo & \checkmark & $\times$ & $\times$ & $\times$ \\   %
tuk & Latn & Turkic & $\times$ & $\times$ & $\times$ & \checkmark \\
tum & Latn & Atlantic-Congo & $\times$ & \checkmark & $\times$ & \checkmark \\
tur & Latn & Turkic & \checkmark & $\times$ & \checkmark (R1) & \checkmark \\
twi & Latn & Atlantic-Congo & \checkmark & $\times$ & $\times$ & \checkmark \\
tyv & Cyrl & Turkic & $\times$ & $\times$ & $\times$ & \checkmark \\
tzh & Latn & Mayan & \checkmark & \checkmark & $\times$ & $\times$ \\
tzm & Tfng & Afro-Asiatic & \checkmark & \checkmark & \checkmark (R2) & $\times$ \\  %
tzo & Latn & Mayan & $\times$ & \checkmark & \checkmark (R2) & $\times$ \\
uig & Arab & Turkic & \checkmark & $\times$ & $\times$ & \checkmark \\
ukr & Cyrl & Indo-European & \checkmark & $\times$ & \checkmark (R1) & \checkmark \\
umb & Latn & Atlantic-Congo & \checkmark & \checkmark & $\times$ & \checkmark \\
urd & Arab & Indo-European & \checkmark & $\times$ & \checkmark (R1) & \checkmark \\
urd & Latn & Indo-European & \checkmark & $\times$ & $\times$ & $\times$ \\
uzn & Latn & Turkic & \checkmark & $\times$ & $\times$ & \checkmark \\
uzs & Arab & Turkic & $\times$ & $\times$ & $\times$ & \checkmark \\
vec & Latn & Indo-European & $\times$ & $\times$ & $\times$ & \checkmark \\
ven & Latn & Atlantic-congo & \checkmark & $\times$ & $\times$ & $\times$ \\
vie & Latn & Austroasiatic & \checkmark & $\times$ & \checkmark (R1) & \checkmark \\
wlv & Latn & Mataguayan & \checkmark & $\times$ & $\times$ & $\times$ \\
vmw & Latn & Atlantic-Congo& \checkmark & \checkmark & $\times$ & \checkmark \\
war & Latn & Aystronesian & \checkmark & $\times$ & \checkmark (R2) & \checkmark \\
wol & Latn & Atlantic-Congo& \checkmark & \checkmark & \checkmark (R2) & \checkmark \\
wsg & Deva & Dravidian & $\times$ & \checkmark & $\times$ & $\times$ \\
wuu & Hans & Sino-Tibetan & \checkmark & $\times$ & $\times$ & \checkmark \\
xho & Latn & Atlantic-Congo & \checkmark & $\times$ & \checkmark (R2) & \checkmark \\
xon & Latn & Atlantic-congo & $\times$ & \checkmark & $\times$ & $\times$ \\
xsr & Deva & Sino-Tibetan & $\times$ & \checkmark & $\times$ & $\times$ \\
xuu & Latn & Khoe-Kwadi & \checkmark & $\times$ & $\times$ & $\times$ \\
yao & Latn & Atlantic-Congo & $\times$ & \checkmark & $\times$ & $\times$ \\
ybb & Latn & Atlantic-Congo & $\times$ & \checkmark & $\times$ & $\times$ \\
ydd & Hebr & Indo-European & \checkmark & $\times$ & $\times$ & \checkmark \\
ydg & Arab & Indo-European & \checkmark & $\times$ & $\times$ & $\times$ \\   %
yor & Latn & Atlantic-Congo & \checkmark & \checkmark & \checkmark (R1, R2) & $\times$ \\
yua & Latn & Mayan & \checkmark & \checkmark & $\times$ & $\times$ \\
yue & Hant & Sino-Tibetan & \checkmark & $\times$ & $\times$ & $\times$ \\
zai & Latn & Otomanguean & \checkmark & $\times$ & $\times$ & $\times$ \\
zgh & Tfng & Afro-Asiatic & $\times$ & $\times$ & $\times$ & \checkmark \\
zsm & Latn & Austronesian & \checkmark & $\times$ & \checkmark (R1) & \checkmark \\
zne & Latn & Atlantic-Congo & \checkmark & \checkmark & $\times$ & $\times$ \\
zul & Latn & Atlantic-Congo & \checkmark & $\times$ & \checkmark (R2) & \checkmark \\
\bottomrule
\caption{Covered Languages for the evaluation dataset contribution in this work (275 in \bouquet, and 120 in Met-\bouquet), the training \medley dataset and our previous and ongoing evaluation community effort of \floresplus.}
\label{tab:languages}
\end{longtable}

\end{document}